\documentclass[10pt,twocolumn,letterpaper]{article}

\usepackage[pagenumbers]{wacv} 

\usepackage{graphicx}
\usepackage{amsmath}
\usepackage{amssymb}
\usepackage{booktabs}
\usepackage{bm}
\usepackage{tikz}
\usepackage{graphicx}
\usepackage{booktabs}
\usepackage{etoolbox}
\usepackage[tracking=true]{microtype}
\usepackage{multirow}
\usepackage{enumitem}
\usepackage{comment}
\usepackage{titling}
\usepackage{colortbl}
\usepackage[accsupp]{axessibility}
\date{\vspace{-2em}}

\DeclareMathOperator*{\argmin}{arg\,min}
\def\PSNRColName{\textls[-20]{PSNR\kern-0.05em$\uparrow$}}
\def\SSIMColName{\textls[-20]{SSIM\kern-0.05em$\uparrow$}}
\def\LPIPSColName{\textls[-40]{LPIPS\kern-0.05em$\downarrow$}}
\def\AlgorithmColName{{\textbf{Algorithm\kern-0.05em}}}

\newlength{\figurewidth}
\newlength{\figureheight}
\newlength{\figurerowheight}

\usepackage[pagebackref,breaklinks,colorlinks=false]{hyperref}

\usepackage[capitalize]{cleveref}
\newcommand{\secref}[1]{Section~\ref{#1}}
\newcommand{\tabref}[1]{Table~\ref{#1}}
\newcommand{\figref}[1]{Fig.~\ref{#1}}
\newcommand{\mywidth}{0.15\textwidth}
\newcommand{\myheight}{0.15\textheight}
\newcommand{\myfont}{\scriptsize}

\newcommand{\boldparagraph}[1]{\noindent{\bf #1} }
\newcommand{\ccgray}{\cellcolor{gray!10}}

\def\wacvPaperID{487} 
\def\confName{WACV}
\def\confYear{2025}

\begin{document}

\title{DN-Splatter: Depth and Normal Priors for \\ Gaussian Splatting and Meshing}

\author{Matias Turkulainen$^*{}^1$ \and Xuqian Ren$^*{}^2$ \and Iaroslav Melekhov$^3$ \and Otto Seiskari$^4$ \and Esa Rahtu$^2$ \and Juho Kannala$^{3,4}$ \\ $^1$~ETH Zurich, $^2$~Tampere University, $^3$~Aalto University, $^4$~Spectacular AI
}

\maketitle

\footnotetext[1]{Denotes equal contribution}
\begin{abstract}
    High-fidelity 3D reconstruction of common indoor scenes is crucial for VR and AR applications. 3D Gaussian splatting, a novel differentiable rendering technique, has achieved state-of-the-art novel view synthesis results with high rendering speeds and relatively low training times. However, its performance on scenes commonly seen in indoor datasets is poor due to the lack of geometric constraints during optimization. In this work, we explore the use of readily accessible geometric cues to enhance Gaussian splatting optimization in challenging, ill-posed, and textureless scenes. We extend 3D Gaussian splatting with depth and normal cues to tackle challenging indoor datasets and showcase techniques for efficient mesh extraction. Specifically, we regularize the optimization procedure with depth information, enforce local smoothness of nearby Gaussians, and use off-the-shelf monocular networks to achieve better alignment with the true scene geometry. We propose an adaptive depth loss based on the gradient of color images, improving depth estimation and novel view synthesis results over various baselines. Our simple yet effective regularization technique enables direct mesh extraction from the Gaussian representation, yielding more physically accurate reconstructions of indoor scenes. Our code will be released in \url{https://github.com/maturk/dn-splatter}.
    \vspace{-1em}
\end{abstract}

\section{Introduction}
\label{sec:intro}

\begin{figure*}[t!]
  \centering\scriptsize
  \includegraphics[width=140mm]{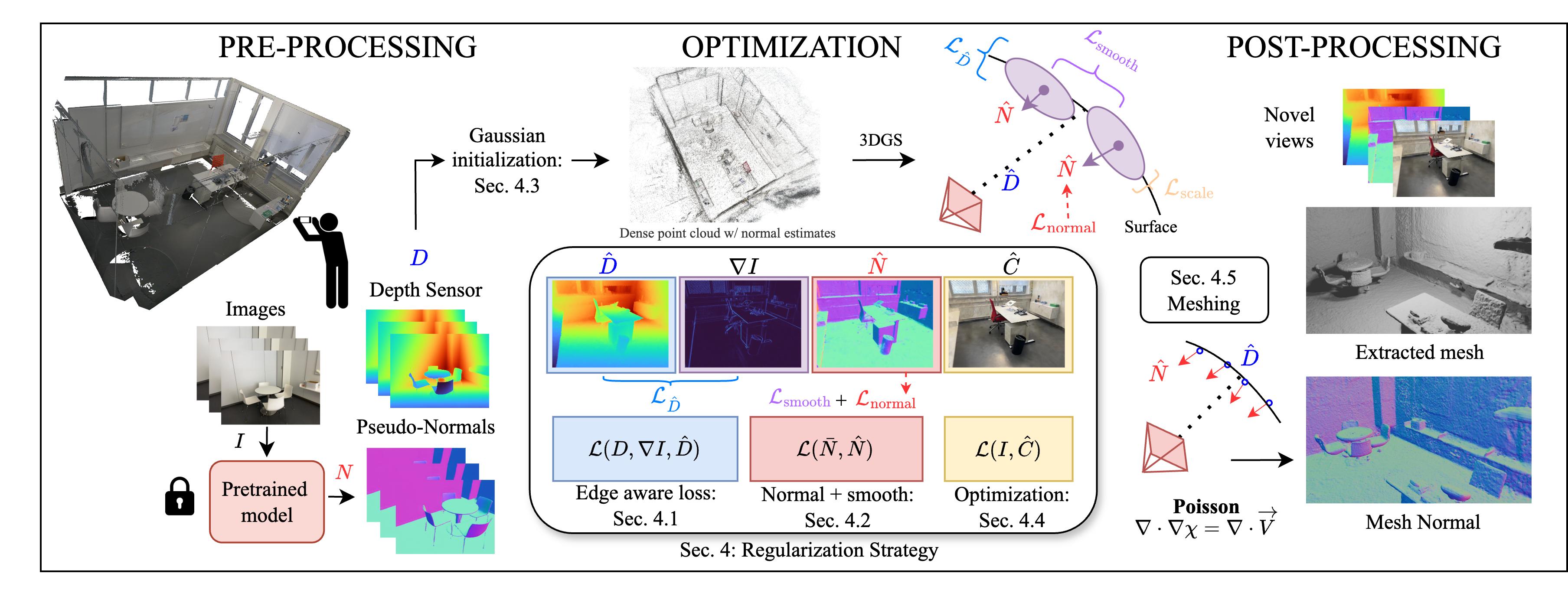}
  \vspace{-1em}
  \caption{\textbf{Overview:} We use depth and normal priors obtained from common handheld devices and general-purpose networks to enhance Gaussian splatting reconstruction quality. By regularizing Gaussian positions, local smoothness, and orientations, we demonstrate improvements in novel-view synthesis and achieve more accurate mesh reconstructions on a variety of challenging indoor room datasets.}
  \vspace{-1em}
  \label{fig:main-pipeline}
  \vspace{-1em}
\end{figure*}

The demand for high-fidelity 3D reconstruction of typical environments is increasing due to VR and AR applications. However, photorealistic and accurate 3D reconstruction of common indoor scenes from casually captured sensor data remains a persistent challenge in 3D computer vision. Textureless and less-observed regions cause ambiguities in reconstructions and do not provide enough constraints for valid geometric solutions. Recently, neural implicit representations have achieved success in high-fidelity 3D reconstruction by representing scenes as continuous volumes with fully differentiable properties \cite{mildenhall2020nerf, wang2021neus, li2023neuralangelo, barron2022mipnerf360}. However, the reconstruction of everyday indoor scenes still poses challenges, even for state-of-the-art methods. These methods rarely achieve good results in both photorealism and geometry reconstruction and often suffer from long training and rendering times, making them inaccessible for general use and VR/AR applications.

3D Gaussian splatting~\cite{kerbl20233d} introduces a novel method for rendering by representing a scene by many differentiable 3D Gaussian primitives with optimizable properties. This explicit representation enables real-time rendering of large, complex scenes--a capability that most neural implicit models lack. 3DGS is a more interoperable scene representation compared to neural methods since it encodes a scene's appearance and geometry directly through the location, shape, and color attributes of Gaussians. However, due to the lack of 3D cues and surface constraints during optimization, artifacts and ambiguities are likely to occur, resulting in floaters and poor surface reconstruction. Scenes can often contain millions of Gaussians, and their properties are directly modified by gradient descent based on photometric losses only. Little focus has been given to exploring better regularization techniques that result in visually and geometrically smoother and more plausible 3D reconstructions that can be converted into meshes, an important downstream application.

Although many modern smartphones are equipped with low-resolution depth sensors, these are rarely used for novel-view synthesis tasks. Motivated by this and advances in depth and normal estimation networks~\cite{bhat2023zoedepth, depthanything, eftekhar2021omnidata, bae2024dsine}, we explore the regularization of 3D Gaussian splatting with these geometric priors. Our goal is to enhance both photorealism and surface reconstruction in challenging indoor scenes. By designing an optimization strategy for 3D Gaussian splatting with depth and normal priors, we improve novel-view synthesis results over baselines whilst better respecting the captured scene geometry. We regularize the position of Gaussians with an edge-aware depth constraint and estimate normals from Gaussians to align them with the real surface boundaries estimated via monocular networks. We show how this simple regularization strategy, illustrated in \cref{fig:main-pipeline}, enables the extraction of meshes from the Gaussian scene representation, resulting in smoother and more geometrically accurate reconstructions. In summary, we make the following contributions:
\vspace{-0.5em}\begin{itemize}[leftmargin=*]
    \item We design an edge-aware depth loss for Gaussian splatting depth regularization to improve reconstruction on indoor scenes with imperfect depth estimates.%
    \vspace{-0.5em}
    \item We use monocular normal priors to align Gaussians with the scene and demonstrate how this aids reconstruction.%
    \vspace{-0.5em}
    \item We show how regularization with depth and normal cues enables efficient mesh extraction directly from the Gaussian scene with improved novel-view synthesis.%
\end{itemize}

\section{Related work}
Here we give a brief overview to image-based rendering (IBR) methods for scene reconstruction and an overview of prior methods utilizing geometry cues for regularization.

\boldparagraph{Traditional IBR.}
Reconstructing 3D geometry from images is a longstanding challenge in computer vision. Traditional methods like Structure-from-Motion (SfM)~\cite{snavely2006photo, schoenberger2016sfmcolmap1} and Multi-view Stereo (MVS)~\cite{Goesele2007MultiViewSF, zhang2020visibility} techniques have focused on reconstructing geometry via a sparse set of multi-view consistent 3D points obtained by triangulation of features from images~\cite{KPLD21,ruckert2022adop}. Learning-based approaches~\cite{Hartmann2017ICCV,patchM,Luo2016CVPR,mishchuk17nips} usually replace parts of the pipeline with learnable modules, leading to improvements in the generalizability of the methods. Other work focus on normal estimation~\cite{schoenberger2016mvscolmap2} and constructing triangle meshes~\cite{poisson, marchingcubes}.

\boldparagraph{Neural implicit IBR.}
Most success obtaining both photorealisic and geometrically accurate 3D reconstruction has been achieved with neural-based inverse rendering methods, most notably that of NeRF~\cite{mildenhall2020nerf}, which represents scenes as volumes with attributes encoded within a neural network and applies volume rendering \cite{Kajiya} to achieve impressive novel-view results. However, the 3D geometry extracted from NeRFs are often poorly defined and suffer from artifacts and floaters. Subsequent work has focused on improving the rendering quality and scene reconstruction through regularization techniques \cite{roessle2022depthpriorsnerf, kangle2021dsnerf} or by adopting other scene representations such as signed distance functions~\cite{Yu2022MonoSDF, wang2021neus, li2023neuralangelo} (SDFs) to improve geometry extraction.

\boldparagraph{Prior regularization.}
Prior regularization of neural implicit models has been an active area of research. Previous NeRF-based approaches add depth regularization to explicitly supervise ray termination~\cite{kangle2021dsnerf, wei2021nerfingmvs, roessle2022depthpriorsnerf} or impose smoothness constraints~\cite{niemeyer2021regnerf} on rendered depth maps. Other works explore regularizing with multi-view consistency~\cite{wang2023sparsenerf, kangle2021dsnerf, lao2023corresnerf} in sparse view settings. For SDF-based models, Manhattan-SDF~\cite{guo2022manhattan} uses planar constraints on walls and flat surfaces to improve indoor reconstruction, and MonoSDF~\cite{Yu2022MonoSDF} uses depth and normal monocular estimates for scene geometry regularization. In this work, we investigate the regularization of 3D Gaussian splatting optimization with depth and normal priors to enhance photometric and geometric reconstruction.

\boldparagraph{Meshable implicit representations.}
Surface extraction as triangle meshes is an important problem since most computer graphics pipelines still rely on triangle rasterization. Watertight meshes also provide a good approximation of scene geometry and surface quality, leading to the development of various metrics for mesh quality. Prior work has focused on extracting meshes from NeRF representations~\cite{Rakotosaona2023THREEDV,tang2022nerf2mesh, nerfstudio} with some success, but these methods often rely on expensive post-refinement stages. Most State-of-the-Art techniques use SDF or occupancy representations~\cite{Yu2022MonoSDF,yariv2021volume,wang2021neus, li2023neuralangelo} combined with marching cubes~\cite{marchingcubes} to achieve finer details. These methods involve querying and evaluating dense 3D volumes, often at multiple levels of detail, and are generally slow to train. In this work, we investigate extracting meshable surfaces directly from a Gaussian scene representation.

\boldparagraph{Meshable 3D Gaussians.}
Extracting meshable surfaces from Gaussian primitives is a relatively new topic. Keselman~\etal~\cite{keselman2023flexible} propose generating an oriented point set from a trained Gaussian scene to be meshed with Poisson reconstruction~\cite{poisson}, using back-projected depth maps and analytically estimated normals. However, without regularization, this approach results in noisy point clouds that are difficult to mesh. NeuSG~\cite{chen2023neusg} addresses this by jointly training a dense SDF neural implicit model \cite{wang2021neus} alongside a Gaussian scene, aligning Gaussians with SDF-estimated normals. The authors improve surface extraction from the Gaussian scene using SDF guidance; however, the approach requires long training times --- over 16 hours on high-end GPUs --- diminishing the appeal of 3DGS.

SuGaR~\cite{guedon2023sugar} proposes treating the positions of Gaussians as intersections of a level set and optimizes a signed-distance loss to converge Gaussians to the surface. Normals are estimated from the derivative of the signed distance, similar to Keselman~\etal. However, due to the lack of geometric priors, the reconstructions remain noisy (we show experimental comparing SuGaR meshing strategy in the supplement). SuGaR refines the coarse mesh with additional optimization, making the process computationally costly. In contrast, 2DGS~\cite{Huang2DGS2024} proposes to use a Gaussian surfel representation to explicitly model planar surfaces. However, we show in our experiments that without prior regularization, results on outward-facing indoor reconstructions are poor.
\section{Preliminaries}
Our work builds on 3D Gaussian splatting (3DGS,~\cite{kerbl20233d}) and we briefly describe the rasterization algorithm. 3DGS represents a scene using differentiable 3D Gaussian primitives, parameterized by their mean $\boldsymbol{\mu} \in \mathbb{R}^3$, a covariance matrix $\boldsymbol{\Sigma} \in \mathbb{R}^{3\times3}$ decomposed into a scaling vector $\boldsymbol{s} \in \mathbb{R}^3$ and a rotation quaternion $\boldsymbol{q} \in \mathbb{R}^4$, along with opacity $o \in \mathbb{R}$ and color $\boldsymbol{c} \in \mathbb{R}^3$, represented via spherical harmonics. Rendering a new view involves projecting 3D Gaussians into 2D Gaussians in camera space. These 2D Gaussians are \textit{z} depth-sorted and alpha-composited using the discrete volume rendering equation to produce a pixel color $\bf \hat{C}$:
\begin{align}
  {\bf \hat{C}}  & = \sum_{i \in N} {\bf{c}}_{i}\alpha_i T_i, \textrm{ where } T_i = \prod_{j = 1}^{i - 1} (1- \alpha_j)
\end{align}
\noindent where $T_i$ is the accumulated transmittance at pixel location $p$ and $\alpha_{i}$ is the blending coefficient for a Gaussian with center $\mu_i$ in screen space:
\begin{align}\label{gaussian-density}
  {\alpha_i}  & = o_i \cdot \exp{\left(-\frac{1}{2}(\boldsymbol{p}-\boldsymbol{\mu}_i)^\intercal \boldsymbol{\Sigma}_i^{-1}(\boldsymbol{p}-\boldsymbol{\mu}_i)\right)}.
\end{align}
The scene is typically initialized with sparse SfM points obtained from a pre-processing step~\cite{schoenberger2016sfmcolmap1, schoenberger2016mvscolmap2}. In this work, we also explore initialization using sensor depth readings. The Gaussian scene is optimized using the Adaptive Density Control (ADC) algorithm~\cite{kerbl20233d}, which progressively culls, splits, and duplicates Gaussians in the scene at fixed intervals based on Gaussian opacity, screen-space size, and the magnitude of the gradient of Gaussian means, respectively.
\section{Method}
We address the problem of achieving high-fidelity reconstruction of common indoor scenes that is both photorealistic and geometrically precise. In~\secref{sec:Depth regularization} we utilize sensor and monocular depth priors to regularize Gaussian positions with an edge-aware loss. Next, we extract normal directions from Gaussians and utilize normal cues for regularization in~\secref{sec:Normal estimation and regularization}. Additionally, we add a smoothing prior on rendered normal maps to better align nearby Gaussians during optimization in~\secref{ssec:Optimization} and initialize the Gaussian scene using dense depth information in~\secref{ssec:initialization}. Lastly, in~\secref{sec:Meshing}, we use the optimized Gaussian scene to directly extract meshes using Poisson surface reconstruction.
\subsection{Leveraging depth cues}\label{sec:Depth regularization}
\boldparagraph{Depth prediction.} Per-pixel z\textendash depth estimates $\hat{D}$ are rendered using the discrete volume rendering approximation similar to color values:
 \begin{align}
  {\hat{D}}  & = \sum_{i \in N} {{d}}_{i}\alpha_i \prod_{j = 1}^{i - 1} (1- \alpha_j)
 \end{align}
\noindent where $\textit{d}_{i}$ is the $i^{\text{th}}$ Gaussian z-depth coordinate in view space. Since 3DGS does not sort Gaussians individually per-pixel along a viewing ray, and instead relies on a single global sort for efficiency; this is only an approximation of per-pixel depth as explained in \cite{radl2024stopthepop}. Nevertheless, this approximation remains effective, particularly for more regular geometries typically encountered in indoor datasets. We normalize depth estimates with the final accumulated transmittance $T_i$ per pixel, ensuring that we correctly estimate a depth value for pixels where the accumulated transmittance does not equal 1. We rasterize color and depths simultaneously per pixel in a single CUDA kernel forward pass, improving inference and training speed compared to separate rendering steps.

\boldparagraph{Sensor depth regularization.} We directly apply depth regularization on predicted depth maps for datasets containing lidar or sensor depth measurements \cite{ren2023mushroom, yeshwanthliu2023scannetpp,replica19arxiv}. Common commercial depth sensors, especially low-resolution variants found in consumer devices like iPhones, often produce non-smooth edges at object boundaries and provide inaccurate readings. Based on this observation and inspired by~\cite{chung2023depth,kosheleva2023edge}, we propose a gradient-aware depth loss for adaptive depth regularization based on the current RGB image. The depth loss is lowered in regions with large image gradients, signifying edges, ensuring that regularization is more enforced on smoother texture-less regions that typically pose challenges for photometric regularization alone. Additionally, our experiments (\cf \tabref{tab:ab_main_depth_losses}) show that using a logarithmic penalty results in smoother reconstructions compared to linear or quadratic penalties. This insight drives our formulation of the gradient-aware depth loss, which effectively balances the regularization across different regions of the image, adapting to the scene's geometry and texture complexity. We define the gradient-aware depth loss as follows:
\begin{align}\label{eq:depth loss}
    \mathcal{L}_{\hat{D}} &= g_\text{rgb} \frac{1}{\vert \hat{D} \vert} \sum \log(1+\| \hat{D} -D\|_1)
\end{align}
\noindent where $g_\text{rgb} = \text{exp}(-\nabla I)$ and $\nabla I$ is the gradient of the current aligned RGB image. $\vert \hat{D} \vert$ indicates the total number of pixels in $\hat{D}$.

\boldparagraph{Monocular depth regularization.} For datasets containing no depth data, we rely on scale-aligned monocular depth estimation networks for regularization. We use off-the-shelf monocular depth networks, such as ZoeDepth~\cite{bhat2023zoedepth} and DepthAnything~\cite{depthanything}, for dense per-pixel depth priors. We address the scale ambiguity between estimated depths and the scene by comparing them with sparse SfM points, similar to prior work~\cite{Yu2022MonoSDF,chung2023depth}. Specifically, for each monocular depth estimate $D_{\textrm{mono}}$, we align the scale to match that of the sparse depth map $D_{\textrm{sparse}}$ obtained by projecting SfM points to the camera view. We solve for a per-image scale $a$ and shift $b$ parameter using the closed-form linear regression solution to:
 \begin{align}
 \label{eq:mono-sparse-alignment}
  \hat{a}, \hat{b}  = \argmin_{a,b} \sum_{ij} \| (a*D_{\text{mono},ij} + b) - D_{\text{sparse},ij}\|_2 ^2,
 \end{align}
\noindent where we denote $D_{\text{sparse},ij}$ and $D_{\text{mono},ij}$ as per-pixel correspondences between the two depth maps. We then apply the same loss as in \cref{eq:depth loss} for regularization.

\subsection{Leveraging normal cues}\label{sec:Normal estimation and regularization}
\boldparagraph{Normal prediction.} During optimization, we expect Gaussians to become flat, disc-like, with one scaling axis much smaller than the other two. This smaller scaling axis serves as an approximation of the normal direction. Specifically, we define a geometric normal for a Gaussian using a rotation matrix $\mathbb{R} \in {SO(3)}$, obtained from its quaternion $\boldsymbol{q}$, and scaling coefficients  $\boldsymbol{s} \in \mathbb{R}^{3}$:
 \begin{align}
      {\boldsymbol{\hat{n}}_i} = R \cdot \textrm{OneHot}{(\argmin(s_1, s_2, s_3))},
 \end{align}
where $\text{OneHot}(.)\in\mathbb{R}^{3}$ returns a unit vector with zeros everywhere except at the position where the scaling $\boldsymbol{s}_i = (s_1, s_2, s_3)$ is minimum. $R$ is obtained from the quaternion $q = (w,x,y,z)^{\top}$ using:
\begin{align}\label{eq:quat-to-rot}
    R=\begin{bmatrix}
    1 - 2 \left( y^2 + z^2 \right) & 2 \left( x y - w z \right) & 2 \left( x z + w y \right) \\
    2 \left( x y + w z \right) & 1 - 2 \left( x^2 + z^2 \right) & 2 \left( y z - w x \right) \\
    2 \left( x z - w y \right) & 2 \left( y z + w x \right) & 1 - 2 \left( x^2 + y^2 \right) \\
    \end{bmatrix}
\end{align}
We minimize one of the scaling axes during training to force Gaussians to become disc-like surfels:
\begin{align}
\begin{split}
\mathcal{L}_{\text{scale}} & = \sum_i \|\argmin(\boldsymbol{s}_i)\|_1.\\
\end{split}
\end{align}

To ensure correct orientations, we flip the direction of the normals at the beginning of training if the dot product between the current camera viewing direction and the Gaussian normal is negative. Normals are transformed into camera space using the current camera transform and alpha-composited according to the rendering equation to provide a single per-pixel normal estimate:
\begin{align}
  {\boldsymbol{\hat{N}}}  & = \sum_{i\in N} {{\boldsymbol{\hat{n}}}}_{i}\alpha_i T_i.
\end{align}

This approach derives normal estimates directly from the geometry of Gaussians. Consequently, during back-propagation, adjustments to scale and rotation parameters, \ie, covariance matrices, directly lead to updates in normal estimates. Therefore, no additional learnable parameters are needed. Intuitively, this results in Gaussians better conforming to the scene's geometry, as their orientations and scales are compelled to align with the surface normal.

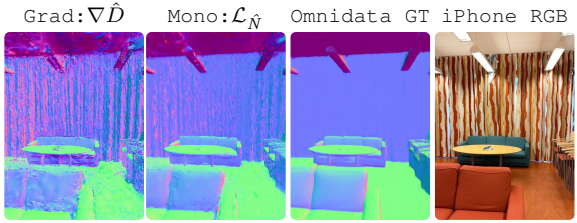
\begin{figure}[t!]
\centering
  \begin{subfigure}[b]{0.12\textwidth}
    \centering
    \subcaption*{\texttt{Grad:{$\nabla\hat{D}$}}}
    \begin{tikzpicture}
    \node [inner sep=0pt,clip,rounded corners=2pt] at (0,0) {\includegraphics[height=2.5cm]{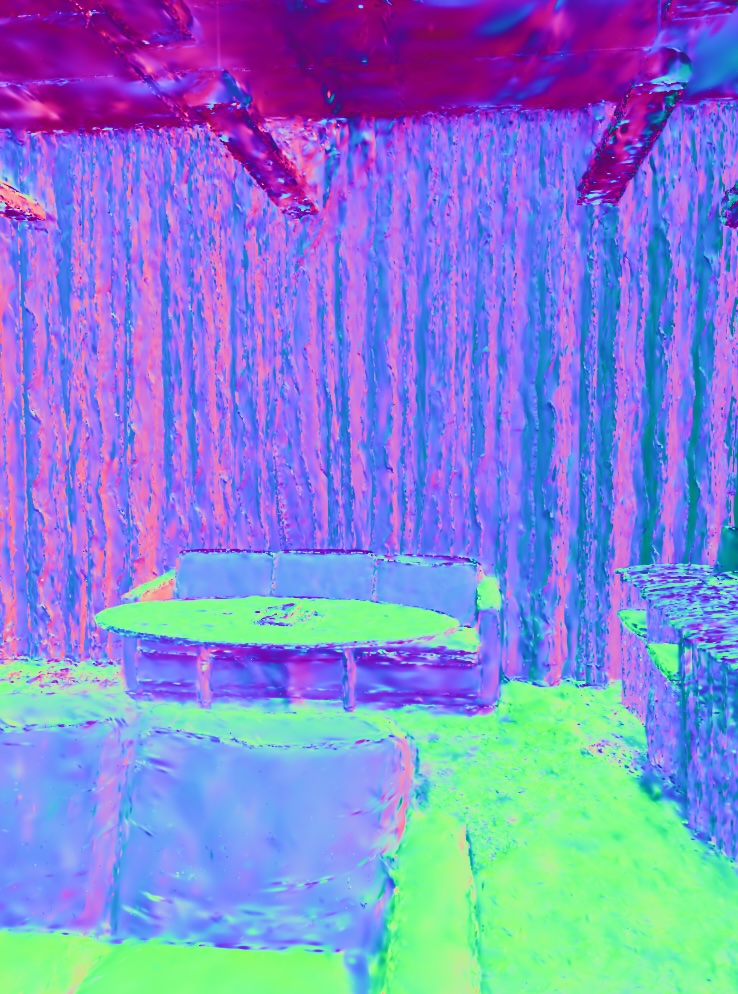}};
    \node[font=\tiny, text=white] at (0.2,-1.1) {};
    \end{tikzpicture}
  \end{subfigure}
  \hspace{-11pt}
  \begin{subfigure}[b]{0.12\textwidth}
    \centering
    \subcaption*{\texttt{Mono:$\mathcal{L}_{\it{\hat{N}}}$}}
    \begin{tikzpicture}
    \node [inner sep=0pt,clip,rounded corners=2pt] at (0,0) {\includegraphics[height=2.5cm]{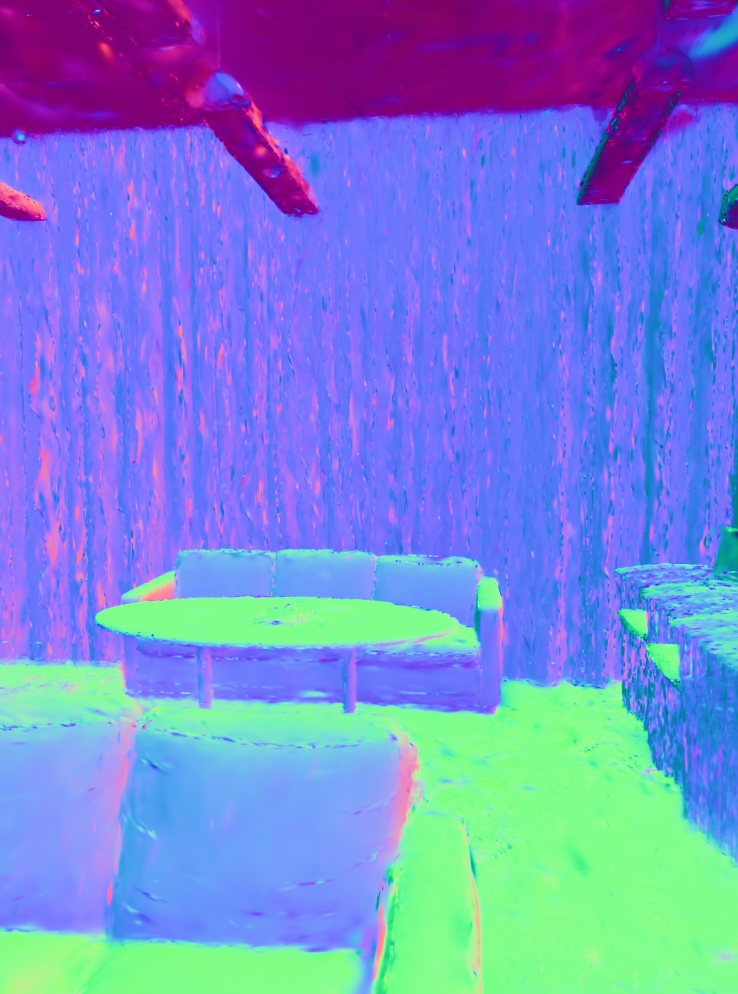}};
    \node[font=\tiny, text=white] at (0.2,-1.1) {};
    \end{tikzpicture}
  \end{subfigure}
  \hspace{-10pt}
  \begin{subfigure}[b]{0.12\textwidth}
    \centering
     \subcaption*{\texttt{Omnidata GT}}
    \begin{tikzpicture}
    \node [inner sep=0pt,clip,rounded corners=2pt] at (0,0) {\includegraphics[height=2.5cm]{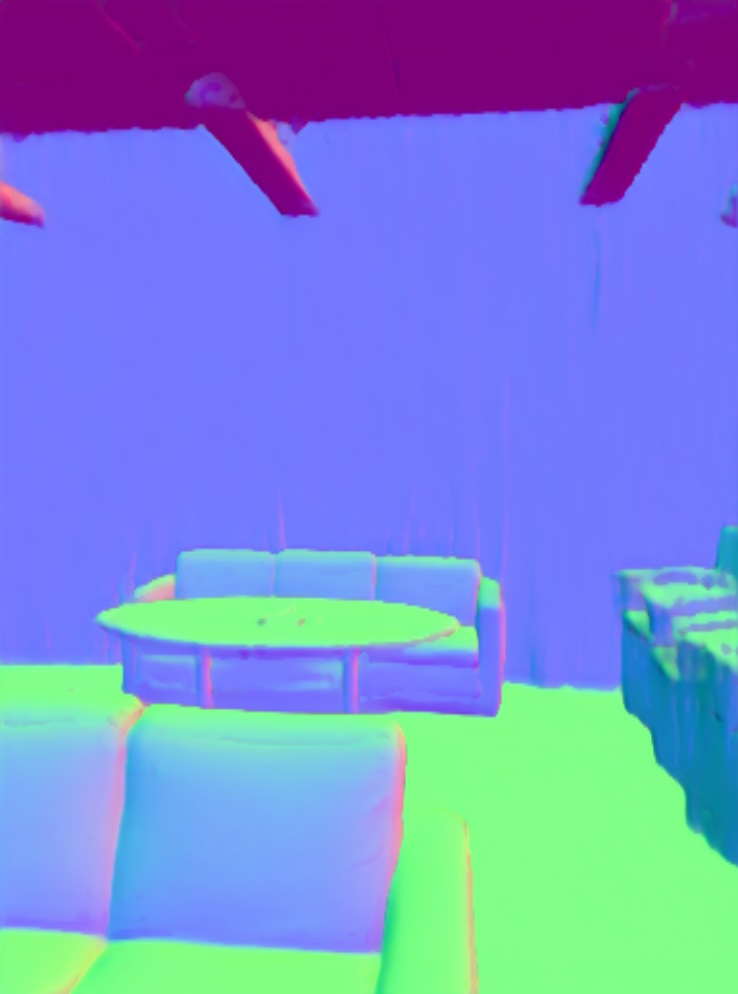}};
    \node[font=\tiny, text=white] at (0.2,-1.1) {};
    \end{tikzpicture}
  \end{subfigure}
  \hspace{-10pt}
  \begin{subfigure}[b]{0.12\textwidth}
    \centering
    \subcaption*{\texttt{iPhone RGB}}
    \begin{tikzpicture}
    \node [inner sep=0pt,clip,rounded corners=2pt] at (0,0) {\includegraphics[height=2.5cm]{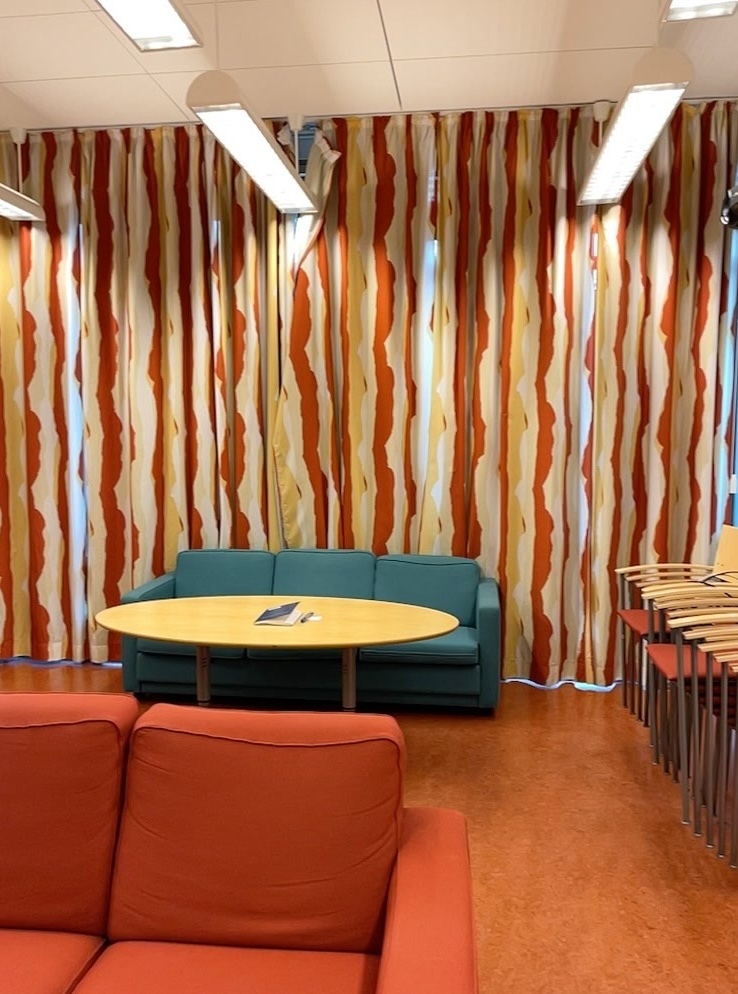}};
    \node[font=\tiny, text=white] at (0.2,-1.1) {};
    \end{tikzpicture}
  \end{subfigure}
  \caption{\textbf{Depth gradient vs. monocular normal supervision strategy.} (a) We observe that using pseudo normal maps derived from the gradient of rendered depths~\cite{R3DG2023} for supervision leads to noisy predicted normals compared to (b) normal supervision by estimates from a pretrained (c) Omnidata model~\cite{eftekhar2021omnidata}.}
  \vspace{-1em}
\label{fig:pseudo-normals}
\end{figure}

\boldparagraph{Monocular normals regularization.} Gao~\etal~\cite{R3DG2023} propose using pseudo-ground truth normal maps estimated from the gradient of rendered depths for supervision, referred to as $\nabla \hat{D}$. However, due to noise in rendered depth maps, especially in complex scenes, this method results in artifacts. Instead, we supervise predicted normals using monocular cues obtained from Omnidata~\cite{eftekhar2021omnidata}, which provide much smoother normal estimates. \cref{fig:pseudo-normals} highlights this difference.
We regularize with an L1 loss:
 \begin{align}
 \label{eq:normal-loss}
  \mathcal{L}_{\it{\hat{N}}} = \frac{1}{\vert \hat{N} \vert}\sum \| \boldsymbol{\hat{N}} - \boldsymbol{N} \|_1.
 \end{align}

We further apply a prior on the total variation of predicted normals, encouraging smooth normal predictions at neighboring pixels with:
\begin{align}\label{eq:tv loss}
\mathcal{L}_{\text{smooth}} = \sum_{i, j} \left( |\nabla_i\boldsymbol{ \hat{N}}_{i,j}| + |\nabla_j \boldsymbol{\hat{N}}_{i, j}|\right),
\end{align}
where $\hat{N}_{ij}$ represents estimated normal values at pixel position (\text{i}, \text{j}) and $\nabla$ represents the finite difference operator that convolves its input with $[-1,1]$ for the $i$-axis and $[-1,1]^\top$ for the $j$-axis. Thus, our normal regularization loss is defined as $\mathcal{L}_\text{normal}=\mathcal{L}_{\it{\hat{N}}}+\mathcal{L}_{\text{smooth}}.$ 

\subsection{Gaussian initialization}\label{ssec:initialization}
Rather than relying on SfM points for initialization, we make use of sensor depth readings, where available, by back-projecting depths into world coordinates for dense scene initialization. Additionally, we initialize Gaussian orientations by estimating normal directions from the initial point cloud using~\cite{Zhou2018}, aligning the Gaussian orientations $q$ with these estimated normals using \cref{eq:quat-to-rot}, and setting one of the scaling axes smaller than the others. This initialization helps normal estimation convergence.

\subsection{Optimization}\label{ssec:Optimization}
The final loss we use for optimization is defined as follows:
\begin{align}
\label{eq:final-loss}
\mathcal{L} = \mathcal{L}_{\hat{C}} + \lambda_\text{d}\mathcal{L}_{\hat{D}} +\mathcal{L}_{\text{scale}}+ (\underbrace{\lambda_\text{n}\mathcal{L}_{\it{\hat{N}}}+\lambda_\text{s}\mathcal{L}_{\text{smooth}}}_{\mathcal{L}_\text{normal}}),
\end{align}
\noindent where $\mathcal{L}_{\hat{C}}$ is the original photometric loss proposed in~\cite{kerbl20233d}. We set $\lambda_\text{d}=0.2$,  $\lambda_\text{n}=0.1$, and $\lambda_\text{s}=0.1$ in our experiments.

\subsection{Meshing}\label{sec:Meshing}
After optimizing with our depth and normal regularization using~\cref{eq:final-loss}, we apply Poisson surface reconstruction~\cite{poisson} to extract a mesh. With our regularization strategy, we ensure that the positions of Gaussians are well-distributed and aligned along the surface of the scene. We directly back-project rendered depth and normal maps from training views to create an oriented point set for meshing. Qualitative and quantitative differences between various Poisson methods and comparisons with the meshing approach proposed in SuGaR \cite{guedon2023sugar} are given in the Supplementary material.

\section{Experiments}
In this section, we demonstrate the proposed regularization strategy on mesh extraction and novel-view synthesis results using indoor datasets. We also give insight regarding various depth supervision strategies and initialization schemes enabled by sensor depth data.

\boldparagraph{Datasets.}
We focus on indoor datasets and consider the following: a) MuSHRoom~\cite{ren2023mushroom}: a real-world indoor dataset containing separate training and evaluation camera trajectories; and b) ScanNet++~\cite{yeshwanthliu2023scannetpp}: a real-world indoor dataset with high fidelity 3D geometry and RGB data.

\boldparagraph{Baselines.}
We consider a range of baselines including implicit NeRF and SDF based representations and explicit Gaussian based methods. We consider a) state-of-the-art NeRF-based method Nerfacto~\cite{nerfstudio}; b) its depth regularized version Depth-Nerfacto with a direct loss on ray termination distribution for depth supervision similar to DS-NeRF \cite{kangle2021dsnerf}; c) Neusfacto \cite{Yu2022SDFStudio} and MonoSDF~\cite{Yu2022MonoSDF} for SDF-based implicit surface reconstruction; d) baseline 3DGS Splatfacto method based on Nerfstudio \texttt{v1.1.3} \cite{nerfstudio}; e) SuGaR~\cite{guedon2023sugar} -- a 3DGS variant for mesh reconstruction; and f) the recent 2DGS~\cite{Huang2DGS2024} method. In addition, for mesh reconstruction we consider g) traditional Poisson meshing utilizing back-projected sensor depth readings.

\boldparagraph{Evaluation metrics.} We follow standard practice and report PSNR, SSIM and LPIPS metrics for color images and common depth metrics, similar to~\cite{wei2021nerfingmvs,kusupati2020normal,luo2020consistent,sinha2020deltas,murez2020atlas,teed2018deepv2d,zhou2017unsupervised}, to analyze depth quality for datasets containing ground truth sensor data. For mesh evaluation, we follow evaluation protocols from ~\cite{ren2023mushroom,wang2022go} and report Accuracy (Acc.), Completion (Comp.), Chamfer-$L_1$ distance (C-$L_1$), Normal Consistency (NC), and F-scores (F1) with a threshold of 5cm.

\boldparagraph{Implementation details.} The proposed method is implemented in PyTorch~\cite{pytorch} and \texttt{gsplat} \cite{ye2024gsplatopensourcelibrarygaussian}~(\texttt{v1.0.0}). We train all models for 30k iterations. To obtain monocular normal cues, we propagate RGB images through the pre-trained Omnidata model~\cite{eftekhar2021omnidata}. We initialize the Gaussian scene using 1M back-projected points from training dataset sensor depths. For Poisson reconstruction, we extract a total of 2 million points and use a depth level of 9 for all methods, unless otherwise stated. All meshes are extracted using back-projection of depth and normal maps besides Neusfacto and MonoSDF (marching cubes) and 2DGS (TSDF). More settings can be seen in the supplementary material.

\boldparagraph{Gaussian initialization.} For all Gaussian based baselines (Splatfacto, SuGaR, 2DGS, and DN-Splatter) we utilize 1M back-projected sensor depth points for initialization of the Gaussian scene unless otherwise stated. We compare sparse COLMAP \cite{schoenberger2016mvscolmap2, schoenberger2016sfmcolmap1} SfM initialization and sensor depth initialization strategies in \cref{tab:ab-sfm-vs-sensor-depth}.

\begin{figure*}[t!]
  \centering\scriptsize

  \begin{tikzpicture}[inner sep=0]

    \setlength{\figurewidth}{0.12\textwidth}
    \setlength{\figureheight}{0.86\figurewidth}
    \setlength{\figurerowheight}{-1.01\figureheight}

    \node[anchor=center, rotate=90] (label-top) at (-2.5em,1\figurerowheight) {ScanNet++: \texttt{8b5caf3398}};
    \node[anchor=center, rotate=90] (label-top) at (-2.5em,3\figurerowheight) {MuSHRoom: \texttt{vr\textunderscore room}};
    \node[anchor=center, rotate=90] (label-top) at (-1em,0.5\figurerowheight) {\bf Normal};
    \node[anchor=center, rotate=90] (label-top) at (-1em,1.5\figurerowheight) {\bf Mesh};
    \node[anchor=center, rotate=90] (label-top) at (-1em,2.5\figurerowheight) {\bf Normal};
    \node[anchor=center, rotate=90] (label-top) at (-1em,3.5\figurerowheight) {\bf Mesh};

    \node[anchor=center, rotate=0] (label-top) at (0.5*\figurewidth,-4.1*\figureheight) {\texttt{Nerfacto}};
    \node[anchor=center, rotate=0] (label-top) at (1.5*\figurewidth,-4.1*\figureheight) {\texttt{Depth-Nerfacto}};
    \node[anchor=center, rotate=0] (label-top) at (2.5*\figurewidth,-4.1*\figureheight) {\texttt{MonoSDF}};
    \node[anchor=center, rotate=0] (label-top) at (3.5*\figurewidth,-4.1*\figureheight) {\texttt{SuGaR}};
    \node[anchor=center, rotate=0] (label-top) at (4.5*\figurewidth,-4.1*\figureheight) {\texttt{2DGS}};
    \node[anchor=center, rotate=0] (label-top) at (5.5*\figurewidth,-4.1*\figureheight) {\texttt{Splatfacto}};
    \node[anchor=center, rotate=0] (label-top) at (6.5*\figurewidth,-4.1*\figureheight) {\texttt{DN-Splatter}};
    \node[anchor=center, rotate=0] (label-top) at (7.5*\figurewidth,-4.1*\figureheight) {\texttt{GT}};
    
    \foreach \dataset/\row in {
      8b5_normal/0,
      8b5_mesh/1,
      vr_room_normal/2,
      vr_room_mesh/3
      } {
          \foreach \file/\label/\col in {
              nerfacto//0,
              depth_nerfacto//1,
              monosdf//2,
              sugar//3,
              2dgs//4,
              splatfacto//5,
              dn//6,
              gt_mesh//7} {
              \begin{scope}
    
              \node[anchor=north west, inner sep=2pt] (label-\row-\col) at (\col*\figurewidth,\row*\figurerowheight) {\label};
              \clip[rounded corners=2pt] (\col*\figurewidth,\figurerowheight*\row) -- (\col*\figurewidth+.99\figurewidth,\figurerowheight*\row) -- (\col*\figurewidth+.99\figurewidth,\figurerowheight*\row-\figureheight) -- (\col*\figurewidth,\figurerowheight*\row-\figureheight) -- cycle;
    
              \node [inner sep=0,minimum width=\figurewidth,minimum height=\figureheight,fill=red!10!white,anchor=north west] at (\col*\figurewidth, \row*\figurerowheight) {%
                  \includegraphics[height=\figureheight]{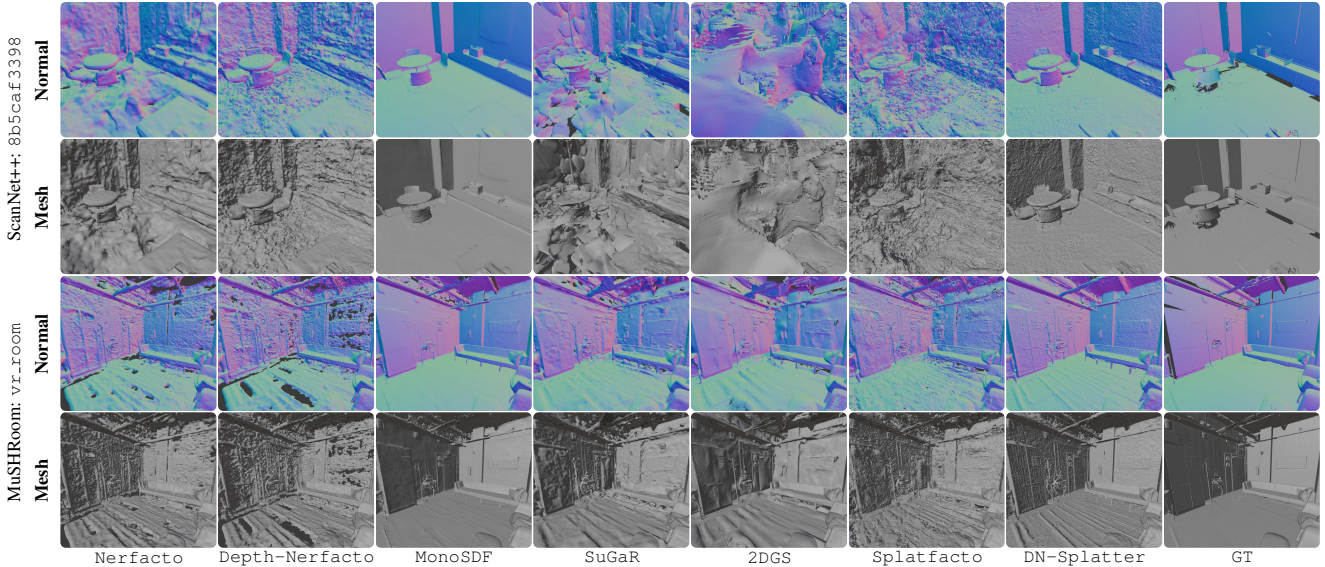}%
              };
              \end{scope}
          }
        }
  \end{tikzpicture}
  \vspace{-1em}
  \caption{\textbf{Mesh reconstruction results}. NeRF variants, even with depth supervision, suffer from artefacts and floaters in reconstruction. The Gaussian based methods Splatfacto, SuGaR, and 2DGS are trained on only photometric losses and thus severly struggle to capture the scene geometry in low texture environments. However, adding depth and normal supervision with DN-Splatter greatly aids reconstruction quality.}
\label{fig:main-figure-mesh}
\end{figure*}
\begin{table*}[th!]
\vspace{-1em}
  \centering\scriptsize
  \setlength{\tabcolsep}{5.5pt}
  \renewcommand*{\arraystretch}{1.05}
\begin{tabular}{clccccccc|c}
\toprule
 & & Sensor Depth Loss & \multicolumn{1}{c}{Algorithm} & Accuracy~$\downarrow$ & Completion~$\downarrow$  & Chamfer-$L_1$~$\downarrow$ & Normal Consistency~$\uparrow$  & F-score~$\uparrow$  & Num GS \\
\midrule
& Traditional Poisson & $\textcolor{red}{-}$ & Poisson & .0399 & .0222 & .0306	& .8688 & .8823 & - \\
\midrule
\multirow{2}{*}{\rotatebox[origin=c]{90}{NeRF}} & Nerfacto \cite{nerfstudio} & $\textcolor{red}{-}$ & Poisson & .0430 & .0578 & .0504 & .7822 & .7212 & -\\
&Depth-Nerfacto~\cite{nerfstudio} & \textcolor{teal}{$\checkmark$} & Poisson & .0447 & .0557 & .0502 & .7614 & .6966 & -\\
\midrule 
\multirow{1}{*}{\rotatebox[origin=c]{90}{\hspace{1ex}\mbox{SDF}\hspace{-1.4ex}}} & \rule{0pt}{3ex}MonoSDF \cite{Yu2022MonoSDF} & \textcolor{teal}{$\checkmark$} & Marching-Cubes & \underline{.0310} & \textbf{.0190} & \underline{.0250} & \textbf{.8846} & \underline{.9211} & -\\
\midrule 
\multirow{4}{*}{\rotatebox[origin=c]{90}{Gaussian}} 
& SuGaR \cite{guedon2023sugar}& $\textcolor{red}{-}$ &  Poisson+IBR & .0656 & .0583 & .0620 & .8031 & .6378 & 700K\\
& 2DGS \cite{Huang2DGS2024}& $\textcolor{red}{-}$ & TSDF & .0731 & .0642 & .0687 & .8008 & .6039 & 2.6M\\
& Splatfacto \cite{nerfstudio} & $\textcolor{red}{-}$ & Poisson & .0749 & .0555 & .0652 & .7727 & .5835 & 1.18M\\
& \ccgray DN-Splatter (Ours) & \ccgray \textcolor{teal}{$\checkmark$} & \ccgray Poisson & \ccgray \textbf{.0239} & \ccgray \underline{.0194} & \ccgray \textbf{.0216} & \ccgray \underline{.8822} & \ccgray \textbf{.9243} & \ccgray 1.18M\\
\bottomrule
\end{tabular}
\vspace{-1em}
\caption{\textbf{Mesh evaluation: MuSHRoom}. Adding depth and normal priors to 3DGS optimization greatly enhances mesh reconstruction on challenging real-world indoor datasets. We report the number of Gaussians in the final optimized scene for Gaussian-based models, with results averaged over 6 scenes: 'coffee\textunderscore room', 'honka', 'kokko', 'sauna', 'computer', and 'vr\textunderscore room'. The best and second-best results are indicated with \textbf{bold} and \underline{underline}.}
\label{tab:main-table-mesh-mushroom}
\vspace{-0.2cm}
\end{table*}

\begin{table*}[th!]
  \centering\scriptsize
  \setlength{\tabcolsep}{5.5pt}
  \renewcommand*{\arraystretch}{1.05}
\begin{tabular}{clccccccc|c}
\toprule
 & & Sensor Depth Loss & \multicolumn{1}{c}{Algorithm} & Accuracy~$\downarrow$ & Completion~$\downarrow$  & Chamfer-$L_1$~$\downarrow$ & Normal Consistency~$\uparrow$  & F-score~$\uparrow$  & Time (min)\\
\midrule
& Traditional Poisson & $\textcolor{red}{-}$ & Poisson & \underline{.0593} & .0574 & \underline{.0564} &.4410 & \underline{.7923} & 2.0 \\
\midrule
\multirow{2}{*}{\rotatebox[origin=c]{90}{NeRF}}
& Nerfacto~\cite{nerfstudio} & $\textcolor{red}{-}$ &  Poisson & .1305 & .1484 & .1394 & .7153 & .4698 & 8.0\\
& Depth-Nerfacto~\cite{nerfstudio} & \textcolor{teal}{$\checkmark$} & Poisson   & .0731 & .1647 & .1189 & .6848 & .5018 & 8.1\\
\midrule 
\multirow{2}{*}{\rotatebox[origin=c]{90}{SDF}}
& Neusfacto~\cite{Yu2022SDFStudio} & $\textcolor{red}{-}$ & Marching-Cubes & .0736 & .1945 & .1340 & .7159 & .4605 & 40.0\\
& MonoSDF~\cite{Yu2022MonoSDF} & \textcolor{teal}{$\checkmark$} & Marching-Cubes & \textbf{.0303} & \underline{.0573} & \textbf{.0438} & \textbf{.8881} & \textbf{.8577} & 47.5\\
\midrule
\multirow{4}{*}{\rotatebox[origin=c]{90}{Gaussian}}
& SuGaR~\cite{guedon2023sugar} & $\textcolor{red}{-}$ & Poisson + IBR   & .0940 & .1011 & .0975 & .7241 & .4367 & 70.0\\
& 2DGS~\cite{Huang2DGS2024} & $\textcolor{red}{-}$ & TSDF & .1272 & .0798 & .1035 & .7799 & .4196 & 33.5\\
& Splatfacto~\cite{nerfstudio} & $\textcolor{red}{-}$ & Poisson & .1934 & .1503 & .1719 & .6741 & .1790 & 8.9\\
& \ccgray DN-Splatter (Ours) & \ccgray \textcolor{teal}{$\checkmark$} & \ccgray Poisson & \ccgray .0940 & \ccgray \textbf{.0395} & \ccgray .0667 & \ccgray \textbf{.8316} & \ccgray .7658 & \ccgray 36.9\\
\bottomrule
\end{tabular}
\vspace{-1em}
\caption{\textbf{Mesh evaluation: ScanNet++}. The results are averaged over the 'b20a261fdf' and '8b5caf3398' scenes. The best and second best results are marked with \textbf{bold} and \underline{underline}. Training time is reported using a Nvidia 4090 GPU.}
\vspace{-2em}
\label{tab:main-table-mesh-scannetpp}
\end{table*}

\subsection{Mesh evaluation}
We demonstrate the effectiveness of the proposed regularization strategy on scene geometry by extracting meshes directly after optimization, without any additional refinement steps. In~\tabref{tab:main-table-mesh-mushroom} and~\tabref{tab:main-table-mesh-scannetpp} we show how incorporating geometric cues enables competitive mesh extraction on scenes from the MuSHRoom~\cite{ren2023mushroom} and ScanNet++~\cite{yeshwanthliu2023scannetpp} datasets. On challenging indoor scenes, neither NeRF~\cite{nerfstudio} nor Gaussian-based methods~\cite{Huang2DGS2024, guedon2023sugar} perform well for indoor reconstruction. Improving depth and normal estimation over the baseline Splatfacto~\cite{nerfstudio} makes our method competitive even against the more computationally expensive SDF approaches~\cite{Yu2022MonoSDF, Yu2022SDFStudio}. NeRF-based methods fail to consistently achieve similar results, with depth supervision sometimes hindering mesh performance. \cref{fig:main-figure-mesh} provides a qualitative comparison of our mesh extraction approach with other baselines on the MuSHRoom and ScanNet++ datasets.

\subsection{Novel-view synthesis and depth estimation}
\begin{figure*}[ht!]
\centering
  \begin{subfigure}[b]{0.14\textwidth}
    \centering
    \begin{tikzpicture}
    \node [inner sep=0pt,outer sep=0pt,clip,rounded corners=2pt] at (0,0) {\includegraphics[height=2.7cm]{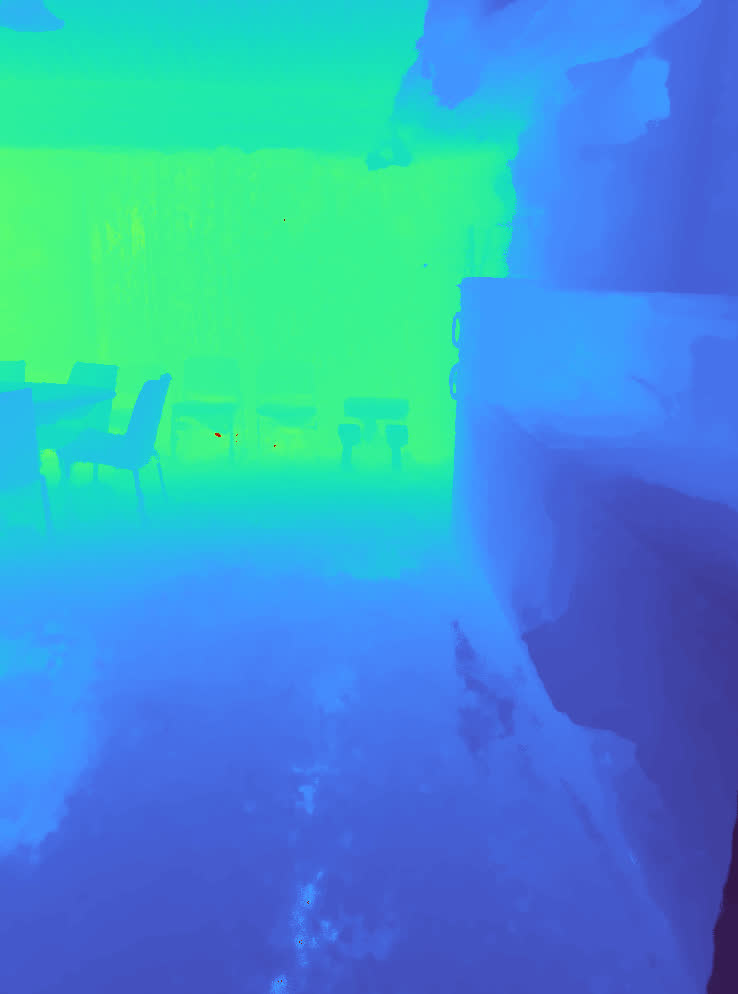}};
    \node[font=\tiny, text=white] at (0.2,-1.1) {};
    \end{tikzpicture}
  \end{subfigure} 
  \hspace{-15pt}
  \begin{subfigure}[b]{0.14\textwidth}
    \centering
    \begin{tikzpicture}
    \node [inner sep=0pt,outer sep=0pt,clip,rounded corners=2pt] at (0,0) {\includegraphics[height=2.7cm]{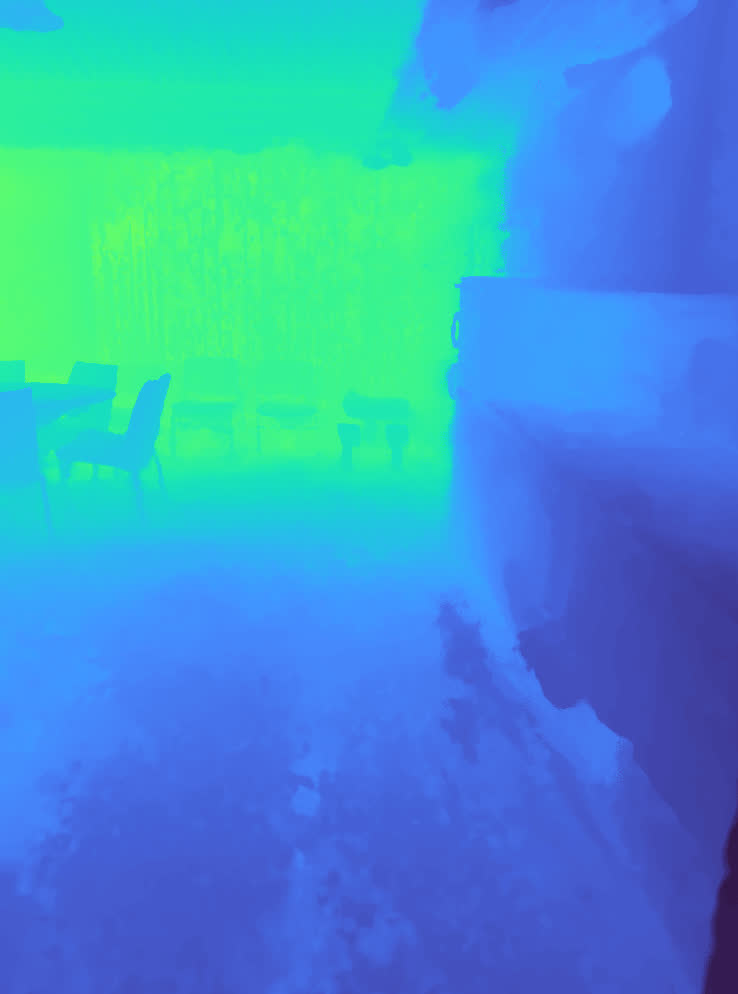}};
    \node[font=\tiny, text=white] at (0.2,-1.1) {};
    \end{tikzpicture}
  \end{subfigure}
  \hspace{-15pt}
    \begin{subfigure}[b]{0.14\textwidth}
    \centering
    \begin{tikzpicture}
    \node [inner sep=0pt,outer sep=0pt,clip,rounded corners=2pt] at (0,0) {\includegraphics[height=2.7cm]{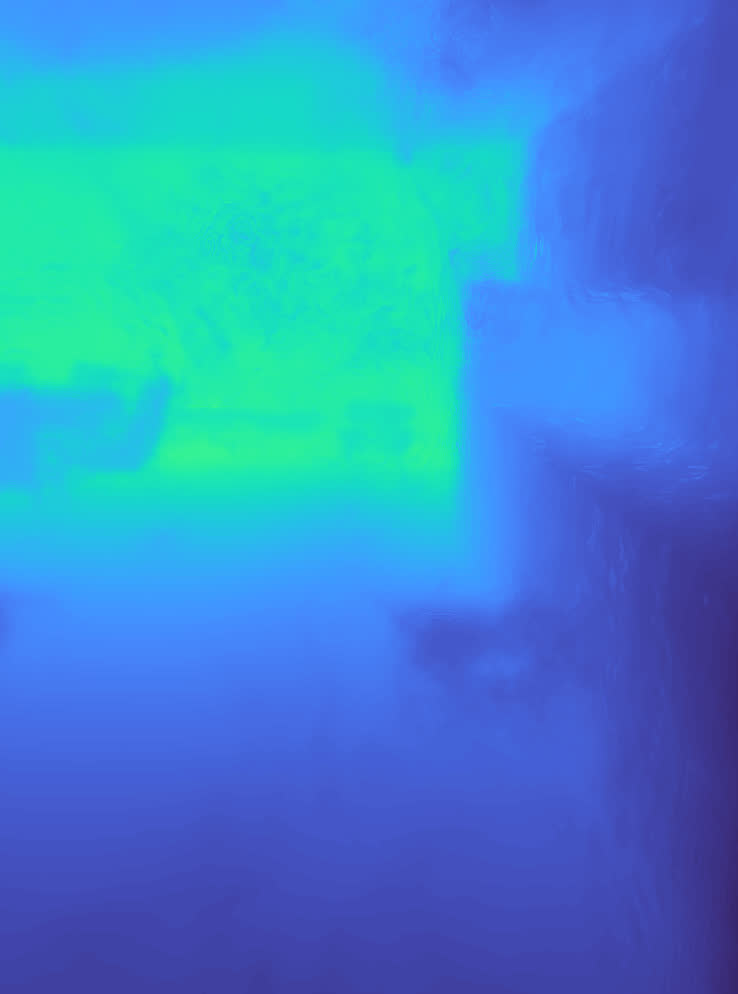}};
    \node[font=\tiny, text=white] at (0.2,-1.1) {};
    \end{tikzpicture}
  \end{subfigure}
  \hspace{-15pt}
    \begin{subfigure}[b]{0.14\textwidth}
\centering
    \begin{tikzpicture}
    \node [inner sep=0pt,outer sep=0pt,clip,rounded corners=2pt] at (0,0) {\includegraphics[height=2.7cm]{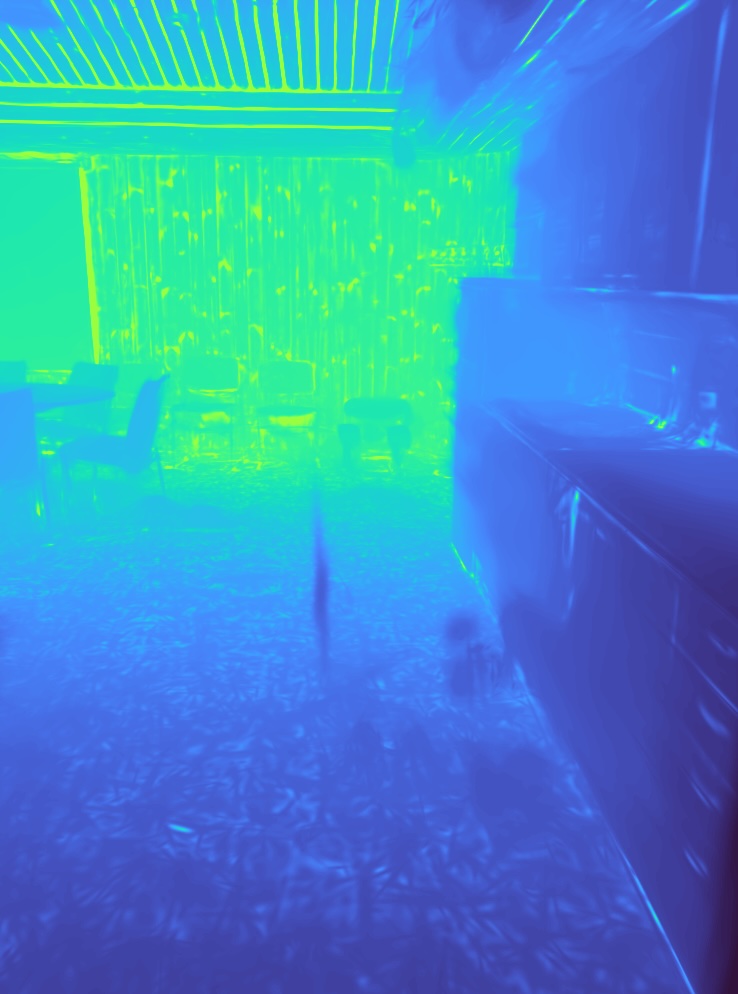}};
    \node[font=\tiny, text=white] at (0.2,-1.1) {};
    \end{tikzpicture}
  \end{subfigure}
  \hspace{-15pt}
  \begin{subfigure}[b]{0.14\textwidth}
    \centering
    \begin{tikzpicture}
    \node [inner sep=0pt,outer sep=0pt,clip,rounded corners=2pt] at (0,0) {\includegraphics[height=2.7cm]{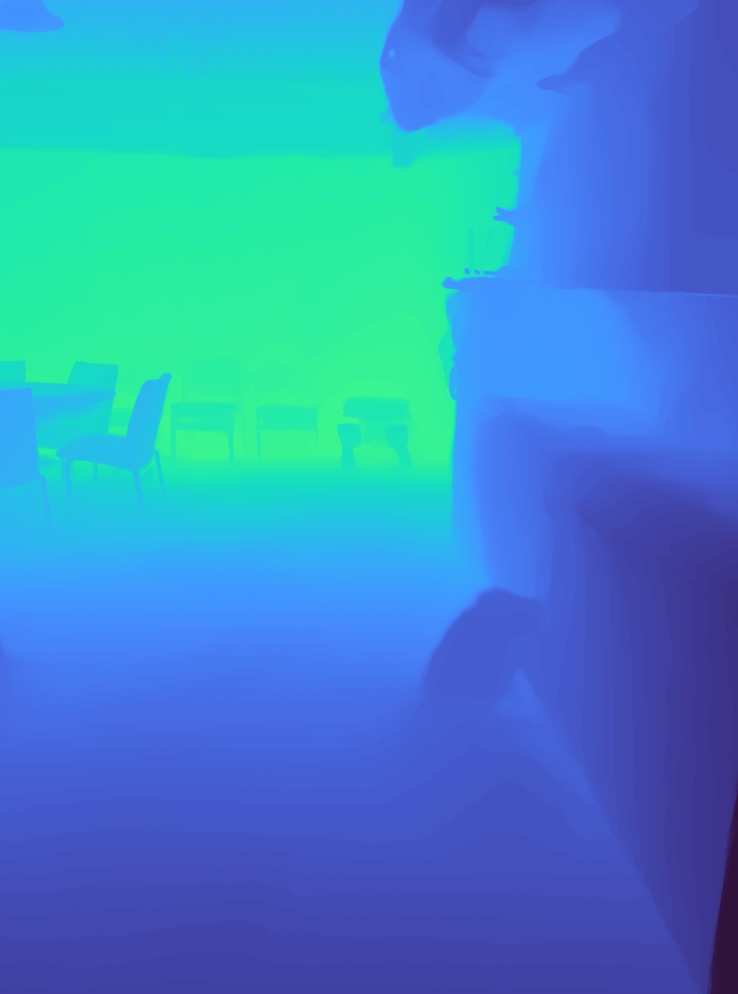}};
    \node[font=\tiny, text=white] at (0.2,-1.1) {};
    \end{tikzpicture}
  \end{subfigure}
  \hspace{-15pt}
  \begin{subfigure}[b]{0.14\textwidth}
    \centering
    \begin{tikzpicture}
    \node [inner sep=0pt,outer sep=0pt,clip,rounded corners=2pt] at (0,0) {\includegraphics[height=2.7cm]{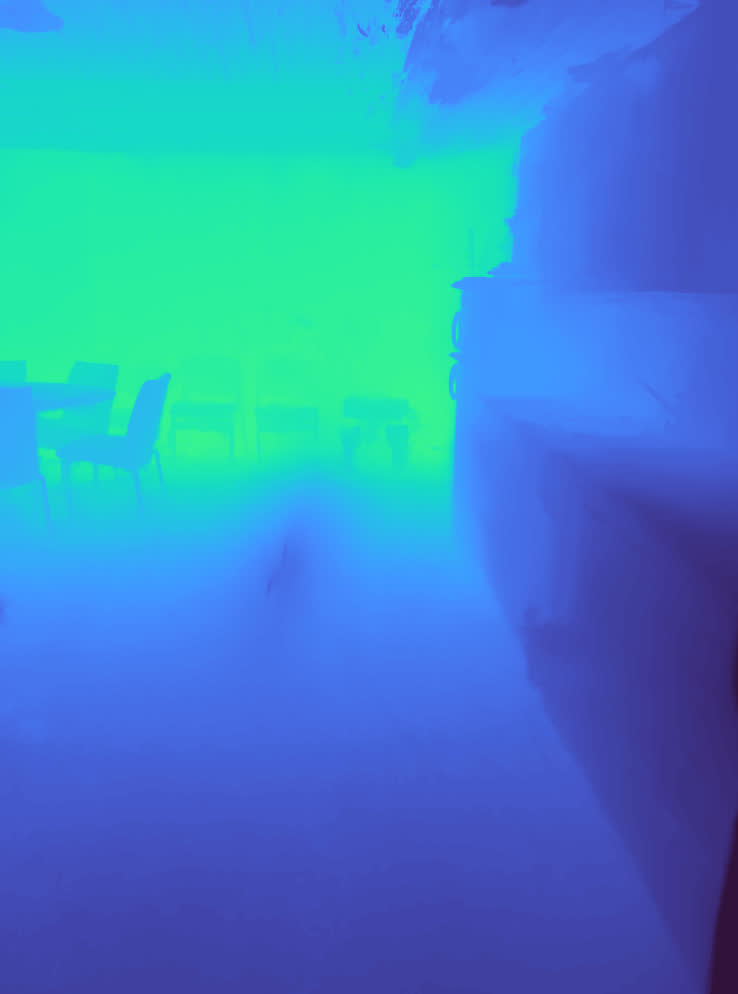}};
    \node[font=\tiny, text=white] at (0.2,-1.1) {};
    \end{tikzpicture}
  \end{subfigure}
  \hspace{-15pt}
  \begin{subfigure}[b]{0.14\textwidth}
    \centering
    \begin{tikzpicture}
    \node [inner sep=0pt,outer sep=0pt,clip,rounded corners=2pt] at (0,0) {\includegraphics[height=2.7cm]{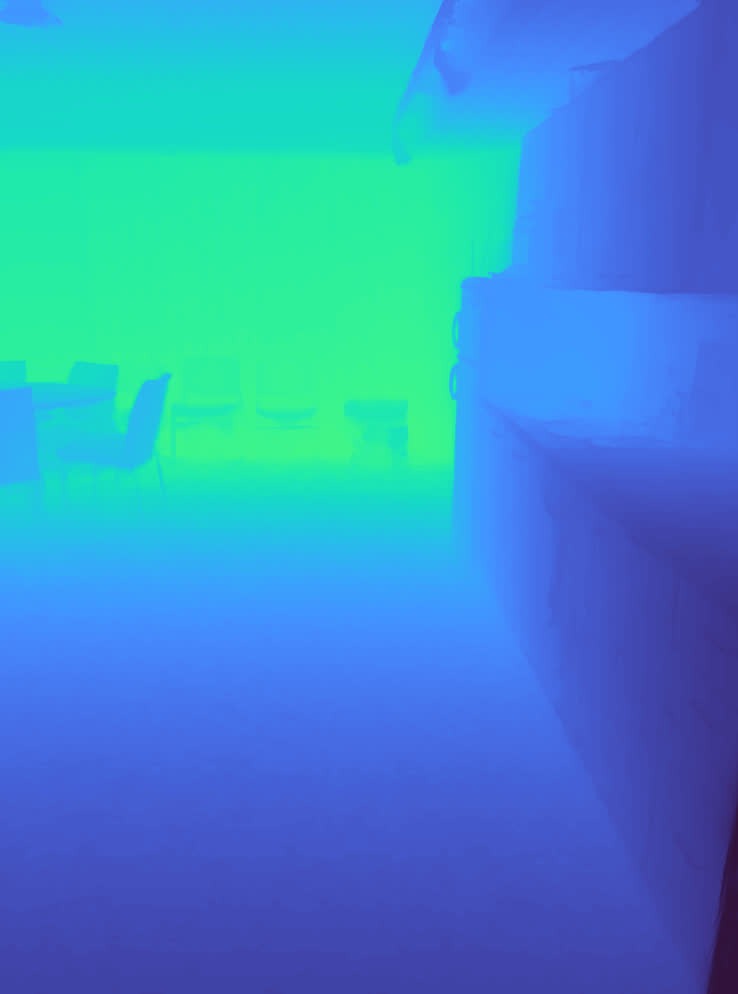}};
    \node[font=\tiny, text=white] at (0.2,-1.1) {};
    \end{tikzpicture}
  \end{subfigure}
    \hspace{-15pt}
    \begin{subfigure}[b]{0.14\textwidth}
    \centering
    \begin{tikzpicture}
    \node [inner sep=0pt,outer sep=0pt,clip,rounded corners=2pt] at (0,0) {\includegraphics[height=2.7cm]{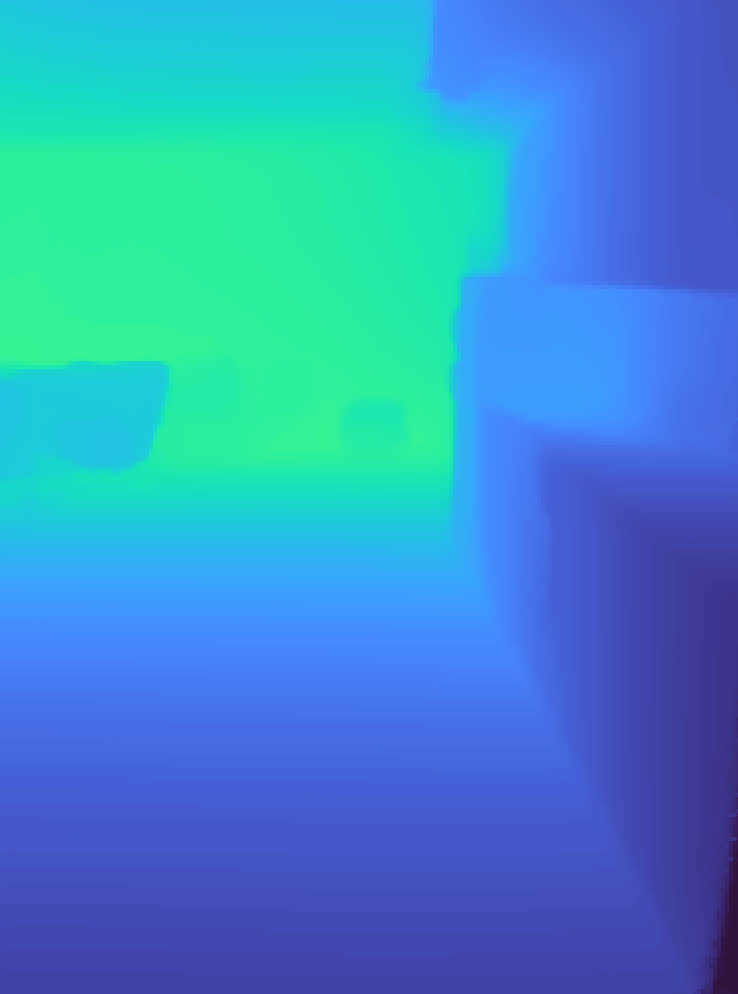}};
    \node[font=\tiny, text=white] at (0.2,-1.1) {};
    \end{tikzpicture}
  \end{subfigure}
  \hspace{-15pt}
  \\
  \begin{subfigure}[b]{0.14\textwidth}
    \centering
    \begin{tikzpicture}
    \node [inner sep=0pt,outer sep=0pt,clip,rounded corners=2pt] at (0,0) {\includegraphics[height=2.7cm]{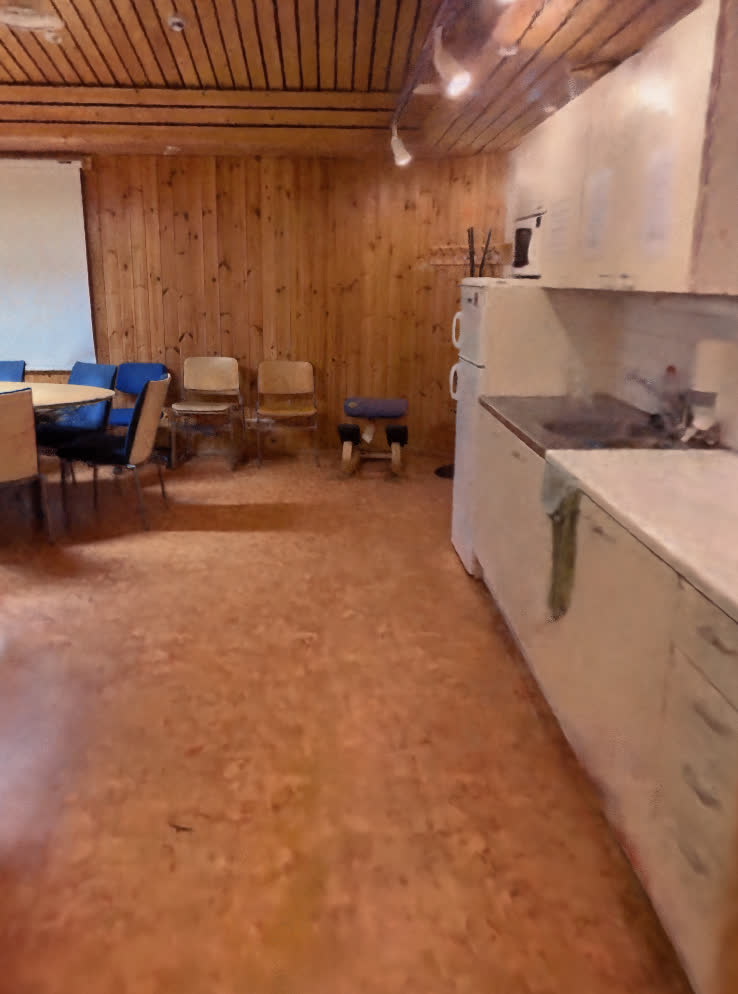}};
    \node[font=\tiny, text=white] at (0.2,-1.1) {PSNR: 20.96};
    \end{tikzpicture}
    \subcaption*{\texttt{Nerfacto}}
  \end{subfigure} 
  \hspace{-15pt}
  \begin{subfigure}[b]{0.14\textwidth}
    \centering
    \begin{tikzpicture}
    \node [inner sep=0pt,outer sep=0pt,clip,rounded corners=2pt] at (0,0) {\includegraphics[height=2.7cm]{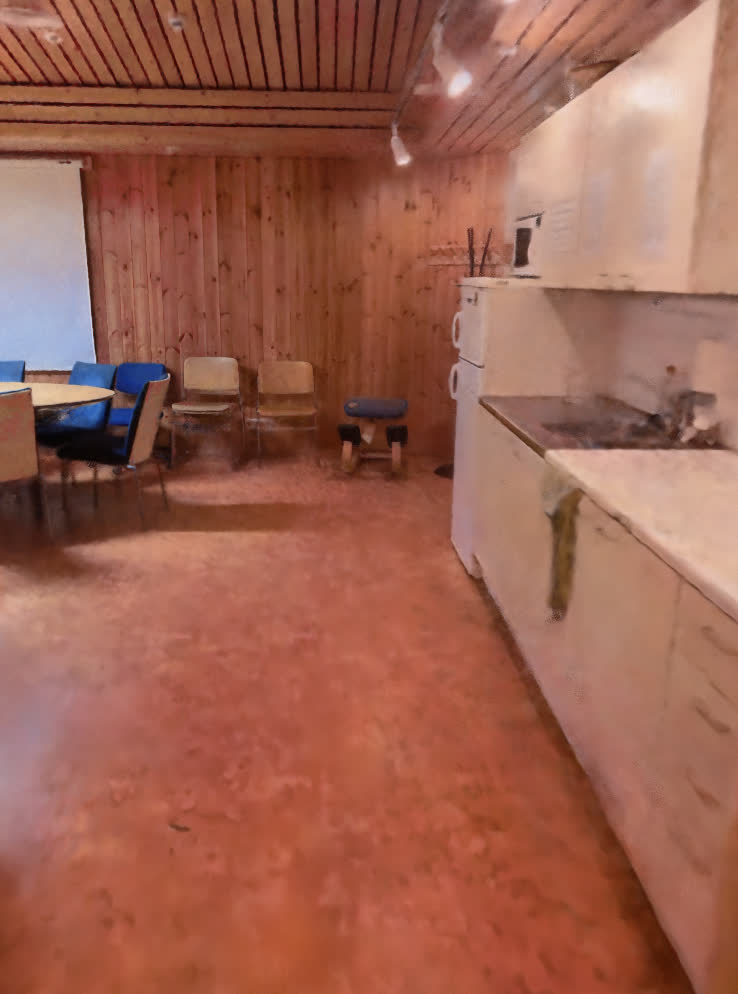}};
    \node[font=\tiny, text=white] at (0.2,-1.1) {PSNR: 20.86};
    \end{tikzpicture}
    \subcaption*{\texttt{D-Nerfacto}}
  \end{subfigure} 
  \hspace{-15pt}
  \begin{subfigure}[b]{0.14\textwidth}
    \centering
    \begin{tikzpicture}
    \node [inner sep=0pt,outer sep=0pt,clip,rounded corners=2pt] at (0,0) {\includegraphics[height=2.7cm]{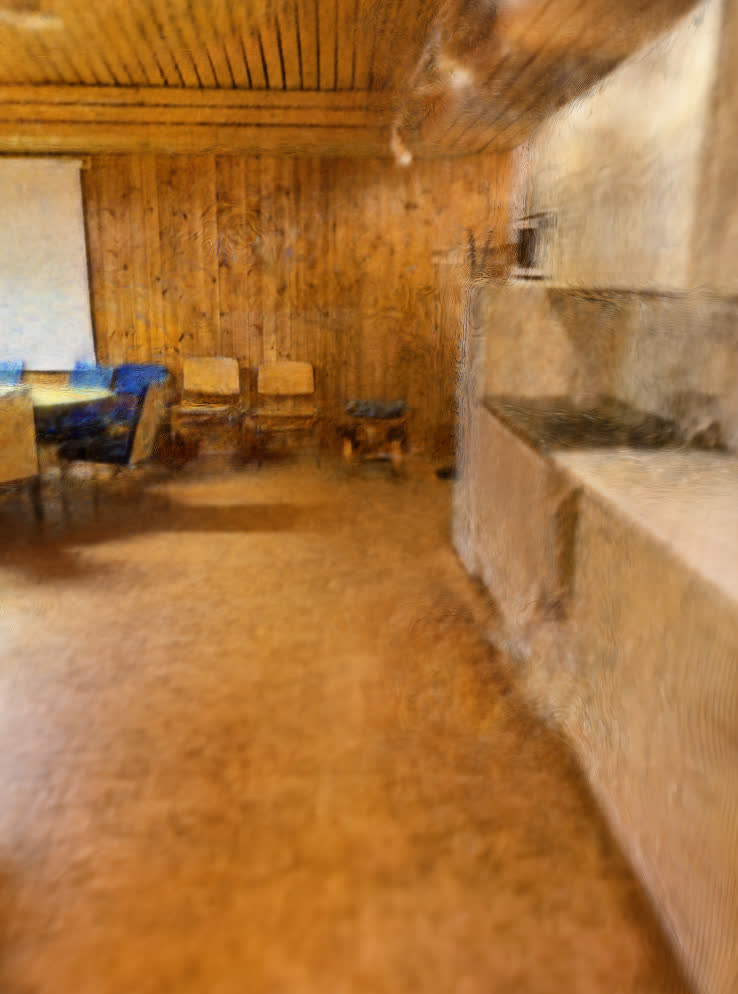}};
    \node[font=\tiny, text=white] at (0.2,-1.1) {PSNR: 20.95};
    \end{tikzpicture}
    \subcaption*{\texttt{MonoSDF}}
  \end{subfigure}
  \hspace{-15pt}
    \begin{subfigure}[b]{0.14\textwidth}
    \centering
    \begin{tikzpicture}
    \node [inner sep=0pt,outer sep=0pt,clip,rounded corners=2pt] at (0,0) {\includegraphics[height=2.7cm]{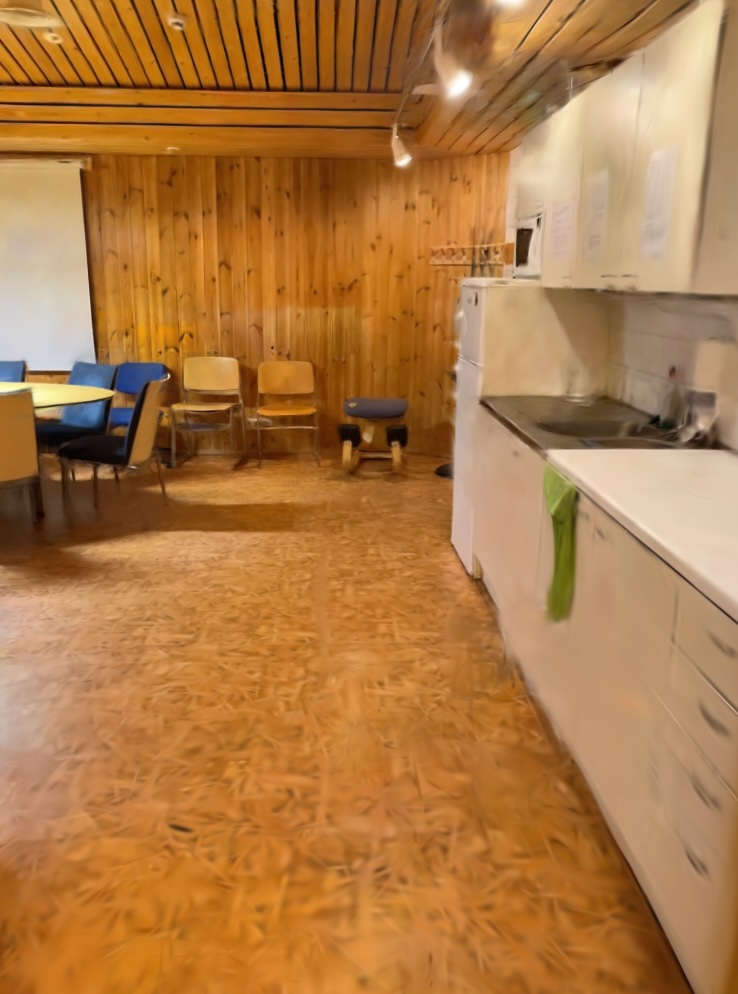}};
    \node[font=\tiny, text=white] at (0.2,-1.1) {PSNR: 22.21};
    \end{tikzpicture}
    \subcaption*{\texttt{SuGaR}}
  \end{subfigure}
  \hspace{-15pt}
    \begin{subfigure}[b]{0.14\textwidth}
    \centering
    \begin{tikzpicture}
    \node [inner sep=0pt,outer sep=0pt,clip,rounded corners=2pt] at (0,0) {\includegraphics[height=2.7cm]{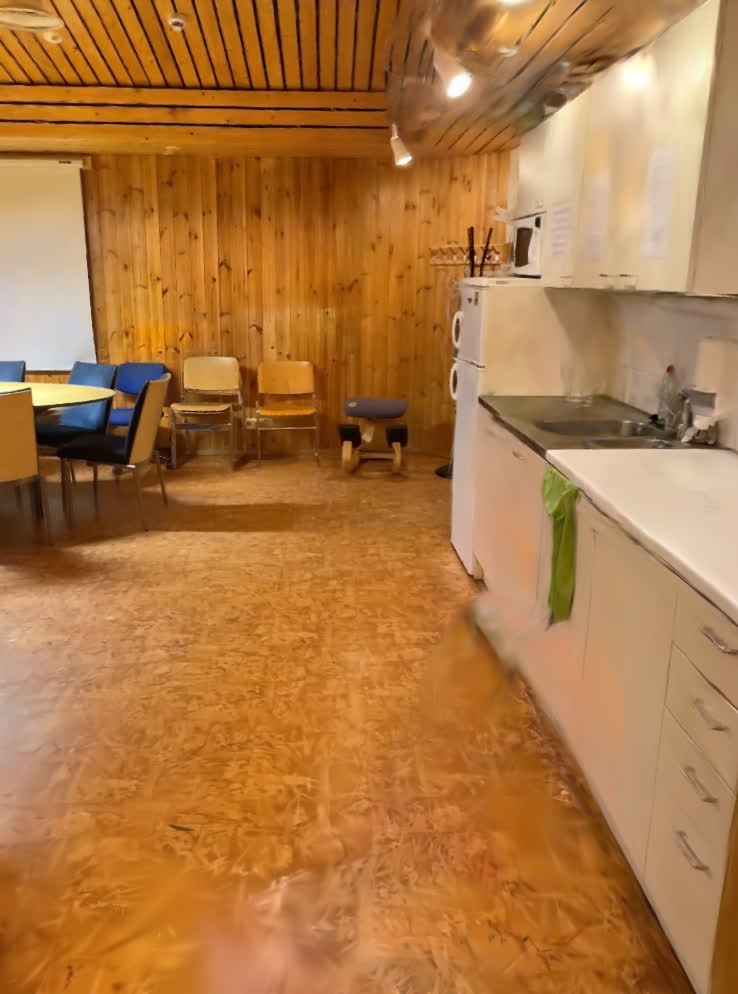}};
    \node[font=\tiny, text=white] at (0.2,-1.1) {PSNR: 21.94};
    \end{tikzpicture}
    \subcaption*{\texttt{2DGS}}
  \end{subfigure}
  \hspace{-15pt}
    \begin{subfigure}[b]{0.14\textwidth}
    \centering
    \begin{tikzpicture}
    \node [inner sep=0pt,outer sep=0pt,clip,rounded corners=2pt] at (0,0) {\includegraphics[height=2.7cm]{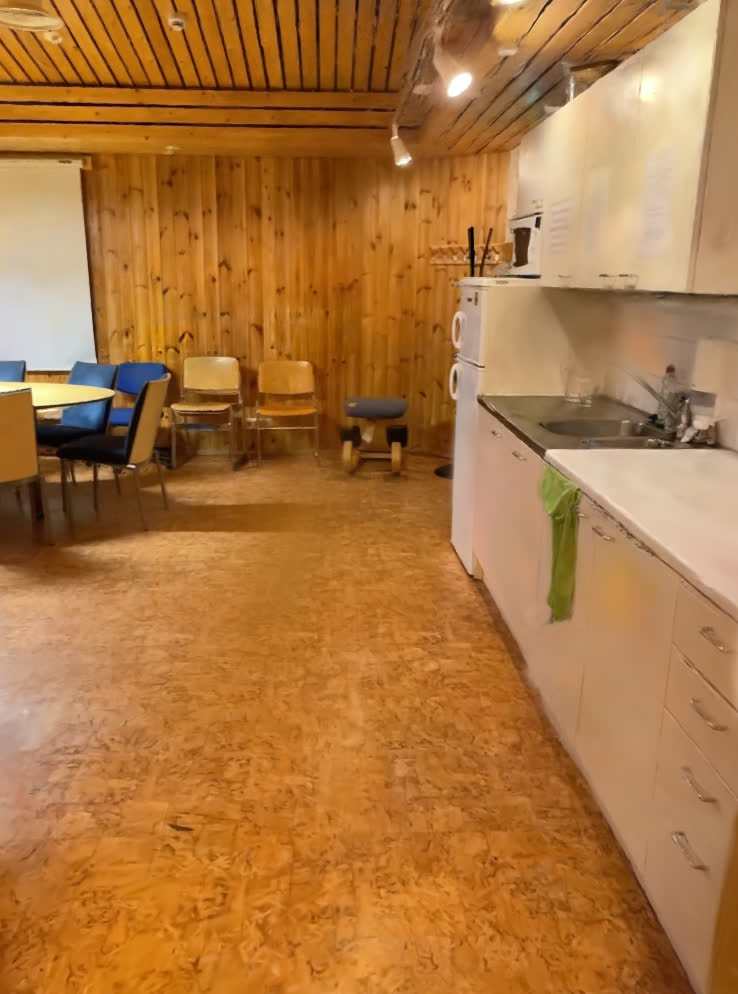}};
    \node[font=\tiny, text=white] at (0.2,-1.1) {PSNR: 22.66};
    \draw[red,thick] (-0.5,-0.5) rectangle (0.2,0.2);
    \end{tikzpicture}
    \subcaption*{\texttt{Splatfacto}}
  \end{subfigure}
   \hspace{-15pt}
  \begin{subfigure}[b]{0.14\textwidth}
    \centering
    \begin{tikzpicture}
    \node [inner sep=0pt,outer sep=0pt,clip,rounded corners=2pt] at (0,0) {\includegraphics[height=2.7cm]{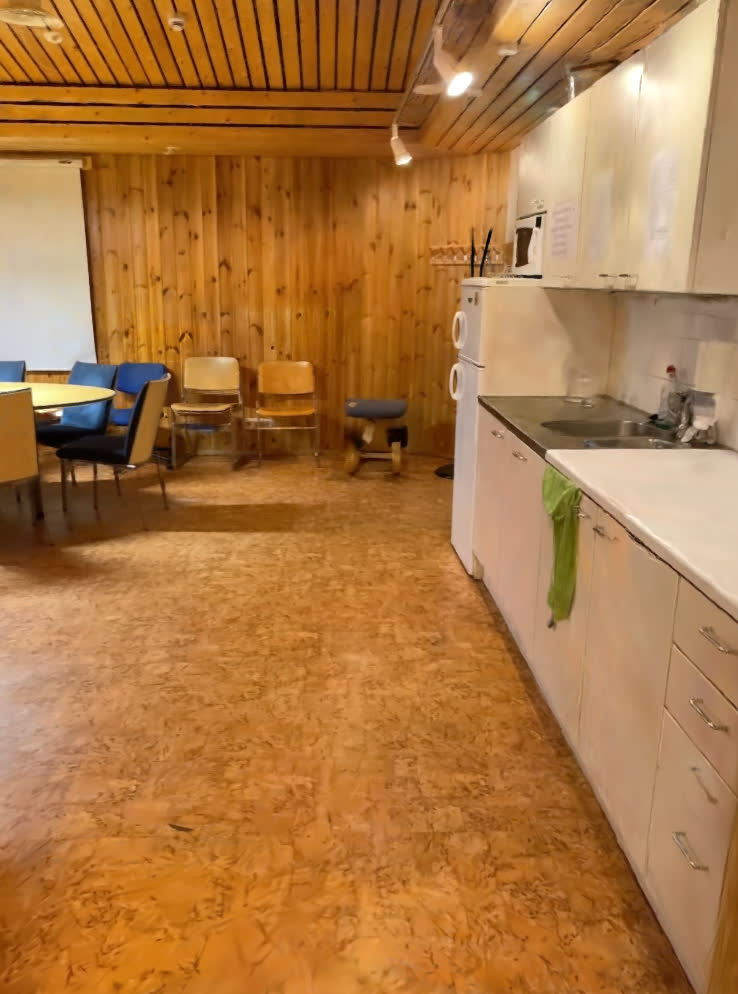}};
    \node[font=\tiny, text=white] at (0.2,-1.1) {PSNR: 23.01};
    \draw[red,thick] (-0.5,-0.5) rectangle (0.2,0.2);
    \end{tikzpicture}
    \subcaption*{\texttt{DN-Splatter}}
  \end{subfigure}
  \hspace{-15pt}
  \begin{subfigure}[b]{0.14\textwidth}
    \centering
    \begin{tikzpicture}
    \node [inner sep=0pt,outer sep=0pt,clip,rounded corners=2pt] at (0,0) {\includegraphics[height=2.7cm]{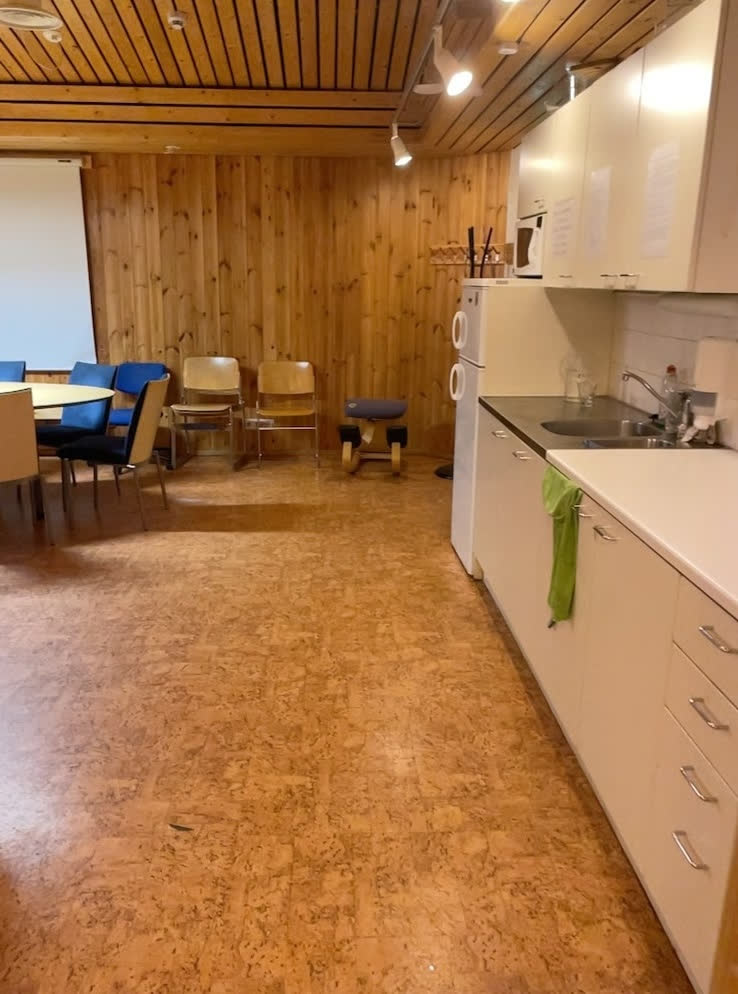}};
    \node[font=\tiny, text=white] at (0.2,-1.1) {};
    \end{tikzpicture}
    \subcaption*{\texttt{iPhone GT}}
  \end{subfigure}
  \vspace{-1em}
  \caption{\textbf{Qualitative comparison of depth and RGB renders against a variety of baselines.} DN-Splatter achieves the highest novel view synthesis results compared to NeRF, SDF, and Gaussian based methods.}
 \label{fig:different-method-cmp}
\end{figure*}

\begin{table*}[t!]
  \centering\scriptsize
  \setlength{\tabcolsep}{5.5pt}
\renewcommand*{\arraystretch}{1.2}
\begin{tabular}{clccccc|ccc}
\toprule
 & & Sensor Depth Loss & Abs Rel $\downarrow$ & Sq Rel $\downarrow$ & RMSE $\downarrow$  & $\delta<1.25$ $\uparrow$ & PSNR $\uparrow$   & SSIM $\uparrow$    & LPIPS $\downarrow$ \\
\midrule
\multirow{2}{*}{\rotatebox[origin=c]{90}{NeRF}} & Nerfacto~\cite{nerfstudio} & $\textcolor{red}{-}$ & .0862 / .0747 & .0293 / .0141 & .0794 / .0667 & .9335 / .9428 & 20.86 / 20.66 & .7859 / .7633 & .2321 / .2702 \\
& Depth-Nerfacto~\cite{nerfstudio} & \textcolor{teal}{$\checkmark$} & .0727 / .0563 & .0155 / .0283 & \textbf{.0583} / .2840 & .9469 / .9389 & 21.24 / 18.93 & .7832 / .7023 & .2414 / .3978 \\
\midrule
\multirow{1}{*}{\rotatebox[origin=c]{90}{\hspace{1ex}\mbox{SDF}\hspace{-1.0ex}}} & \rule{0pt}{3ex}MonoSDF~\cite{Yu2022MonoSDF} & \textcolor{teal}{$\checkmark$} & .0555 / .0911 & .0263 / .0804 & .2493 / .4535 & .9353 / .8825 & 20.68 / 20.16 & .7357 / .7653 & .3590 / .2261 \\
\midrule
\multirow{6}{*}{\rotatebox[origin=c]{90}{Explicit}} & 
SuGaR\cite{guedon2023sugar} & $\textcolor{red}{-}$ & .1213 / .1174  & .1258 / .1474 & .5562 / .5820 & .8412 / .8564 & 20.52 / 18.18 & .7740 / .7125 & .2427 / .2959 \\
& 2DGS~\cite{Huang2DGS2024} & $\textcolor{red}{-}$ & .0864 / .0923 & .0612 / .0583 & .3799 / .3132 & .8820 / .8927 & 22.52 / 21.73 & .8185 / .7898 & .1773 / .1911 \\
& Splatfacto~\cite{nerfstudio} & $\textcolor{red}{-}$ & .0787 / .0817 & .0364 / .0521 & .2407 / .2941 & .9072 / .9061 & 24.44 / 21.33 & .8486 / .7821 & .1387 / .2240 \\
\cline{2-10}
& Splatfacto + $\mathcal{L}_{\hat{D}}$ &  \textcolor{teal}{$\checkmark$} & \underline{.0234} / .0364 & .0092 / \underline{.0145} & .1293 / \underline{.1486} & \underline{.9849} / \textbf{.9745} & \textbf{24.77} / \underline{21.95} & \underline{.8538} / \underline{.7948} & \textbf{.1238} / \underline{.1852} \\
& Splatfacto + $\mathcal{L}_{\hat{D}}$ + $\mathcal{L}_{\text{normal}}$ & \textcolor{teal}{$\checkmark$} & .0241 / \textbf{.0340} & .0094 / \textbf{.0123} & .1308 / \textbf{.1472} & .9848 / \textbf{.9777} & \underline{24.67} / \textbf{21.99} & .8517 / .7941 & \underline{.1275} / .1864 \\
\rowcolor{gray!10} & DN-Splatter (Ours) & \textcolor{teal}{$\checkmark$} & \textbf{.0228} / \underline{.0354} & \textbf{.0089} / .0214 & \underline{.1280} / .2032 & \textbf{.9854} / .9683 & 24.58 / 21.89 & \textbf{.8558} / \textbf{.7984} & .1293 / \textbf{.1797} \\
\bottomrule
\end{tabular}
\vspace{-1em}
\caption{\textbf{Depth estimation and novel view synthesis: MuSHRoom}. The reported results are reported for two distinct evaluation datasets: \texttt{left/right} where \texttt{left} is a test set obtained by sampling uniformly every 10 frames within the training sequence and \texttt{right} is a test split obtained from a different camera trajectory with no overlap with the training sequence. Results are averaged over 6 scenes.}
\vspace{-2em}
\label{tab:main-table-nvs-mushroom}
\end{table*}

We show an extensive study on depth and normal supervision and its effect on novel-view synthesis and depth metrics on challenging real-world scenes from the MuSHRoom and ScanNet++ datasets. \tabref{tab:main-table-nvs-mushroom} demonstrates that incorporating sensor depth supervision into 3DGS enhances both depth and RGB metrics compared to scenarios without geometric supervision. We qualitatively highlight the improvements in novel-view synthesis and depth quality, including the reduction of floaters and artifacts in~\figref{fig:different-method-cmp}.

\newcommand{\fivew}{0.162\textwidth}
\begin{figure*}[t]
\setlength{\tabcolsep}{0pt}
    \begin{tabular}{c}%
        \begin{minipage}{1\textwidth}%
            \centering
            \setlength{\tabcolsep}{0pt}%
            \renewcommand{\arraystretch}{1}
            \footnotesize
            \begin{tabular}{cccccc}
            \begin{subfigure}[b]{\fivew}
                \begin{tikzpicture}
                \node [inner sep=0pt,outer sep=0pt,clip,rounded corners=2pt] at (0,0) {\includegraphics[width=1\textwidth]{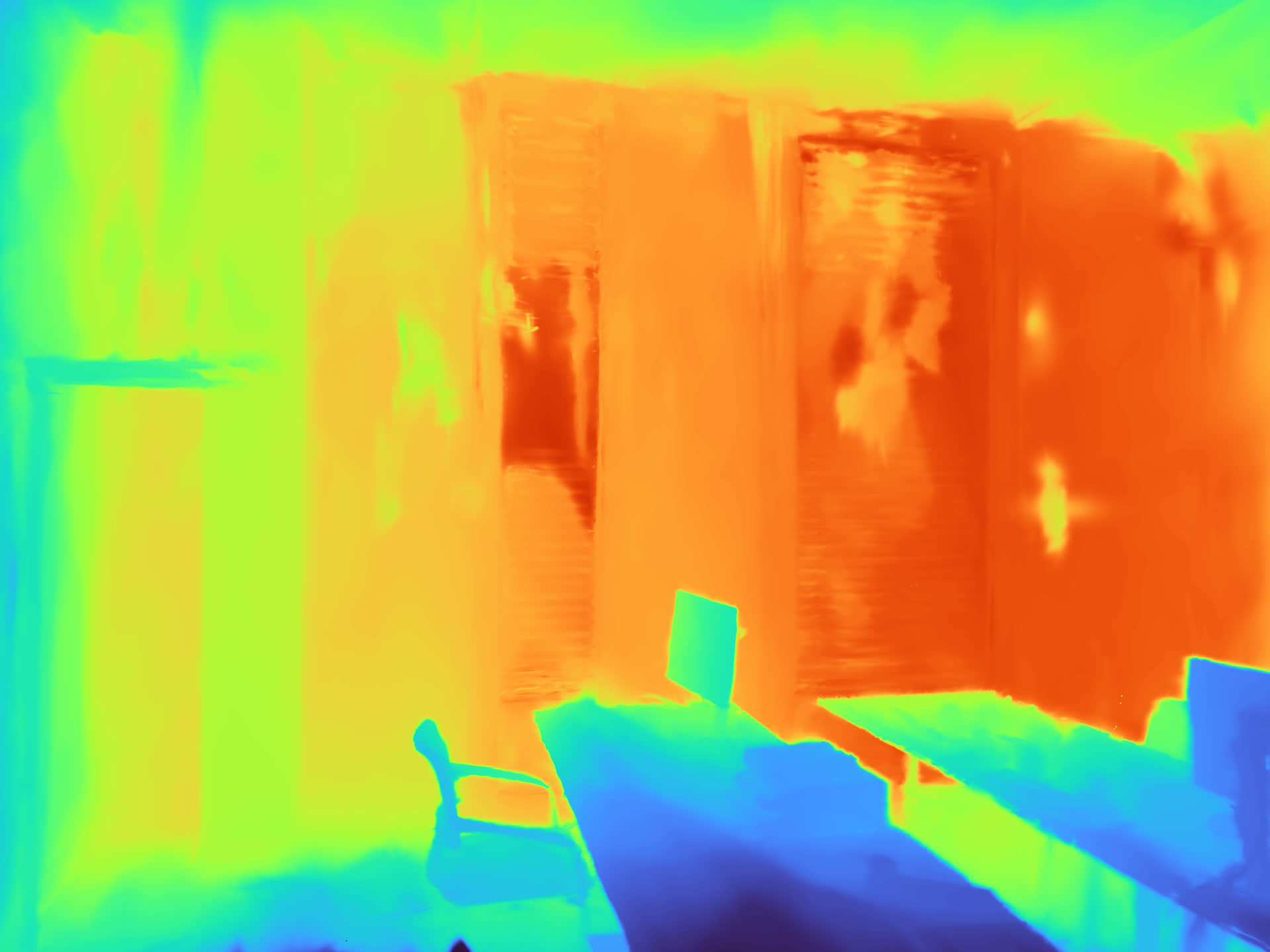}};
                \end{tikzpicture}
                \subcaption*{\texttt{baseline}}
          \end{subfigure} 
          \begin{subfigure}[b]{\fivew}
                \begin{tikzpicture}
                \node [inner sep=0pt,outer sep=0pt,clip,rounded corners=2pt] at (0,0) {\includegraphics[width=1\textwidth]{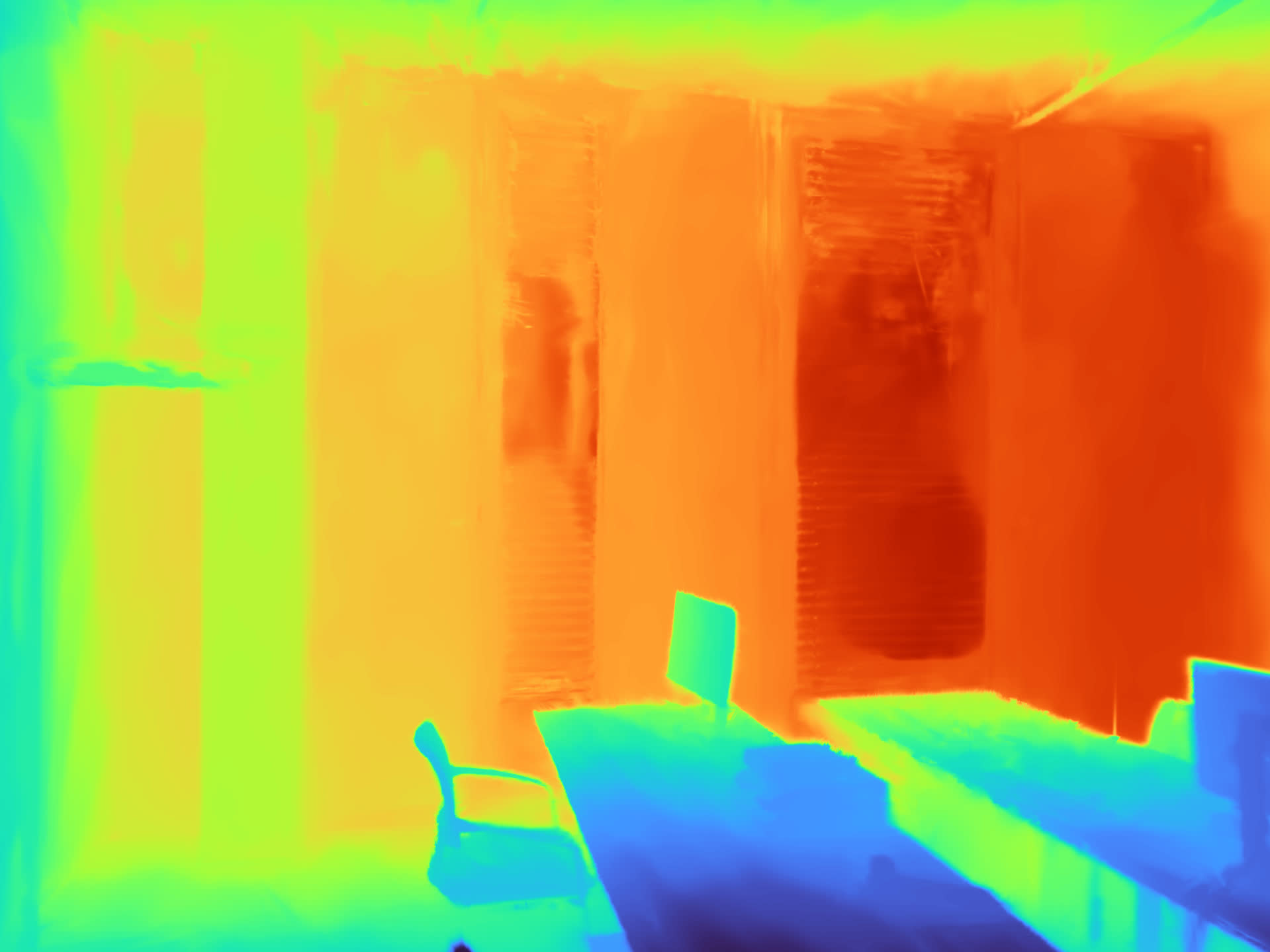}};
                \end{tikzpicture}
                \subcaption*{\texttt{$\mathcal{L}_\text{MSE}$}}
          \end{subfigure} 
          \begin{subfigure}[b]{\fivew}
                \begin{tikzpicture}
                \node [inner sep=0pt,outer sep=0pt,clip,rounded corners=2pt] at (0,0) {\includegraphics[width=1\textwidth]{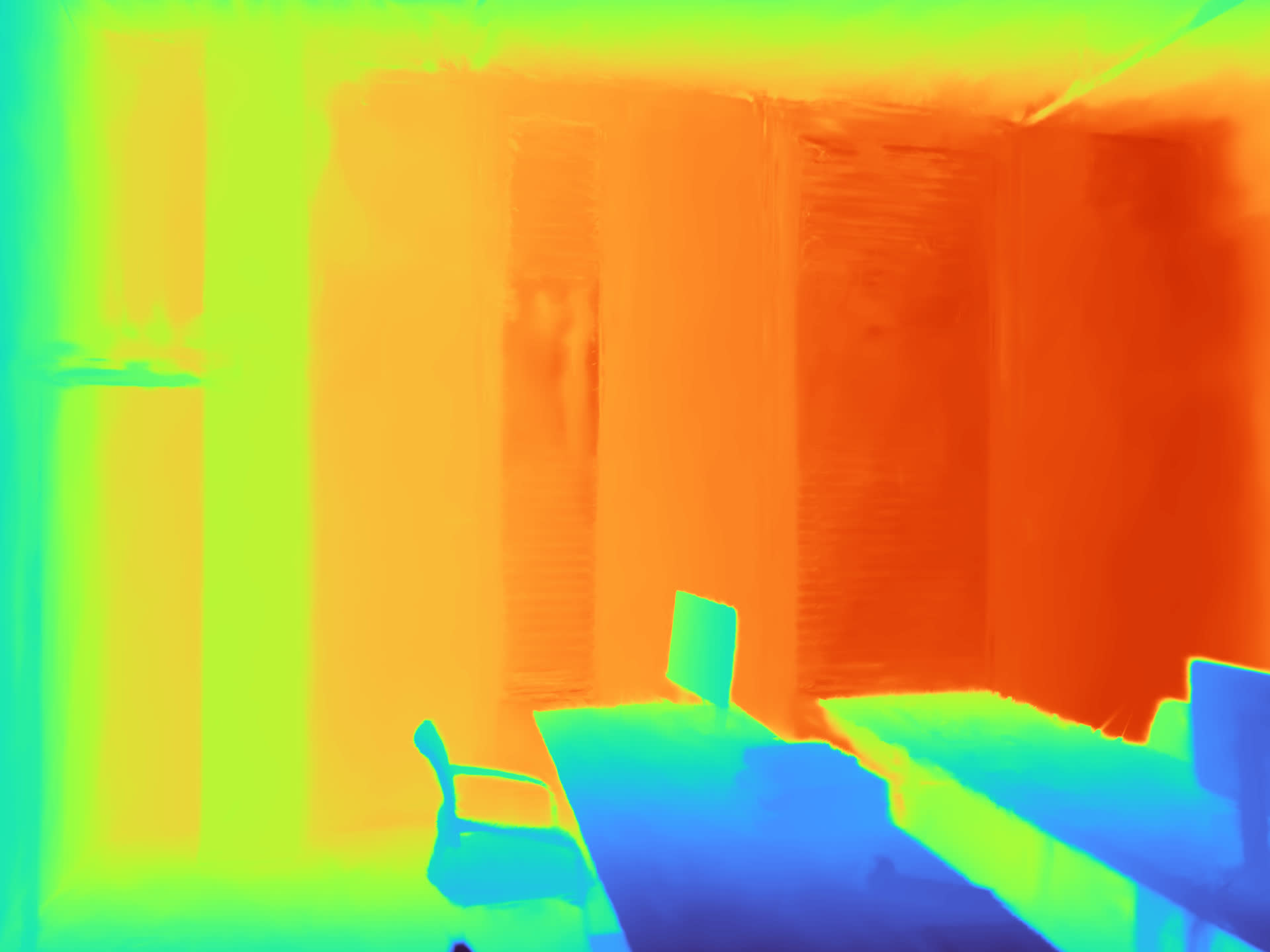}};
                \end{tikzpicture}
                \subcaption*{\texttt{$\mathcal{L}_1$}}
          \end{subfigure} 
          \begin{subfigure}[b]{\fivew}
                \begin{tikzpicture}
                \node [inner sep=0pt,outer sep=0pt,clip,rounded corners=2pt] at (0,0) {\includegraphics[width=1\textwidth]{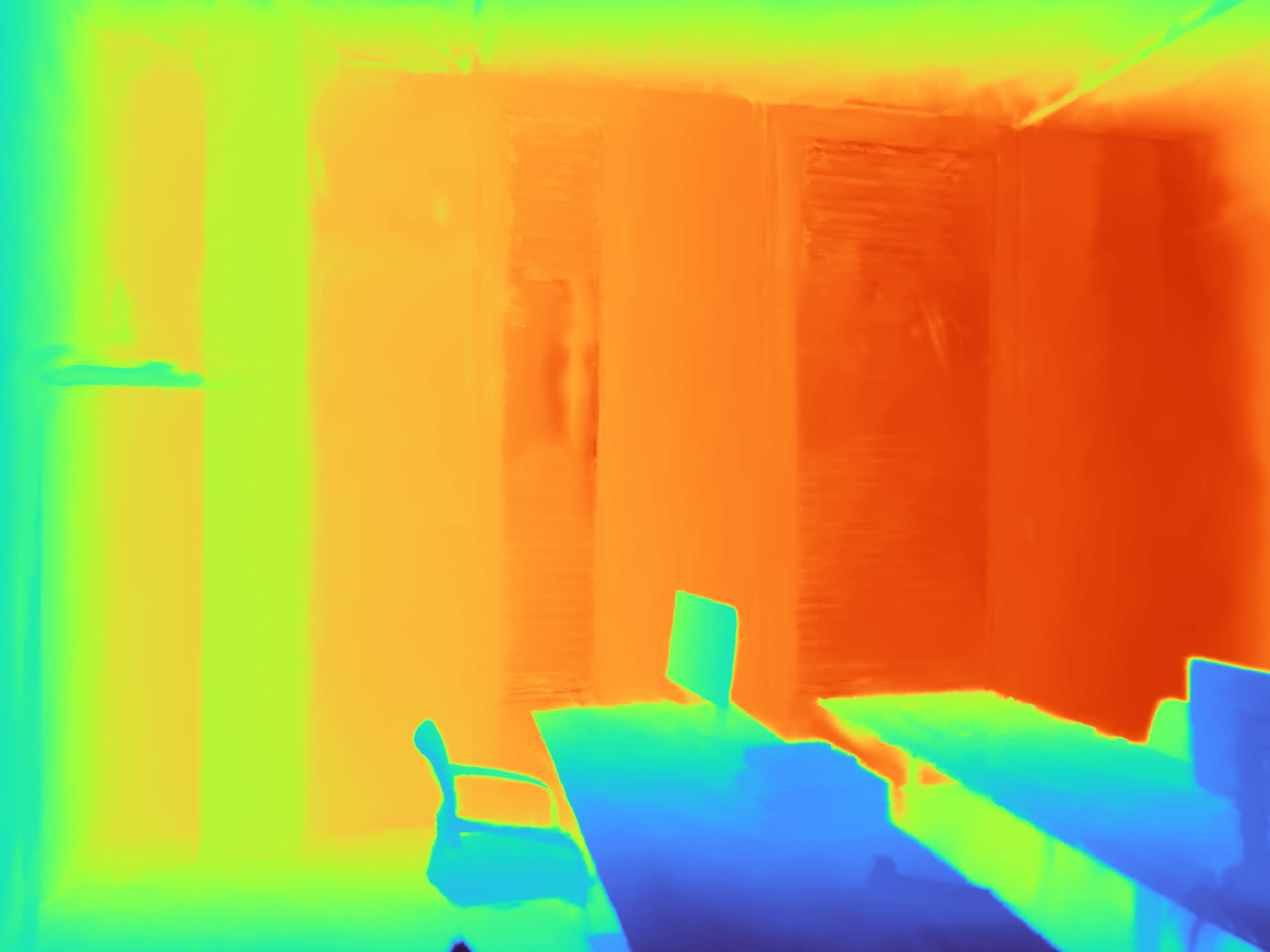}};
                \end{tikzpicture}
                \subcaption*{\texttt{$\mathcal{L}_\text{LogL1}$}}
          \end{subfigure} 
          \begin{subfigure}[b]{\fivew}
                \begin{tikzpicture}
                \node [inner sep=0pt,outer sep=0pt,clip,rounded corners=2pt] at (0,0) {\includegraphics[width=1\textwidth]{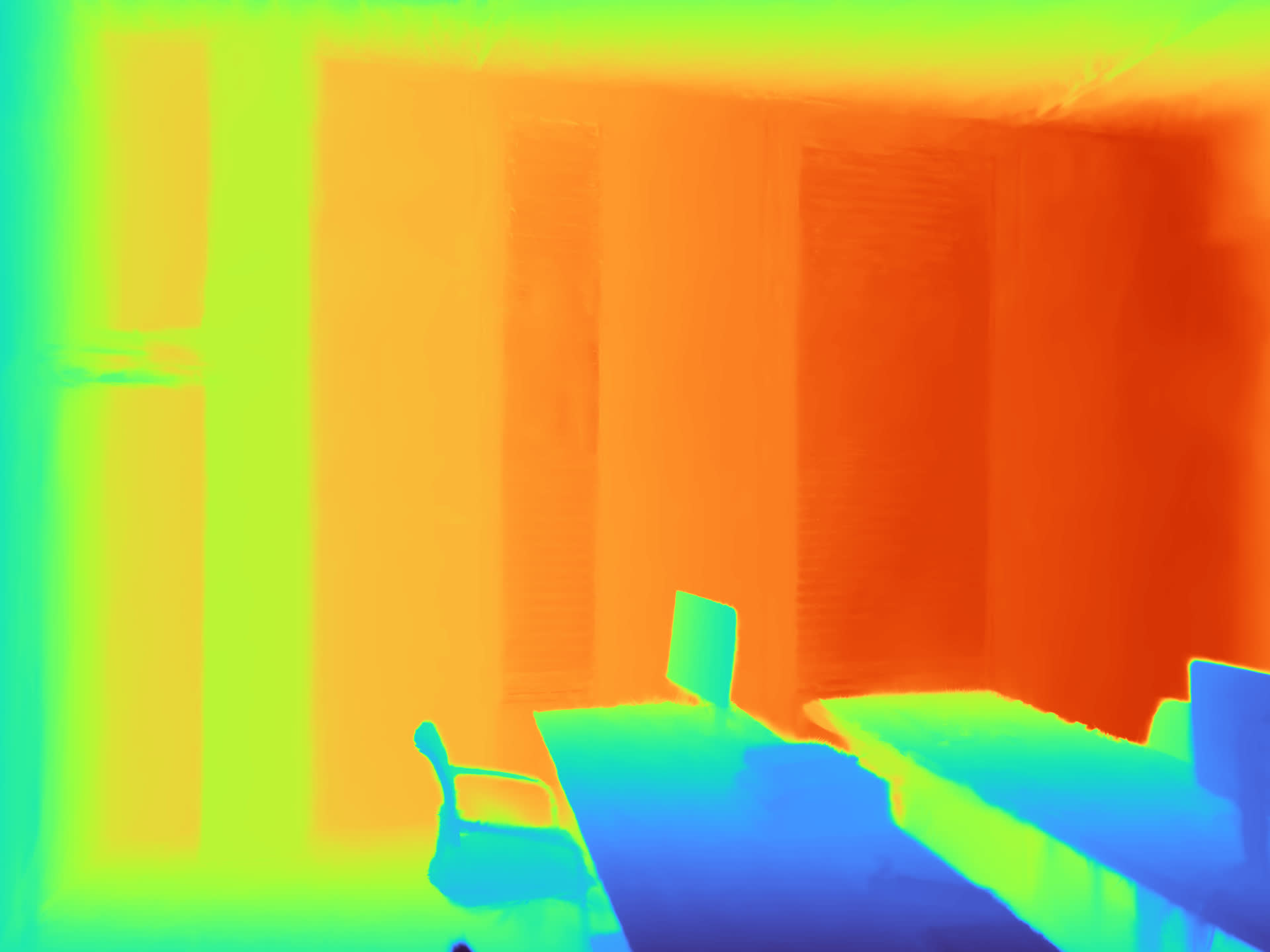}};
                \end{tikzpicture}
                \subcaption*{$\mathcal{L}_{\hat{D}}$ (\texttt{Ours})}
          \end{subfigure} 
          \begin{subfigure}[b]{\fivew}
                \begin{tikzpicture}
                \node [inner sep=0pt,outer sep=0pt,clip,rounded corners=2pt] at (0,0) {\includegraphics[width=1\textwidth]{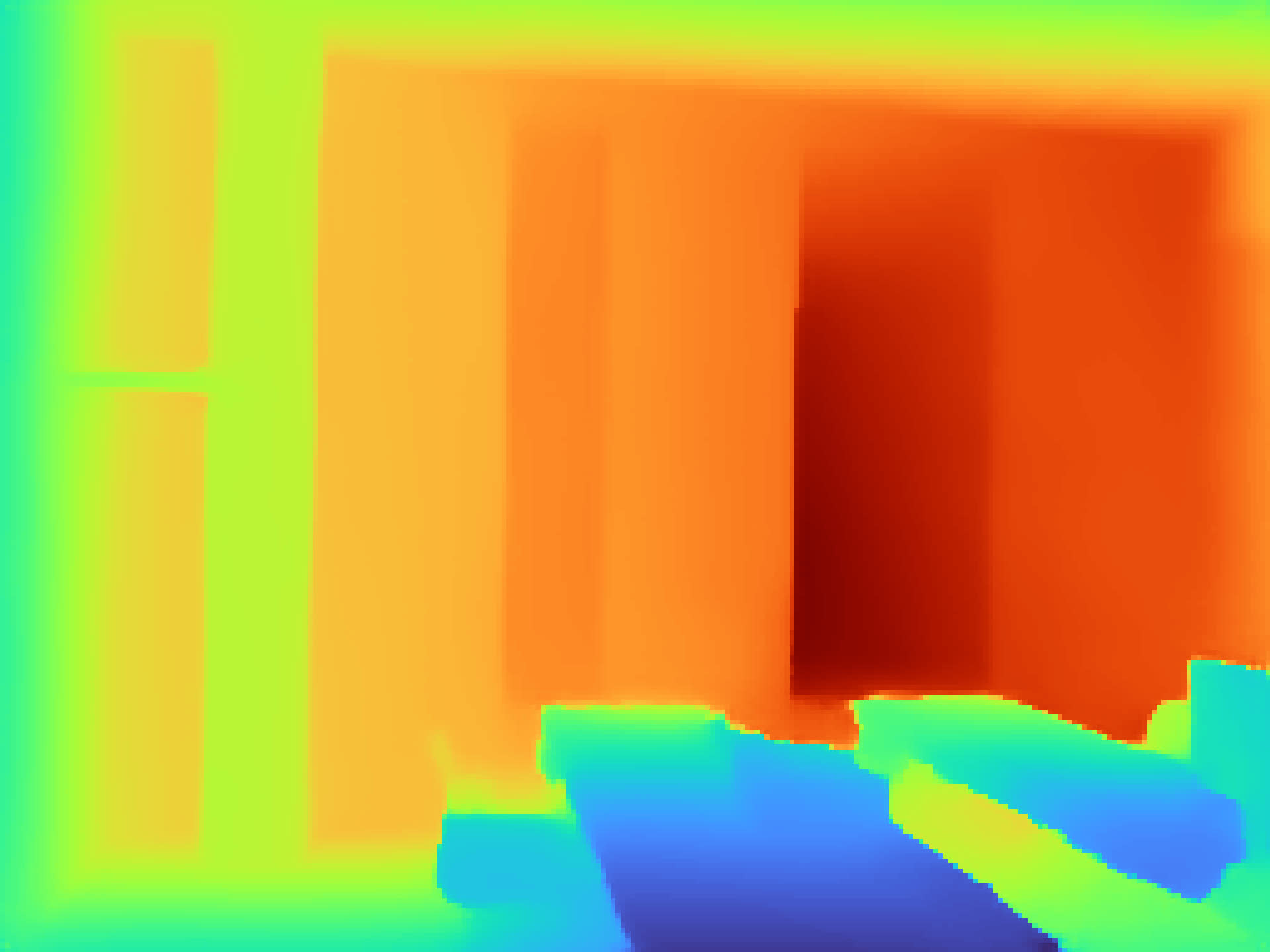}};
                \end{tikzpicture}
                \subcaption*{\texttt{iPhone GT}}
          \end{subfigure} 
        \end{tabular}
    \end{minipage}%
    \\%
\end{tabular}
\vspace{-1em}
\caption{\textbf{Qualitative comparison of depth losses: ScanNet++.} We observe that the proposed gradient aware $\mathcal{L_{\hat{D}}}$ regularizer obtains the best qualitative results, mitigating uncertainties at edges from the raw iPhone depth captures. Zoom in to see the details.}
\label{fig:ab-depth-loss}
\vspace{-1em}
\end{figure*}

\begin{table}[t]
    \centering
    \scriptsize
    \setlength{\tabcolsep}{1.7pt}
    \renewcommand*{\arraystretch}{1}
    \begin{tabular}{l|ccc|cc|c}
\toprule
    Method & C-$L_1$ $\downarrow$ & NC $\uparrow$& F1$\uparrow$ & PSNR $\uparrow$ &SSIM $\uparrow$ & Time (min)\\
    \midrule
    Splatfacto ($\mathcal{L}_{\text{rgb}}$)~\cite{kerbl20233d, nerfstudio} & 7.0 & 78.0 & 56.3 & 26.6 & .870 & 12.9 \\
    + Normal ($\mathcal{L}_{\hat{N}}$) & 7.7 & 81.3 & 60.2 & 26.7 & .874 & 17.2 \\
    + Normal ($\mathcal{L}_{\hat{N}}$ + $\mathcal{L}_{\text{smooth}}$) & 6.7 & 82.6 & 58.9 & 26.4 & .866 & 17.3 \\
    + Depth ($\mathcal{L}_{\hat D}$) & 2.6 & 86.8 & 86.6 & \textbf{27.1} & \textbf{.885} & 13.1 \\
    + Both ($\mathcal{L}_{\hat D}$ + $\mathcal{L}_{\hat{N}}$) & \textbf{2.4} & \textbf{90.0} & \textbf{87.7} & \underline{26.9} & \underline{.883} & 17.5 \\
    \rowcolor{gray!10} + Both ($\mathcal{L}_{\hat D}$ + $\mathcal{L}_{\hat{N}}$ + $\mathcal{L}_{\text{smooth}}$) & \underline{2.5} & \underline{89.6} & 87.1 & 26.8 & .879 & 17.9 \\
\bottomrule
\end{tabular}
\vspace{-1em}
\caption{\textbf{Comparison of geometric supervision strategies}. Geometric cues significantly improve reconstruction quality. We report mesh, novel view, and training time metrics (Nvidia 4090 GPU, 30k iterations) on the 'VR room' sequence from MuSHRoom}
\vspace{-2em}
\label{tab:ab-supervision-strat}
\end{table}

\begin{table}[t]
    \centering
    \scriptsize
    \setlength{\tabcolsep}{3.4pt}
    \renewcommand*{\arraystretch}{1}
    \begin{tabular}{l|c|ccc|cc}
    \toprule
    Initialization Method & \# Init GS & C-$L_1$ $\downarrow$ & NC $\uparrow$& F1$\uparrow$ & PSNR $\uparrow$ &SSIM $\uparrow$\\
    \midrule
    COLMAP SfM \cite{schoenberger2016mvscolmap2} & $\sim$54.8K & .0242 & .8713 & .9066 & 24.23 & .8411\\
    \midrule
    Sensor Depth & 50 K &.0242 & .8725 & .9074 & 24.16 & .8389  \\
    Sensor Depth & 100 K & .0241 & .8714 & .9079 & 24.19 & .8395\\
    Sensor Depth & 500 K & .0239 &.8730 &.9091 &24.20 &.8399 \\
    \rowcolor{gray!10} Sensor Depth & 1 M & .0215 & .8390 & .9381 & 24.58 & .8558 \\
    Sensor Depth & 1.5 M & .0238 & .8730 & .9171 & 24.31 & .8491 \\
    \bottomrule
    \end{tabular}
    \vspace{-1em}
\caption{\textbf{Comparison of SfM and sensor depth initialization.} We compare Gaussian scene initialization using COLMAP \cite{schoenberger2016mvscolmap2} SfM points and those obtained from back-projecting training dataset sensor depth readings on the MuSHRoom dataset. We note that sensor depth initialization can improve overall reconstruction results compared to SfM initialization. We use 1M points in our experiments. Results are averaged over 6 scenes.}
\label{tab:ab-sfm-vs-sensor-depth}
\vspace{-1em}
\end{table}
\begin{table}[t]
    \centering\scriptsize
    \setlength{\tabcolsep}{1.1pt}
    \renewcommand*{\arraystretch}{1}
    \begin{tabular}{lcccc|ccc}%
    \toprule
    Method & Abs Rel $\downarrow$ & Sq Rel $\downarrow$ & RMSE $\downarrow$ & $\delta<1.25$ $\uparrow$ & PSNR $\uparrow$ & SSIM $\uparrow$\\ 
    \midrule
    Splatfacto~\cite{nerfstudio} & .1481 & .1345 & .5122 & .7602 & 23.41 & .9111\\ \midrule
    Splatfacto + $\mathcal{L}_{\text{MSE}}$ & .0551 & .0180 & .2104 & .9647 & \textbf{23.84} & \textbf{.9140}\\
    Splatfacto + $\mathcal{L}_{\mathrm{1}}$ & \underline{.0351} & \underline{.0170} & \underline{.1889} & \underline{.9758} & 23.74 & .9138 &\\
    Splatfacto + $\mathcal{L}_{\mathrm{LogL1}}$ & .0364 & .0180 & .1955 & .9744 & \underline{23.77} & \underline{.9139}\\
    \rowcolor{gray!10} Splatfacto + $\mathcal{L}_{\hat{D}}$ & \textbf{.0285} & \textbf{.0162} & \textbf{.1790} & \textbf{.9782} & 23.61 & .9122 \\
    \bottomrule
    \end{tabular}%
    \vspace{-1em}
    \caption{\textbf{Quantitative comparison of depth losses: ScanNet++.} We compare various depth supervision strategies and the resulting novel-view and mesh reconstruction results. See \cref{fig:ab-depth-loss} for qualitative comparisons of the losses.}
    \vspace{-1em}
    \label{tab:ab_main_depth_losses}
\end{table}%
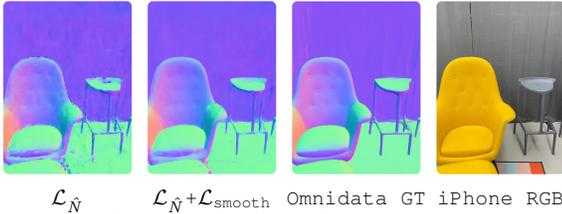
\begin{figure}[t]
\centering
  \begin{subfigure}[b]{0.12\textwidth}
    \centering
    \begin{tikzpicture}
    \node [inner sep=0pt,clip,rounded corners=2pt] at (0,0) {\includegraphics[height=2.3cm]{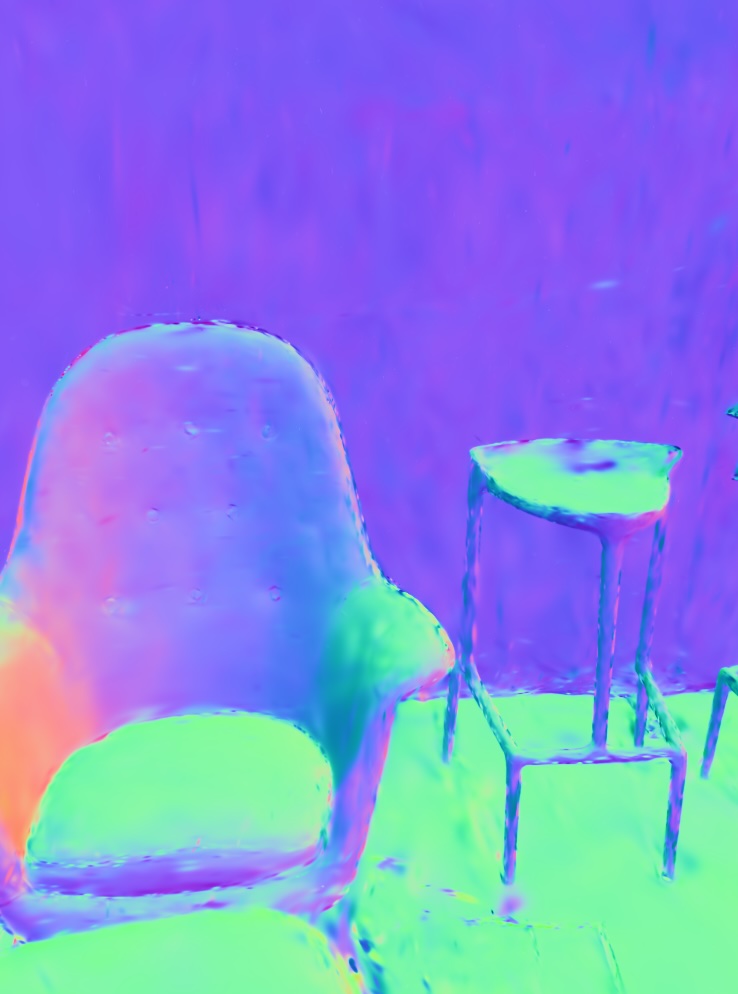}};
    \node[font=\tiny, text=white] at (0.2,-1.1) {};
    \end{tikzpicture}
    \subcaption*{\texttt{$\mathcal{L}_{\it{\hat{N}}}$}}
  \end{subfigure}
  \hspace{-10pt}
  \begin{subfigure}[b]{0.12\textwidth}
    \centering
    \begin{tikzpicture}
    \node [inner sep=0pt,clip,rounded corners=2pt] at (0,0) {\includegraphics[height=2.3cm]{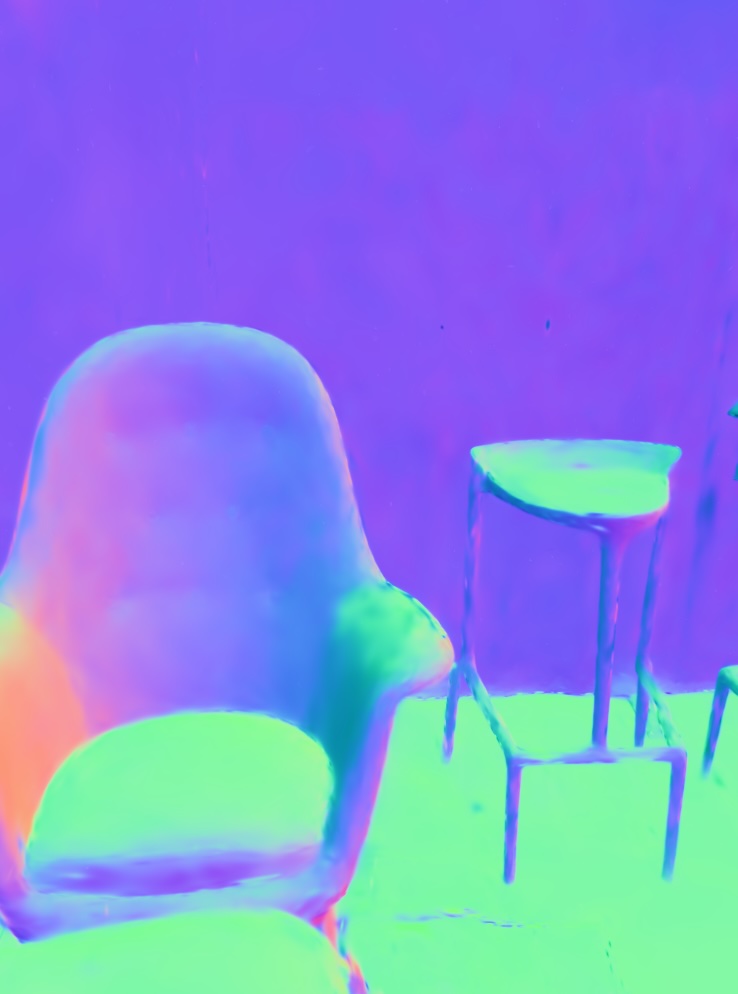}};
    \node[font=\tiny, text=white] at (0.2,-1.1) {};
    \end{tikzpicture}
    \subcaption*{\texttt{$\mathcal{L}_{\it{\hat{N}}}$+$\mathcal{L}_{\text{smooth}}$}}
  \end{subfigure}
  \hspace{-10pt}
  \begin{subfigure}[b]{0.12\textwidth}
    \centering
    \begin{tikzpicture}
    \node [inner sep=0pt,clip,rounded corners=2pt] at (0,0) {\includegraphics[height=2.3cm]{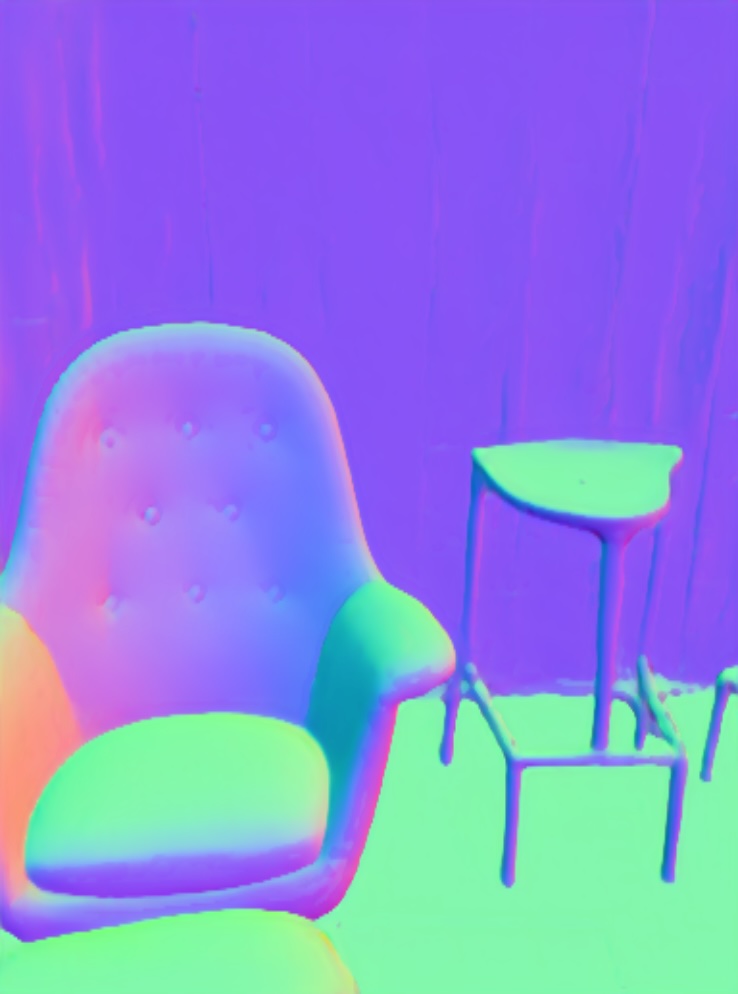}};
    \node[font=\tiny, text=white] at (0.2,-1.1) {};
    \end{tikzpicture}
     \subcaption*{\texttt{Omnidata GT}}
  \end{subfigure}
  \hspace{-10pt}
  \begin{subfigure}[b]{0.12\textwidth}
    \centering
    \begin{tikzpicture}
    \node [inner sep=0pt,clip,rounded corners=2pt] at (0,0) {\includegraphics[height=2.3cm]{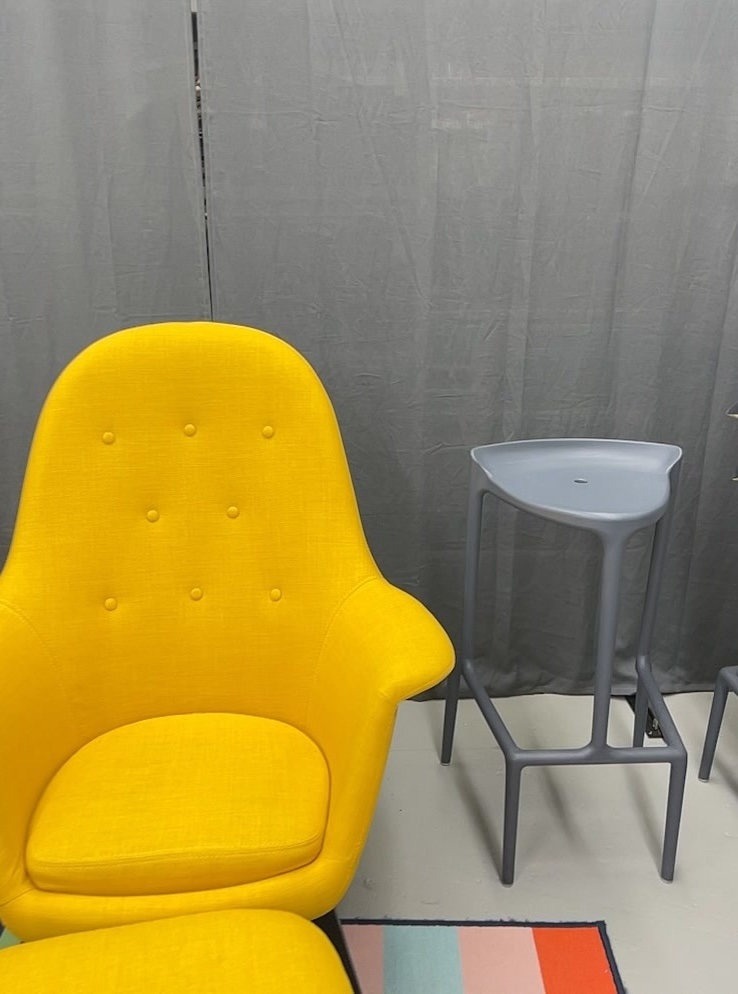}};
    \node[font=\tiny, text=white] at (0.2,-1.1) {};
    \end{tikzpicture}
     \subcaption*{\texttt{iPhone RGB}}
  \end{subfigure}
  \vspace{-1em}
  \caption{\textbf{Qualitative comparison of normal supervision.} We observe that using a direct normal loss $\mathcal{L}_{\hat{N}}$ (a) results in non-smooth surfaces whereas using a normal smoothing prior $\mathcal{L}_{\text{smooth}}$ (b) significantly improves normal estimates according to (c) Omnidata ~\cite{eftekhar2021omnidata} predictions.}
  \vspace{-1em}
  \label{fig:normal-supervision}
\end{figure}
\begin{table}[t]
    \centering
    \scriptsize
    \setlength{\tabcolsep}{5.2pt}
    \renewcommand*{\arraystretch}{1}
    \begin{tabular}{lccc|cc}
    \toprule
     Method & C-$L_1$ $\downarrow$ & NC $\uparrow$& F1$\uparrow$ & PSNR $\uparrow$ &SSIM $\uparrow$ \\
    \midrule
    2DGS              & .0687 & .8008 & .6039 & 22.52 & .8185  \\
    2DGS w/ Monodepth \cite{bhat2023zoedepth} & .0795 & .8558 & .5446 & 21.85 & .8009  \\
    2DGS w/ Sensor Depth & \underline{.0275} & \textbf{.8830} & \underline{.8886} & 23.02 & .8250 \\
    \midrule
    Splatfacto & .0652 & .7727 & .5835 & 24.44 & .8486  \\
    Ours w/ Monodepth \cite{bhat2023zoedepth} & .0477 & .0860 & .6963 & \underline{24.48} & \underline{.8509} \\
    \rowcolor{gray!10} Ours w/ Sensor Depth & \textbf{.0216} & \underline{.8822} & \textbf{.9243} & \textbf{24.58} & \textbf{.8558}  \\
    \bottomrule
\end{tabular}
\vspace{-1em}
\caption{\textbf{2DGS vs. DN-Splatter}. We implement depth supervision to the recent 2DGS \cite{Huang2DGS2024} method and compare with monocular and sensor depth regularization on MuSHRoom. We use the same loss and initialization strategy for fair comparison. Results are averaged over 6 scenes. NVS is reported within the train sequence.}
\label{tab:ab-2dgs-vs-dn-splatter}
\vspace{-2em}
\end{table}

\subsection{Ablation studies}

\boldparagraph{Proposed regularization strategy.} We evaluate various design choices of our method in~\tabref{tab:ab-supervision-strat}. The normal loss $\mathcal{L}_{\hat{N}}$ (\cref{eq:normal-loss}) helps align Gaussians along the scene geometry improving the resulting mesh normal-completeness and F-scores. The depth loss $\mathcal{L}_{\hat D}$ (\cref{eq:depth loss}) significantly improves reconstruction quality and novel-view synthesis in ambiguous, textureless regions which are common in indoor scenes. The normal smoothing prior $\mathcal{L}_{\text{smooth}}$ (\cref{eq:tv loss}) further helps normal-completeness for Poisson meshing with a minimal impact on other metrics. Although the smoothing prior's effect on quantitative metrics is minimal, its importance is evident in the qualitative renders illustrated in~\cref{fig:normal-supervision}.

\boldparagraph{SfM vs. sensor depth initialization.} We compare initialization strategies using sparse COLMAP \cite{schoenberger2016mvscolmap2, schoenberger2016sfmcolmap1} SfM points and those obtained from back-projecting sensor depth readings. We note that using dense sensor depth data for initialization enhances both mesh and novel-view synthesis metrics, demonstrating the value of incorporating available sensor depth data.

\boldparagraph{2DGS vs. DN-Splatter.} We evaluate our method against a variant of 2DGS \cite{Huang2DGS2024} with the same depth regularization strategy in~\tabref{tab:ab-2dgs-vs-dn-splatter}. Our comparison includes both monocular depth supervision with the Pearson correlation loss \cite{xiong2023sparsegs} and sensor depth supervision. We observe that 2DGS also faces difficulties with textureless regions, which are common in indoor datasets, leading to challenges with planar surfaces.

\boldparagraph{Depth supervision and losses.} We assess the impact of the proposed depth loss \cref{eq:depth loss} on ScanNet++, with qualitative renders shown in~\figref{fig:ab-depth-loss} and quantitative results in \tabref{tab:ab_main_depth_losses}. We compare common depth losses: $\mathcal{L}_\text{MSE}$, $\mathcal{L}_{\mathrm{1}}$, $\mathcal{L}_{\mathrm{LogL1}}$~\cite{Hu_Ozay_Zhang_Okatani_2018}, and our edge-aware $\mathcal{L}_{\hat{D}}$. Results indicate that $\mathcal{L}_{\mathrm{1}}$ and $\mathcal{L}_{\mathrm{LogL1}}$ generally yield the best color metrics, with the logarithmic variant providing the smoothest reconstructions. Further details and comparisons with ground truth Faro lidar scans are available in the supplementary material.

Lastly, \tabref{tab:ab-sensor-vs-mono} evaluates our regularization strategy against common alternatives. We compare reconstruction with SotA monocular~\cite{bhat2023zoedepth, hu2024metric3dv2} and multi-view~\cite{zhang2020visibility} networks using the alignment strategy outlined in~\secref{sec:Depth regularization} using~\cref{eq:mono-sparse-alignment}. 
We also examine the patch-based Pearson correlation loss~\cite{xiong2023sparsegs} for relative depth supervision. While it performs better than naive monocular depth supervision, reconstruction quality is still inferior to using iPhone depths. Despite their low resolution and inaccuracies (mainly at object edges), sensor depths remain practical for real-world indoor scenes. Further research is needed to improve monocular depth supervision performance.
\boldparagraph{}
\begin{table}[t]
    \centering
    \scriptsize
    \setlength{\tabcolsep}{2.7pt}
    \renewcommand*{\arraystretch}{1}
    \begin{tabular}{lccccc}
    \toprule
     Method & Acc. $\downarrow$ & Comp. $\downarrow$& C-$L_1$ $\downarrow$ & NC $\uparrow$& F-score$\uparrow$\\
    \midrule
    No supervision & .2627 & .2091 & .2359 & .6511 & .1343 \\
    \midrule
    Monodepth: Zoe-Depth~\cite{bhat2023zoedepth}& .1751 & .2084 & .1918 & .7420 & .1455 \\
    Monodepth: Metric3D~\cite{hu2024metric3dv2} & .1798 & .2079 & .1938 & .7358 & .1439 \\
    Multi-view Stereo (MVSNet)~\cite{zhang2020visibility} & .3120 & .2375 & .2748 & .6408 & .1903 \\
    Pearson Loss ~\cite{xiong2023sparsegs} w/ Metric3D & .1183 & .1766 & .1474 & .7975 & .2236 \\
    \midrule
    \rowcolor{gray!10} Sensor Depth (iPhone) & \textbf{.0609} & \textbf{.1433} & \textbf{.1021} & \textbf{.8130} & \textbf{.5833} \\
    \bottomrule
\end{tabular}
\vspace{-1em}
\caption{\textbf{Alternative depth regularization strategies}. We compare regularization with monocular~\cite{bhat2023zoedepth,hu2024metric3dv2} and Multi-view Stereo depth estimates ~\cite{hu2024metric3dv2} using our $\mathcal{L}_{\hat{D}}$ loss, the Pearson correlation loss (patch-based)~\cite{xiong2023sparsegs} for relative depth supervision with Zoe-Depth estimates, as well as utilizing iPhone sensor depth supervision on the 'b20a261fdf' scene from ScanNet++.}
\vspace{-2em}
\label{tab:ab-sensor-vs-mono}
\end{table}

\section{Conclusion}
We presented DN-Splatter, a method for depth and normal regularization of 3DGS to address photorealistic and geometrically accurate reconstruction of challenging indoor datasets. This simple yet effective strategy enhances novel-view metrics and significantly improves the surface quality extracted from a Gaussian scene. We demonstrated that prior regularization is essential for achieving more geometrically valid and consistent reconstructions in challenging indoor scenes.
\section{Acknowledgments}
We thank Tobias Fischer, Songyou Peng, and Philipp Lindenberger for their fruitful discussions. We thank Jiepeng Wang and Marcus Klasson for their help in proof reading. We acknowledge funding from the Academy of Finland (grant No. 327911, 353138, 324346, and 353139) and support from the Wallenberg AI, Autonomous Systems and Software Program (WASP) funded by the Knut and Allice Wallenberg Foundation. MT was funded by the Finnish Center for Artificial Intelligence (FCAI).

{\small
\bibliographystyle{ieee_fullname}
\bibliography{egbib}
}

\clearpage
\title{Supplementary Material}
\maketitle

\appendix
\setcounter{page}{1}

In this \textbf{supplementary material}, we provide further details regarding our baseline methods and datasets in \cref{supp:implementation}, definitions for our evaluation metrics and losses in \cref{supp:metrics}, and further quantitative and qualitative results in \cref{supp_sec:additional_results} and \cref{supp_sec:qualitative} respectively.

\section{Implementation details}
\label{supp:implementation}
\subsection{Baselines}\label{supp_sec:baselines}
We compare a variety of baseline methods for novel view synthesis, depth estimation, and mesh reconstruction.

\boldparagraph{\texttt{Nerfacto}.} We use the Nerfacto model from Nerfstudio \cite{nerfstudio} version 1.0.2 in our experiments. We use default settings, disable pose optimization, and predict normals using the proposed method from Ref-NeRF \cite{verbin2022refnerf}. We use rendered normal and depth maps for Poisson surface reconstruction.

\boldparagraph{\texttt{Depth-Nerfacto}.} We use the depth supervised variant of Nerfacto with a direct loss on ray termination distribution for sensor depth supervision as described in DS-NeRF \cite{kangle2021dsnerf}. Besides this, we use the same settings as for Nerfacto.

\boldparagraph{\texttt{Neusfacto}.} We use default settings provided by Neusfacto from SDFStudio \cite{Yu2022SDFStudio} and use the default marching cubes algorithm for meshing.

\boldparagraph{\texttt{MonoSDF}.} We use the recommended settings from MonoSDF~\cite{Yu2022MonoSDF} and with sensor depth and monocular normal supervision. We set the sensor depth loss multiplier to 0.1 and normal loss multiplier to 0.05. Normal predictions are obtained from Omnidata \cite{eftekhar2021omnidata}.

\boldparagraph{\texttt{Splatfacto}.} The Splatfacto model from Nerfstudio version 1.1.3 and \texttt{gsplat}~\cite{Ye_gsplat} version 1.0.0 serves as our baseline 3DGS model. This is a faithful re-implementation of the original 3DGS work \cite{kerbl20233d}. We keep all the default settings for the baseline comparison.

\boldparagraph{\texttt{SuGaR}.} We use the official SuGaR~\cite{guedon2023sugar} source-code. The original code-base, written as an extension to the original 3DGS work \cite{kerbl20233d}, supports only COLMAP based datasets (that is, datasets containing a COLMAP database file). We made slight modifications to the original source-code to support non-COLMAP based formats to import camera information and poses directly from a pre-made .json files. We use default settings for training as described in \cite{guedon2023sugar}. We use the SDF trained variant in all experiments. We extract both the coarse and refined meshes for evaluation, although the difference in geometry metrics are small between them. We found a small inconsistency in SuGaR's normal directions for outward facing indoor datasets, which we corrected in our experiments. 

In addition, we have modified the original source-code to support depth rendering, which was not possible in the original author's code release. This is achieved by replacing the CUDA backend with a variant that also includes depth rendering support.

\boldparagraph{\texttt{2DGS}.} We use the official 2DGS~\cite{Huang2DGS2024} source-code. Similar to our SuGaR implementation, we made slight modifications to the original source-code to support non-COLMAP based formats to import camera information and poses directly from a pre-made .json files. We use default settings for training as described in \cite{Huang2DGS2024} and the default meshing strategy using TSDF fusion.

\boldparagraph{\texttt{2DGS + $\mathcal{L_{\hat{D}}}$ variant}.} We implement the proposed depth regularization strategy into the official 2DGS code release. Specifically, we enable supervision and gradient flow to depths within the CUDA backend rasterizer and supervise with sensor or monocular depth estimates. Our overall optimization loss becomes $\mathcal{L} = \mathcal{L}_{\text{rgb}} + \lambda_d \mathcal{L}_{d}$ where $\lambda_d$ is set to 0.2 and $\mathcal{L}_{\text{rgb}}$ is the original loss from \cite{Huang2DGS2024}.

\subsection{Datasets}\label{supp_sub:datasets}
\boldparagraph{MuSHRoom.} We use the official train and evaluation splits from the MuSHRoom \cite{ren2023mushroom} dataset. We report evaluation metrics on a) images obtained from uniformly sampling every 10 frames from the training camera trajectory and b) images obtained from a different camera trajecotry. We use the globally optimized COLMAP~\cite{schoenberger2016sfmcolmap1} for both evaluation sequences. We use a total of 5 million points for mesh extraction for Poisson surface reconstruction.

\boldparagraph{ScanNet++.} We use the "b20a261fdf" and "8b5caf3398" scenes in our experiments. We use the iPhone sequences with COLMAP registered poses. The sequences contain 358 and 705 registered images respectively. We uniformly load every 5th frame from the sequences from which we reserve every 10th frame for evaluation.

\section{Definitions for metrics and losses}
\label{supp:metrics}
\subsection{Depth evaluation metrics}\label{supp_subsec:depth_metrics}
For the ScanNet++ and MuSHRoom datasets, we follow
~\cite{wei2021nerfingmvs,kusupati2020normal,luo2020consistent,sinha2020deltas,murez2020atlas,teed2018deepv2d,zhou2017unsupervised} and report depth evaluation metrics, defined in~\tabref{tab:depth_metrics_definition}. We use the Absolute Relative Distance (\textit{Abs Rel}), Squared Relative Distance (\textit{Sq Rel}), Root Mean Squared Error \textit{RMSE} and its logarithmic variant \textit{RMSE log}, and the \textit{Threshold Accuracy} $(\delta<t)$ metrics. The \textit{Abs Rel} metric provides a measure of the average magnitude of the relative error between the predicted depth values and the ground truth depth values. Unlike the \textit{Abs Rel} metric, the \textit{Sq Rel} considers the squared relative error between the predicted and ground truth depth values. The \textit{RMSE} metric calculates the square root of the average of the squared differences between the predicted and the ground-truth values, giving a measure of the magnitude of the error made by the predictions. The \textit{RMSE log} metric is similar to \textit{RMSE} but applied in the logarithmic domain, which can be particularly useful for very large depth values. The \textit{Threshold accuracy} measures the percentage of predicted depth values within a certain threshold factor, $\delta$ of the ground-truth depth values.

\begin{table}[t!]
    \centering\scriptsize
    \setlength{\tabcolsep}{3.2pt}
    \renewcommand*{\arraystretch}{2}
    \begin{tabular}{lc}
\toprule
Metric & Definition \\
\midrule
\rowcolor{gray!10} Abs Rel & $\frac{1}{N} \sum_{i=1}^{N} \frac{\left|d_{i}^{\text{pred}} - d_{i}^{\text{gt}}\right|}{d_{i}^{\text{gt}}}$ \\
\rowcolor{gray!10} Sq Rel & $\frac{1}{N} \sum_{i=1}^{N} \frac{\left(d_{i}^{\text{pred}} - d_{i}^{\text{gt}}\right)^2}{d_{i}^{\text{gt}}}$ \\
\rowcolor{gray!10} RMSE & $\sqrt{\frac{1}{N} \sum_{i=1}^{N} \left(d_{i}^{\text{pred}} - d_{i}^{\text{gt}}\right)^2}$ \\
RMSE log & $\sqrt{\frac{1}{N} \sum_{i=1}^{N} \left(\log d_{i}^{\text{pred}} - \log d_{i}^{\text{gt}}\right)^2}$ \\
\rowcolor{gray!10} Threshold accuracy, $\delta$ & $\frac{1}{N} \sum_{i=1}^{N}\left[ \max\left(\frac{d_{i}^{\text{pred}}}{d_{i}^{\text{gt}}}, \frac{d_{i}^{\text{gt}}}{d_{i}^{\text{pred}}}\right) < \delta \right]$\\
\bottomrule
\end{tabular}
    
    \caption{\textbf{Depth Evaluation Metrics.} 
    We show definitions for our depth evaluation metrics. $d_{i}^{\text{pred}}$ and $d_{i}^{\text{gt}}$ are predicted and ground-truth depths for the $i$-th pixel. $\delta$ is the threshold factor (\eg, $\delta<1.25$, $\delta<1.25^2$, $\delta<1.25^3$).}
    \label{tab:depth_metrics_definition}
    \vspace*{-1em}
\end{table}

\subsection{Mesh evaluation metrics}\label{supp_sec:mesh_evaluation_metrics}
In \tabref{tab:supp_mesh_evaluation_metrics} we provide the definitions for mesh evaluation used throughout the text for comparing predicted and ground truth meshes. We use a threshold of $5cm$ for precision, recall, and F-scores. Furthermore, we evaluate mesh quality only within the visibility of the training camera views.
\begin{table}[ht]
      \centering\scriptsize
      \setlength{\tabcolsep}{3.2pt}
      \renewcommand*{\arraystretch}{1.4}
      \begin{tabular}{lc}
\toprule
 Metric & Definition \\
\midrule
\rowcolor{gray!10} Accuracy & 
$\frac{1}{|P|}\sum_{\mathbf{p} \in P}\left(\min _{\mathbf{p}^* \in P^*}\left\|\mathbf{p}-\mathbf{p}^*\right\|_1\right)$
\\
\rowcolor{gray!10} Completion & 
$\frac{1}{|P^*|}\sum_{\mathbf{p}^* \in P^*}\left(\min _{\mathbf{p} \in P}\left\|\mathbf{p}-\mathbf{p}^*\right\|_1\right)$
\\
\rowcolor{gray!10} Chamfer-$L_1$ & 
$\frac{\text { Accuracy + Completion }}{2}$
\\
 Normal Completion & 
$\frac{1}{|P^*|}\sum_{\mathbf{p}^* \in P^*}\left(\mathbf{n}_{\mathbf{p}}^T \mathbf{n}_{\mathbf{p}^*}\right)$ s.t. $\mathbf{p}=\underset{p \in P}{\text{argmin}}\left\|\mathbf{p}-\mathbf{p}^*\right\|_1$
\\
\rowcolor{gray!10} Normal-Consistency & 
$\frac{\text { Normal-Acc+Normal-Comp }}{2}$
\\
 Precision & 
$\frac{1}{|P|}\sum_{\mathbf{p} \in P}\left(\min _{\mathbf{p}^* \in P^*}\left\|\mathbf{p}-\mathbf{p}^*\right\|_1<5cm\right)$
\\
 Recall & 
$\frac{1}{|P^*|}\sum_{\mathbf{p}^* \in P^*}\left(\min _{\mathbf{p} \in P}\left\|\mathbf{p}-\mathbf{p}^*\right\|_1<5cm\right)$
\\
 \rowcolor{gray!10} F-score & 
$\frac{2 \cdot \text { Precision } \cdot \text { Recall }}{\text { Precision }+ \text { Recall }}$\\
\bottomrule
\end{tabular}
      \caption{\textbf{Mesh Evaluation Metrics}. $P$ and $P^*$
    are the point clouds sampled from the predicted and the
    ground truth mesh. $n_p$ is the normal vector at point $\mathbf{p}$.}
\label{tab:supp_mesh_evaluation_metrics}
\vspace*{-1em}
\end{table}

\subsection{Depth losses}\label{supp_subsec:depth_losses}
\label{sec:depth-losses}
For depth supervision, we compare the following variants of loss functions defined in~\tabref{tab:depth_objectives_definition}

\begin{table}[t!]
    \centering\scriptsize
    \setlength{\tabcolsep}{14pt}
    \renewcommand*{\arraystretch}{1.4}
  \begin{tabular}{lc}
    \toprule
    Loss & Definition \\
    \midrule
   $\mathcal{L}_{\text{MSE}}$ & $\frac{1}{\vert \hat{D} \vert} \sum (\hat{D} -D)^2$ \\
    $\mathcal{L}_{\mathrm{1}}$ & $\frac{1}{\vert \hat{D} \vert} \sum \| \hat{D} -D\|_1$ \\
    $\mathcal{L}_{\text{LogL1}}$ & $\frac{1}{\vert \hat{D} \vert} \sum \log(1+\| \hat{D} -D\|_1)$ \\
    $\mathcal{L}_{\text{HuberL1}}$ & $\begin{cases}\|D-\hat{D}\|_1, & \text { if }\|D-\hat{D}\|_1 \leq \delta, \\ \frac{(D-\hat{D})^2+\delta^2}{2 \delta}, & \text { otherwise. }\end{cases}$ \\
    $\mathcal{L}_{\text{DSSIML1}}$ & $\alpha \frac{1-\operatorname{SSIM}(I, \hat{I})}{2}+(1-\alpha)|I-\hat{I}|$\\
    $\mathcal{L}_{\text{EAS}}$ & $g_\text{rgb} \frac{1}{\vert \hat{D} \vert} \sum \| \hat{D} -D\|_1$\\
    \rowcolor{gray!10} $\mathcal{L}_{\hat{D}}$ & $g_\text{rgb} \frac{1}{\vert \hat{D} \vert} \sum \log(1+\| \hat{D} -D\|_1)$\\
    \bottomrule
\end{tabular}
\caption{\textbf{Depth Regularization Objectives.} We show the definitions for various depth objectives. Here, $\delta = 0.2\max(\|D-\hat{D}\|_1)$, $g_\text{rgb} = \text{exp}(-\nabla I)$, $D$/$\hat{D}$ are the ground truth and rendered depths, and $I$/$\hat{I}$ is the ground truth/rendered RGB image.}
\label{tab:depth_objectives_definition}
\vspace{-1em}
\end{table}

We compare the performance of these losses as supervision in~\tabref{tab:supp_ap_depth_losses}.

\section{Additional quantitative results}
\label{supp_sec:additional_results}
Here we provide additional quantitative results for DN-Splatter. We provide a comparison of Poisson meshing strategies, comparison of depth estimation quality with ground truth Faro scanner data, as well as further ablations on depth loss variants.
\subsection{Mesh extraction techniques} We investigate various Poisson meshing techniques. In~\cref{tab:meshing-ablation-replica}, we demonstrate that extracting oriented point sets from optimized depth and normal maps results in smoother and more realistic reconstructions compared to other methods. We report mesh evaluation metrics for these different techniques. We compare several approaches: directly using trained Gaussian means and normals for Poisson meshing (total of 512k Gaussians); extraction of surface density at levels 0.1 and 0.5 by projecting rays from camera views and querying scene intersections based on local density values, as proposed in SuGaR~ \cite{guedon2023sugar}; and back-projection of optimized depth and normal maps. All models were trained with our depth and normal regularization. To ensure a fair comparison, we set the total number of extracted points to 500k for both the surface density and back-projection methods.
\begin{table}[t!]
    \centering
    \setlength{\tabcolsep}{0pt}
        \begin{tabular}{c}
        \begin{minipage}{0.5\textwidth}
        \centering
            \setlength{\tabcolsep}{0pt}
            \renewcommand*{\arraystretch}{0.5}
            \footnotesize
            \begin{tabular}{ccccc}
            \begin{subtable}[b]{0.19\textwidth}
                \centering
                \subcaption*{\texttt{Gaussians}}
                \begin{tikzpicture}
                    \node [inner sep=0pt,clip,rounded corners=2pt] at (0,0) {\includegraphics[width=1\textwidth]{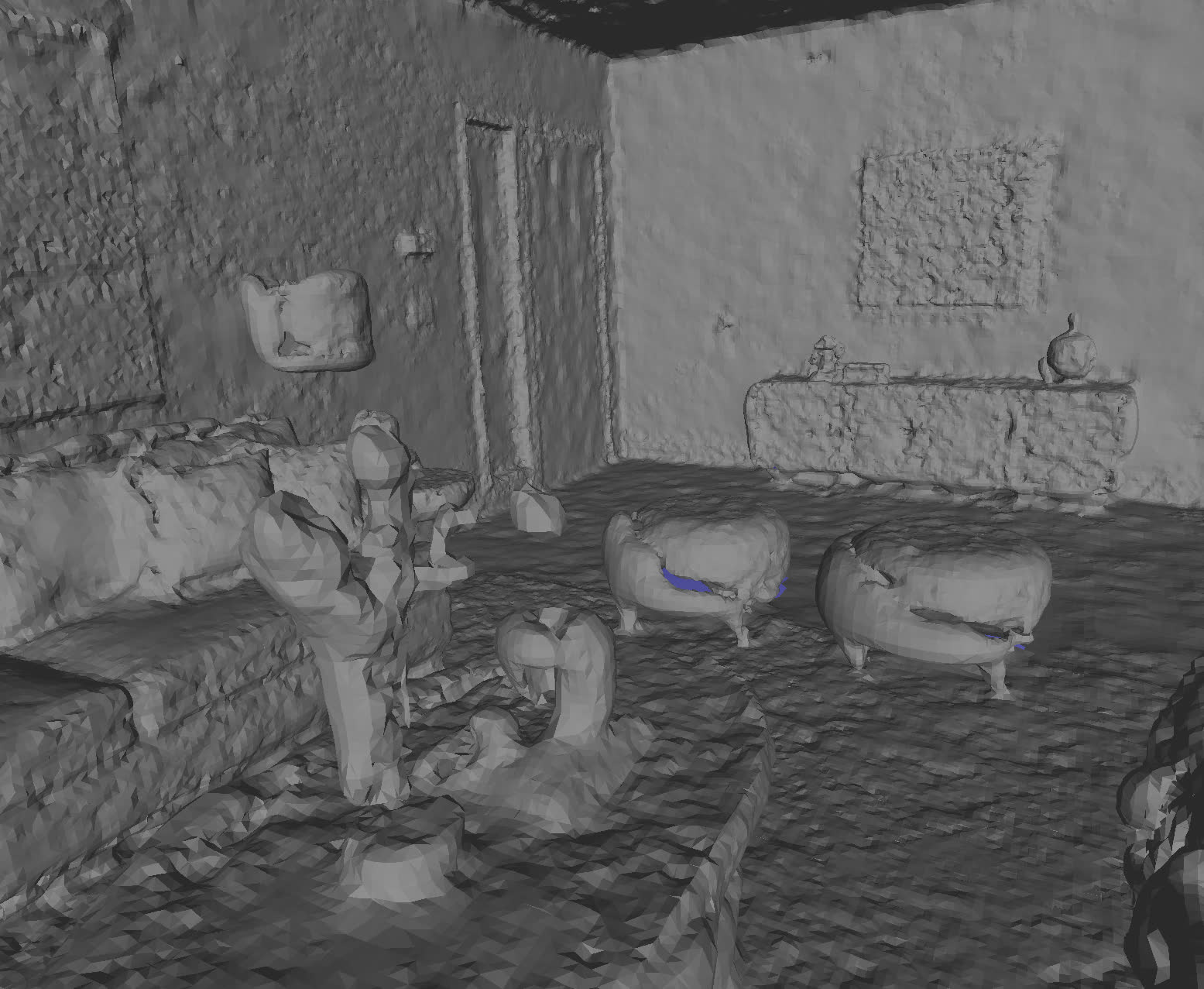}};
                    \node[font=\tiny, text=white] at (0.2,-1.1) {};
                \end{tikzpicture}
            \end{subtable}
            \hspace{-3pt}
            \begin{subtable}[b]{0.19\textwidth}
                \centering
                \subcaption*{\texttt{Density 0.1}}
                \begin{tikzpicture}
                    \node [inner sep=0pt,clip,rounded corners=2pt] at (0,0) {\includegraphics[width=1\textwidth]{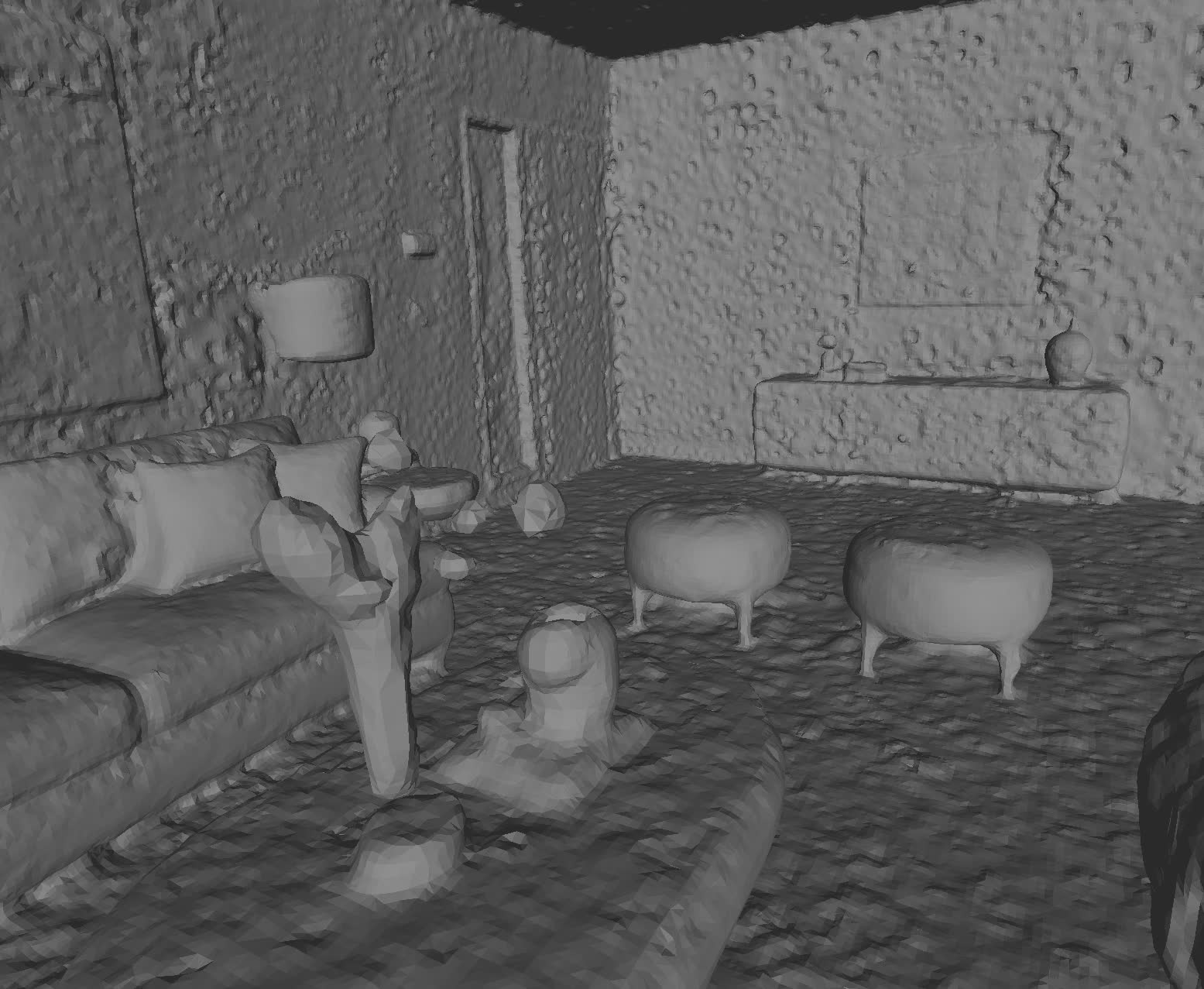}};
                    \node[font=\tiny, text=white] at (0.2,-1.1) {};
                \end{tikzpicture}
            \end{subtable}
            \begin{subtable}[b]{0.19\textwidth}
                \centering
                \subcaption*{\texttt{Density 0.5}}
                \begin{tikzpicture}
                    \node [inner sep=0pt,clip,rounded corners=2pt] at (0,0) {\includegraphics[width=1\textwidth]{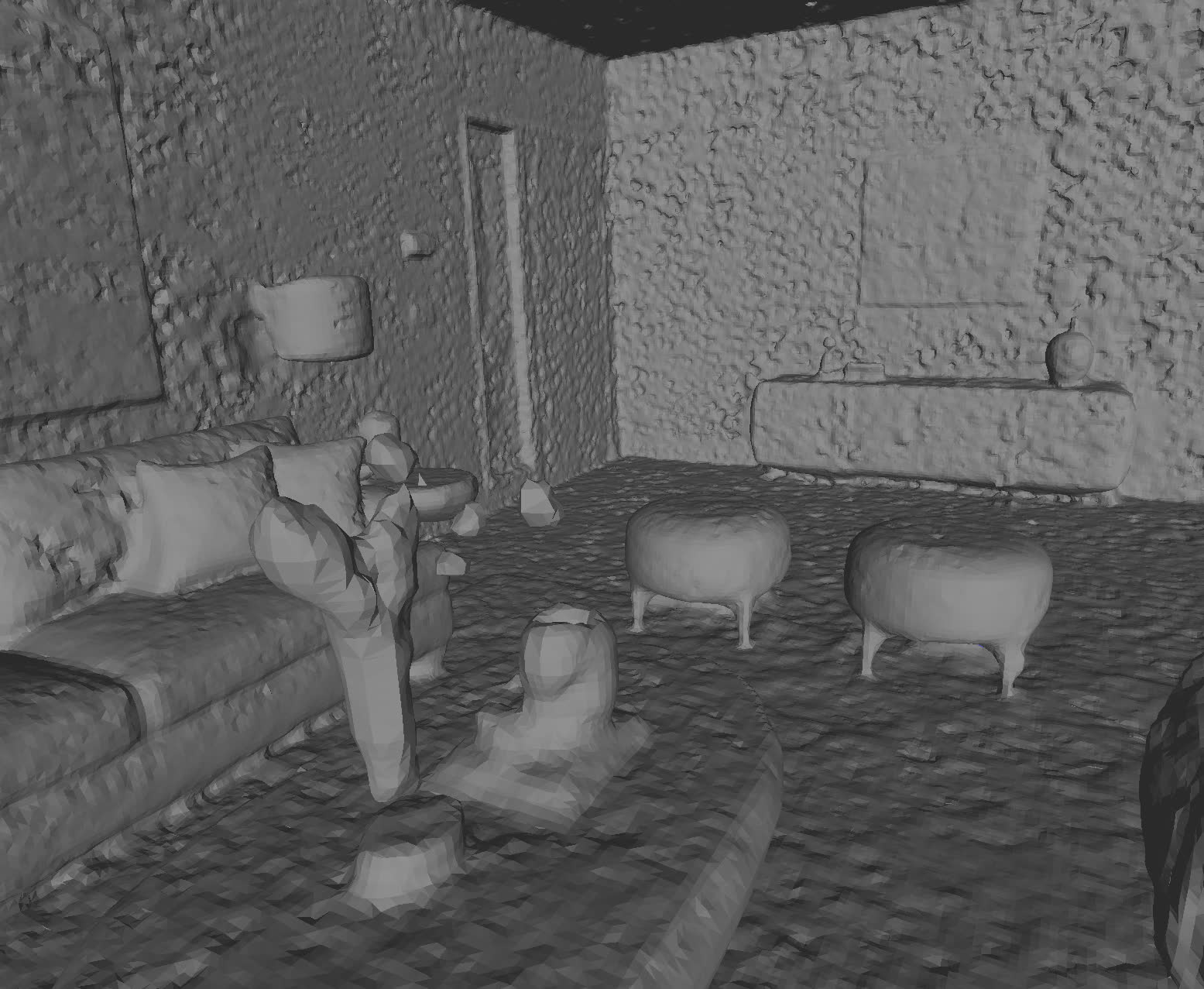}};
                    \node[font=\tiny, text=white] at (0.2,-1.1) {};
                \end{tikzpicture}
            \end{subtable}
            \hspace{-3pt}
            \begin{subtable}[b]{0.19\textwidth}
                \centering
                \subcaption*{\texttt{Ours}}
                \begin{tikzpicture}
                    \node [inner sep=0pt,clip,rounded corners=2pt] at (0,0) {\includegraphics[width=1\textwidth]{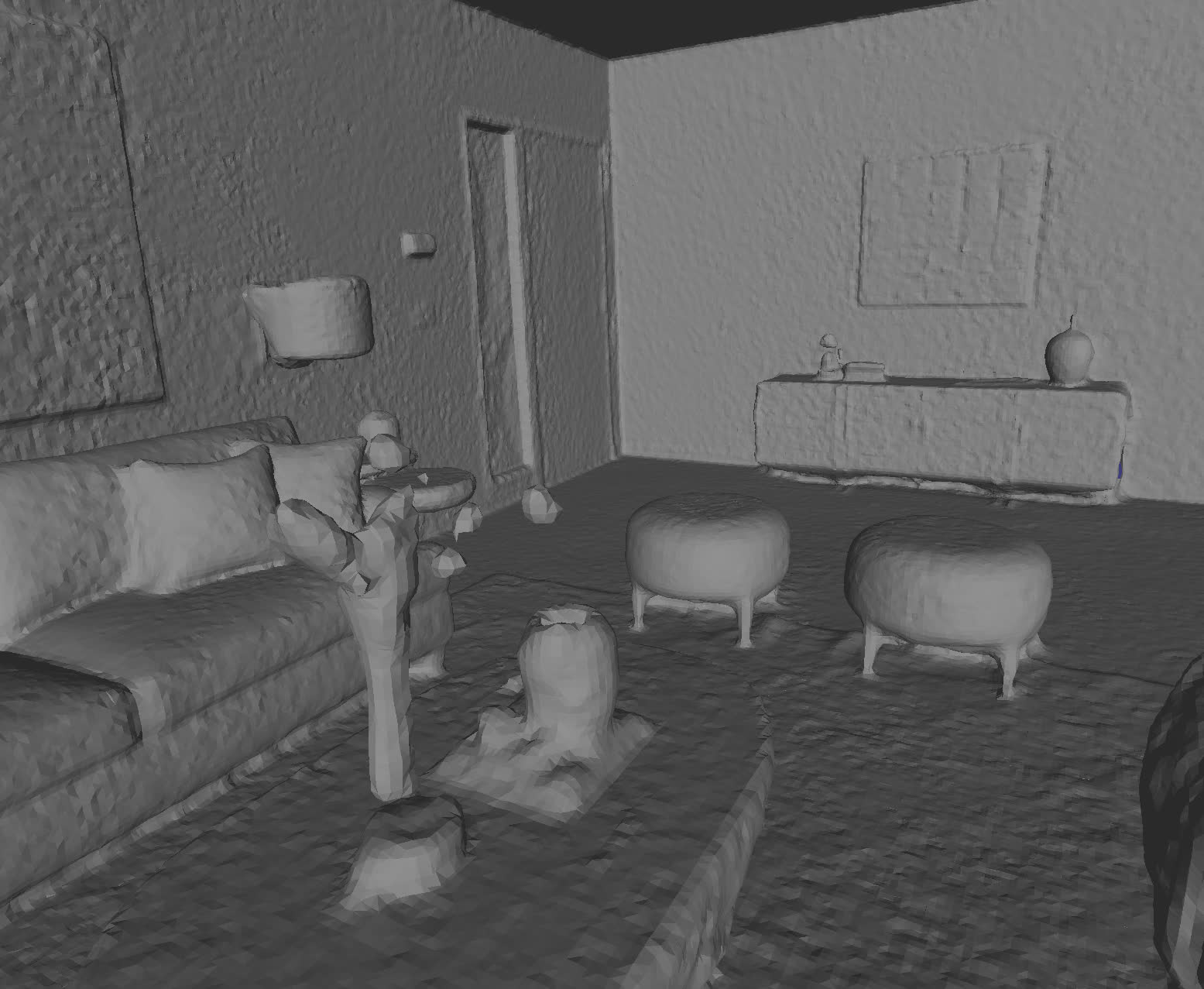}};
                    \node[font=\tiny, text=white] at (0.2,-1.1) {};
                \end{tikzpicture}
            \end{subtable}
            \hspace{-3pt}
            \begin{subtable}[b]{0.19\textwidth}
                \centering
                \subcaption*{\texttt{GT}}
                \begin{tikzpicture}
                    \node [inner sep=0pt,clip,rounded corners=2pt] at (0,0) {\includegraphics[width=1\textwidth]{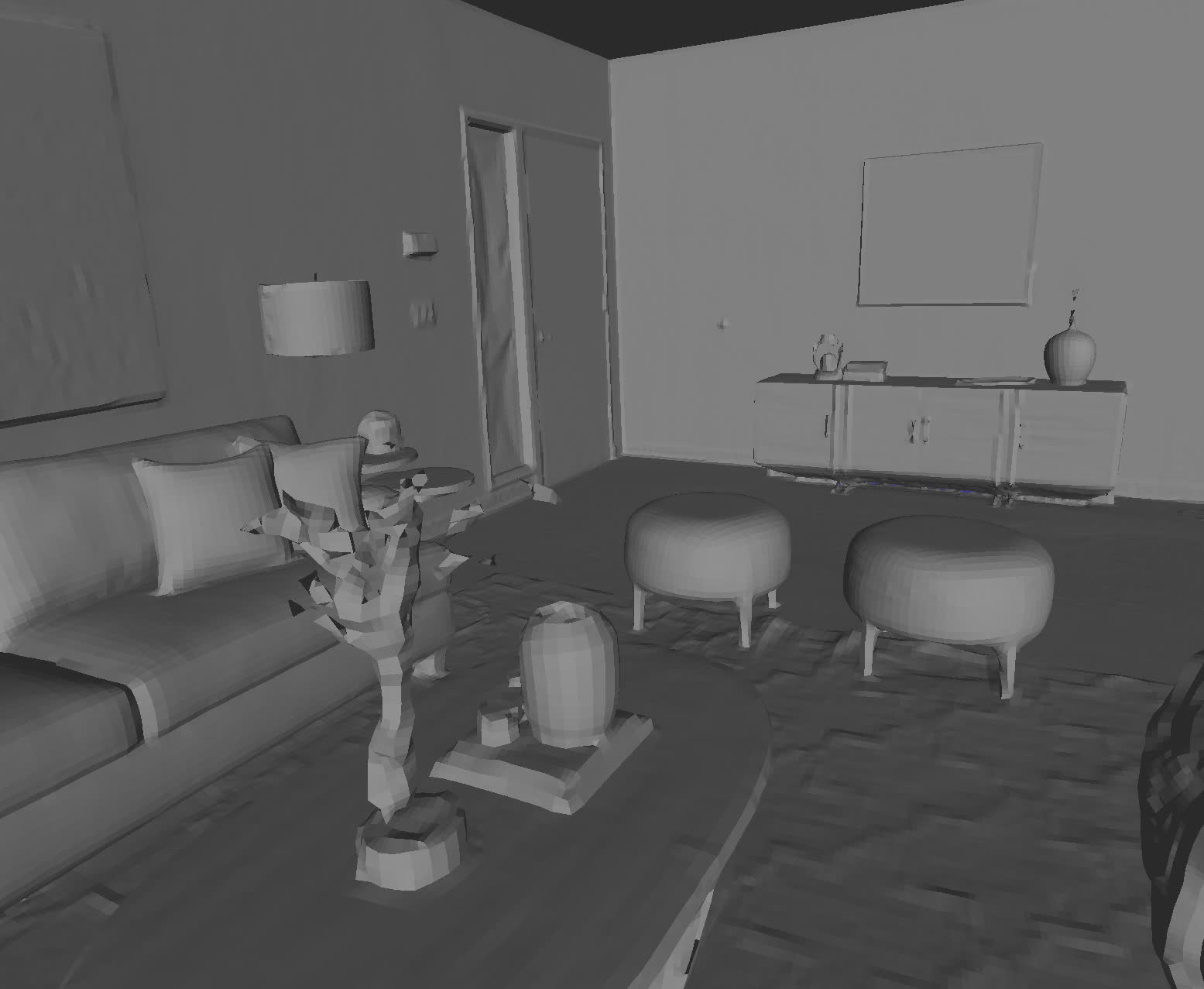}};
                    \node[font=\tiny, text=white] at (0.2,-1.1) {};
                \end{tikzpicture}
            \end{subtable}
            \vspace{-1em}
        \end{tabular}
    	\end{minipage}
    	\\
            \centering
    	\begin{minipage}{0.5\textwidth}
                \scriptsize
                \centering
                \setlength{\tabcolsep}{5pt}
                \renewcommand*{\arraystretch}{1}
                \resizebox{\textwidth}{!}{%
                \begin{tabular}{lccccc}
                \toprule
                & Acc. $\downarrow$ & Comp. $\downarrow$ & C-$L_1$ $\downarrow$ & NC $\uparrow$ & F-score $\uparrow$ \\
                \midrule
                Gaussians & .0206 & .0412 & .0309 & .9091 & .9117 \\
                SuGaR~\cite{guedon2023sugar}: density 0.1 & .0130 & .0357 & .0243 & .9301 & .9275 \\
                SuGaR~\cite{guedon2023sugar}: density 0.5 & .0083 & \textbf{.0304} & \textbf{.0193} & .9309 & \textbf{.9325} \\
                \rowcolor{gray!10} Back-projection (ours) & \textbf{.0074} & .0312 & .0194 & \textbf{.9428} & .9310 \\
                \bottomrule
                \end{tabular}
                }
            \end{minipage}
    \end{tabular}
    \caption{\textbf{Ablation of Poisson mesh extraction techniques: Replica.} We compare naive Gaussian-based meshing, the meshing strategy proposed in SuGaR~\cite{guedon2023sugar}, and our back-projection approach. All models were trained using the proposed depth and normal objectives.}\label{tab:meshing-ablation-replica}
    \vspace{-1em}
\end{table}

\begin{table*}[ht]
  \centering\scriptsize
  \setlength{\tabcolsep}{2.5pt}
  \renewcommand*{\arraystretch}{1}
  \vspace*{-1em}
  \begin{tabular}{lcccccc|ccccc}
    \toprule
    & \multicolumn{1}{c}{} & \multicolumn{5}{c|}{(a) Test within a sequence} & \multicolumn{5}{c}{(b) Test with a different sequence}\\
    & Sensor Depth & Abs Rel $\downarrow$ & Sq Rel $\downarrow$ & RMSE $\downarrow$ & RMSE log $\downarrow$ & $\delta<1.25$ $\uparrow$ & Abs Rel $\downarrow$ & Sq Rel $\downarrow$ & RMSE $\downarrow$ & RMSE log $\downarrow$ & $\delta<12.5$ $\uparrow$ \\ \midrule
    Nerfacto~\cite{nerfstudio} & $\textcolor{red}{-}$ & 14.72 & 19.79 & 61.05 & 13.26 & 88.25 & 14.52 & 18.32 & 63.85 & 13.13 & 88.41 \\
    Depth-Nerfacto~\cite{nerfstudio} & \textcolor{teal}{$\checkmark$} & 13.90 & 11.71 & 50.21 & 12.98 & 88.46 & 13.49 & 10.76 & 51.63 & 12.62 & 89.23 \\
    MonoSDF~\cite{Yu2022MonoSDF} & \textcolor{teal}{$\checkmark$} & 10.90 & 9.87 & 48.74 & 11.27 & 83.48 & 11.00 & 10.98 & 50.92 & 11.37 & 82.62 \\
    Splatfacto (no cues)~\cite{kerbl20233d} & $\textcolor{red}{-}$ & 8.32 & 5.45 & 38.47 & 10.23 & 89.75 & 8.06 & 5.39 & 38.61 & 10.05 & 90.51\\ \midrule
    Splatfacto + $\mathcal{L}_{\hat{D}}$ (Ours) & \textcolor{teal}{$\checkmark$} & \underline{3.71} & \underline{3.08} & \underline{30.80} & \underline{4.27} & \underline{95.52} & \underline{3.78} & \underline{3.08} & \underline{31.35} & \underline{4.26} & \underline{95.47}\\
    \rowcolor{gray!10} Splatfacto + $\mathcal{L}_{\hat{D}}$ + $\mathcal{L}_{\hat{N}}$ (Ours) & \textcolor{teal}{$\checkmark$} & \textbf{3.64} & \textbf{3.02} & \textbf{30.33} & \textbf{4.17} & \textbf{95.60} & \textbf{3.69} & \textbf{2.97} & \textbf{30.57} & \textbf{4.15} & \textbf{95.64} \\
    \bottomrule
    \end{tabular}
    \caption{\textbf{Depth evaluation metrics compared to ground truth Faro scanner data} for the MuSHRoom dataset. Instead of evaluating on noisy captured iPhone depth maps for evaluation, we rely on more accurate depth maps reconstructed from a Faro lidar scanner. We show that our depth regularization strategy, utilizing low-resolution iPhone depths, greatly outperforms other baselines. Results are averaged over 10 scenes.}
  \label{tab:supp_depth_losses_faro}
\end{table*}

\begin{table*}[t!]
    \begin{tabular}{c}
    \begin{minipage}{1\textwidth}
        \centering \footnotesize
        \setlength{\tabcolsep}{3.4pt}
        \renewcommand*{\arraystretch}{1}
        \vspace{-1em}
        \subcaption{We load every 3/5/8/12 views from the whole training sequence (around 260). Results are evaluated on "Courtroom" from Tanks \& Temples.}
        \begin{tabular}{lccc|ccc|ccc|ccc}
            \toprule
            \multirow{2}{*}{Methods} & \multicolumn{3}{c|}{load every 3} & \multicolumn{3}{c|}{load every 5} & \multicolumn{3}{c|}{load every 8} & \multicolumn{3}{c}{load every 12} \\ 
            & PSNR $\uparrow$ & SSIM $\uparrow$ & LPIPS $\downarrow$ & PSNR $\uparrow$ & SSIM $\uparrow$ & LPIPS $\downarrow$ & PSNR $\uparrow$ & SSIM $\uparrow$ & LPIPS $\downarrow$ & PSNR $\uparrow$ & SSIM $\uparrow$ & LPIPS $\downarrow$ \\ \midrule
            Splatfacto & 20.68 & .7445 & .1921 & 18.50 & .6991 & .2110 & 16.86 & .6459 & .2474 & 14.76 & .5580 & .3332 \\
            Ours + Zoe-Depth~\cite{bhat2023zoedepth} & 20.88 & .7518 & .1833 & 19.58 & .7118 & .2007 & \textbf{17.60} & \textbf{.6568} & \textbf{.2433} & 15.90 & .5835 & .2971 \\
            Ours + DepthAnything~\cite{depthanything} & \textbf{20.91} & \textbf{.7528} & \textbf{.1830} & \textbf{19.60} & \textbf{.7153} & \textbf{.1997} & 17.44 & \textbf{.6568} & .2456 & \textbf{16.24} & \textbf{.5902} & \textbf{.2924} \\ 
            \bottomrule
        \end{tabular}
    \end{minipage}
    \\
    \begin{minipage}{1\textwidth}
        \centering\footnotesize
        \setlength{\tabcolsep}{3.4pt}
        \renewcommand*{\arraystretch}{1}
        \vspace*{1em}
        \subcaption{We load every 5/8/12/20 views from the whole training sequence (around 270). Results are evaluated on "8b5caf3398" from ScanNet++ DSLR sequence.}
        \begin{tabular}{lccc|ccc|ccc|ccc}
            \toprule
            \multirow{2}{*}{Methods} & \multicolumn{3}{c|}{load every 5} & \multicolumn{3}{c|}{load every 8} & \multicolumn{3}{c|}{load every 12} & \multicolumn{3}{c}{load every 20} \\ 
            & PSNR $\uparrow$ & SSIM $\uparrow$ & LPIPS $\downarrow$ & PSNR $\uparrow$ & SSIM $\uparrow$ & LPIPS $\downarrow$ & PSNR $\uparrow$ & SSIM $\uparrow$ & LPIPS $\downarrow$ & PSNR $\uparrow$ & SSIM $\uparrow$ & LPIPS $\downarrow$ \\ \midrule
            Splatfacto & 24.68 & .8810 & .1169 & 22.81 & .8568 & .1559 & 21.08 & .8357 & .1816 & 18.90 & .8059 & .2375 \\
            Ours + Zoe-Depth~\cite{bhat2023zoedepth} & \textbf{24.72} & .8821 & \textbf{.1163} & 23.04 & .8591 & .1521 & \textbf{21.81} & \textbf{.8415} & .1755 & 19.10 & .8059 & .2332 \\
            Ours + DepthAnything~\cite{depthanything} & 24.66 & \textbf{.8826} & .1194 & \textbf{23.21} & \textbf{.8595} & \textbf{.1507} & 21.76 & .8406 & \textbf{.1751} & \textbf{19.51} & \textbf{.8101} & \textbf{.2321} \\ 
            \bottomrule
        \end{tabular}
    \end{minipage}
\end{tabular}
    \caption{\textbf{Comparison of DN-Splatter performance with monocular depth supervision}. We ablate the Zoe-Depth\cite{bhat2023zoedepth} and DepthAnything\cite{depthanything} monocular estimators with sparse views on the "Courtroom" sequence of Tanks \& Temples advanced dataset. Monocular depth supervision aids in novel-view synthesis under sparse settings.}
    \vspace*{-1em}
    \label{tab:supp_sparse_mono}
\end{table*}
\subsection{Depth estimation compared to Faro scanner ground truth} In~\tabref{tab:supp_depth_losses_faro}, we show the depth evaluation performance of our proposed regularization scheme on the MuSHRoom dataset, evaluated against ground truth Faro lidar scanner data instead of the low-resolution iPhone depths. This corresponds to Table 3 from the main paper, which compares depth metrics on iPhone depth captures for the same scenes and baselines. When comparing to laser scanner depths, our method still out performs other baseline methods on depth estimation.
\subsection{Additional depth comparisons}
We consider the performance of DN-Splatter within sparse view setting guided by only monocular depth estimates. We test on the large scale Tanks \& Temples scene in~\tabref{tab:supp_sparse_mono}. We consider training with dense and sparse captures and conclude that although monocular depth supervision in dense captures provides minimal improvements, the increase in novel view synthesis metrics under sparse view settings is notable.
Lastly, in~\tabref{tab:supp_ap_depth_losses} we compare the performance of various depth losses described in~\secref{sec:depth-losses} on depth estimation and novel view synthesis. There are several interesting observations. First, the logarithmic depth loss $L_\text{LogL1}$ outperforms other popular variants like $L_\text{L1}$ or $L_\text{MSE}$ on depth and RGB synthesis. Second, the gradient-aware logarithmic depth variant $L_{\hat{D}}$ outperforms the simpler variant, validating our assumption that captured sensor depths, like those from iPhone cameras, tend to contain noise and inaccuracies at edges or sharp boundaries. Therefore, the gradient-aware variant mitigates these inaccurate sensor readings.
\begin{table*}[t!]
    \centering
    \begin{tabular}{c}
    \begin{minipage}{\textwidth}
        \centering\footnotesize
        \setlength{\tabcolsep}{10pt}
        \renewcommand*{\arraystretch}{1}
        \vspace*{0em}
        \subcaption{Test split obtained by sampling uniformly every 10 frames within the training sequence.}
        \begin{tabular}{lccccc|ccc}
        \toprule
         & \multicolumn{5}{c|}{Depth estimation} & \multicolumn{3}{c}{Novel view synthesis} \\
         & Abs Rel $\downarrow$ & Sq Rel $\downarrow$ & RMSE $\downarrow$ & RMSE log $\downarrow$ & $\delta<1.25$ $\uparrow$ & PSNR $\uparrow$ & SSIM $\uparrow$ & LPIPS $\downarrow$ \\ \midrule
        $\mathcal{L}_{\text{MSE}}$ & .0587 & .0229 & .2313 & .0618 & .9534 & 22.32 & .7995 & .1653 \\
        $\mathcal{L}_{\mathrm{1}}$ & .0419 & .0233 & .2286 & .0435 & .9629 & 22.46 & .8041 & .1594 \\
        $\mathcal{L}_{\text{DSSIML1}}$ & .0476 & .0331 & .2773 & .0523 & .9476 & 21.77 & .7802 & .1879 \\
        $\mathcal{L}_{\text{LogL1}}$ & .0430 & .0267 & .2414 & .0444 & .9609 & 22.48 & \textbf{.8053} & \textbf{.1580} \\
        $\mathcal{L}_{\text{HuberL1}}$ & .0536 & .0239 & .2335 & .0561 & .9579 & 22.39 & .8017 & .1625 \\ 
        $\mathcal{L}_{\text{EAS}}$ & .0954 & .0572 & .3581 & .1103 & .8726 & 22.18 & .7951 & .1780 \\
        \rowcolor{gray!10} $\mathcal{L}_{\hat{D}}$ (Ours)  & \textbf{.0338} & \textbf{.0212} & \textbf{.2170} & \textbf{.0350} & \textbf{.9691} & \textbf{22.49} & .8031 & .1630 \\
        \bottomrule
        \end{tabular}
\end{minipage}
\\
\begin{minipage}{\textwidth}
        \centering\footnotesize
        \setlength{\tabcolsep}{10pt}
        \renewcommand*{\arraystretch}{1}
        \vspace*{1em}
        \subcaption{Test split obtained from a different camera trajectory with no overlap with the training sequence.}
        \begin{tabular}{lccccc|ccc}
        \toprule
         & \multicolumn{5}{c|}{Depth estimation} & \multicolumn{3}{c}{Novel view synthesis} \\ 
        & Abs Rel $\downarrow$ & Sq Rel $\downarrow$ & RMSE $\downarrow$ & RMSE log $\downarrow$ & $\delta<1.25$ $\uparrow$ & PSNR $\uparrow$ & SSIM $\uparrow$ & LPIPS $\downarrow$ \\ \midrule
        $\mathcal{L}_{\text{MSE}}$ & .0572 & .0282 & .2506 & .0570 & .9585 & 19.37 & .7088 & .2329 \\
        $\mathcal{L}_{\mathrm{1}}$ & .0449 & \textbf{.0248} & .2364 & .0449 & \textbf{.9639} & 19.45 & .7164 & .2253 \\
        $\mathcal{L}_{\text{DSSIML1}}$ & .0482 & .0330 & .2775 & .0527 & .9495 & 18.98 & .7040 & .2430 \\
        $\mathcal{L}_{\text{LogL1}}$ & .0451 & .0269 & .2454 & .0453 & .9629 & 19.50 & .7183 & \textbf{.2228} \\
        $\mathcal{L}_{\text{HuberL1}}$ & .0526 & .0267 & .2483 & .0533 & .9617 & 19.45 & .7128 & .2285 \\ 
        $\mathcal{L}_{\text{EAS}}$ & .0724 & .0442 & .3142 & .0819 & .9329 & 19.30 & .7108 & .2351 \\
        \rowcolor{gray!10} $\mathcal{L}_{\hat{D}}$ (Ours) & \textbf{.0427} & .0252 & \textbf{.2335} & \textbf{.0420} & .9632 & \textbf{19.53} & \textbf{.7187} & .2286 \\
        \bottomrule
        \end{tabular}
    \end{minipage}
\end{tabular}
    \caption{\textbf{Ablation on depth losses} on the MuSHRoom dataset. We consider various depth losses as defined in \cref{sec:depth-losses} and their impact on depth estimation and novel view synthesis. We achieve the best performance with our proposed edge-aware $\mathcal{L}_{\hat{D}}$ loss.}
    \label{tab:supp_ap_depth_losses}
    \vspace{-1em}
\end{table*}
\section{Additional qualitative results}
\label{supp_sec:qualitative}

\subsection{Normal smoothing loss}
\begin{figure}[t]
\centering
\begin{subfigure}[b]{0.15\textwidth}
    \centering
    \subcaption*{RGB}
    \begin{tikzpicture}
        \node [inner sep=0pt,clip,rounded corners=2pt] at (0,0) {\includegraphics[height=2.5cm]{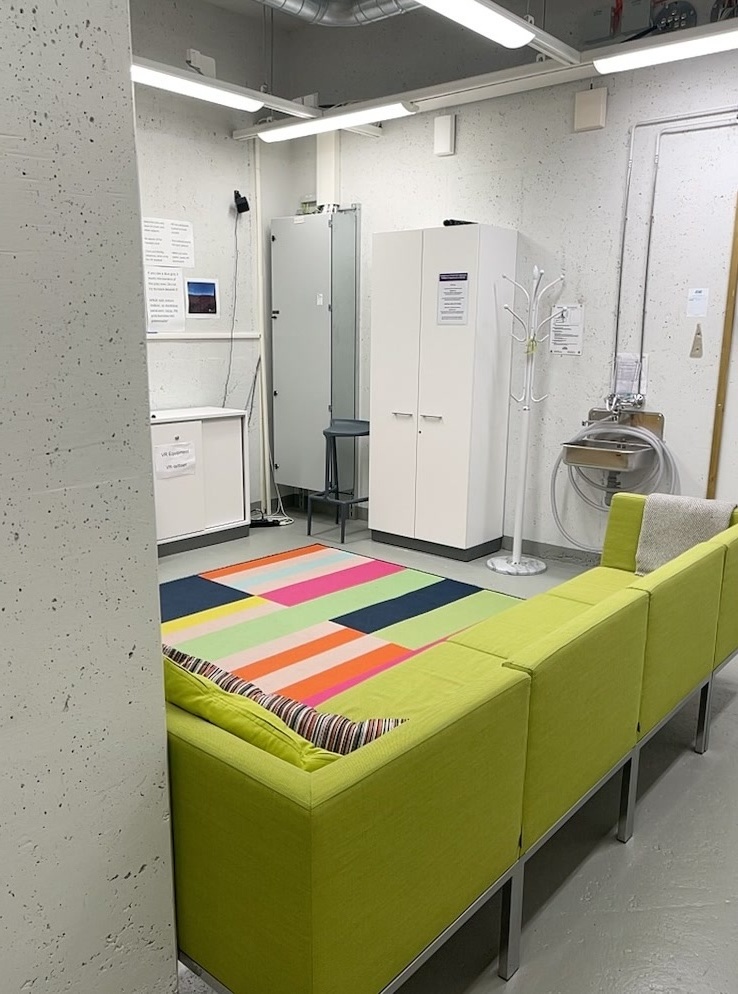}};
        \node[font=\tiny, text=white] at (0.2,-1.1) {};
    \end{tikzpicture}
\end{subfigure}
\hspace{-25pt}
\begin{subfigure}[b]{0.15\textwidth}
    \centering
    \subcaption*{w/o $\mathcal{L}_{\text{smooth}}$}
    \begin{tikzpicture}
        \node [inner sep=0pt,clip,rounded corners=2pt] at (0,0) {\includegraphics[height=2.5cm]{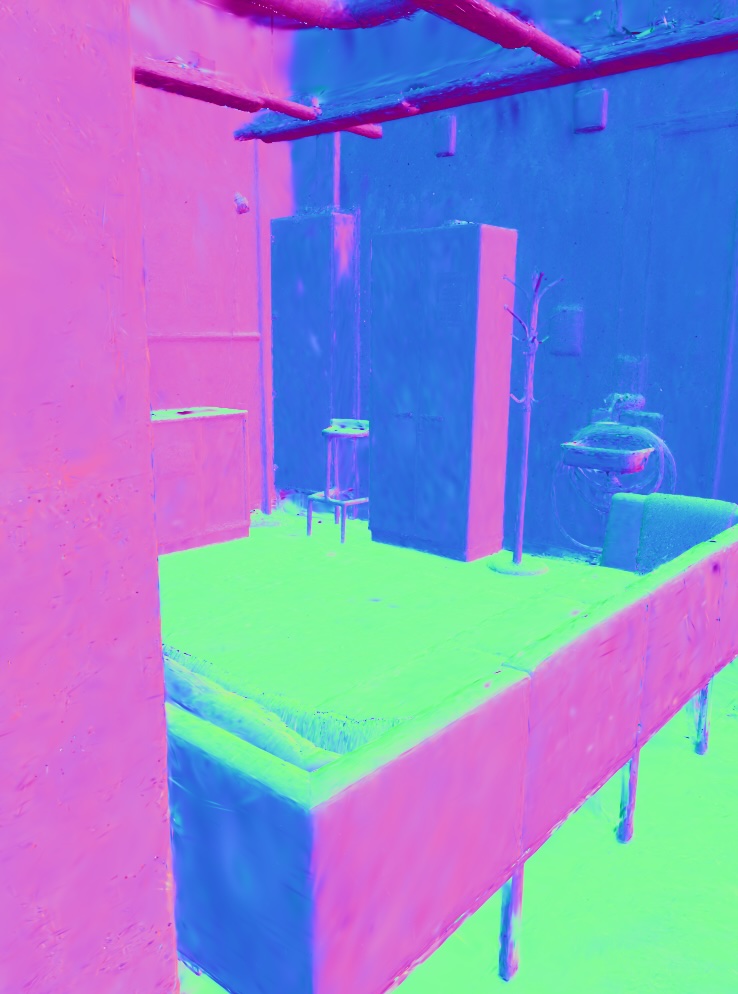}};
        \node[font=\tiny, text=white] at (0.2,-1.1) {};
    \end{tikzpicture}
\end{subfigure}
\hspace{-25pt}
\begin{subfigure}[b]{0.15\textwidth}
    \centering
    \subcaption*{w/ $\mathcal{L}_{\text{smooth}}$}
    \begin{tikzpicture}
        \node [inner sep=0pt,clip,rounded corners=2pt] at (0,0) {\includegraphics[height=2.5cm]{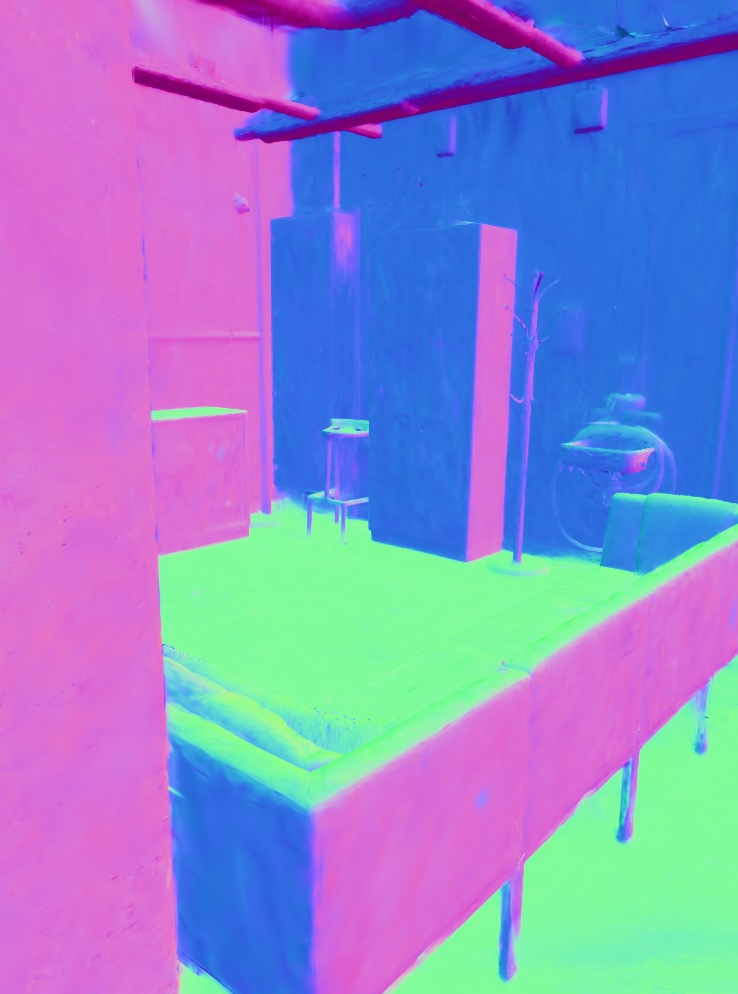}};
        \node[font=\tiny, text=white] at (0.2,-1.1) {};
    \end{tikzpicture}
\end{subfigure}
\hspace{-25pt}
\begin{subfigure}[b]{0.15\textwidth}
    \centering
    \subcaption*{Omnidata GT}
    \begin{tikzpicture}
        \node [inner sep=0pt,clip,rounded corners=2pt] at (0,0) {\includegraphics[height=2.5cm]{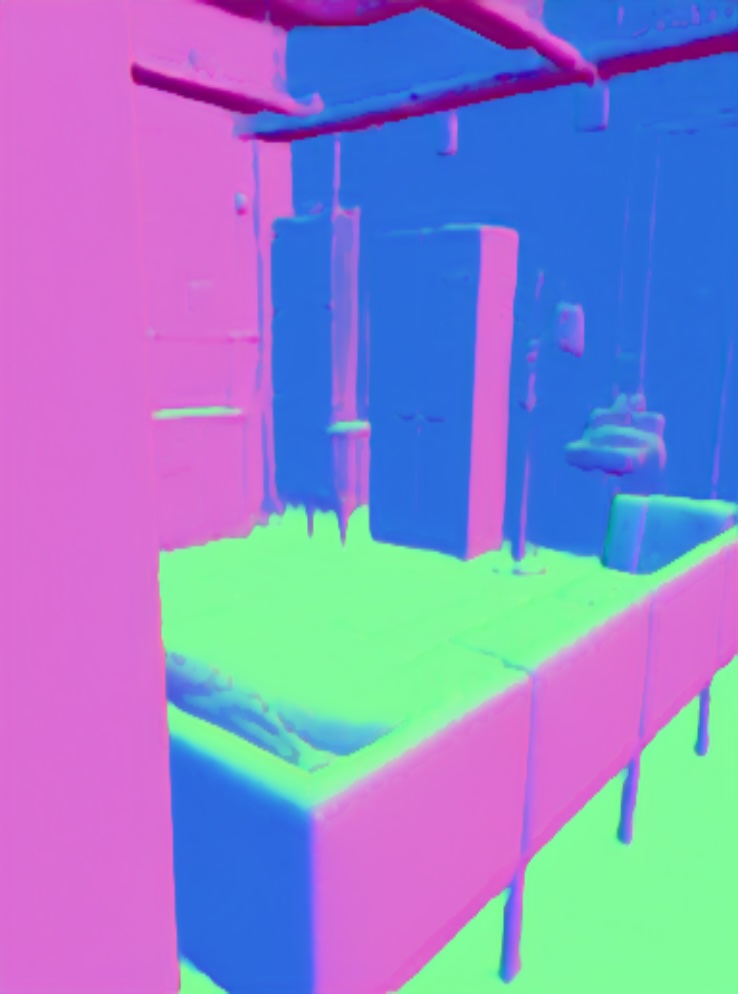}};
        \node[font=\tiny, text=white] at (0.2,-1.1) {};
    \end{tikzpicture}
\end{subfigure}

\begin{subfigure}[b]{0.15\textwidth}
    \centering
    \begin{tikzpicture}
        \node [inner sep=0pt,clip,rounded corners=2pt] at (0,0) {\includegraphics[height=2.5cm]{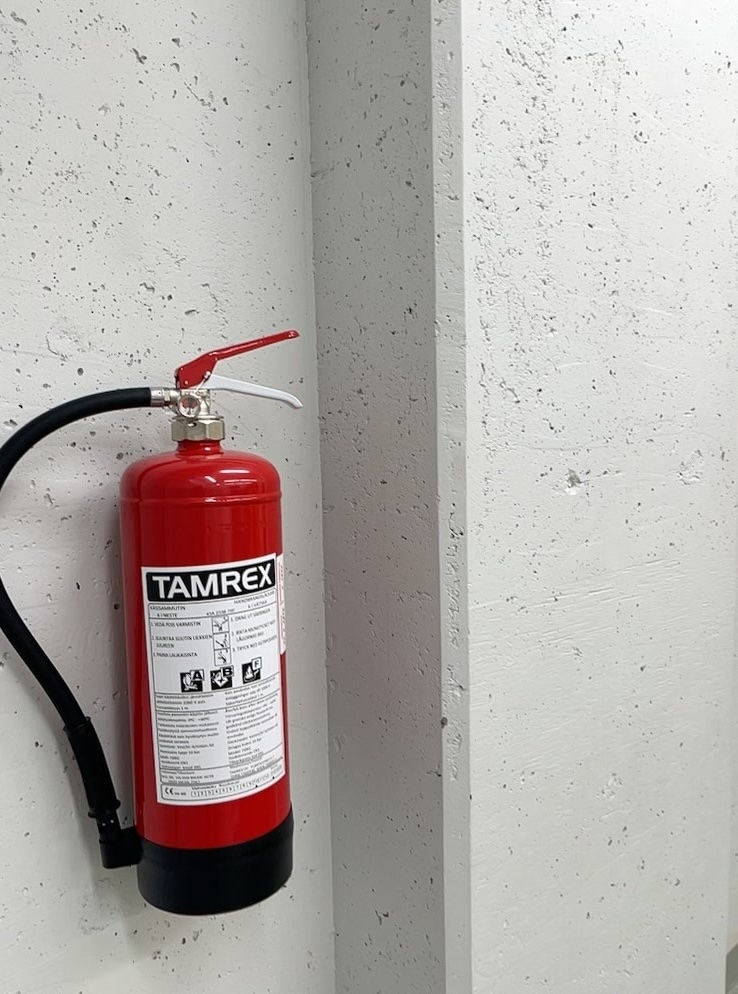}};
        \node[font=\tiny, text=white] at (0.2,-1.1) {};
    \end{tikzpicture}
\end{subfigure}
\hspace{-25pt}
\begin{subfigure}[b]{0.15\textwidth}
    \centering
    \begin{tikzpicture}
        \node [inner sep=0pt,clip,rounded corners=2pt] at (0,0) {\includegraphics[height=2.5cm]{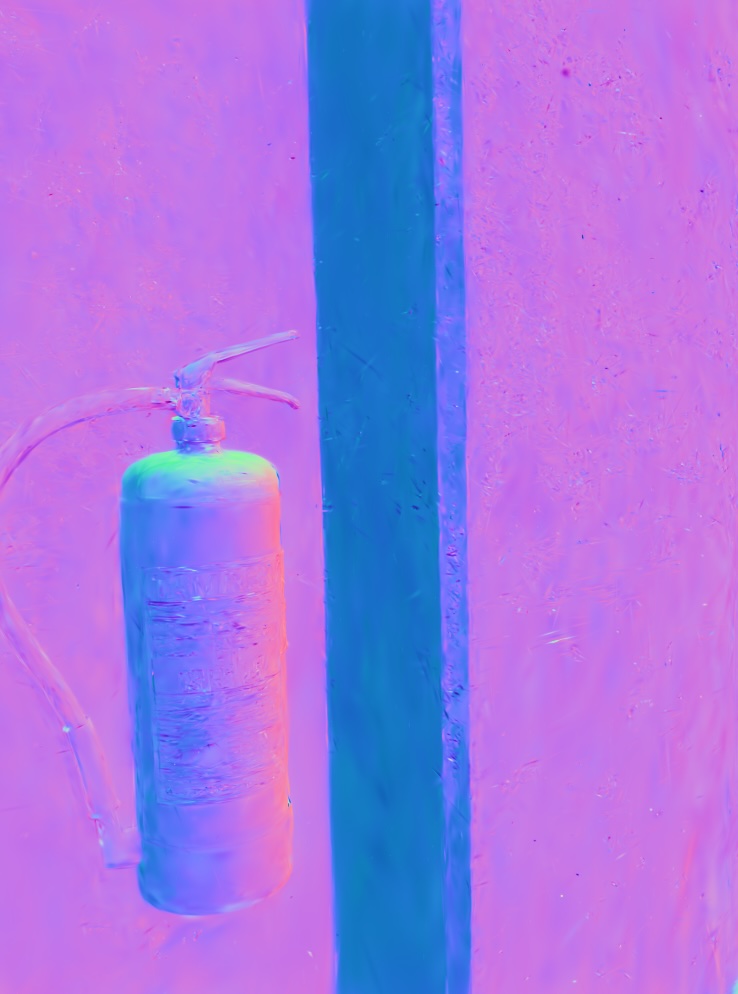}};
        \node[font=\tiny, text=white] at (0.2,-1.1) {};
    \end{tikzpicture}
\end{subfigure}
\hspace{-25pt}
\begin{subfigure}[b]{0.15\textwidth}
    \centering
    \begin{tikzpicture}
        \node [inner sep=0pt,clip,rounded corners=2pt] at (0,0) {\includegraphics[height=2.5cm]{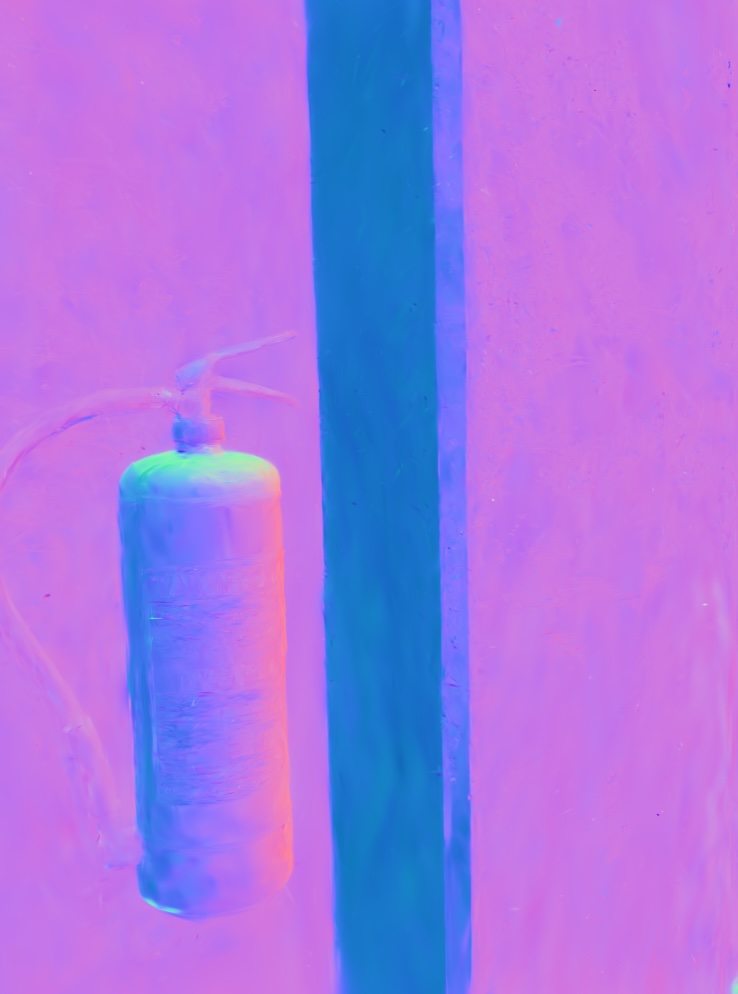}};
        \node[font=\tiny, text=white] at (0.2,-1.1) {};
    \end{tikzpicture}
\end{subfigure}
\hspace{-25pt}
\begin{subfigure}[b]{0.15\textwidth}
    \centering
    \begin{tikzpicture}
        \node [inner sep=0pt,clip,rounded corners=2pt] at (0,0) {\includegraphics[height=2.5cm]{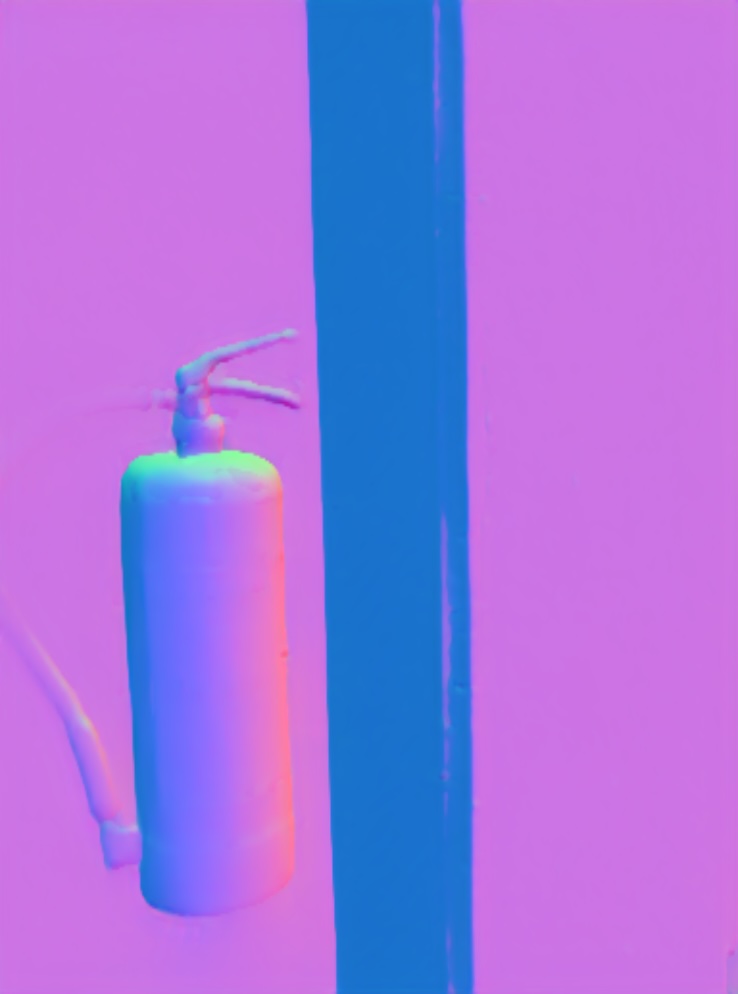}};
        \node[font=\tiny, text=white] at (0.2,-1.1) {};
    \end{tikzpicture}
\end{subfigure}

\begin{subfigure}[b]{0.15\textwidth}
    \centering
    \begin{tikzpicture}
        \node [inner sep=0pt,clip,rounded corners=2pt] at (0,0) {\includegraphics[height=2.5cm]{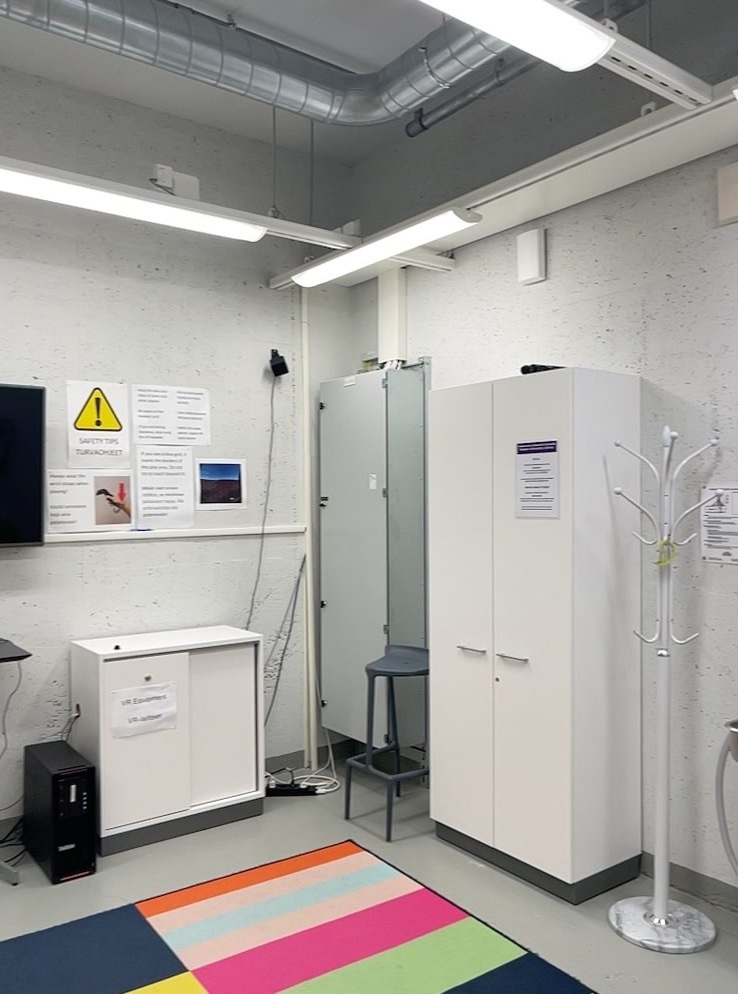}};
        \node[font=\tiny, text=white] at (0.2,-1.1) {};
    \end{tikzpicture}
\end{subfigure}
\hspace{-25pt}
\begin{subfigure}[b]{0.15\textwidth}
    \centering
    \begin{tikzpicture}
        \node [inner sep=0pt,clip,rounded corners=2pt] at (0,0) {\includegraphics[height=2.5cm]{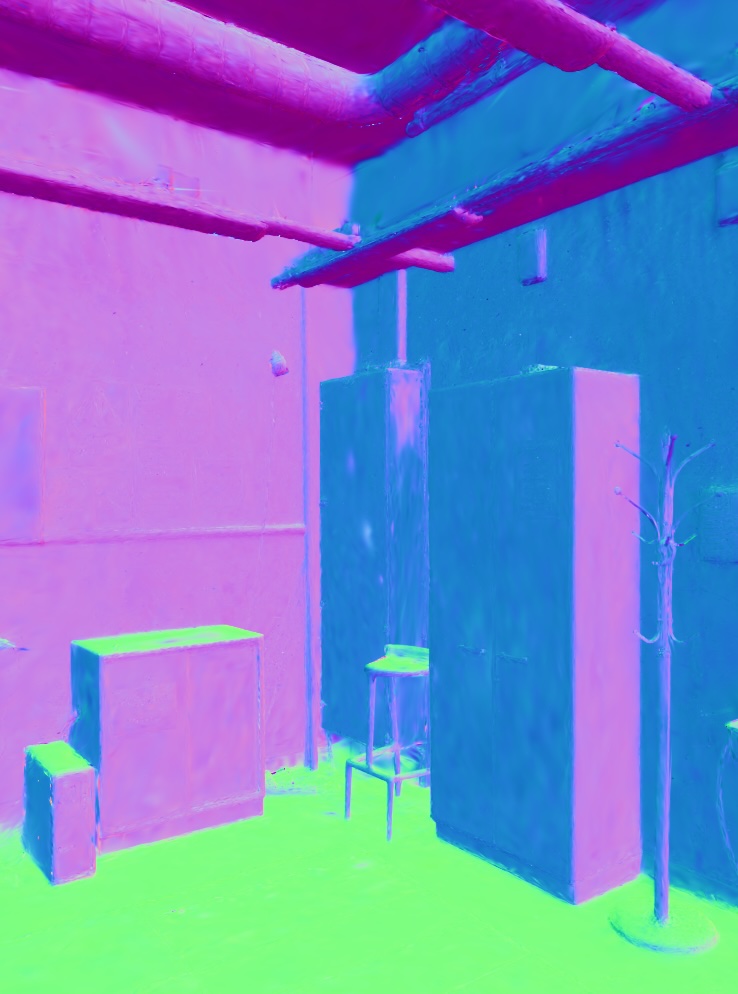}};
        \node[font=\tiny, text=white] at (0.2,-1.1) {};
    \end{tikzpicture}
\end{subfigure}
\hspace{-25pt}
 \begin{subfigure}[b]{0.15\textwidth}
    \centering
    \begin{tikzpicture}
        \node [inner sep=0pt,clip,rounded corners=2pt] at (0,0) {\includegraphics[height=2.5cm]{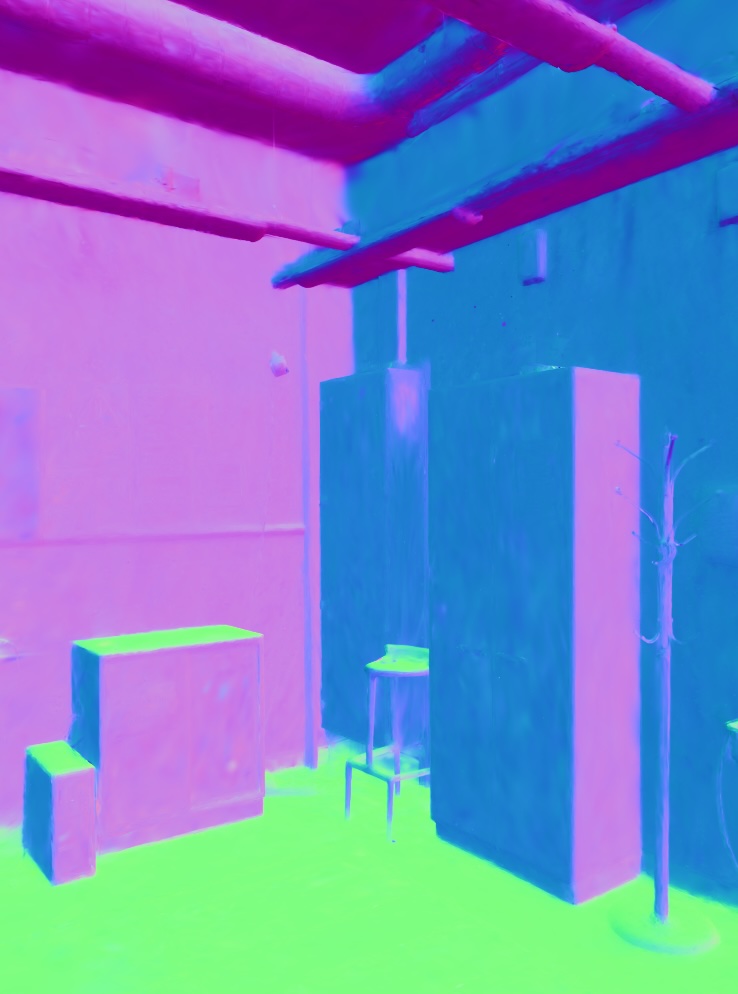}};
        \node[font=\tiny, text=white] at (0.2,-1.1) {};
    \end{tikzpicture}
\end{subfigure}
\hspace{-25pt}
\begin{subfigure}[b]{0.15\textwidth}
    \centering
    \begin{tikzpicture}
        \node [inner sep=0pt,clip,rounded corners=2pt] at (0,0) {\includegraphics[height=2.5cm]{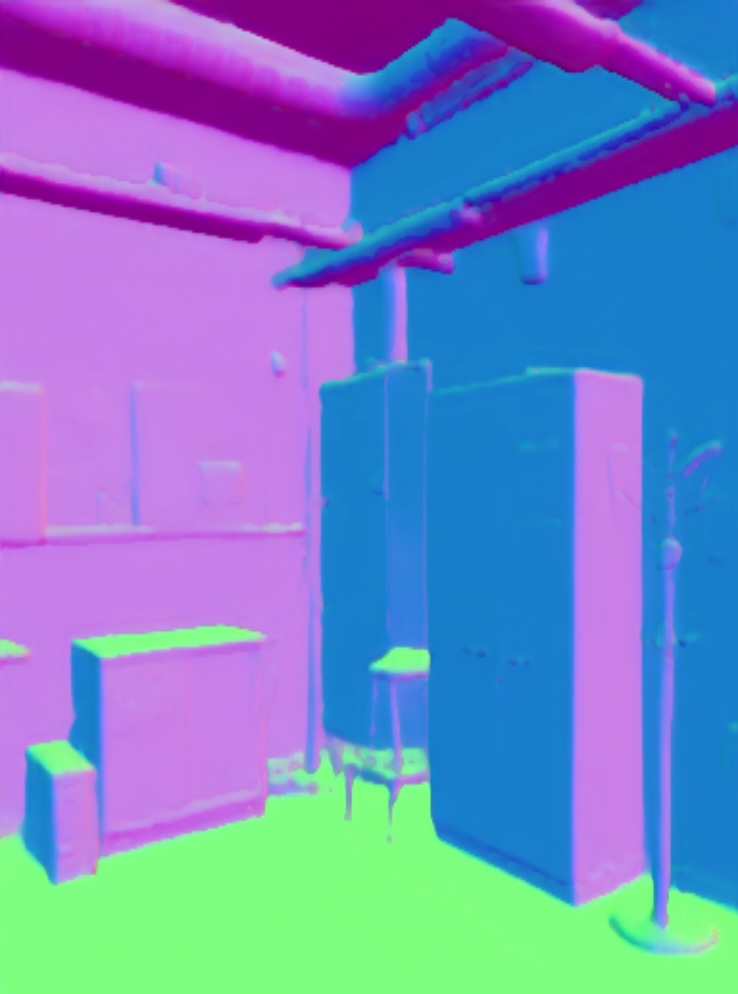}};
        \node[font=\tiny, text=white] at (0.2,-1.1) {};
    \end{tikzpicture}
\end{subfigure}
\caption{\textbf{Qualitative comparison of normal smoothing prior.} We visualize normal estimates with and without our $\mathcal{L}_{\text{smooth}}$ smoothing prior on the 'VR Room' scene from MuSHRoom dataset.}
\vspace{-1em}
\label{fig:normal_smoothing_imgs}
\end{figure}
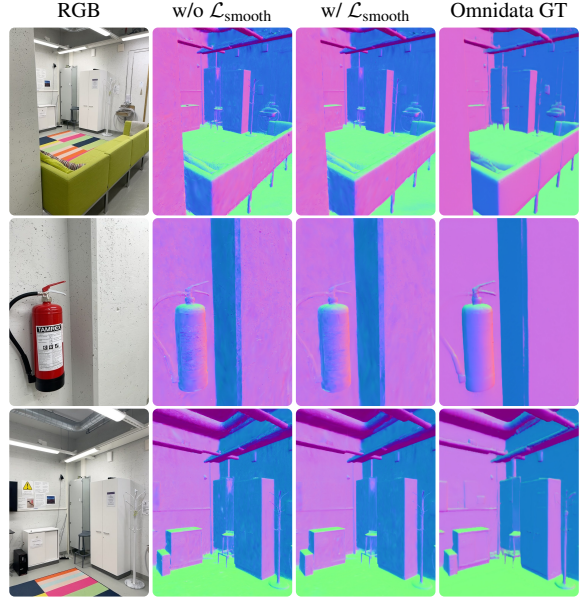

We visualize the impact of $\mathcal{L}_{\text{smooth}}$ prior on rendered normal estimates in \cref{fig:normal_smoothing_imgs}. We achieve smoother predictions with the prior.

\subsection{2DGS vs. DN-Splatter renders}
\begin{figure*}[t!]
\centering
\begin{subfigure}[b]{0.2\textwidth}
    \centering
    \subcaption*{iPhone RGB}
    \begin{tikzpicture}
        \node [inner sep=0pt,clip,rounded corners=2pt] at (0,0) {\includegraphics[height=3.5cm]{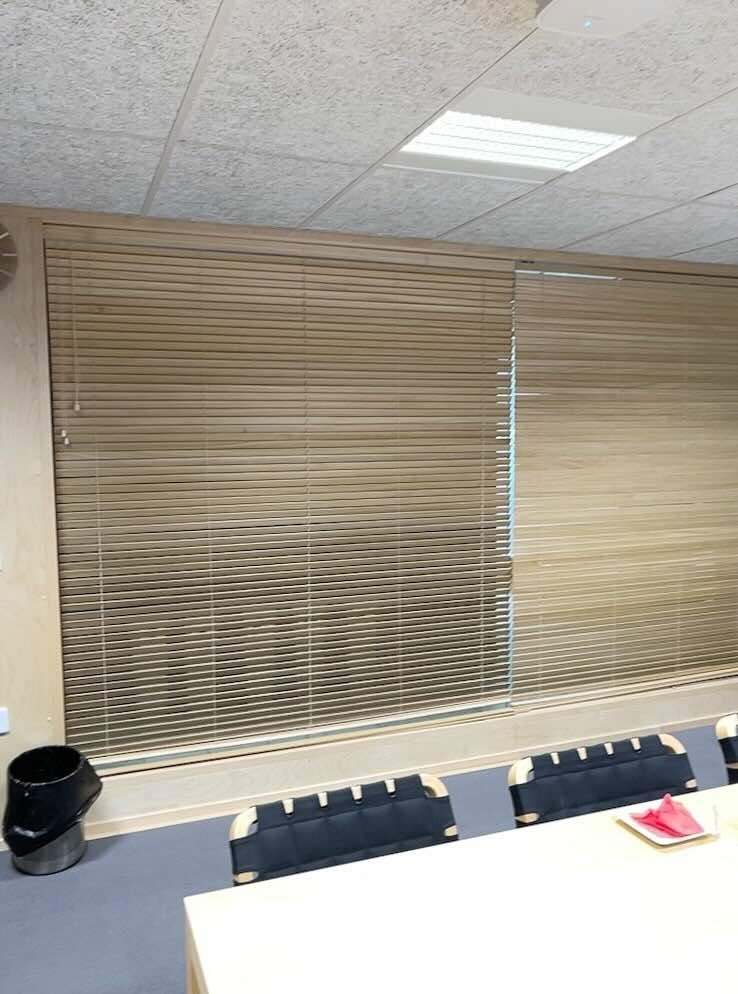}};
        \node[font=\tiny, text=white] at (0.2,-1.1) {};
    \end{tikzpicture}
\end{subfigure}
\hspace{-25pt}
\begin{subfigure}[b]{0.2\textwidth}
    \centering
    \subcaption*{Splatfacto}
    \begin{tikzpicture}
        \node [inner sep=0pt,clip,rounded corners=2pt] at (0,0) {\includegraphics[height=3.5cm]{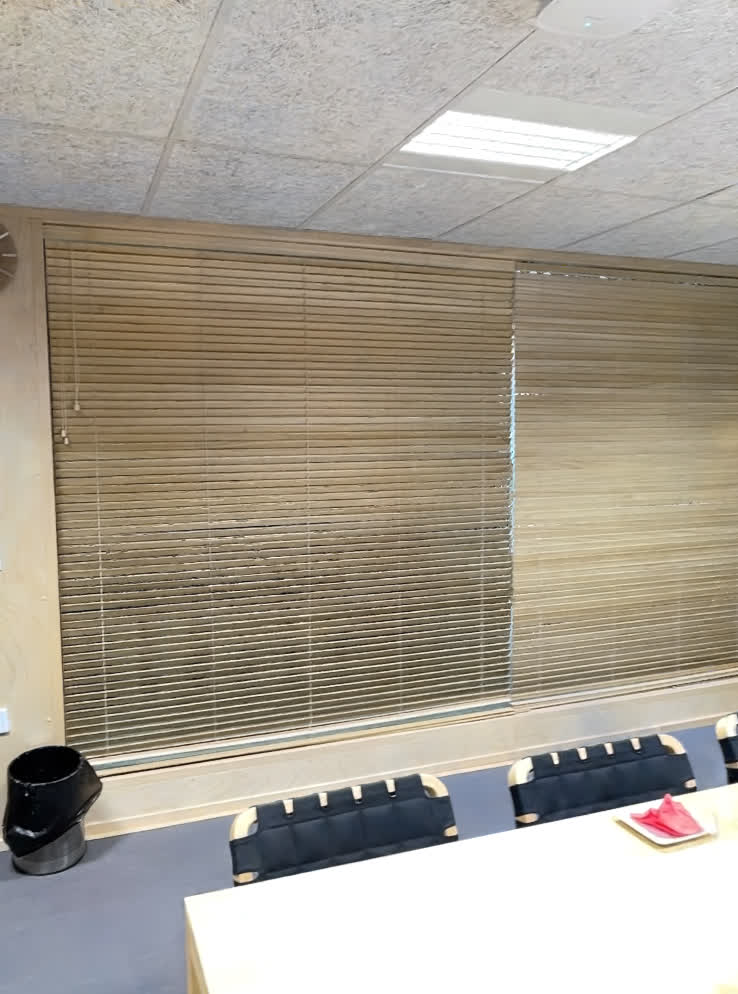}};
        \node[font=\tiny, text=white] at (0.2,-1.1) {};
    \end{tikzpicture}
\end{subfigure}
\hspace{-25pt}
\begin{subfigure}[b]{0.2\textwidth}
    \centering
    \subcaption*{2DGS}
    \begin{tikzpicture}
        \node [inner sep=0pt,clip,rounded corners=2pt] at (0,0) {\includegraphics[height=3.5cm]{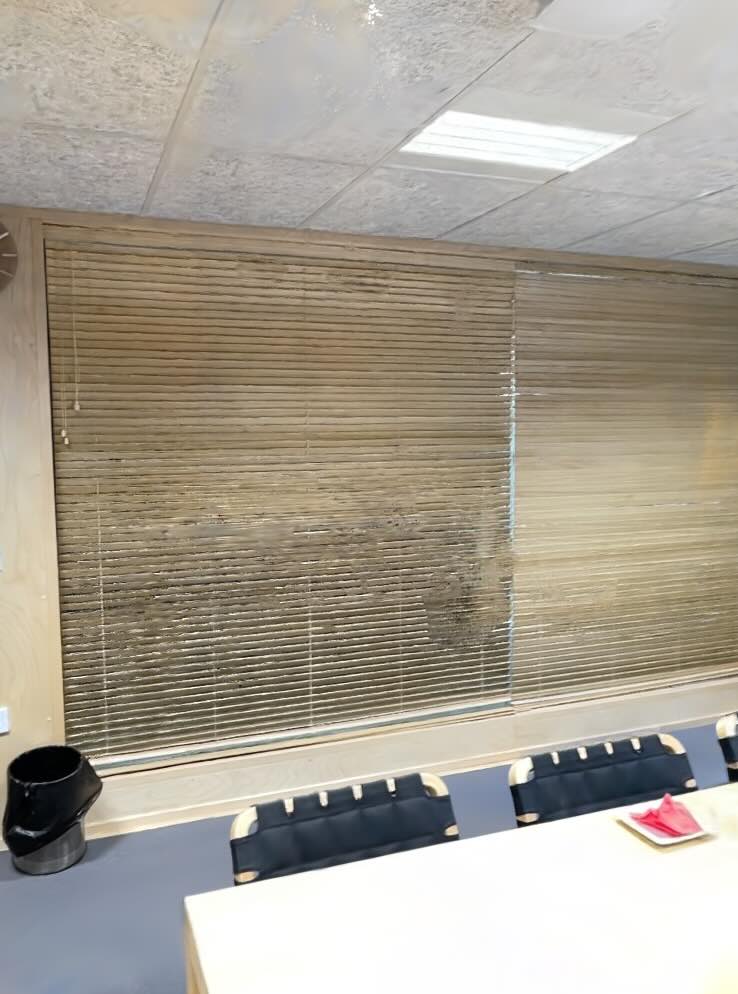}};
        \node[font=\tiny, text=white] at (0.2,-1.1) {};
    \end{tikzpicture}
\end{subfigure}
\hspace{-25pt}
\begin{subfigure}[b]{0.2\textwidth}
    \centering
    \subcaption*{2DGS w/ Sensor Depth}
    \begin{tikzpicture}
        \node [inner sep=0pt,clip,rounded corners=2pt] at (0,0) {\includegraphics[height=3.5cm]{figures/supplement/2dgs_vs_dn/2dgs_long_frame_00203.jpg}};
        \node[font=\tiny, text=white] at (0.2,-1.1) {};
    \end{tikzpicture}
\end{subfigure}
\hspace{-25pt}
\begin{subfigure}[b]{0.2\textwidth}
    \centering
    \subcaption*{Ours w/ Sensor Depth}
    \begin{tikzpicture}
        \node [inner sep=0pt,clip,rounded corners=2pt] at (0,0) {\includegraphics[height=3.5cm]{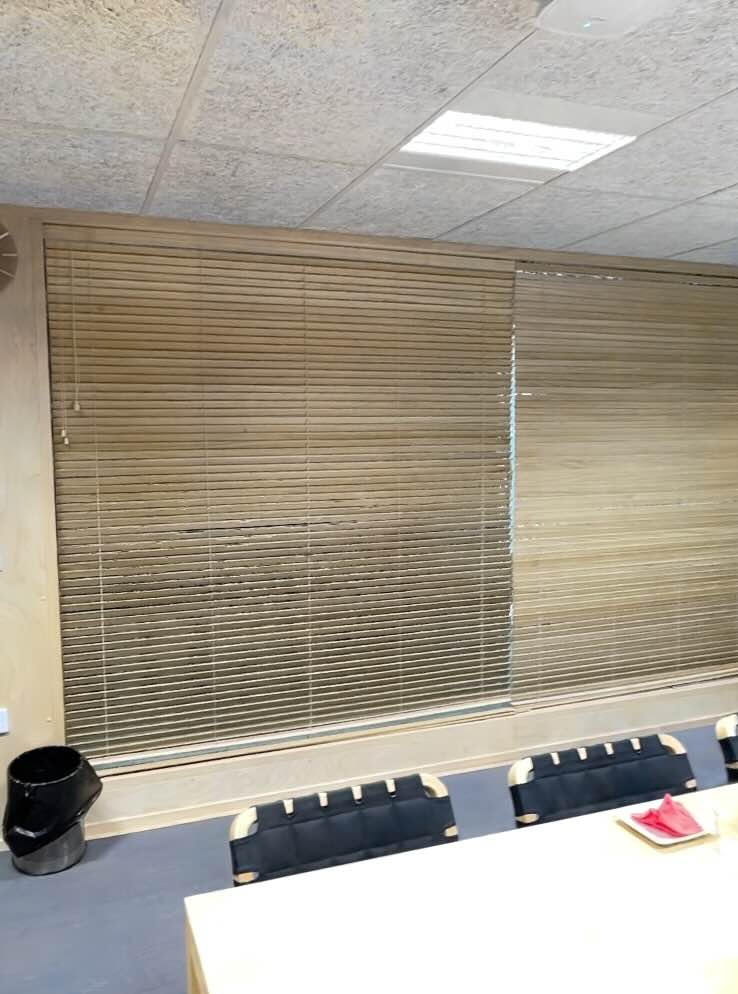}};
        \node[font=\tiny, text=white] at (0.2,-1.1) {};
    \end{tikzpicture}
\end{subfigure}
\\
\begin{subfigure}[b]{0.2\textwidth}
    \centering
    \subcaption*{iPhone Depth}
    \begin{tikzpicture}
        \node [inner sep=0pt,clip,rounded corners=2pt] at (0,0) {\includegraphics[height=3.5cm]{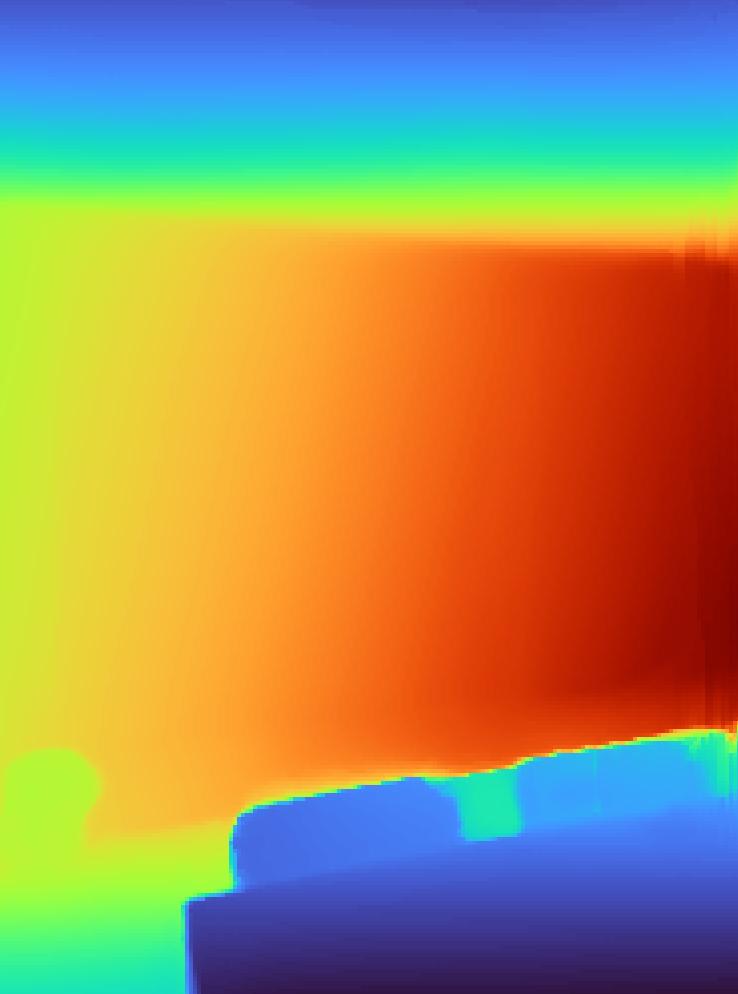}};
        \node[font=\tiny, text=white] at (0.2,-1.1) {};
    \end{tikzpicture}
\end{subfigure}
\hspace{-25pt}
\begin{subfigure}[b]{0.2\textwidth}
    \centering
    \subcaption*{Splatfacto}
    \begin{tikzpicture}
        \node [inner sep=0pt,clip,rounded corners=2pt] at (0,0) {\includegraphics[height=3.5cm]{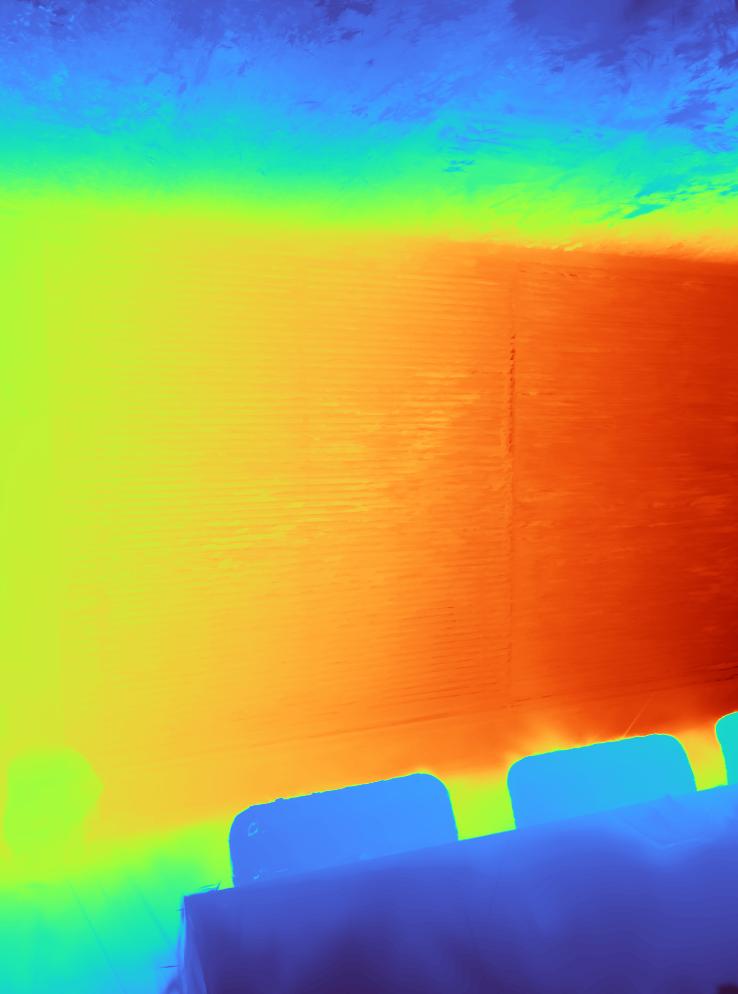}};
        \node[font=\tiny, text=white] at (0.2,-1.1) {};
    \end{tikzpicture}
\end{subfigure}
\hspace{-25pt}
\begin{subfigure}[b]{0.2\textwidth}
    \centering
    \subcaption*{2DGS}
    \begin{tikzpicture}
        \node [inner sep=0pt,clip,rounded corners=2pt] at (0,0) {\includegraphics[height=3.5cm]{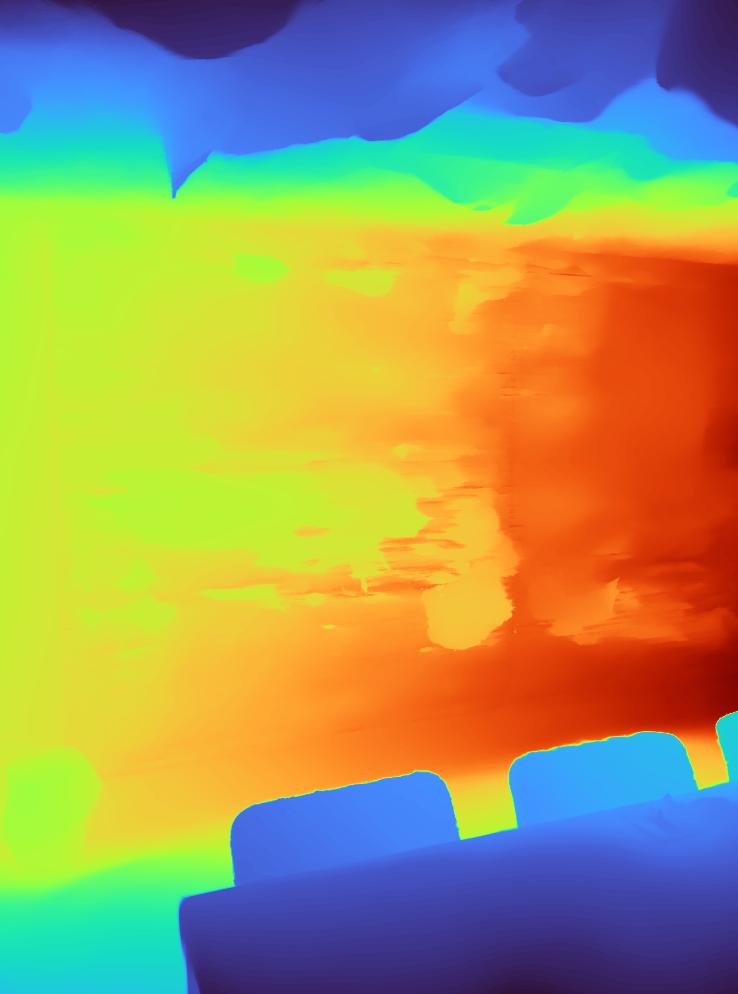}};
        \node[font=\tiny, text=white] at (0.2,-1.1) {};
    \end{tikzpicture}
\end{subfigure}
\hspace{-25pt}
\begin{subfigure}[b]{0.2\textwidth}
    \centering
    \subcaption*{2DGS w/ Sensor Depth}
    \begin{tikzpicture}
        \node [inner sep=0pt,clip,rounded corners=2pt] at (0,0) {\includegraphics[height=3.5cm]{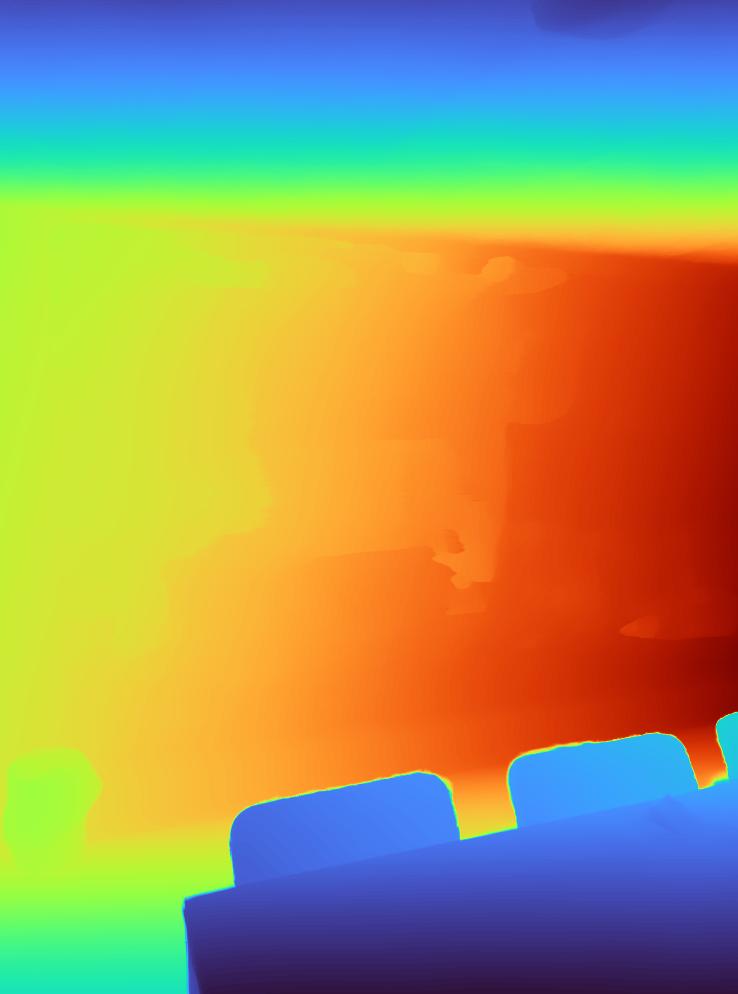}};
        \node[font=\tiny, text=white] at (0.2,-1.1) {};
    \end{tikzpicture}
\end{subfigure}
\hspace{-25pt}
\begin{subfigure}[b]{0.2\textwidth}
    \centering
    \subcaption*{Ours w/ Sensor Depth}
    \begin{tikzpicture}
        \node [inner sep=0pt,clip,rounded corners=2pt] at (0,0) {\includegraphics[height=3.5cm]{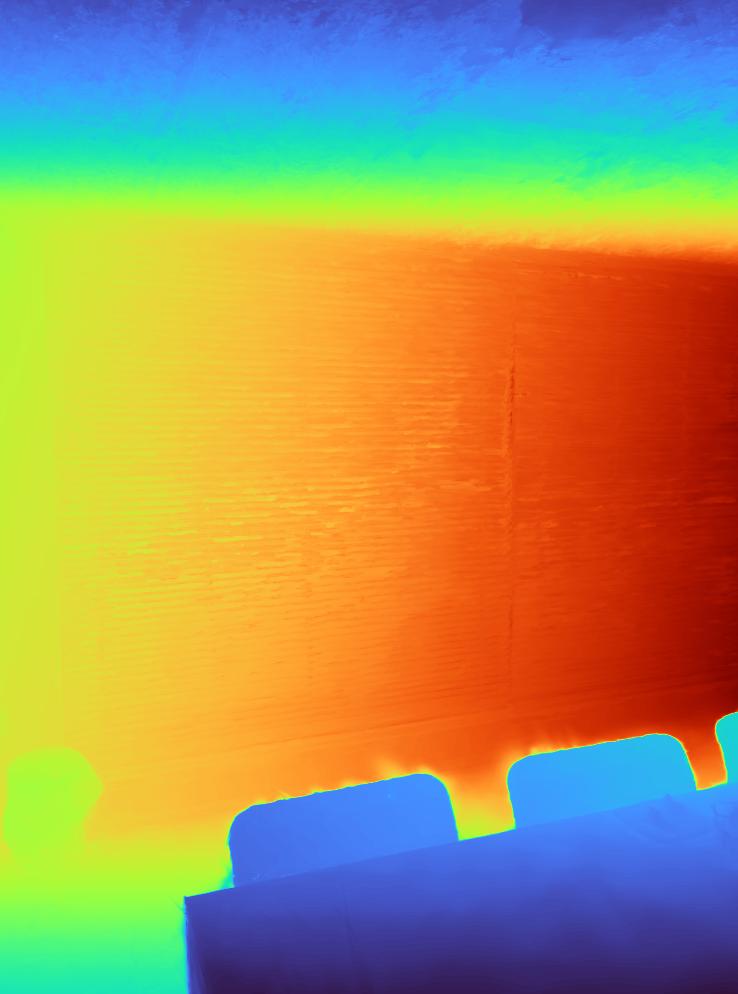}};
        \node[font=\tiny, text=white] at (0.2,-1.1) {};
    \end{tikzpicture}
\end{subfigure}
\hspace{-25pt}
\\
\begin{subfigure}[b]{0.2\textwidth}
    \centering
    \subcaption*{iPhone RGB}
    \begin{tikzpicture}
        \node [inner sep=0pt,clip,rounded corners=2pt] at (0,0) {\includegraphics[height=3.5cm]{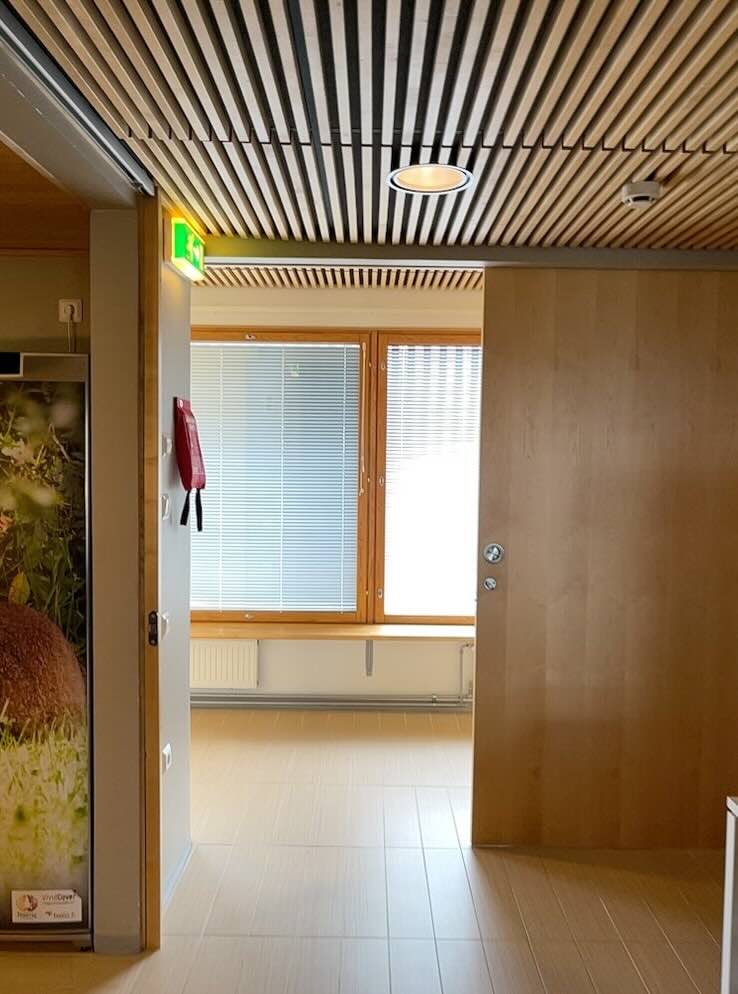}};
        \node[font=\tiny, text=white] at (0.2,-1.1) {};
    \end{tikzpicture}
\end{subfigure}
\hspace{-25pt}
\begin{subfigure}[b]{0.2\textwidth}
    \centering
    \subcaption*{Splatfacto}
    \begin{tikzpicture}
        \node [inner sep=0pt,clip,rounded corners=2pt] at (0,0) {\includegraphics[height=3.5cm]{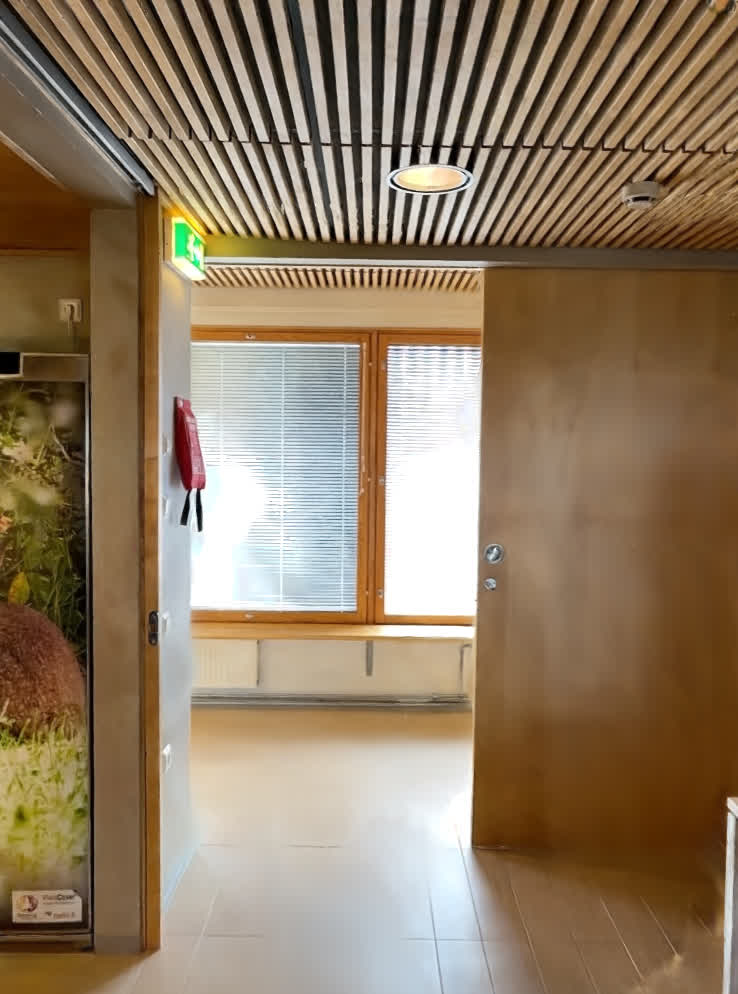}};
        \node[font=\tiny, text=white] at (0.2,-1.1) {};
    \end{tikzpicture}
\end{subfigure}
\hspace{-25pt}
\begin{subfigure}[b]{0.2\textwidth}
    \centering
    \subcaption*{2DGS}
    \begin{tikzpicture}
        \node [inner sep=0pt,clip,rounded corners=2pt] at (0,0) {\includegraphics[height=3.5cm]{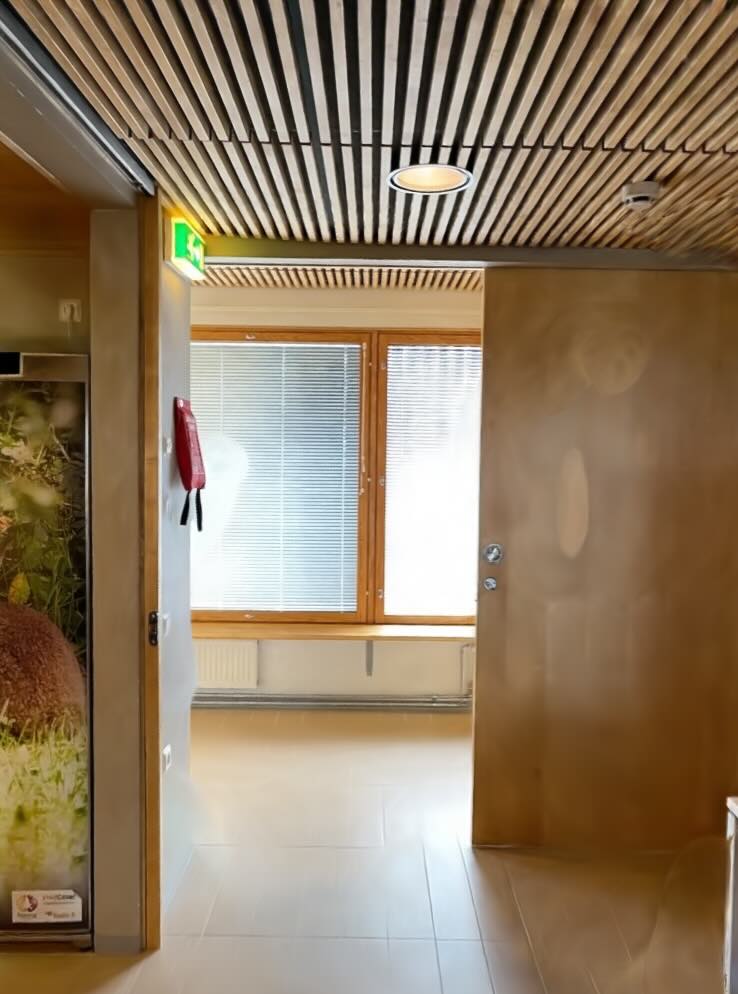}};
        \node[font=\tiny, text=white] at (0.2,-1.1) {};
    \end{tikzpicture}
\end{subfigure}
\hspace{-25pt}
\begin{subfigure}[b]{0.2\textwidth}
    \centering
    \subcaption*{2DGS w/ Sensor Depth}
    \begin{tikzpicture}
        \node [inner sep=0pt,clip,rounded corners=2pt] at (0,0) {\includegraphics[height=3.5cm]{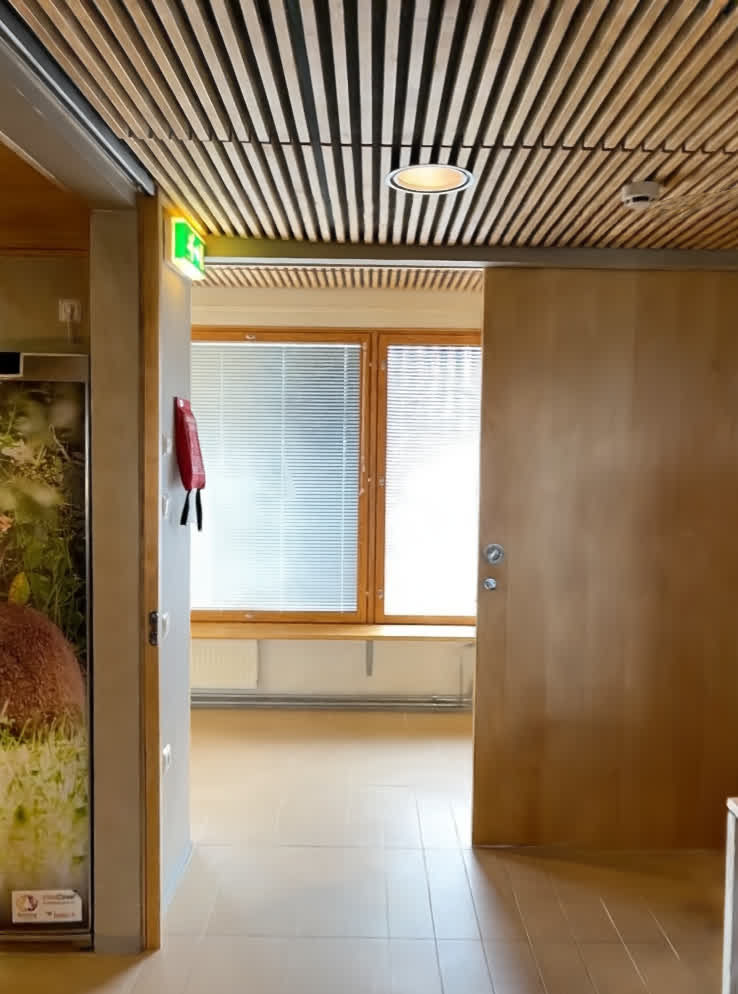}};
        \node[font=\tiny, text=white] at (0.2,-1.1) {};
    \end{tikzpicture}
\end{subfigure}
\hspace{-25pt}
\begin{subfigure}[b]{0.2\textwidth}
    \centering
    \subcaption*{Ours w/ Sensor Depth}
    \begin{tikzpicture}
        \node [inner sep=0pt,clip,rounded corners=2pt] at (0,0) {\includegraphics[height=3.5cm]{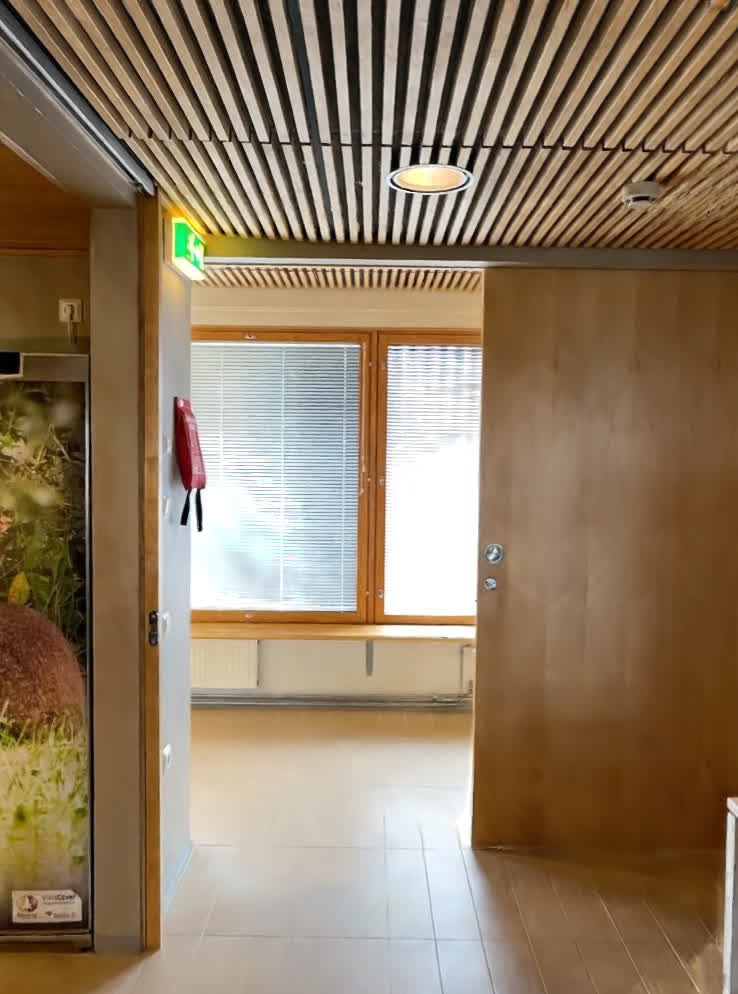}};
        \node[font=\tiny, text=white] at (0.2,-1.1) {};
    \end{tikzpicture}
\end{subfigure}
\hspace{-25pt}
\\
\begin{subfigure}[b]{0.2\textwidth}
    \centering
    \subcaption*{iPhone Depth}
    \begin{tikzpicture}
        \node [inner sep=0pt,clip,rounded corners=2pt] at (0,0) {\includegraphics[height=3.5cm]{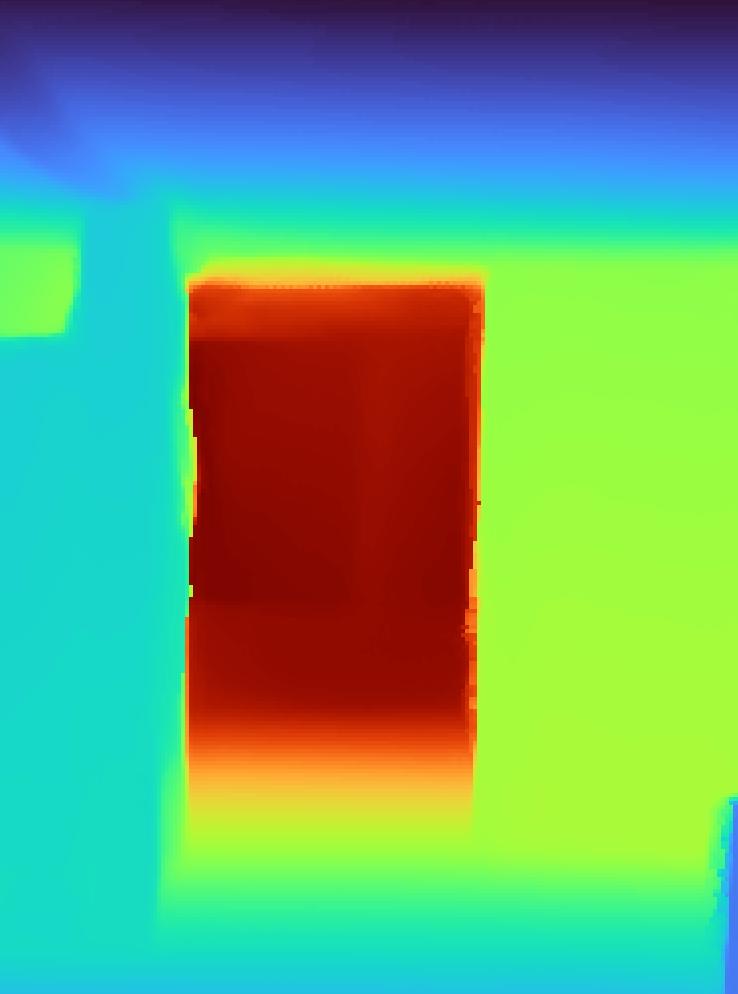}};
        \node[font=\tiny, text=white] at (0.2,-1.1) {};
    \end{tikzpicture}
\end{subfigure}
\hspace{-25pt}
\begin{subfigure}[b]{0.2\textwidth}
    \centering
    \subcaption*{Splatfacto}
    \begin{tikzpicture}
        \node [inner sep=0pt,clip,rounded corners=2pt] at (0,0) {\includegraphics[height=3.5cm]{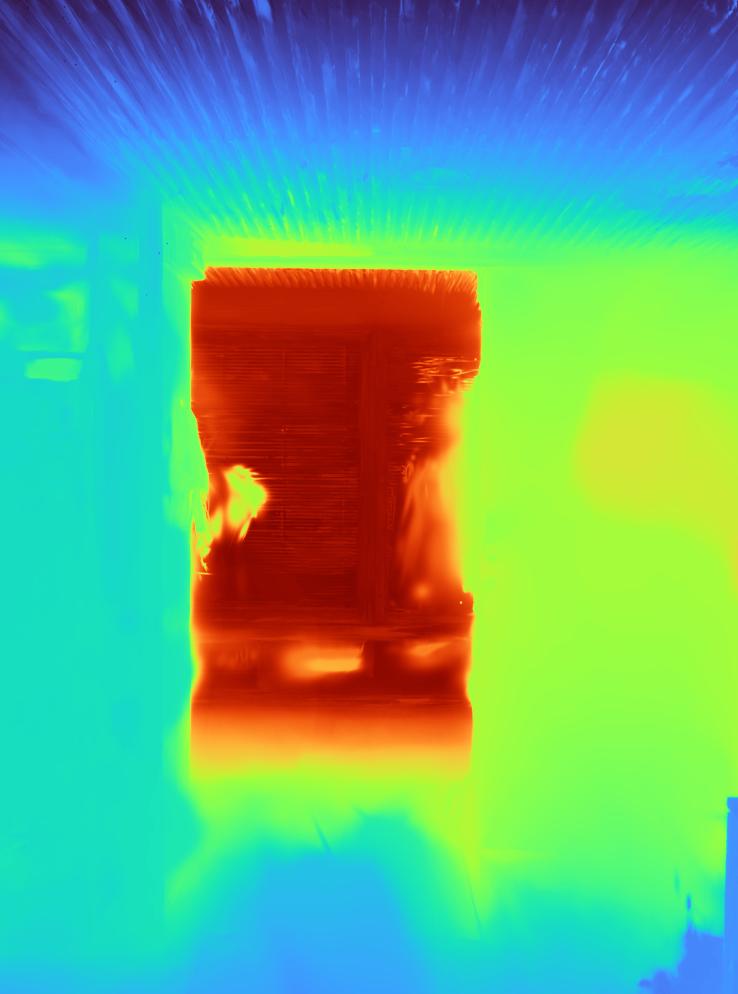}};
        \node[font=\tiny, text=white] at (0.2,-1.1) {};
    \end{tikzpicture}
\end{subfigure}
\hspace{-25pt}
\begin{subfigure}[b]{0.2\textwidth}
    \centering
    \subcaption*{2DGS}
    \begin{tikzpicture}
        \node [inner sep=0pt,clip,rounded corners=2pt] at (0,0) {\includegraphics[height=3.5cm]{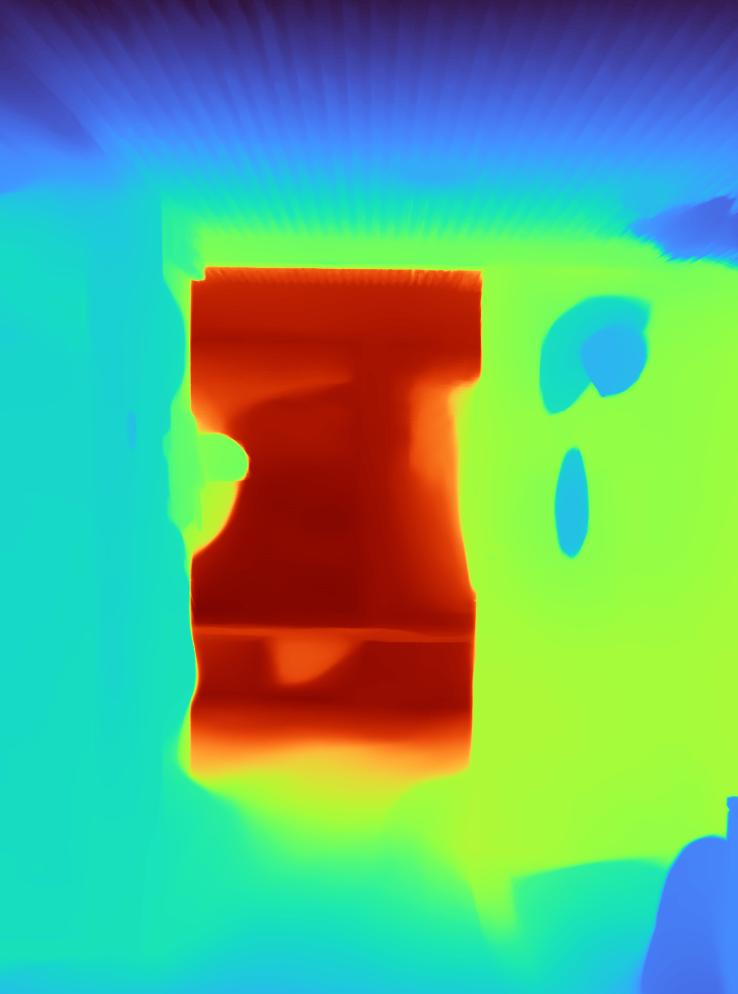}};
        \node[font=\tiny, text=white] at (0.2,-1.1) {};
    \end{tikzpicture}
\end{subfigure}
\hspace{-25pt}
\begin{subfigure}[b]{0.2\textwidth}
    \centering
    \subcaption*{2DGS w/ Sensor Depth}
    \begin{tikzpicture}
        \node [inner sep=0pt,clip,rounded corners=2pt] at (0,0) {\includegraphics[height=3.5cm]{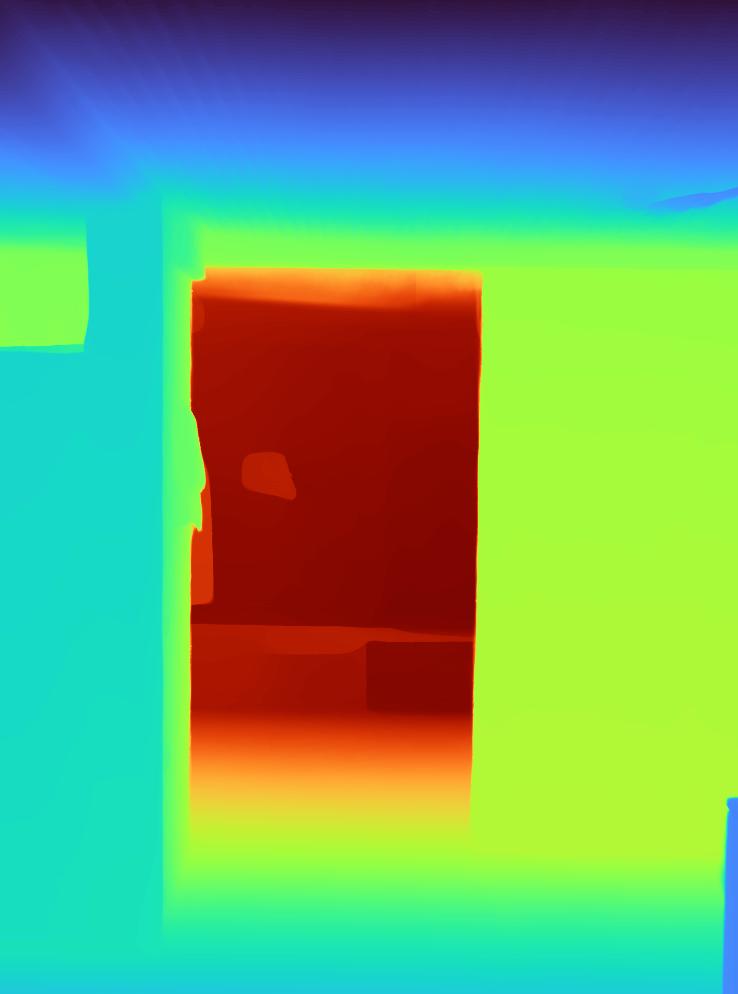}};
        \node[font=\tiny, text=white] at (0.2,-1.1) {};
    \end{tikzpicture}
\end{subfigure}
\hspace{-25pt}
\begin{subfigure}[b]{0.2\textwidth}
    \centering
    \subcaption*{Ours w/ Sensor Depth}
    \begin{tikzpicture}
        \node [inner sep=0pt,clip,rounded corners=2pt] at (0,0) {\includegraphics[height=3.5cm]{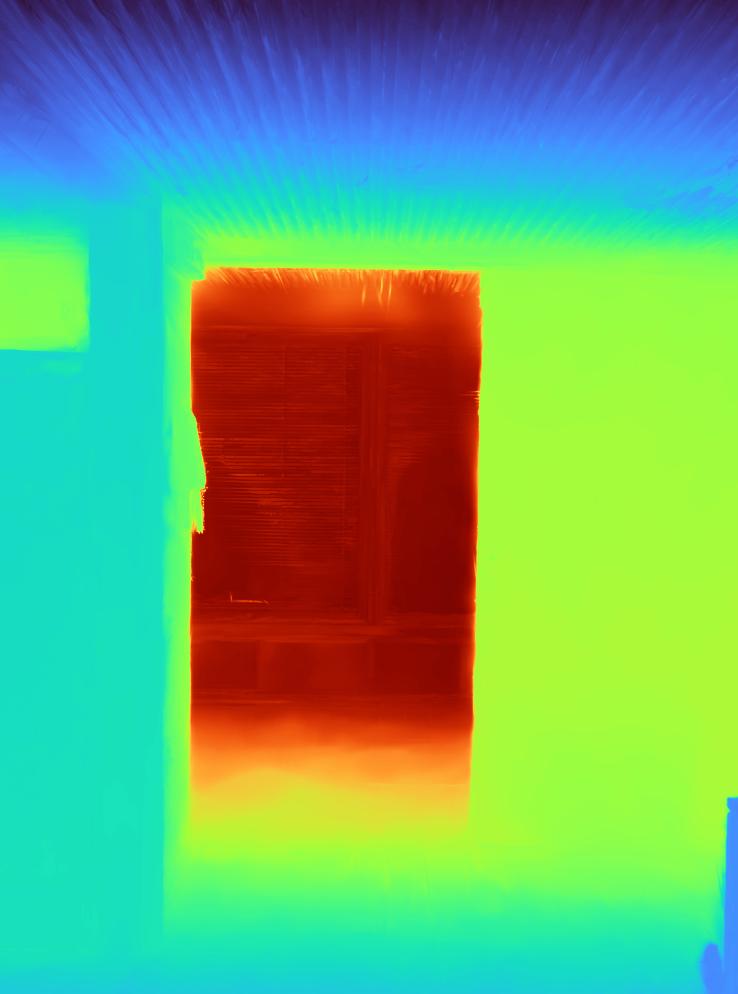}};
        \node[font=\tiny, text=white] at (0.2,-1.1) {};
    \end{tikzpicture}
\end{subfigure}
\caption{\textbf{Qualitative comparison between 2DGS and DN-Splatter.} We supervise both 2DGS and our method with the $\mathcal{L}_{\hat{D}}$ regularization loss and visualize novel-view and depth renders from the 'Honka' and 'Kokko' scenes from the MuSHRoom dataset. We note that DN-Splatter obtains higher metrics in both novel-view and mesh reconstruction metrics; whilst 2DGS obtains more smooth depth renders.}
\vspace{-1em}
\label{fig:2dgs_dn_imgs}
\end{figure*}

In \cref{fig:2dgs_dn_imgs} we compare novel-view and depth estimation renders using baseline Splatfacto and 2DGS models as well as a variant of 2DGS with depth supervision enabled and our method.

\subsection{Mesh and NVS renders}
Lastly, we provide additional qualitative results for mesh performance in~\figref{fig:supp-different-method-mesh} as well as depth and novel view renders in~\figref{fig:supp-different-method-cmp1},~\figref{fig:supp-different-method-cmp2}, and~\figref{fig:supp-different-method-cmp3}, respectively.

\newcommand{\mywidthmesh}{0.16\textwidth}
\newcommand{\myheightmesh}{0.09\textheight}
\begin{figure*}[t!]
\centering
  \setlength{\tabcolsep}{0.1em}
    \renewcommand{\arraystretch}{0.5}
    \hfill{}\hspace*{-0.5em}
    \footnotesize
    \begin{tabular}{cccccc}
        \includegraphics[width=\mywidthmesh, height=\myheightmesh]{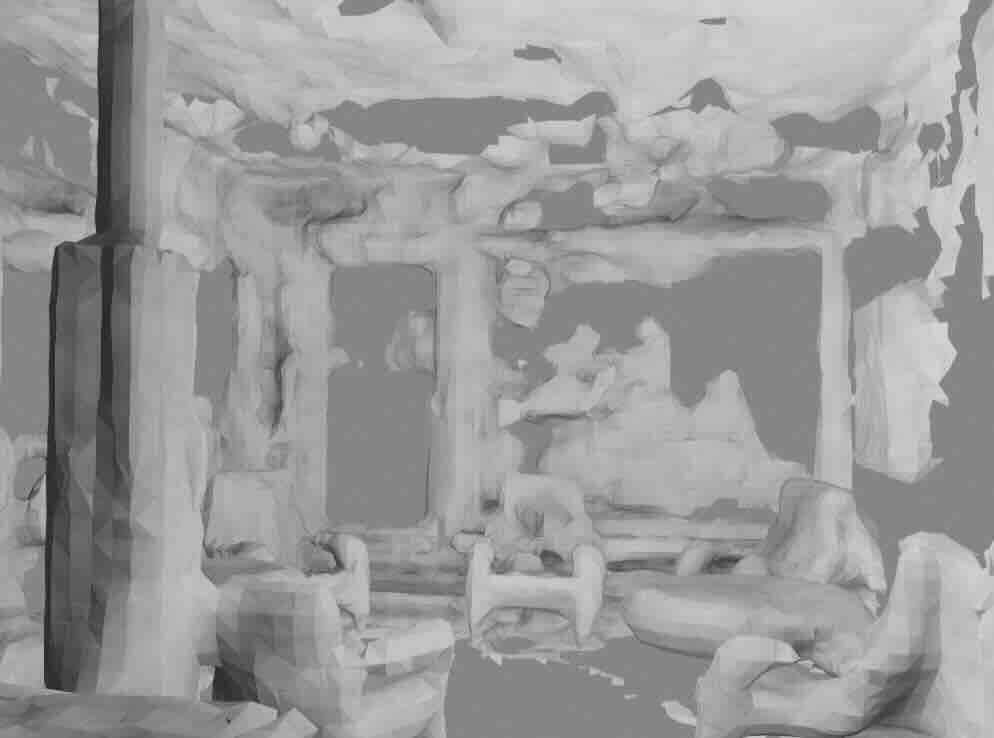}&
        \includegraphics[width=\mywidthmesh, height=\myheightmesh]{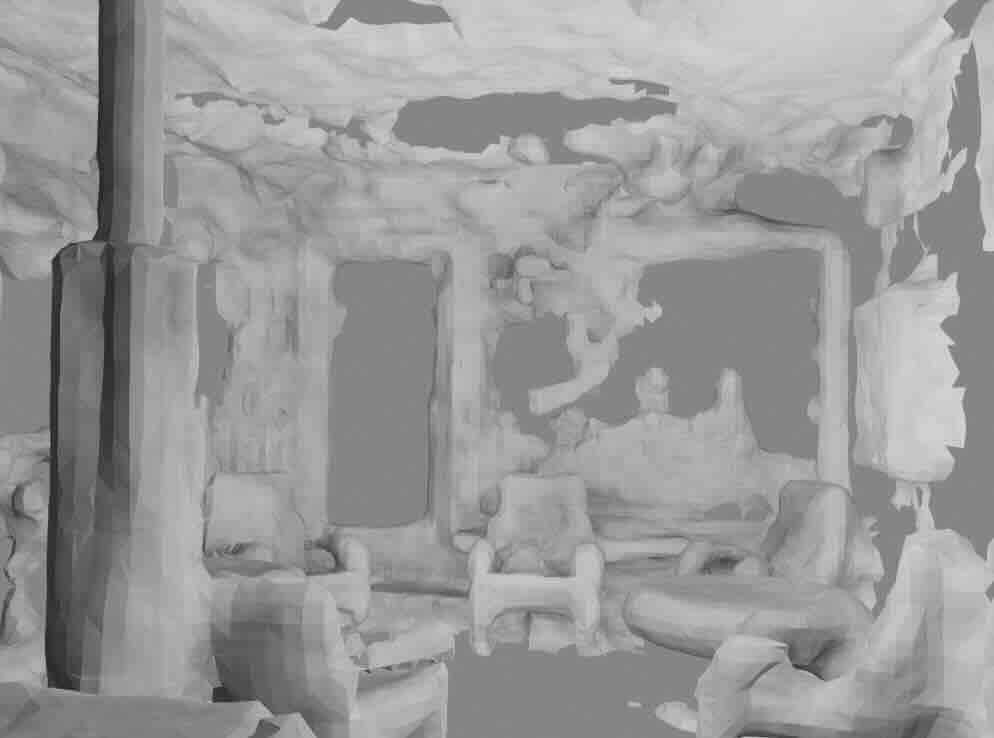}&
        \includegraphics[width=\mywidthmesh, height=\myheightmesh]{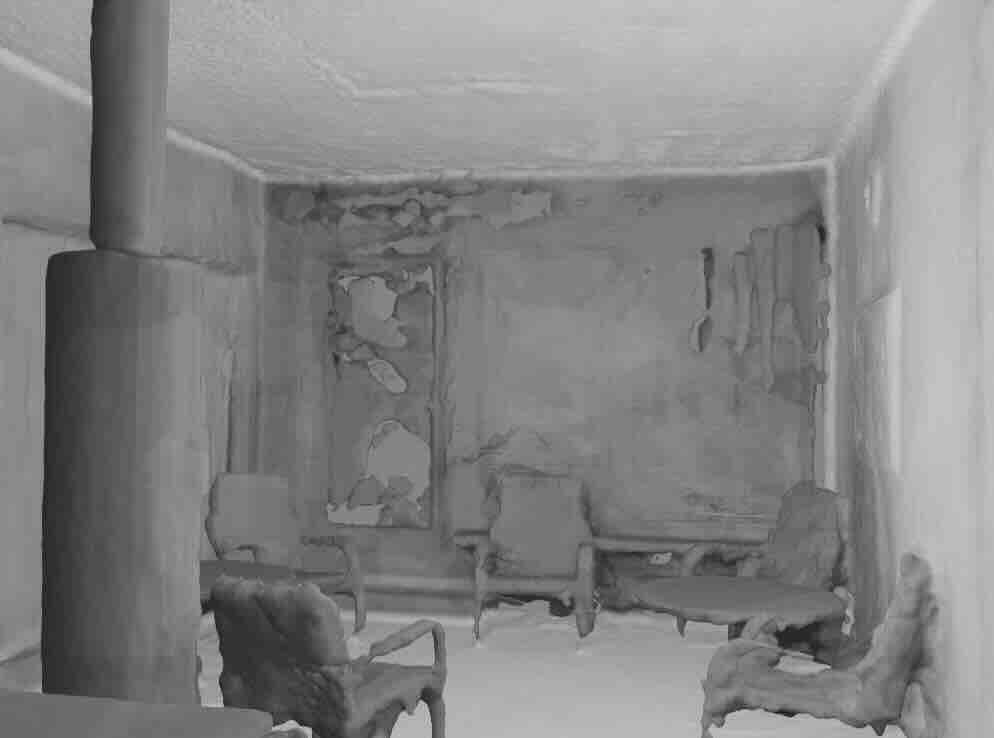}&
        \includegraphics[width=\mywidthmesh, height=\myheightmesh]{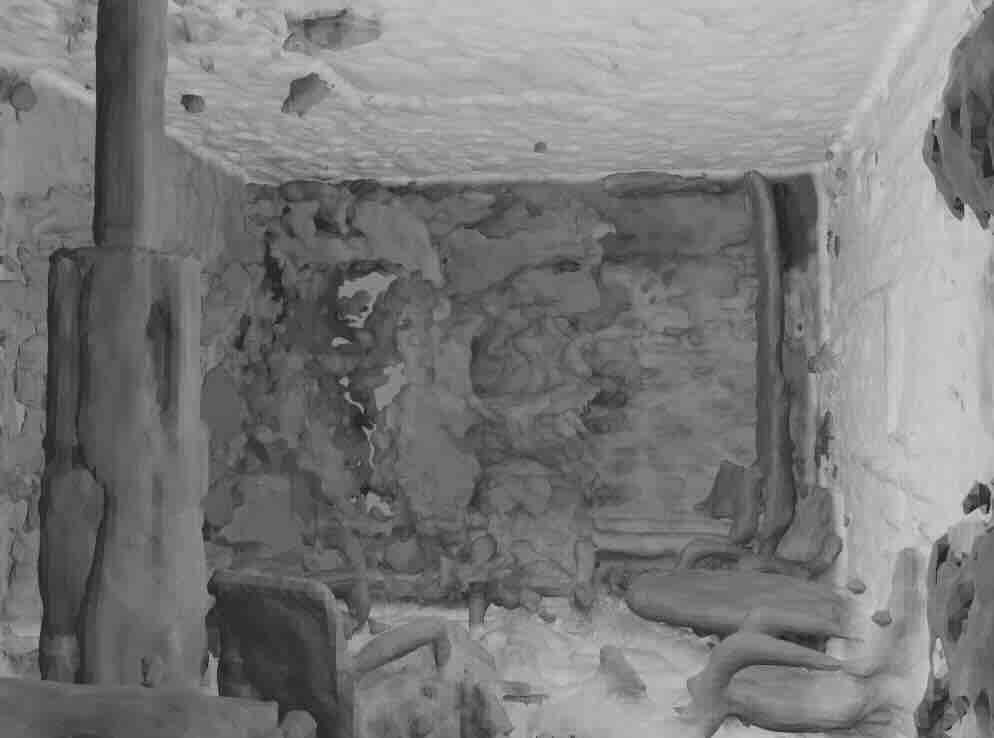}&
        \includegraphics[width=\mywidthmesh, height=\myheightmesh]{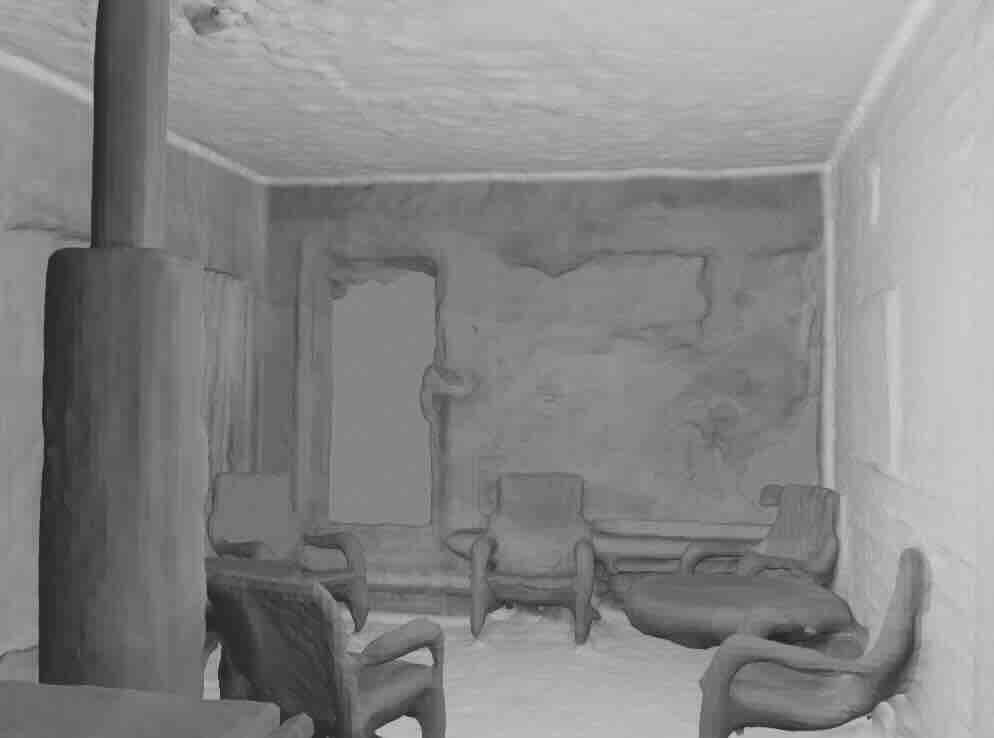}&
        \includegraphics[width=\mywidthmesh, height=\myheightmesh]{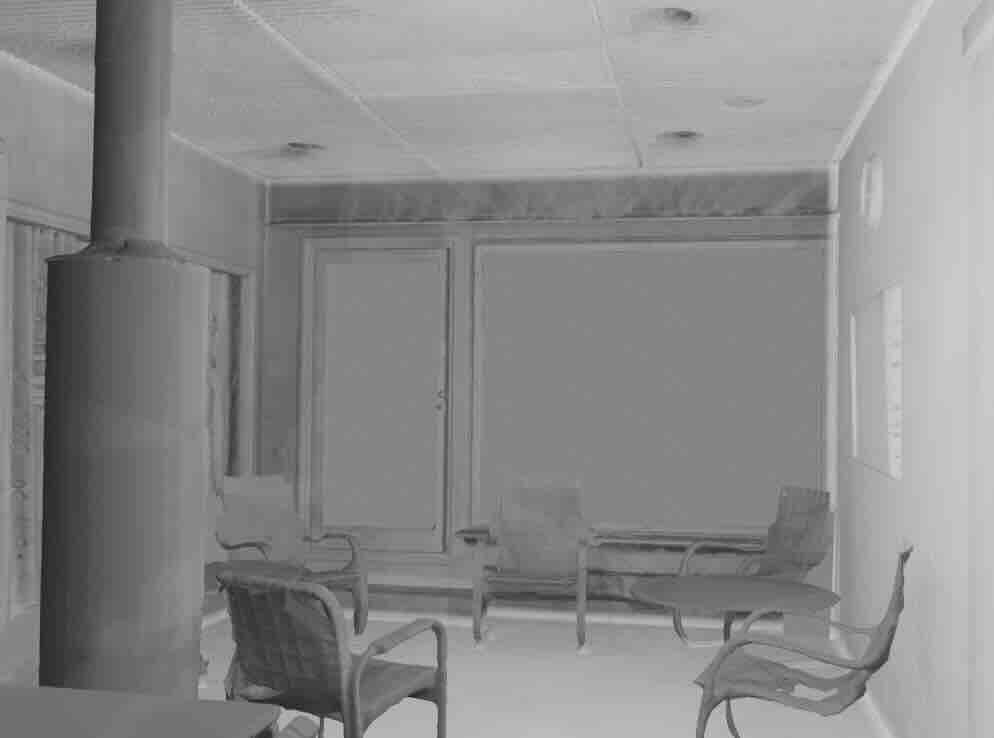}\\
        \includegraphics[width=\mywidthmesh, height=\myheightmesh]{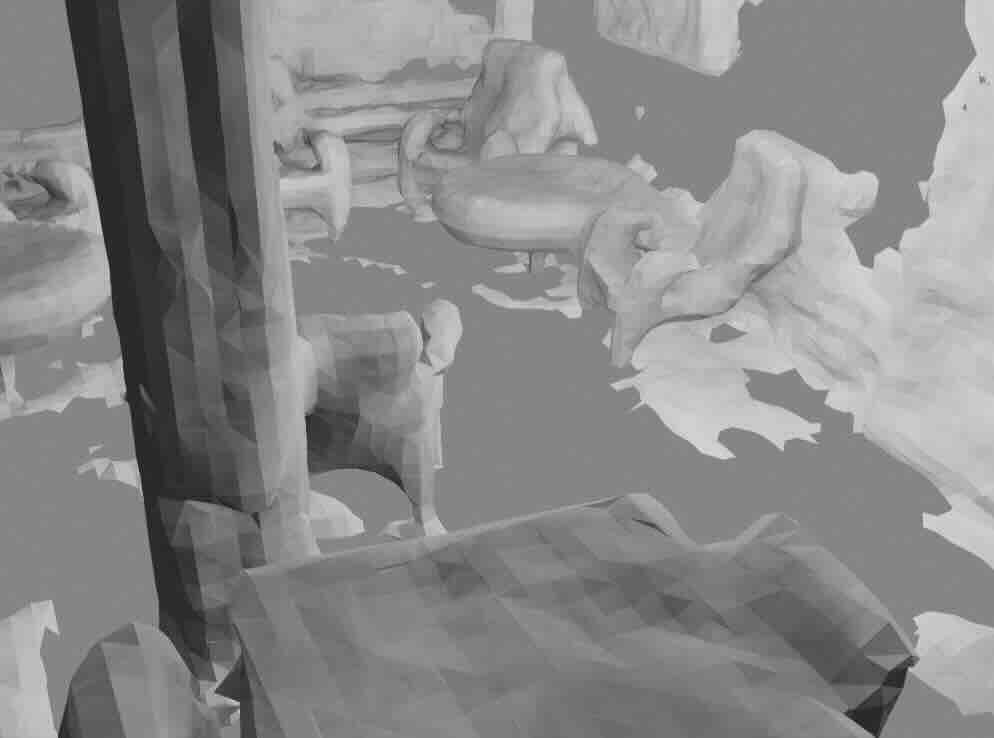}&
        \includegraphics[width=\mywidthmesh, height=\myheightmesh]{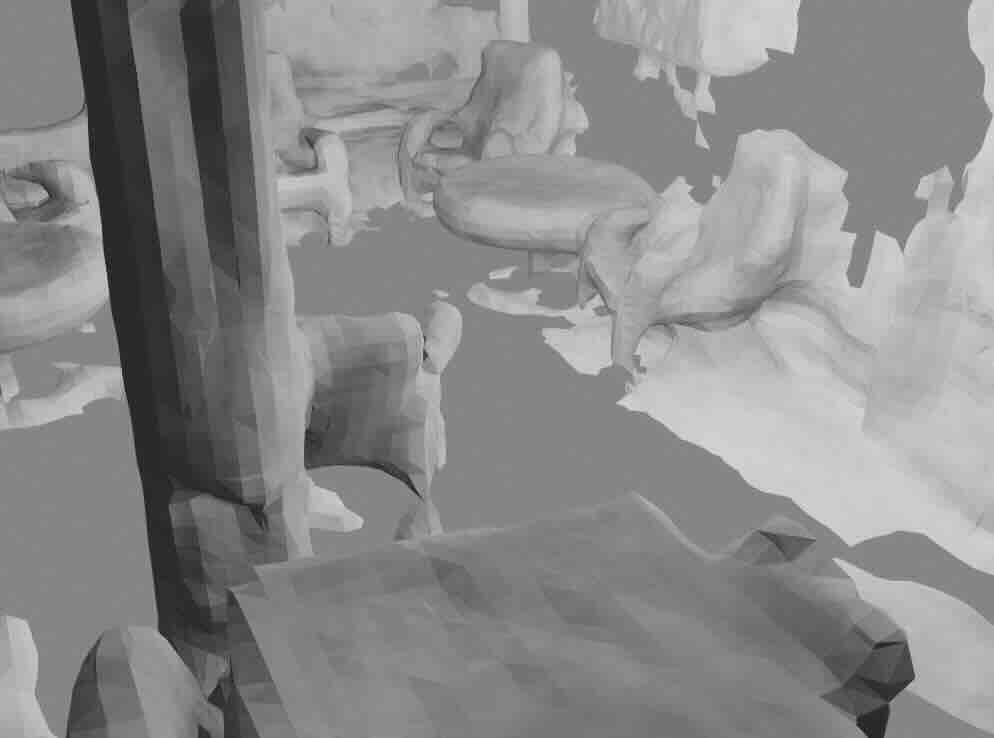}&
        \includegraphics[width=\mywidthmesh, height=\myheightmesh]{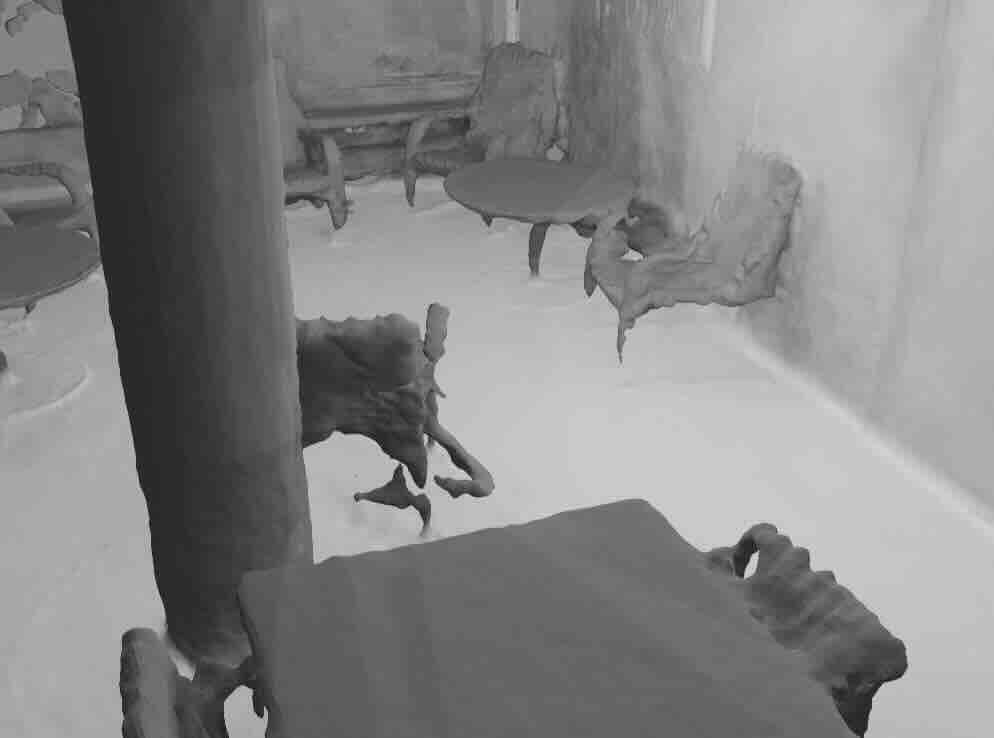}&
        \includegraphics[width=\mywidthmesh, height=\myheightmesh]{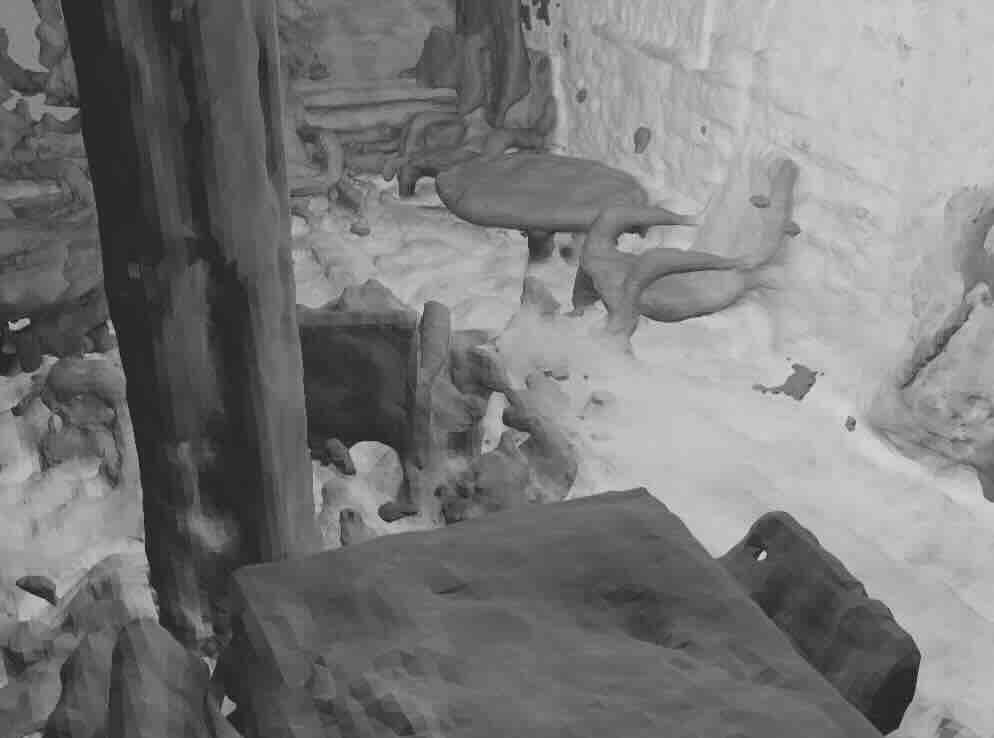}&
        \includegraphics[width=\mywidthmesh, height=\myheightmesh]{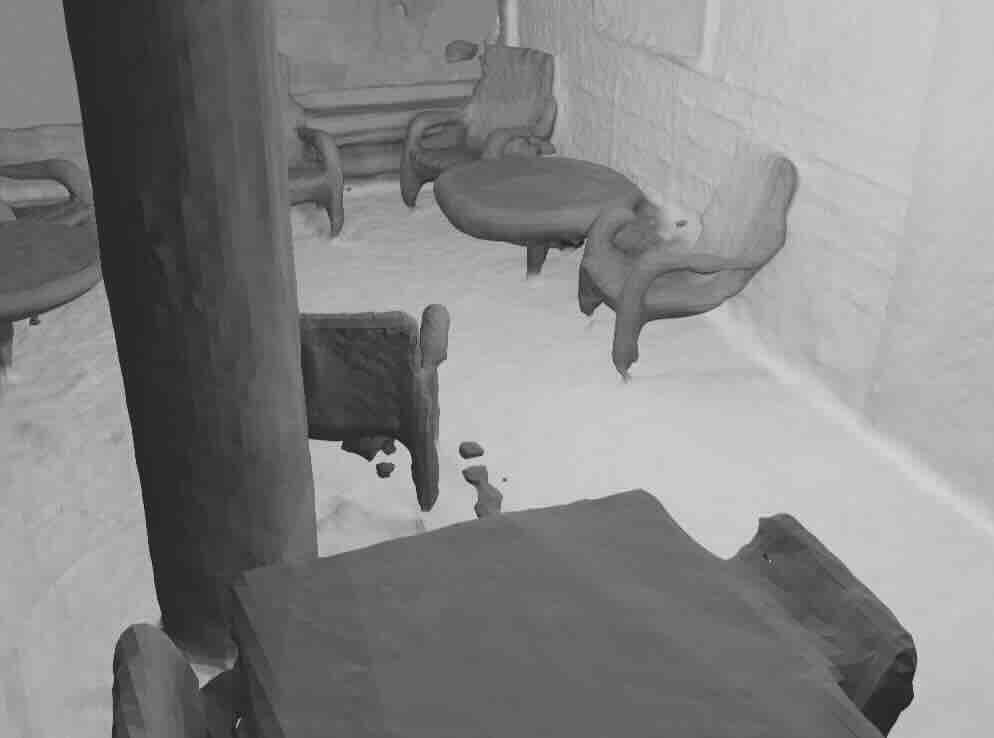}&
        \includegraphics[width=\mywidthmesh, height=\myheightmesh]{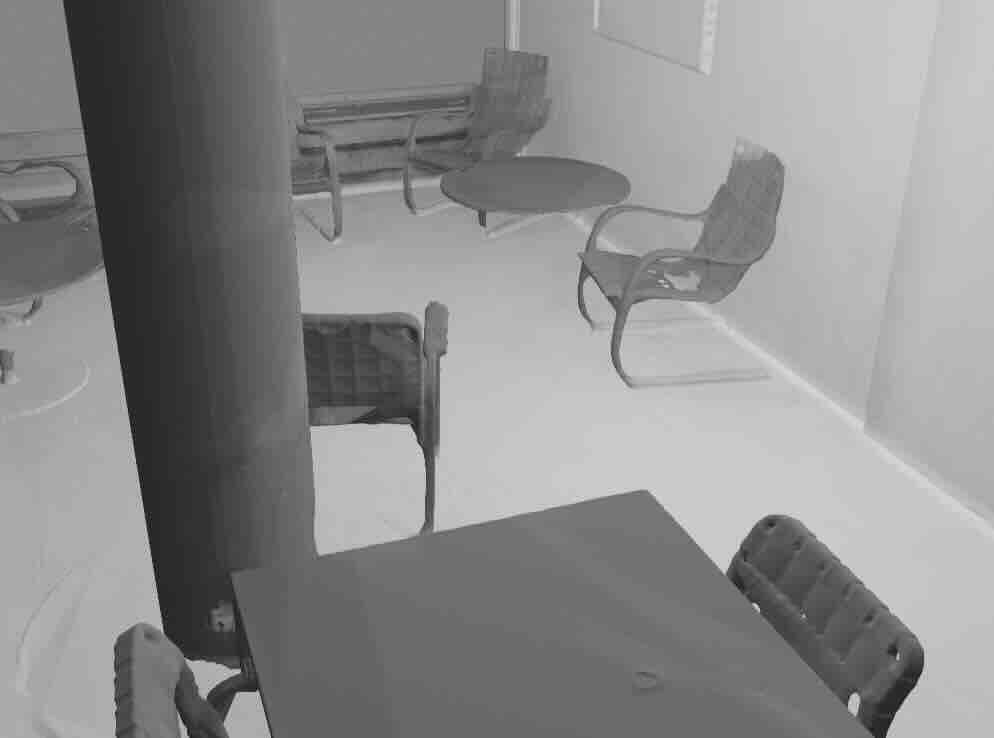}\\
        \includegraphics[width=\mywidthmesh, height=\myheightmesh]{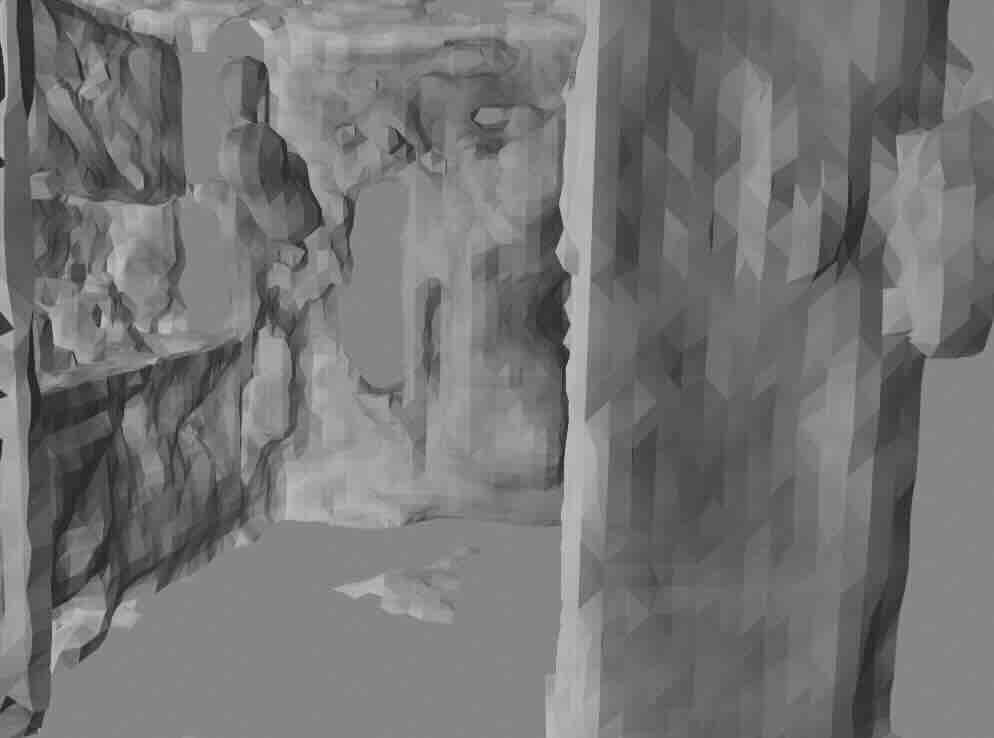}&
        \includegraphics[width=\mywidthmesh, height=\myheightmesh]{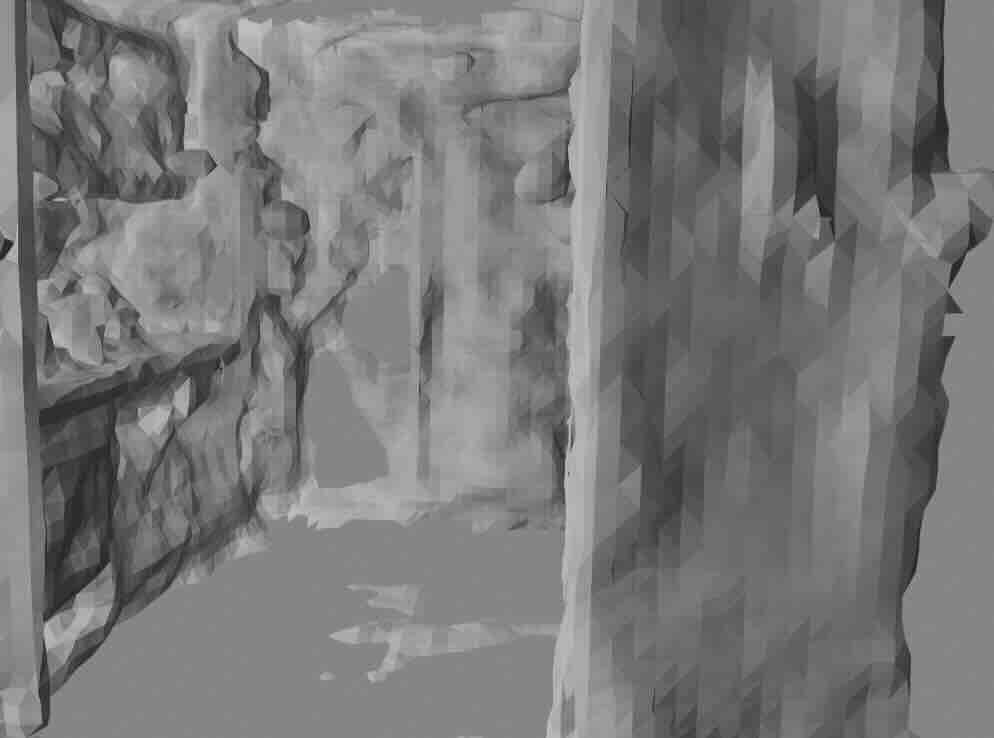}&
        \includegraphics[width=\mywidthmesh, height=\myheightmesh]{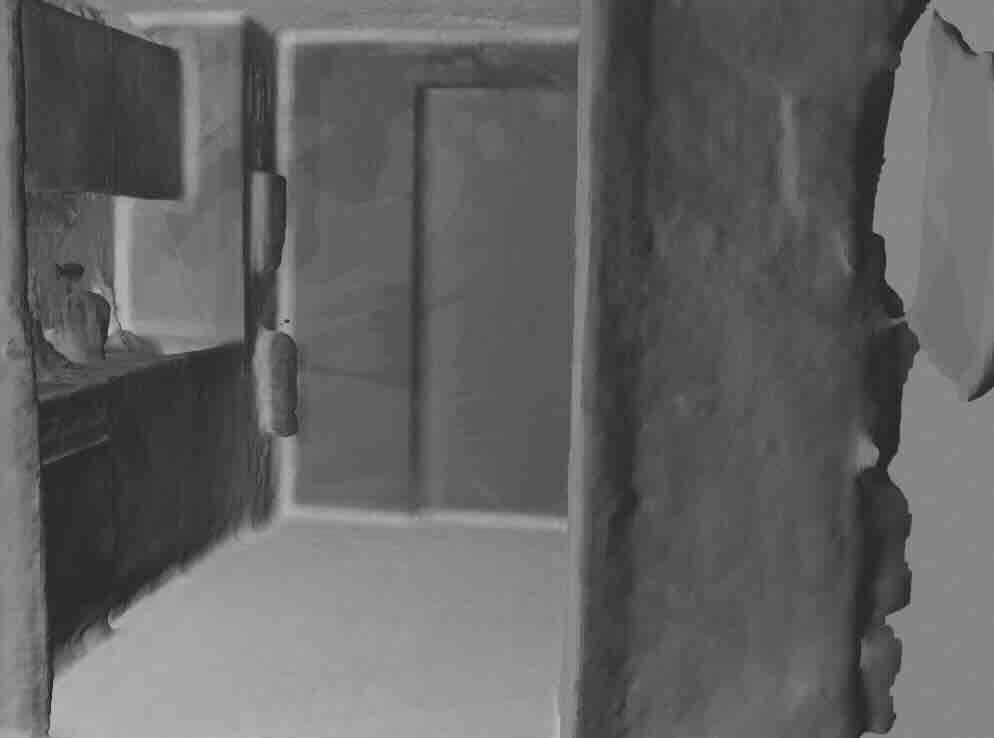}&
        \includegraphics[width=\mywidthmesh, height=\myheightmesh]{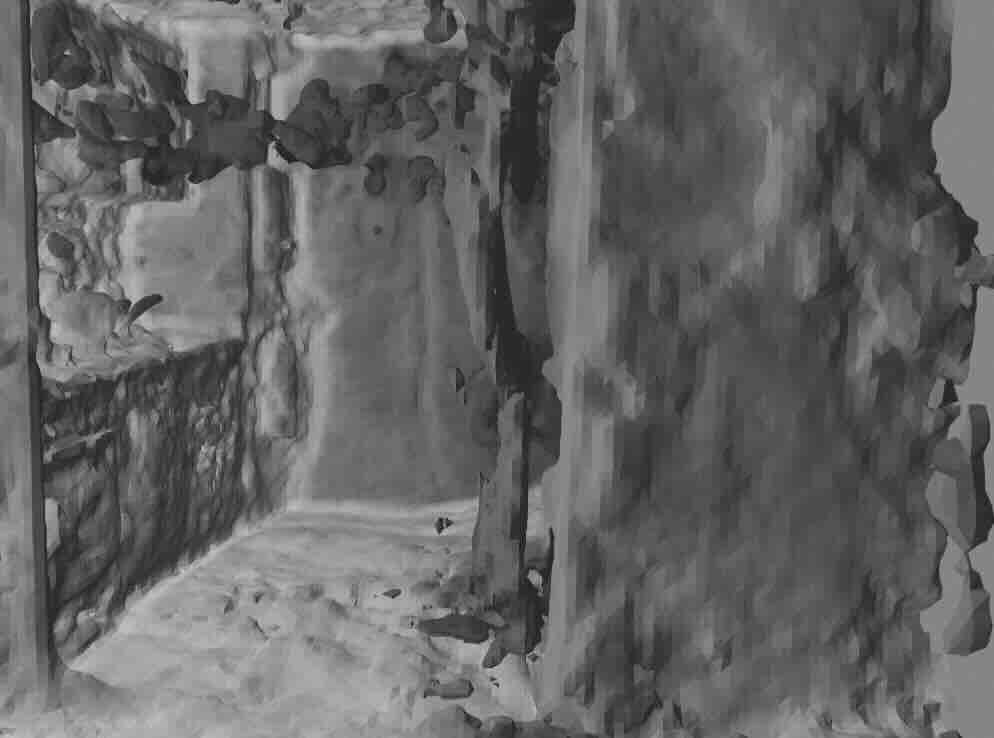}&
        \includegraphics[width=\mywidthmesh, height=\myheightmesh]{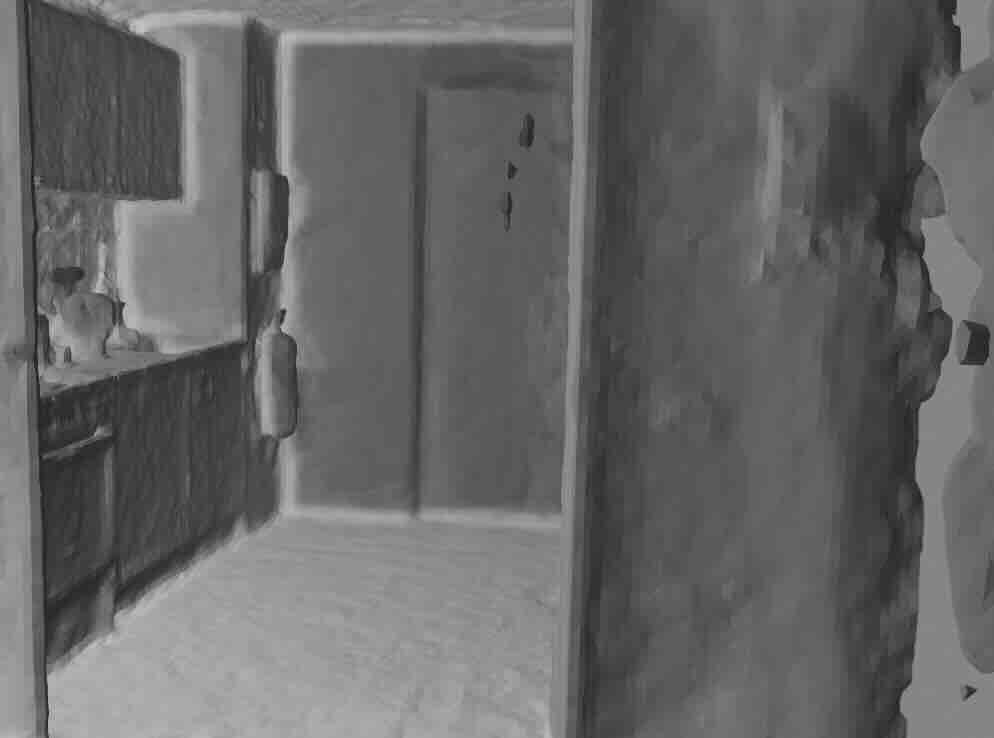}&
        \includegraphics[width=\mywidthmesh, height=\myheightmesh]{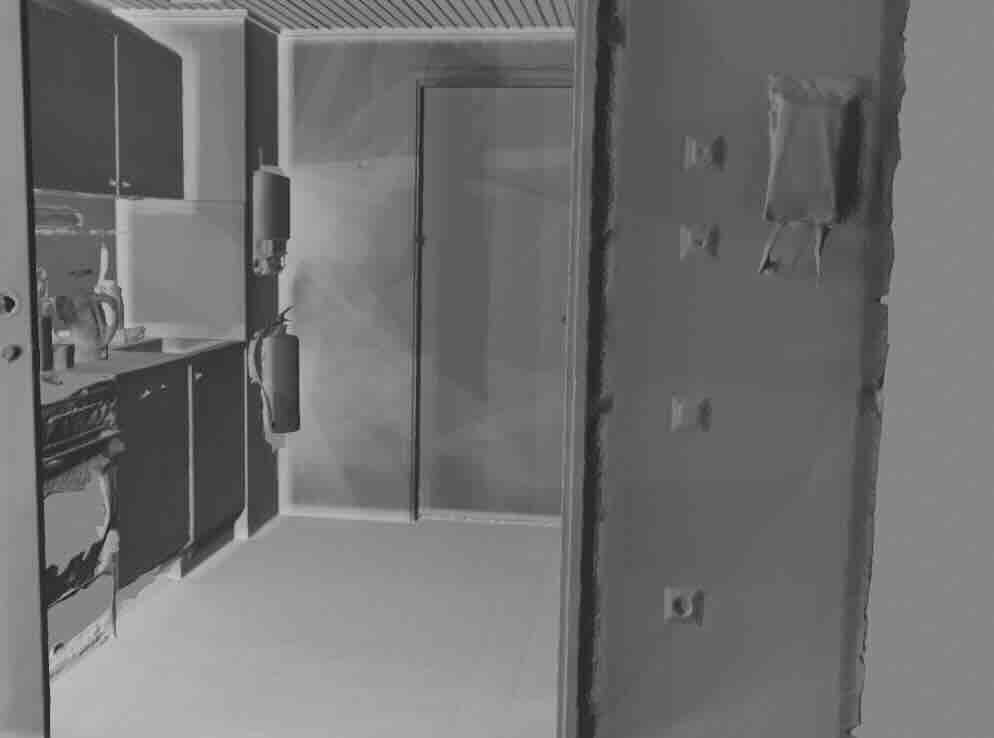}\\
        \includegraphics[width=\mywidthmesh, height=\myheightmesh]{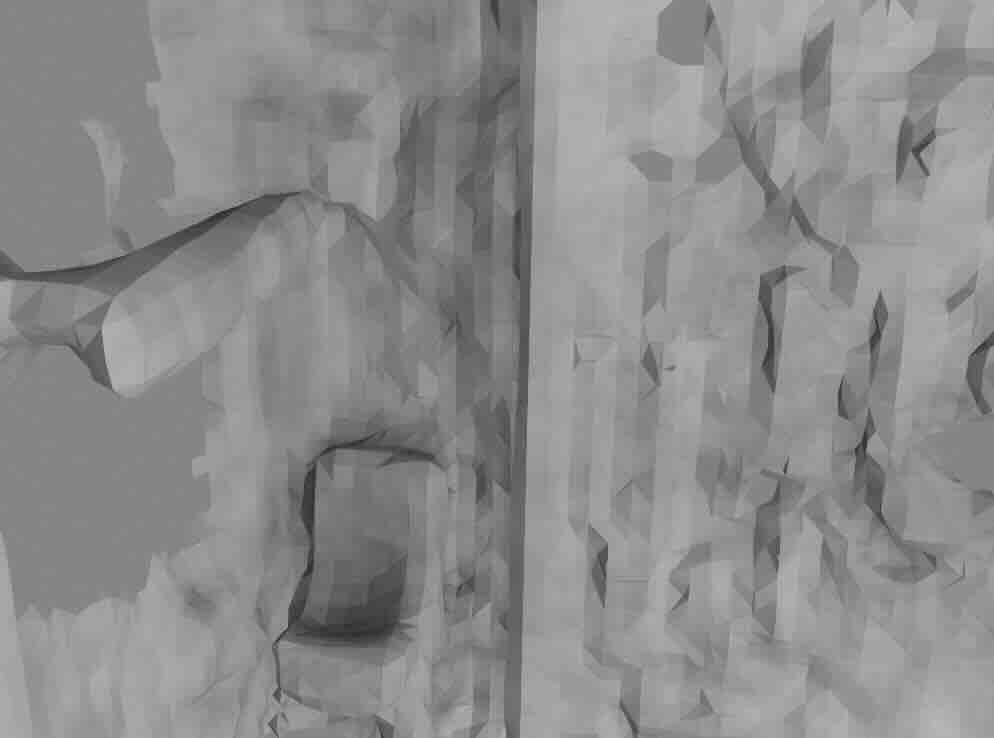}&
        \includegraphics[width=\mywidthmesh, height=\myheightmesh]{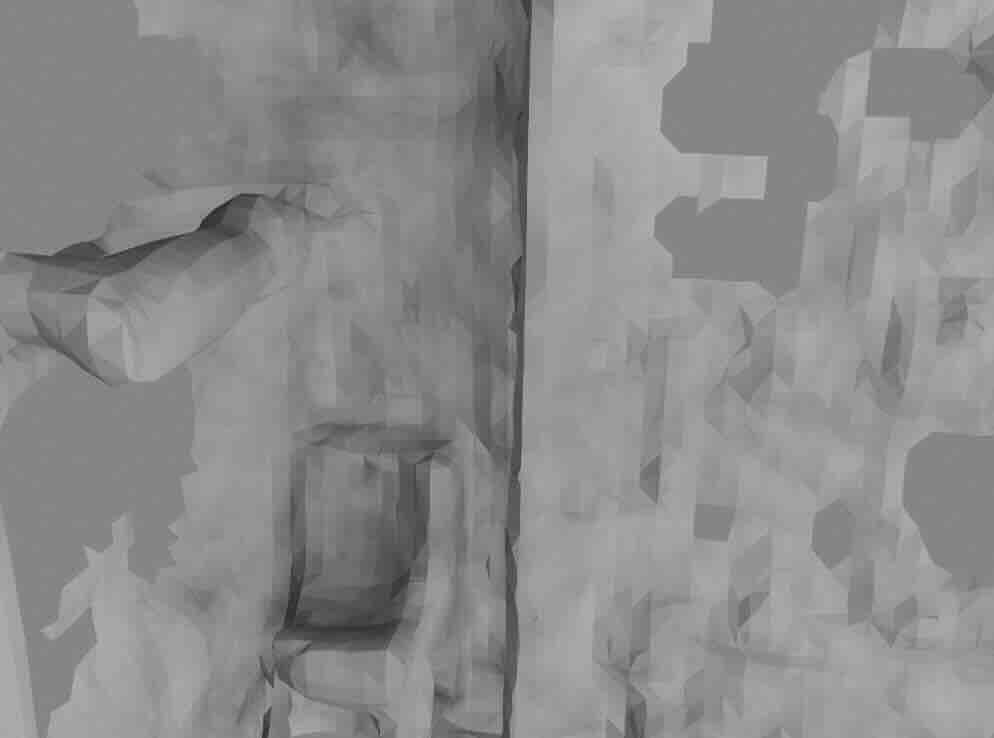}&
        \includegraphics[width=\mywidthmesh, height=\myheightmesh]{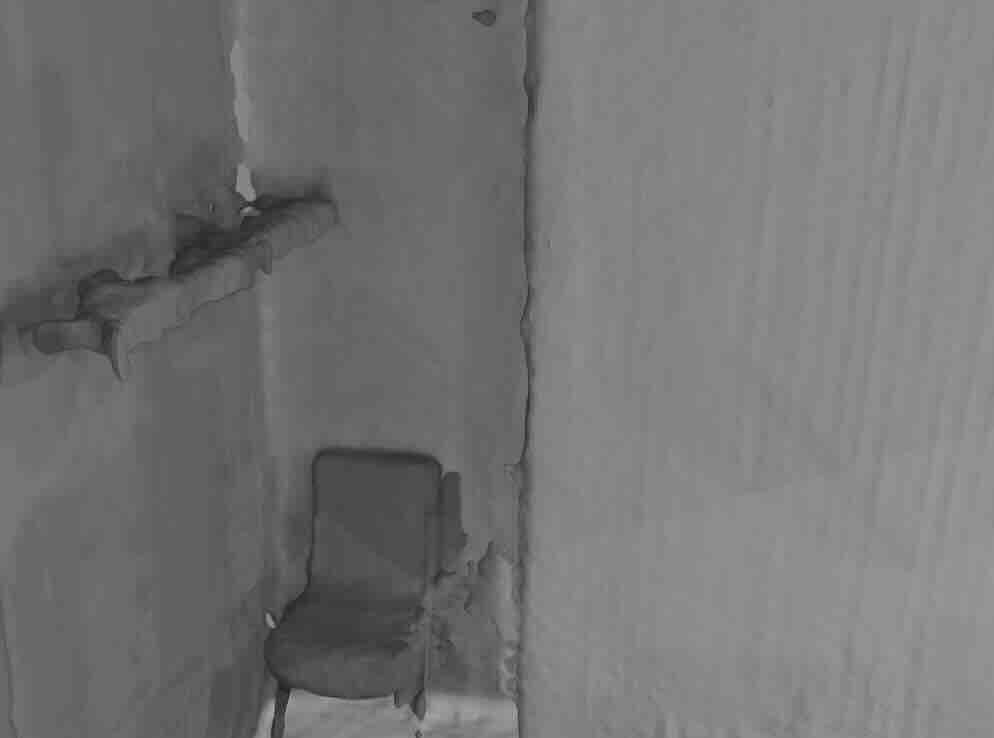}&
        \includegraphics[width=\mywidthmesh, height=\myheightmesh]{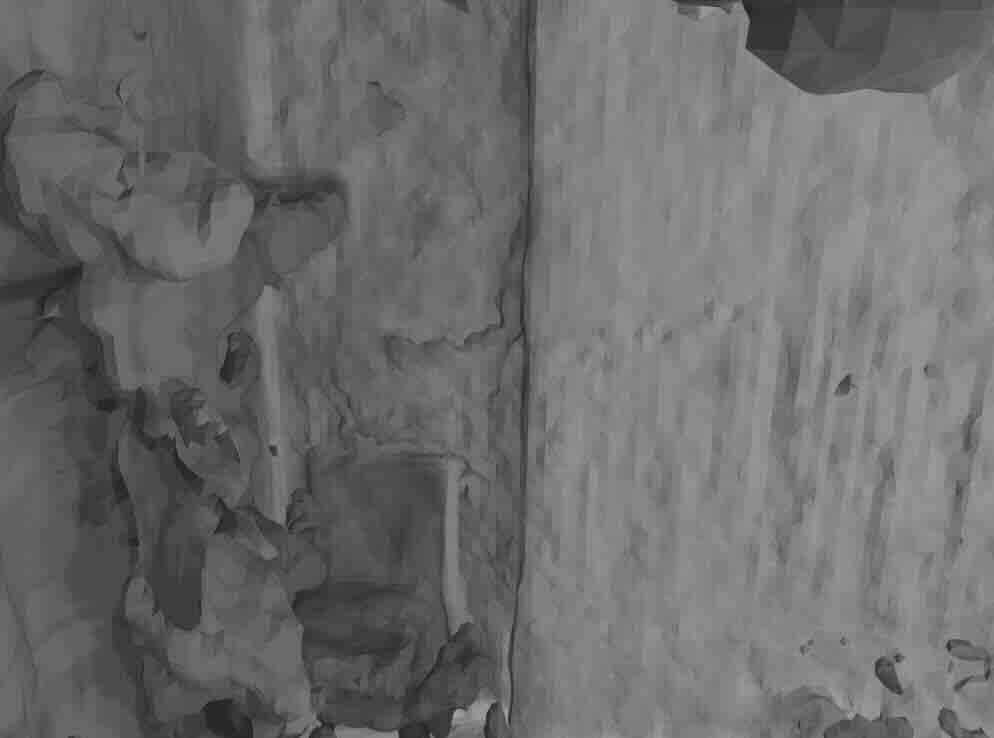}&
        \includegraphics[width=\mywidthmesh, height=\myheightmesh]{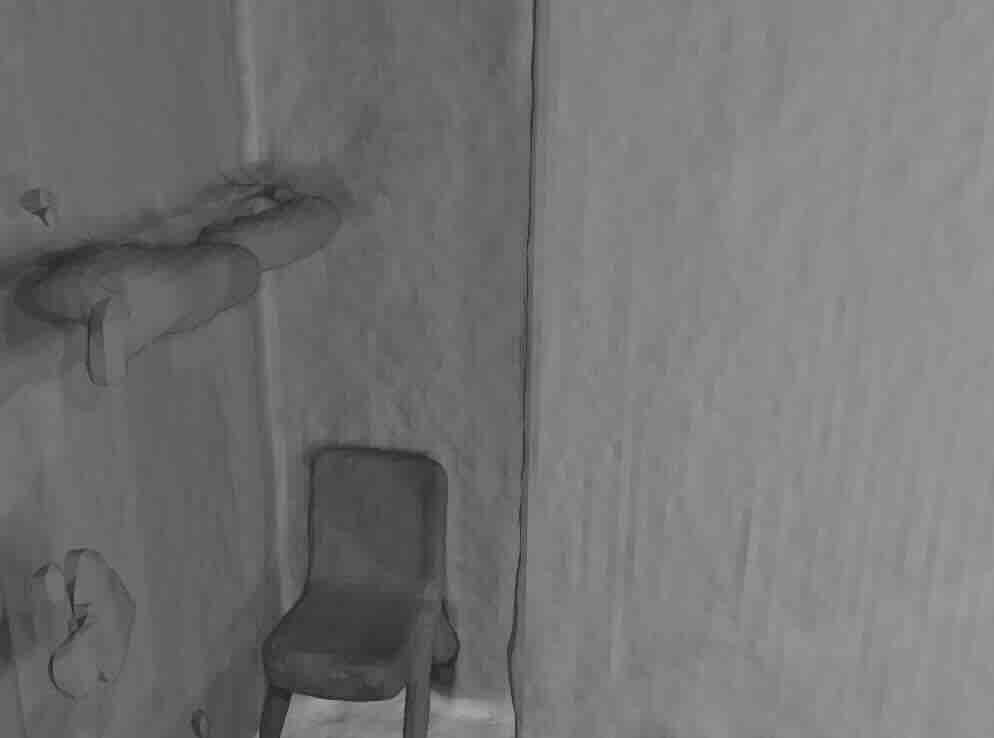}&
        \includegraphics[width=\mywidthmesh, height=\myheightmesh]{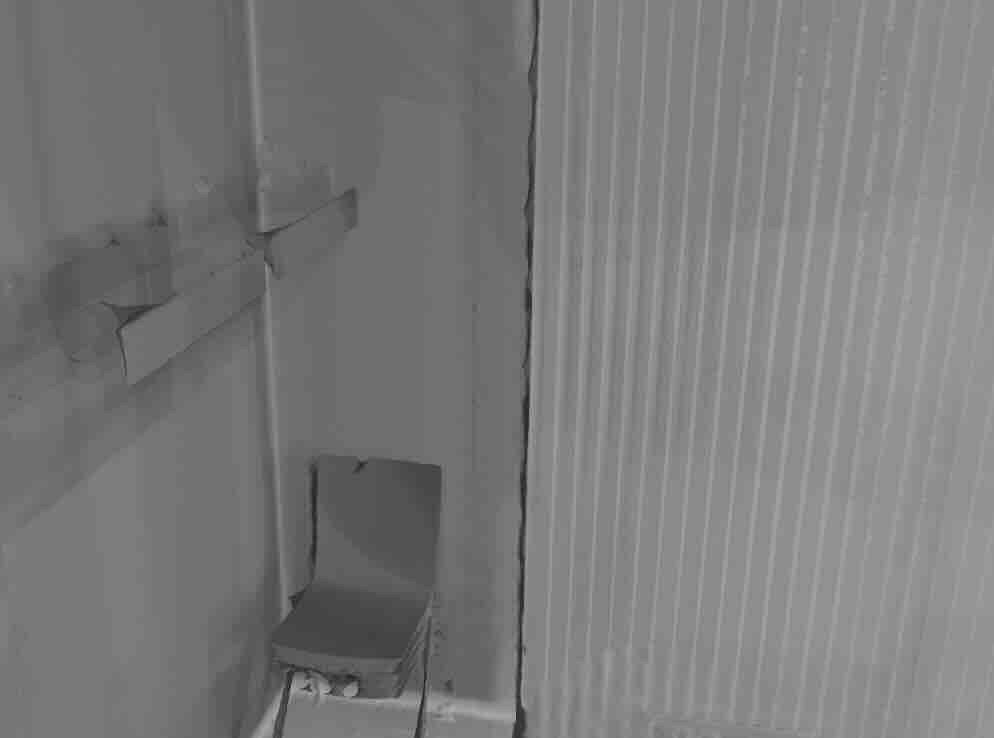}\\
        \includegraphics[width=\mywidthmesh, height=\myheightmesh]{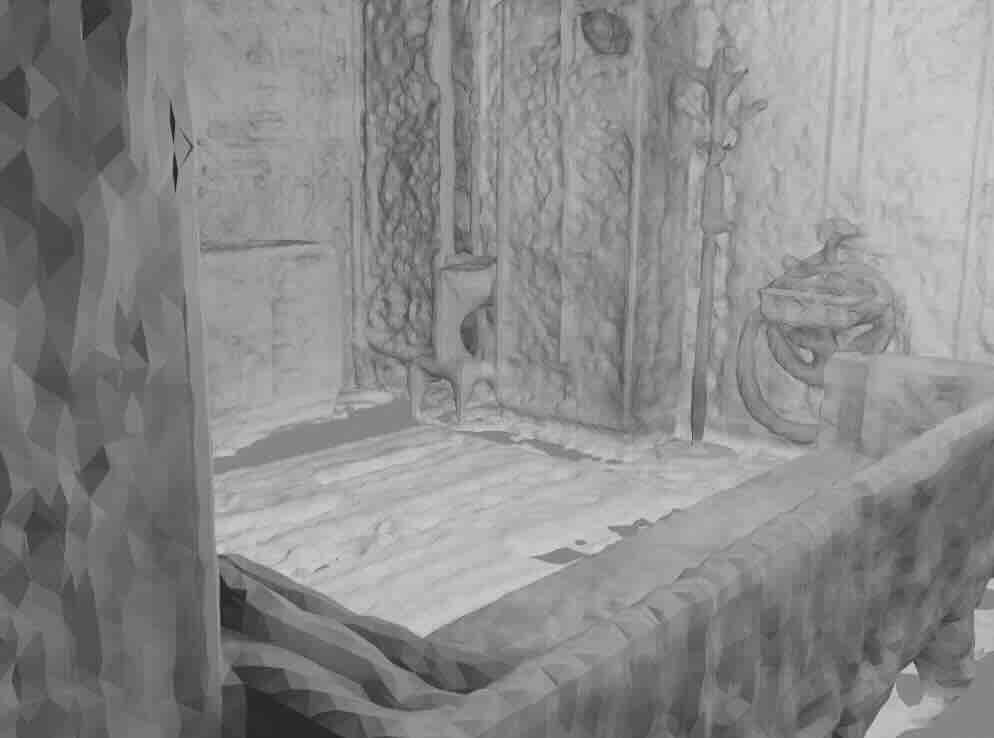}&
        \includegraphics[width=\mywidthmesh, height=\myheightmesh]{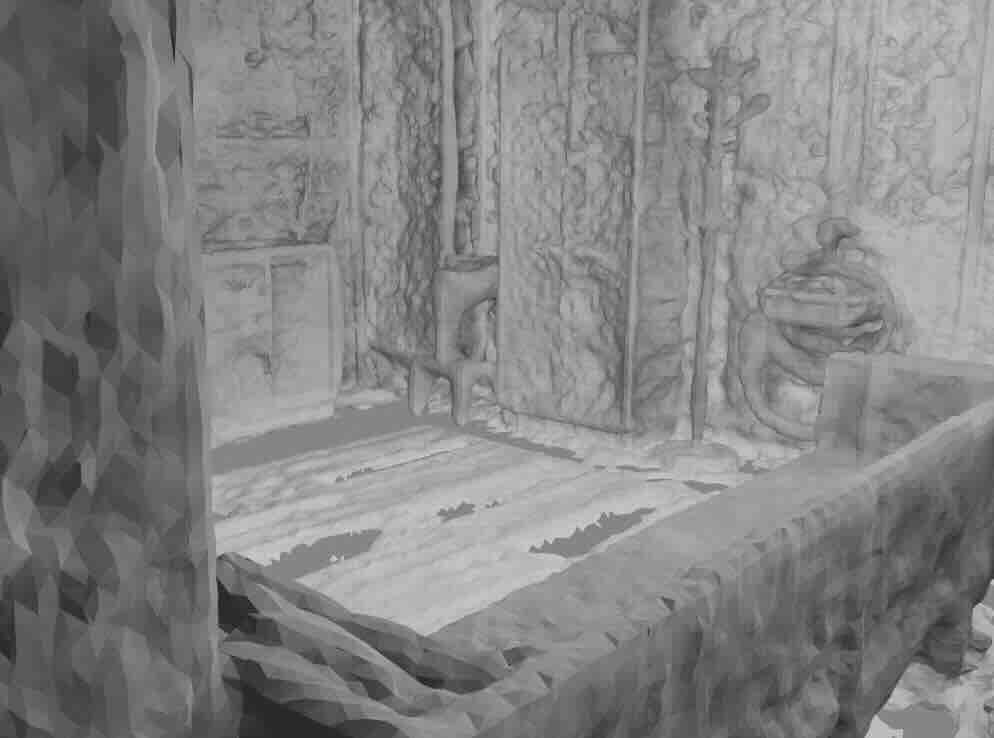}&
        \includegraphics[width=\mywidthmesh, height=\myheightmesh]{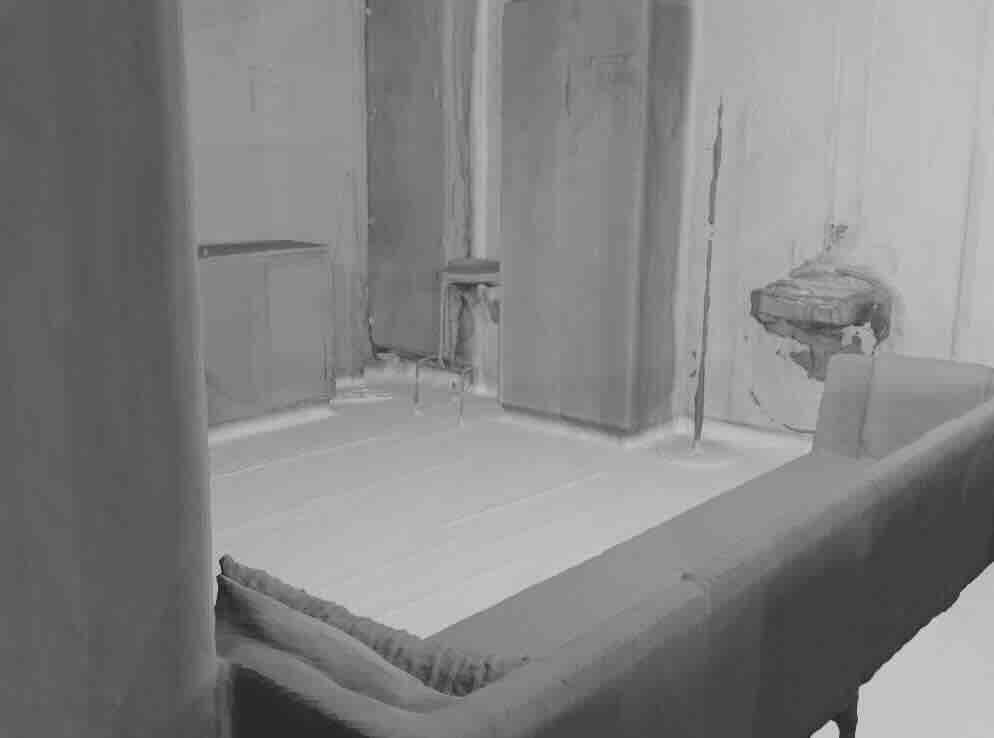}&
        \includegraphics[width=\mywidthmesh, height=\myheightmesh]{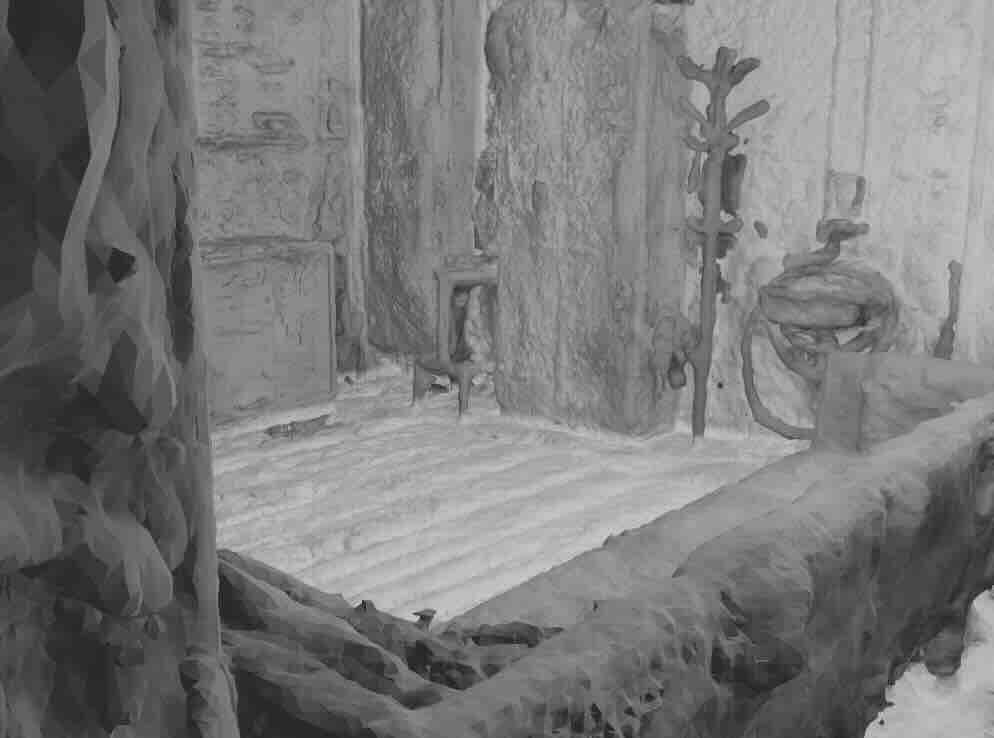}&
        \includegraphics[width=\mywidthmesh, height=\myheightmesh]{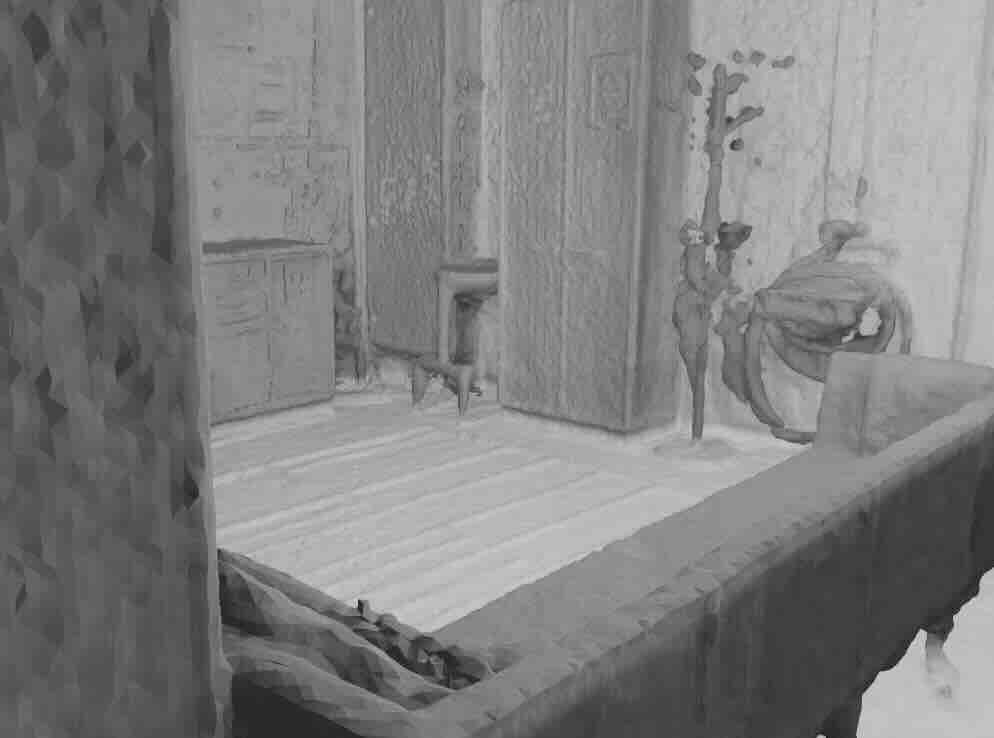}&
        \includegraphics[width=\mywidthmesh, height=\myheightmesh]{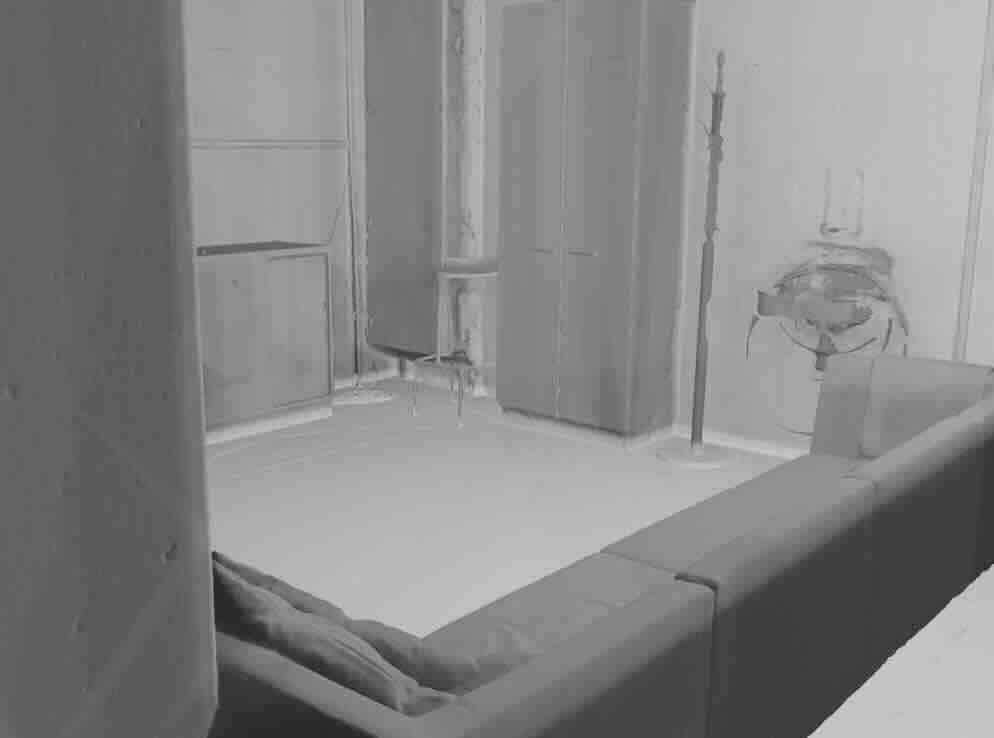}\\
        \includegraphics[width=\mywidthmesh, height=\myheightmesh]{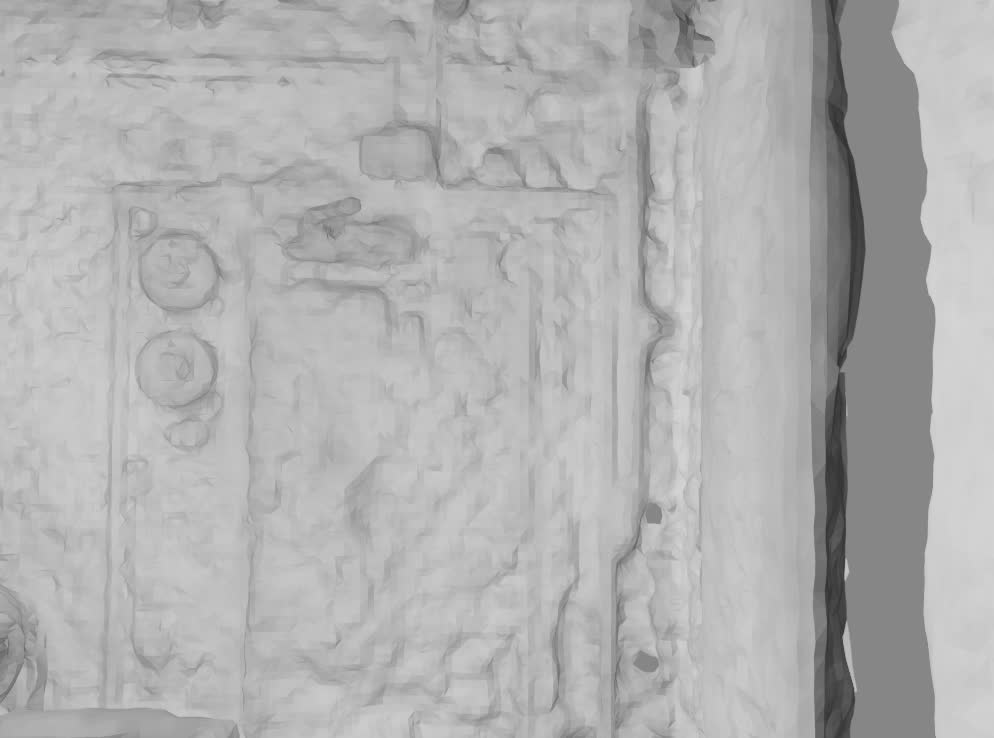}&
        \includegraphics[width=\mywidthmesh, height=\myheightmesh]{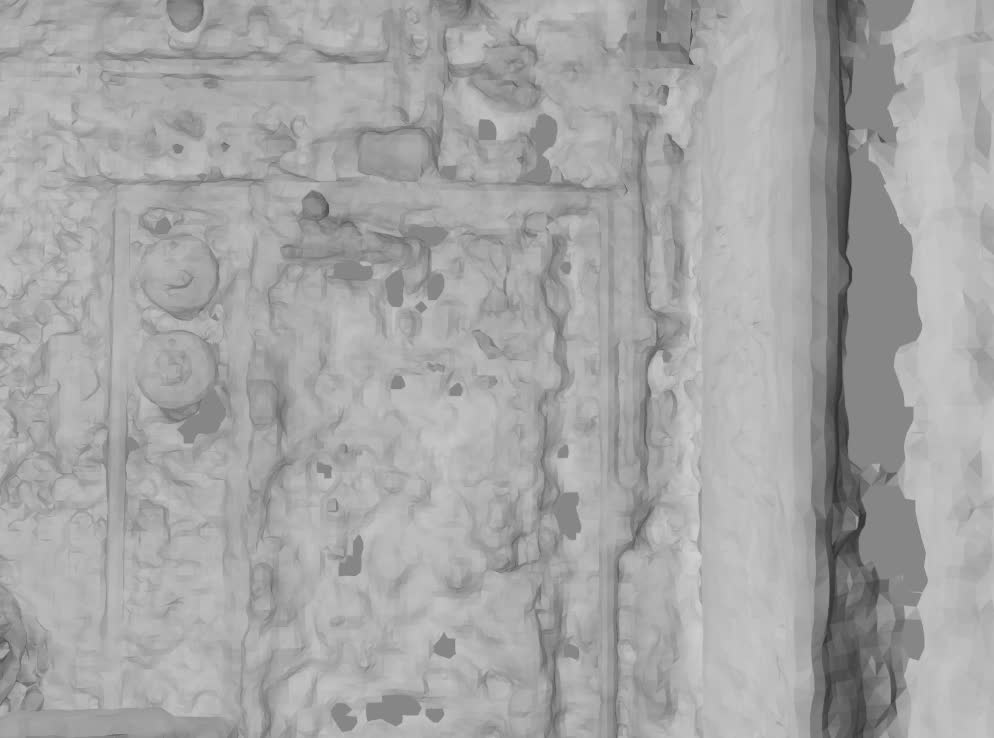}&
        \includegraphics[width=\mywidthmesh, height=\myheightmesh]{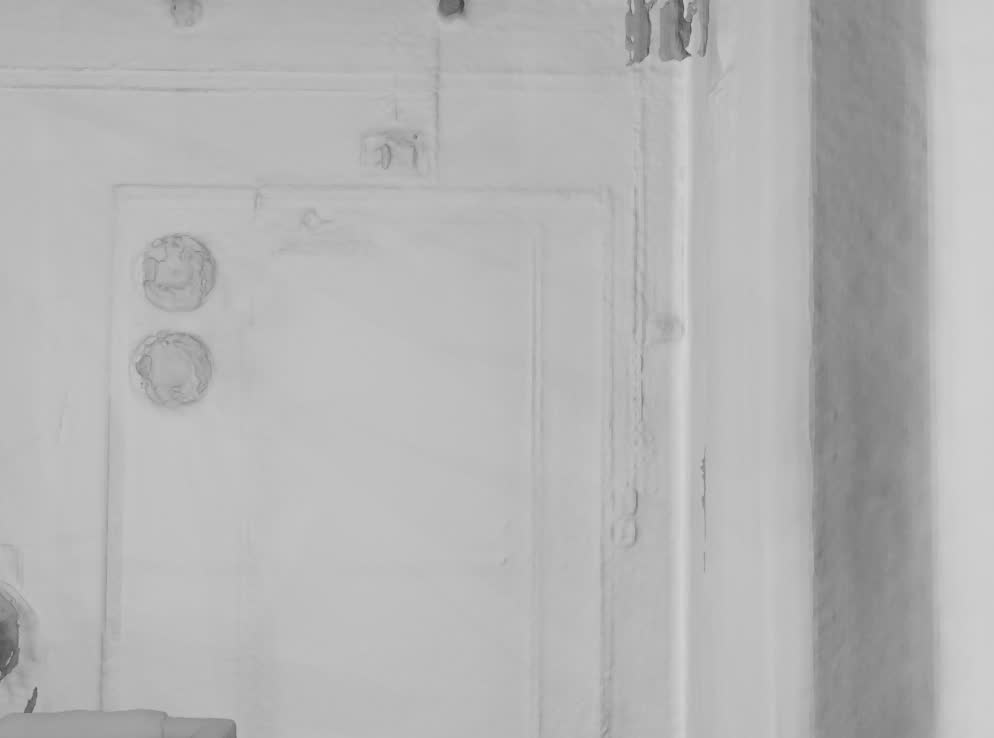}&
        \includegraphics[width=\mywidthmesh, height=\myheightmesh]{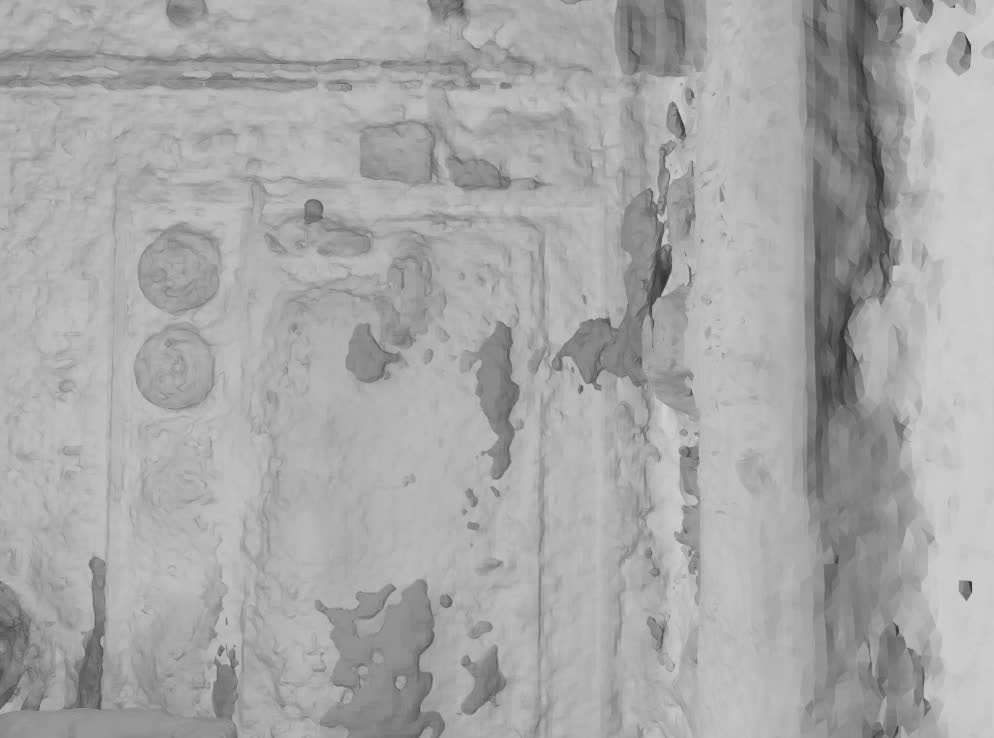}&
        \includegraphics[width=\mywidthmesh, height=\myheightmesh]{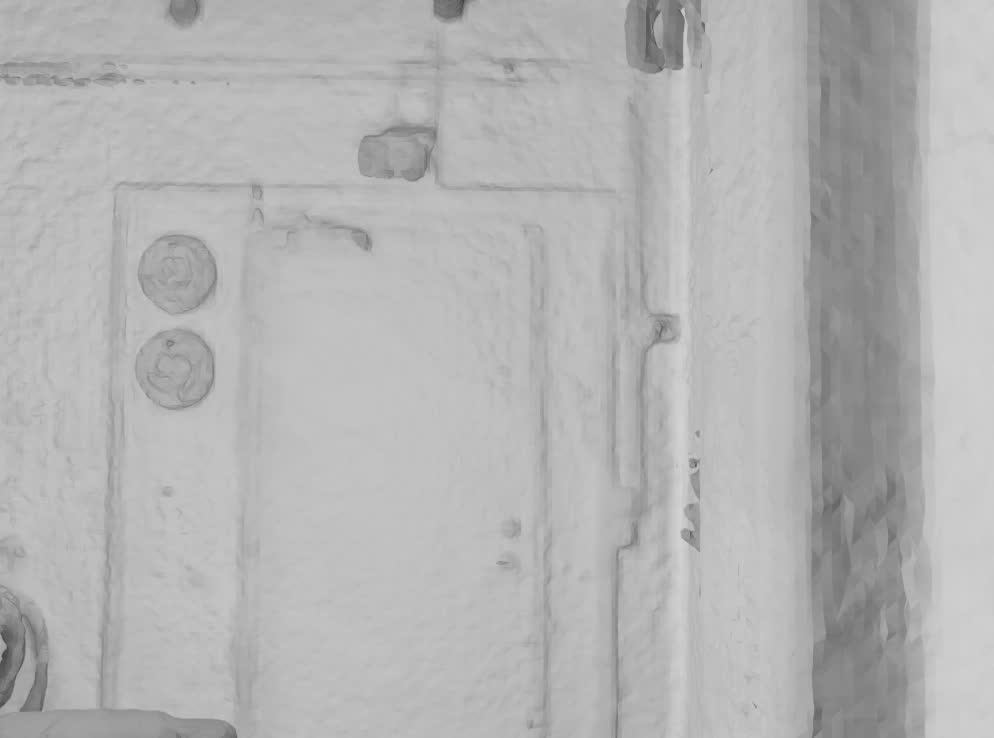}&
        \includegraphics[width=\mywidthmesh, height=\myheightmesh]{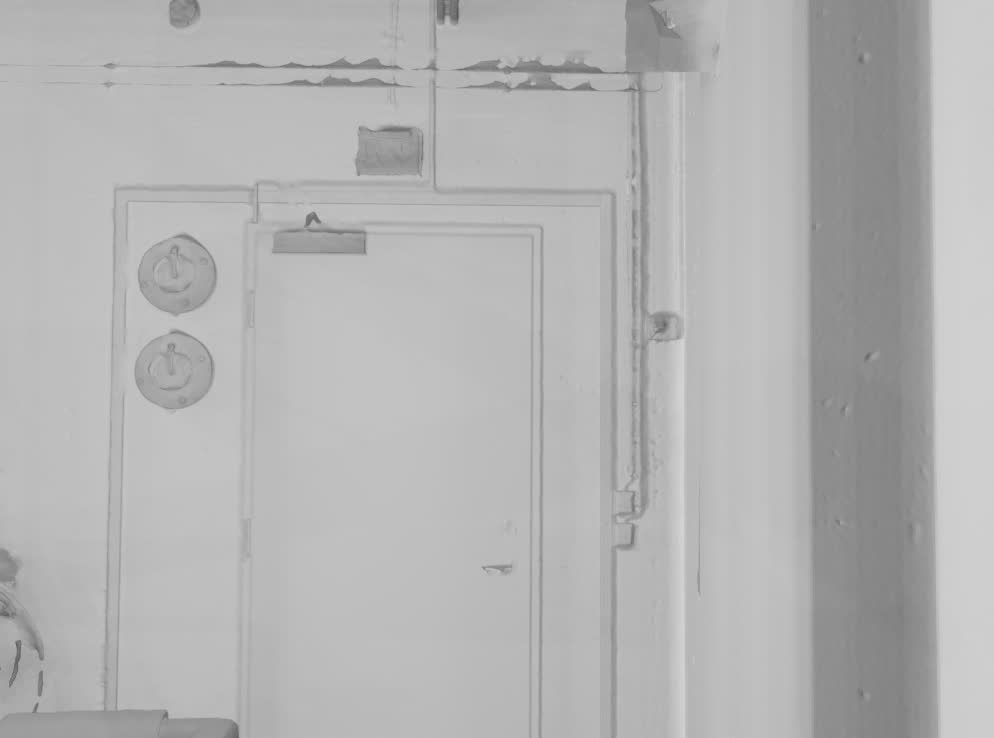}\\
        \includegraphics[width=\mywidthmesh, height=\myheightmesh]{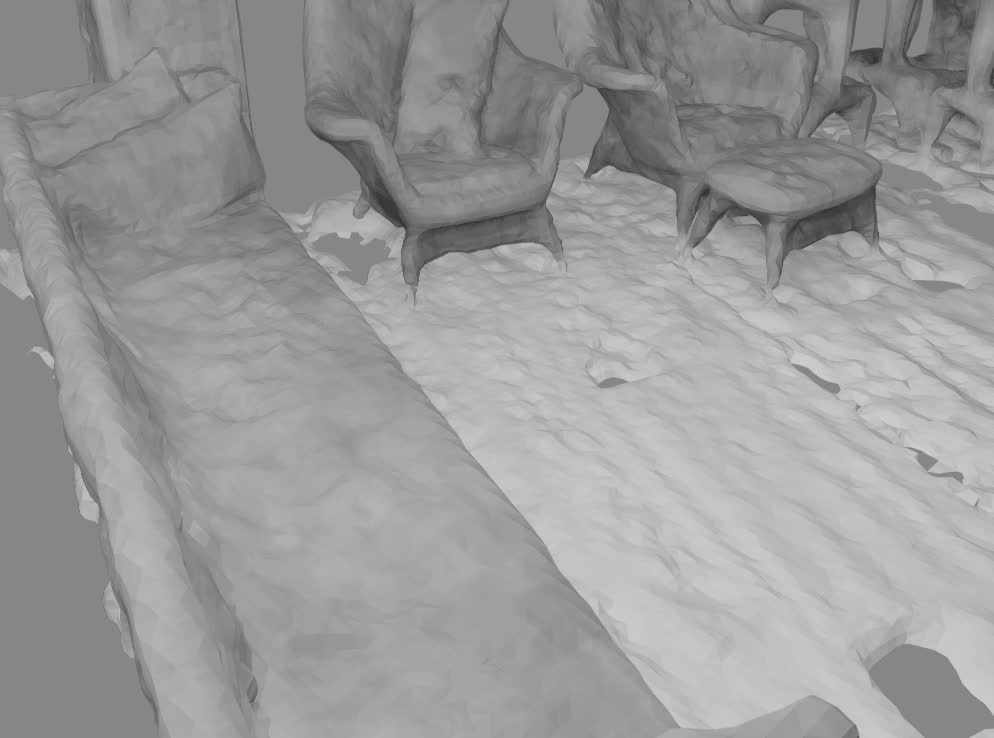}&
        \includegraphics[width=\mywidthmesh, height=\myheightmesh]{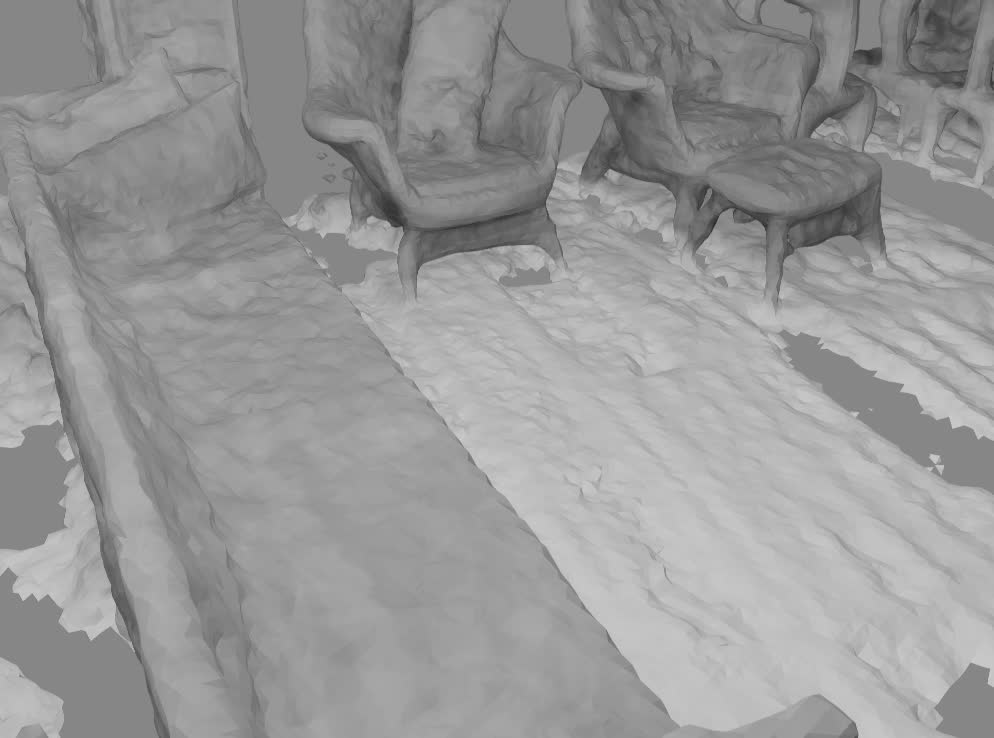}&
        \includegraphics[width=\mywidthmesh, height=\myheightmesh]{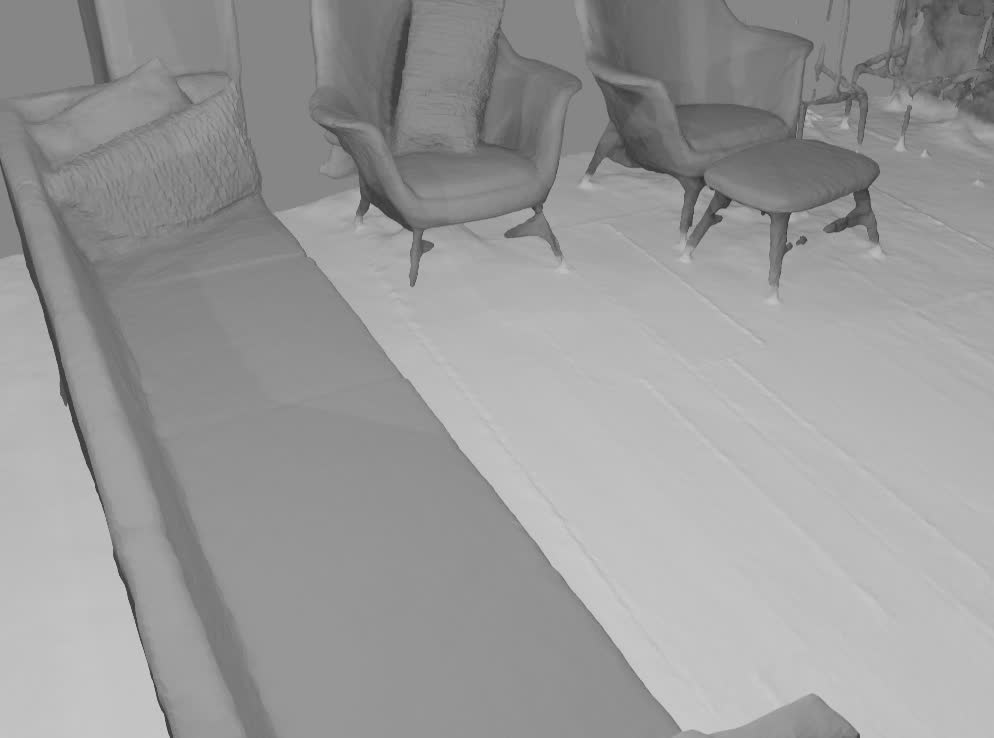}&
        \includegraphics[width=\mywidthmesh, height=\myheightmesh]{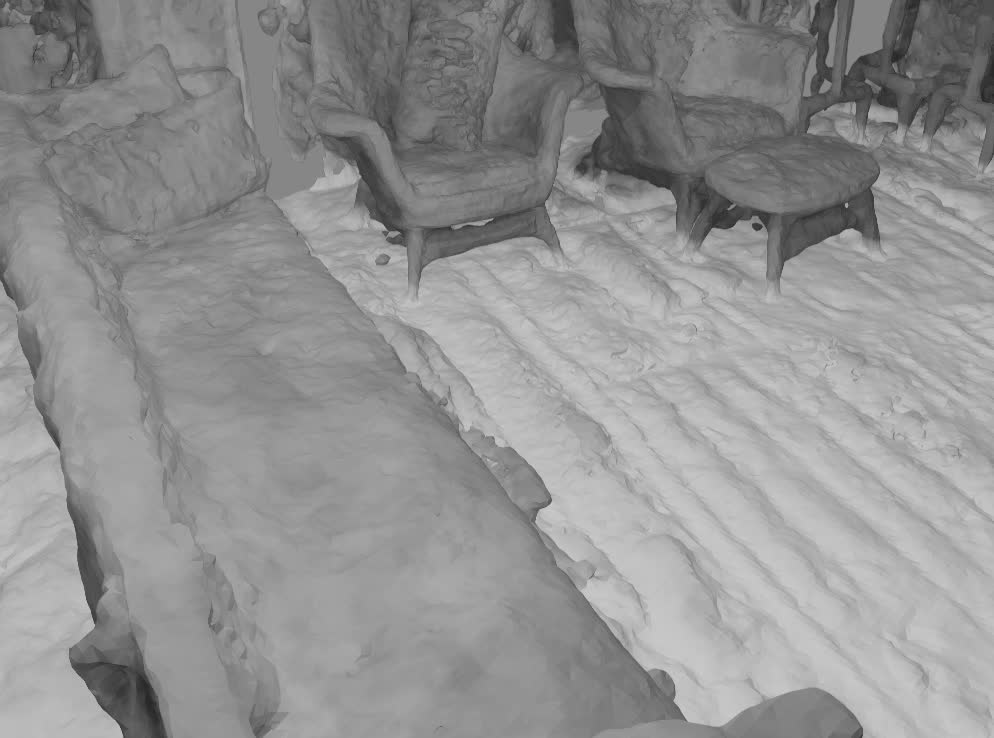}&
        \includegraphics[width=\mywidthmesh, height=\myheightmesh]{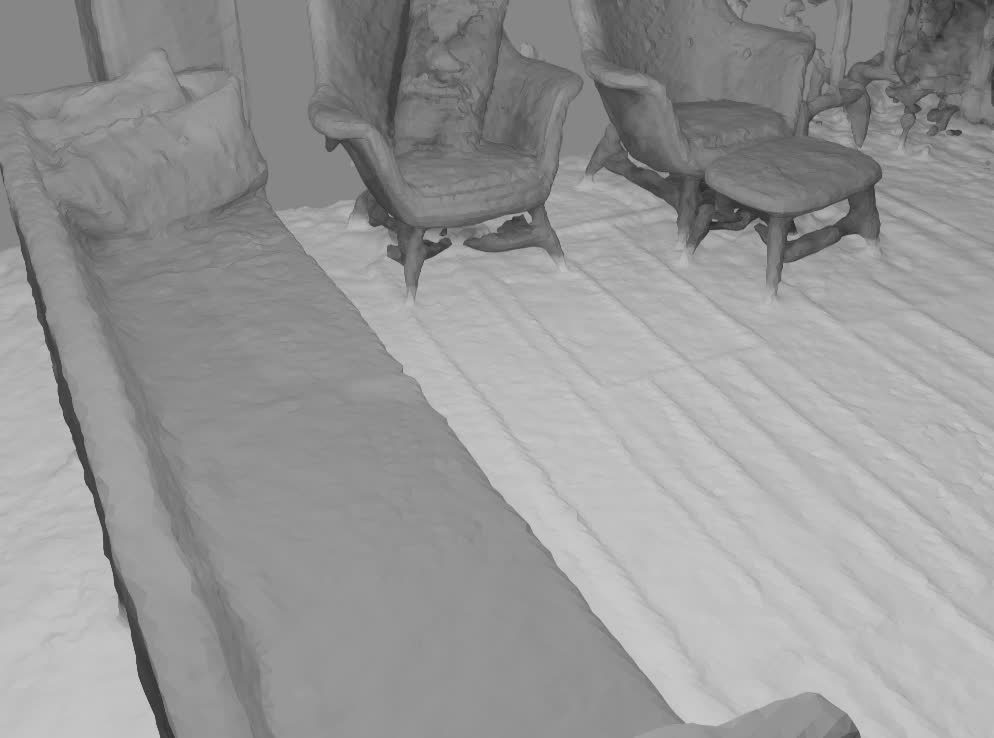}&
        \includegraphics[width=\mywidthmesh, height=\myheightmesh]{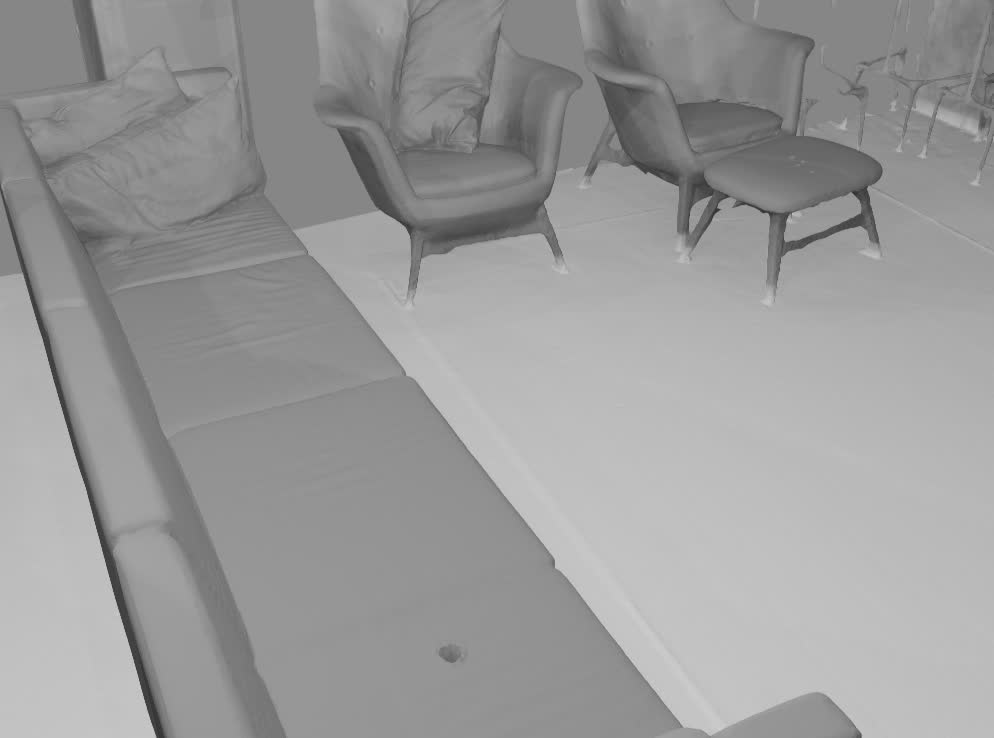}\\
        \includegraphics[width=\mywidthmesh, height=\myheightmesh]{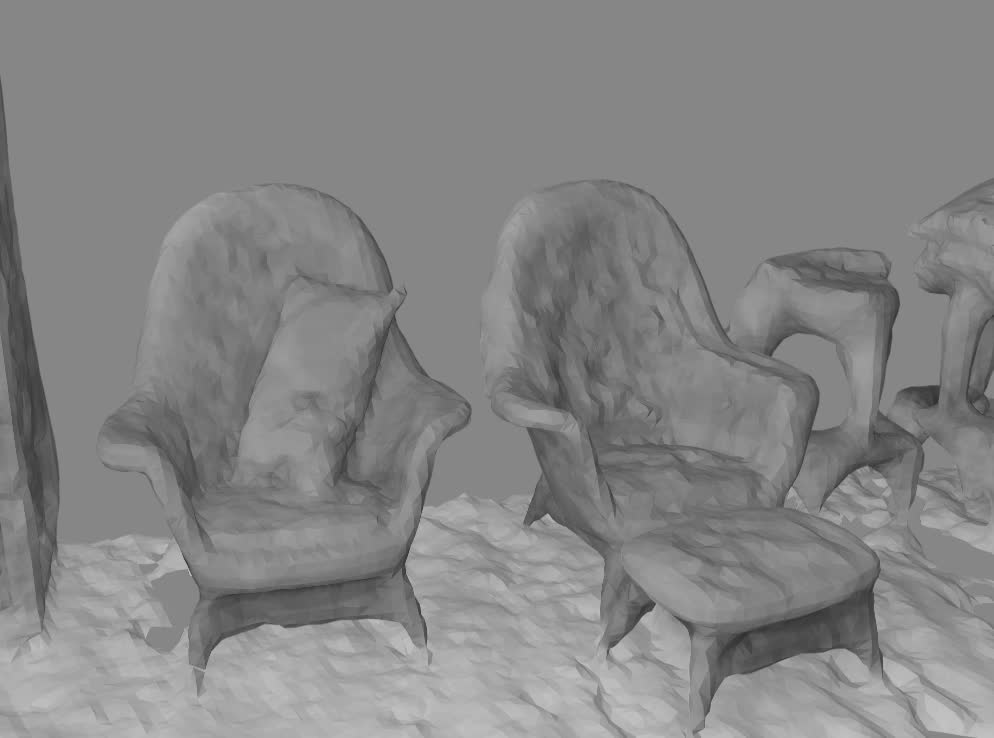}&
        \includegraphics[width=\mywidthmesh, height=\myheightmesh]{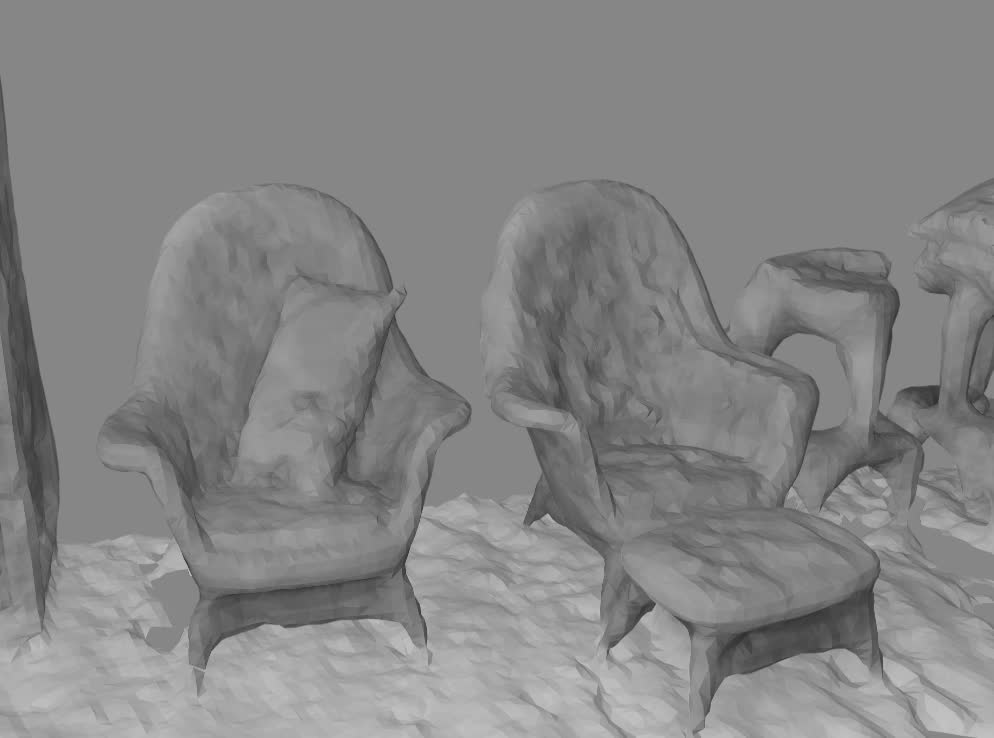}&
        \includegraphics[width=\mywidthmesh, height=\myheightmesh]{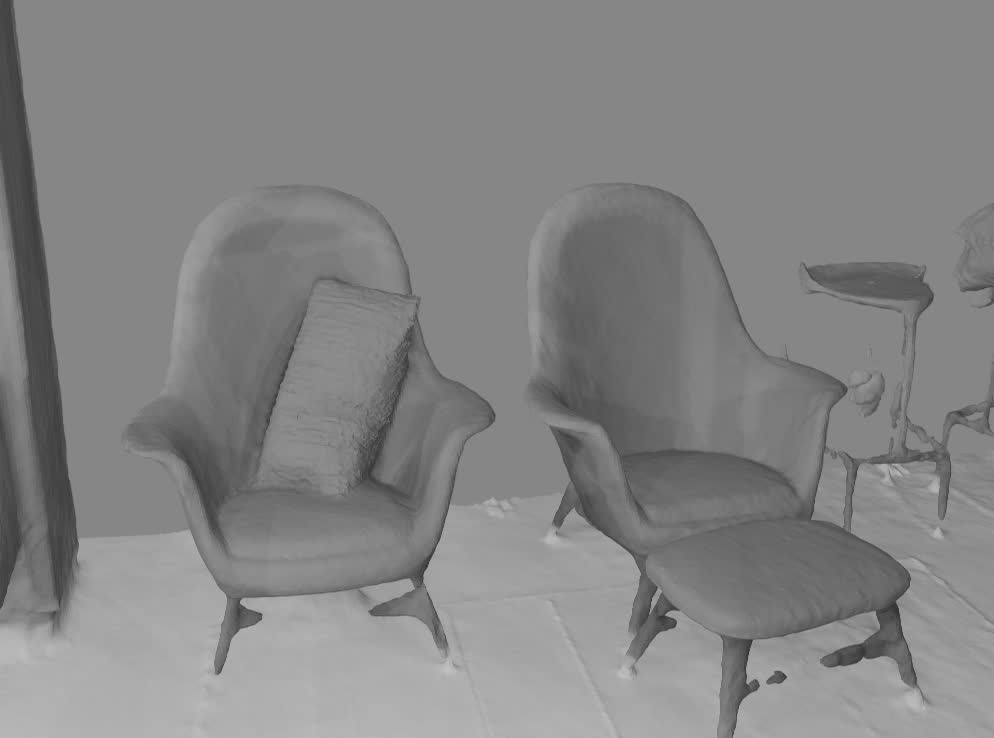}&
        \includegraphics[width=\mywidthmesh, height=\myheightmesh]{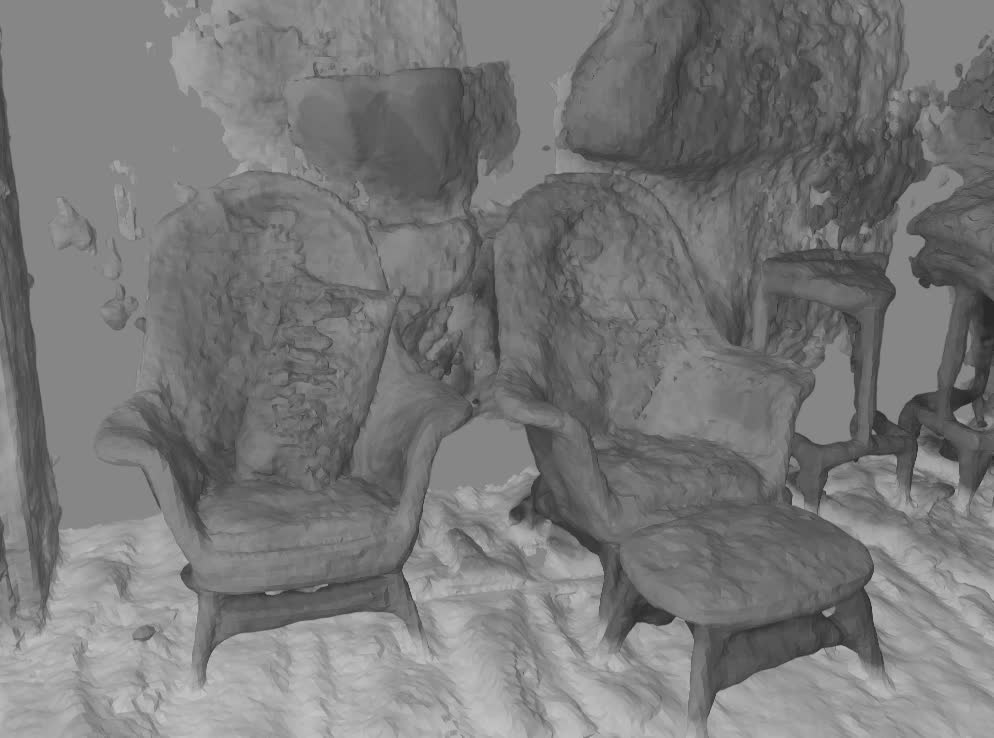}&
        \includegraphics[width=\mywidthmesh, height=\myheightmesh]{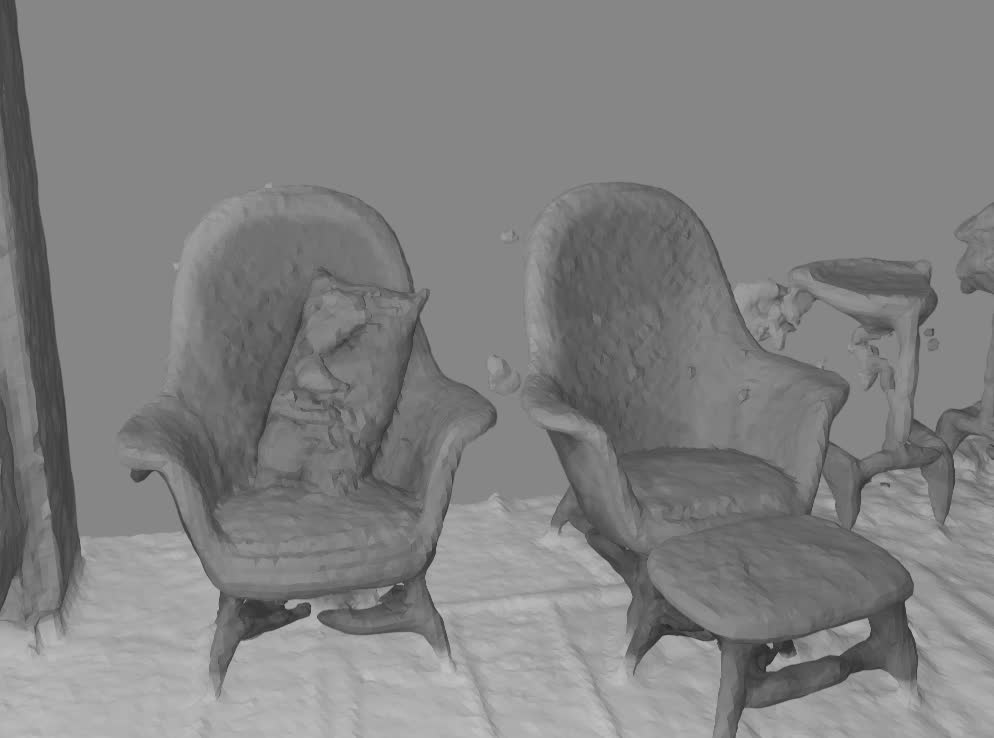}&
        \includegraphics[width=\mywidthmesh, height=\myheightmesh]{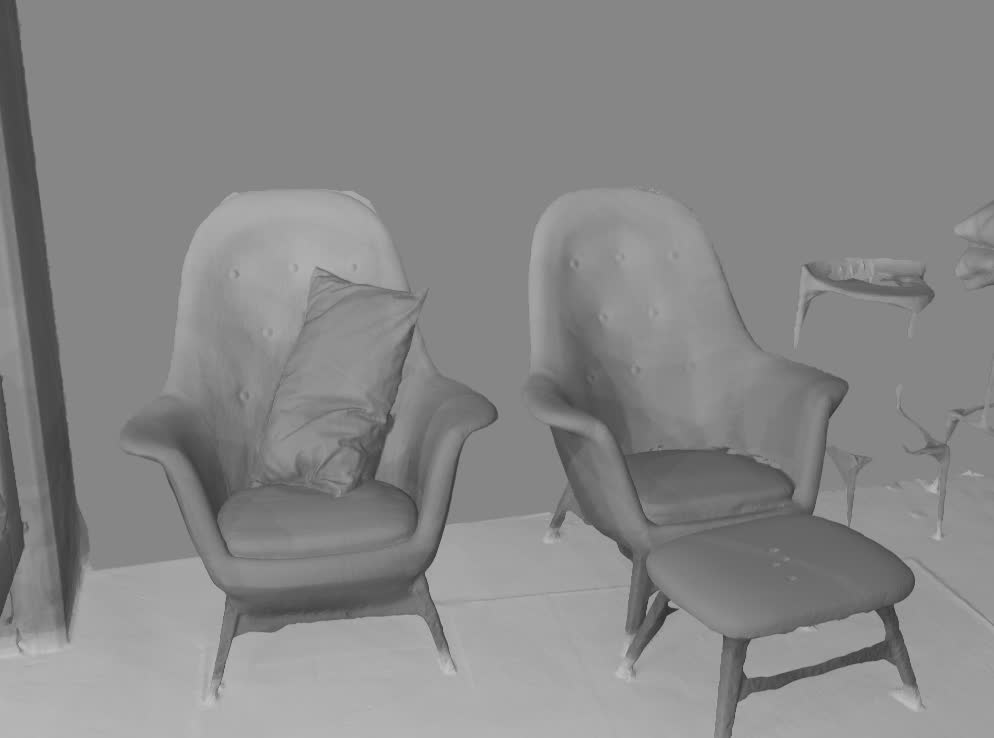}\\
        \includegraphics[width=\mywidthmesh, height=\myheightmesh]{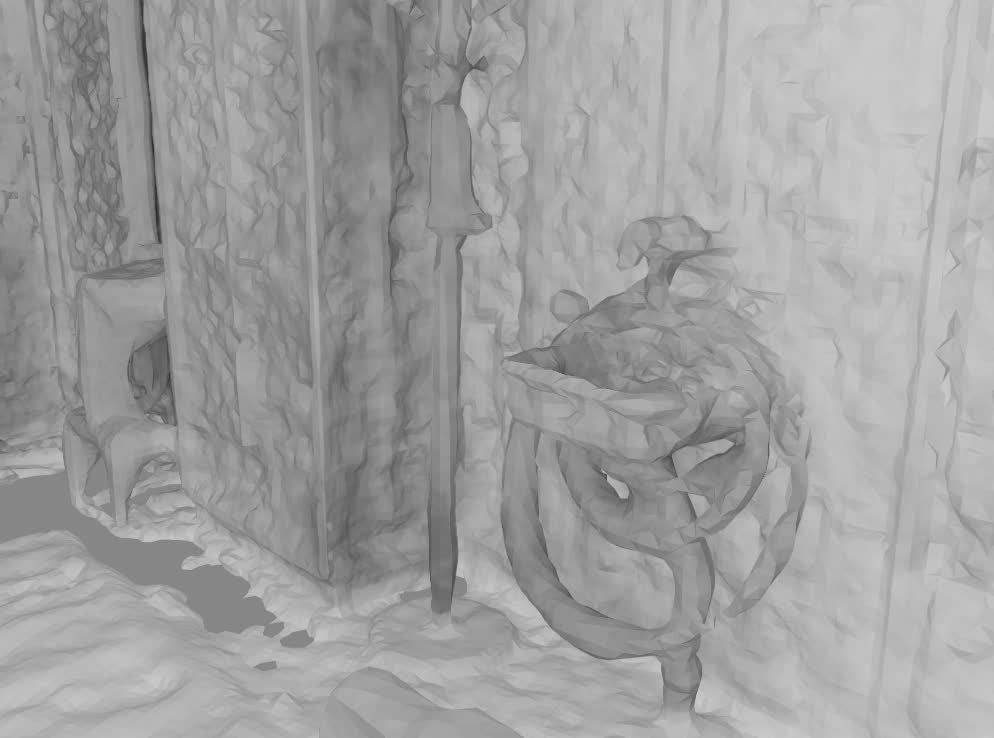}&
        \includegraphics[width=\mywidthmesh, height=\myheightmesh]{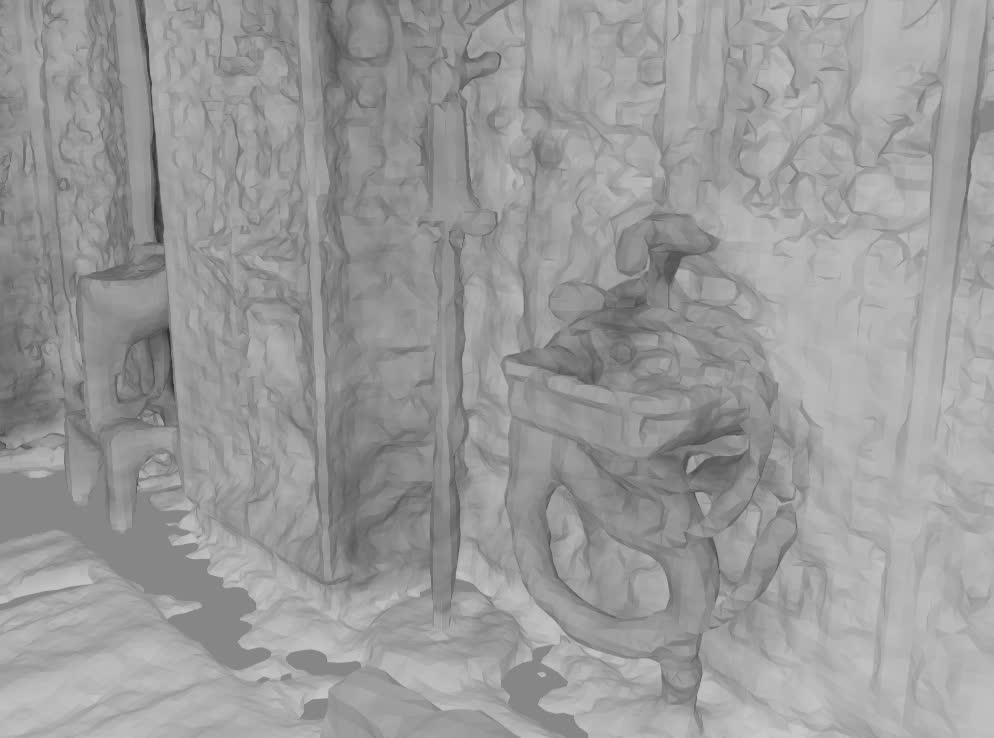}&
        \includegraphics[width=\mywidthmesh, height=\myheightmesh]{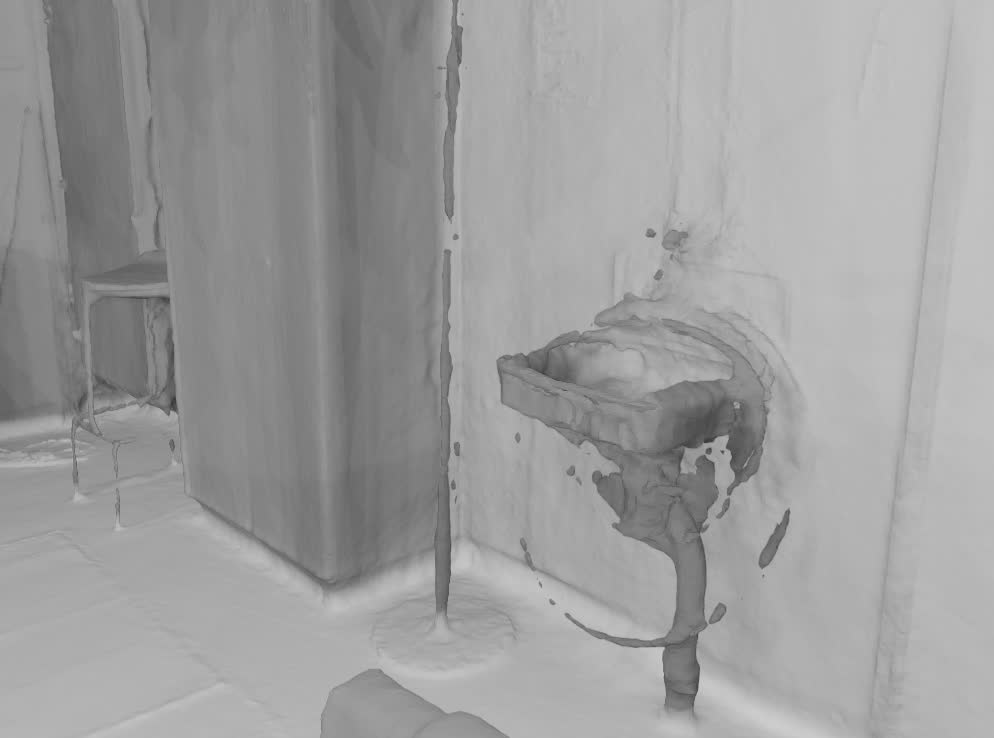}&
        \includegraphics[width=\mywidthmesh, height=\myheightmesh]{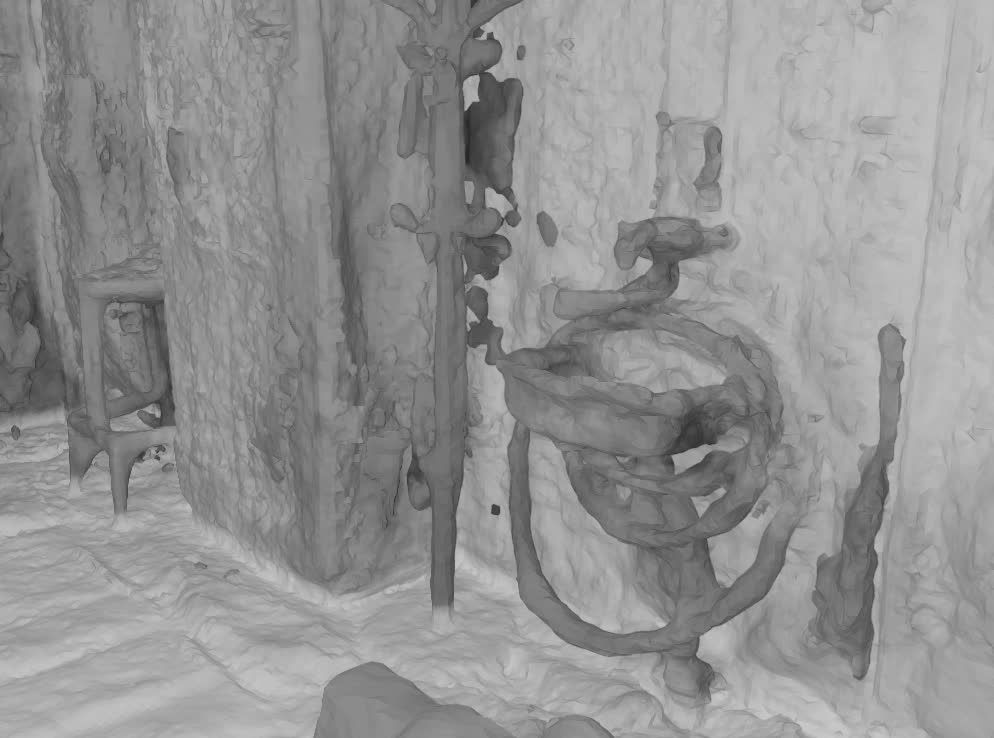}&
        \includegraphics[width=\mywidthmesh, height=\myheightmesh]{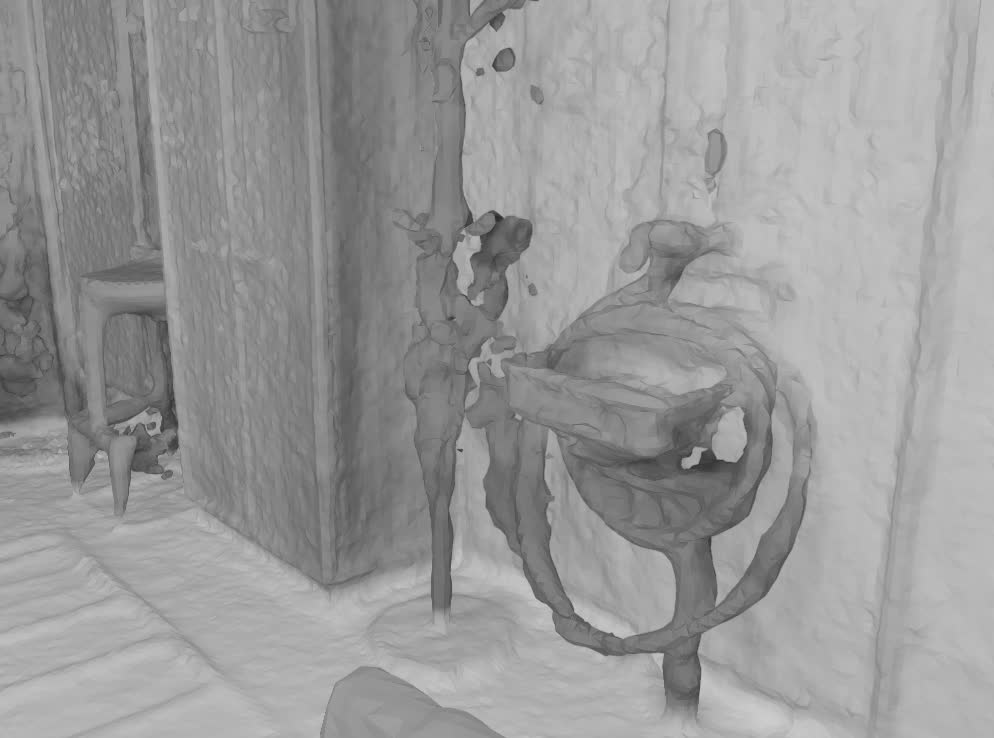}&
        \includegraphics[width=\mywidthmesh, height=\myheightmesh]{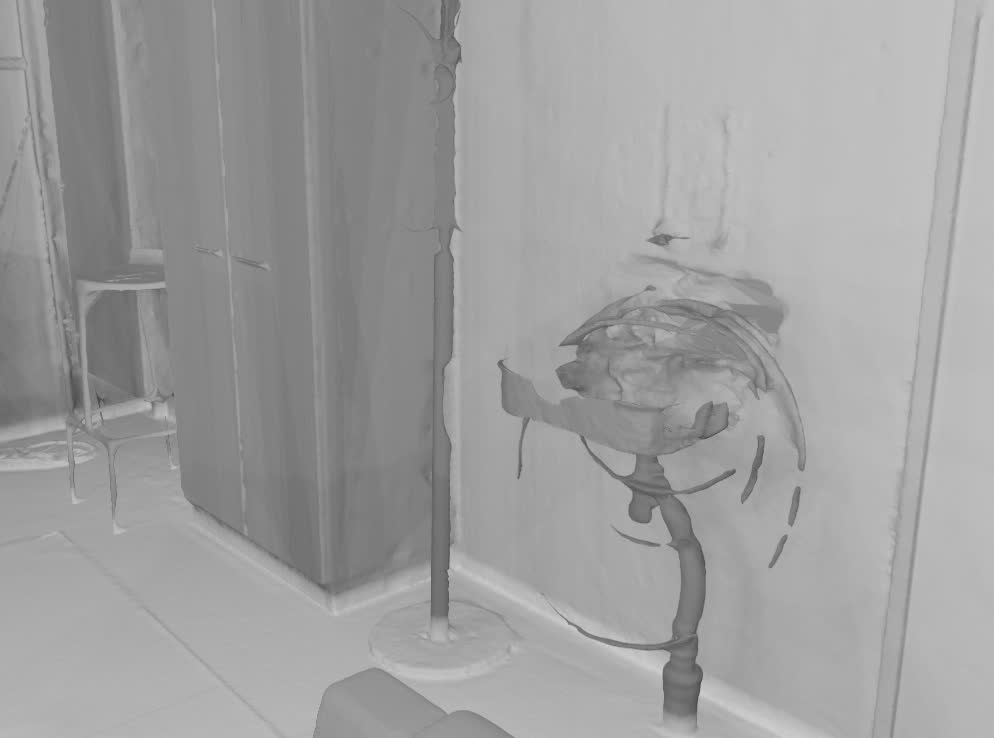}\\
        \includegraphics[width=\mywidthmesh, height=\myheightmesh]{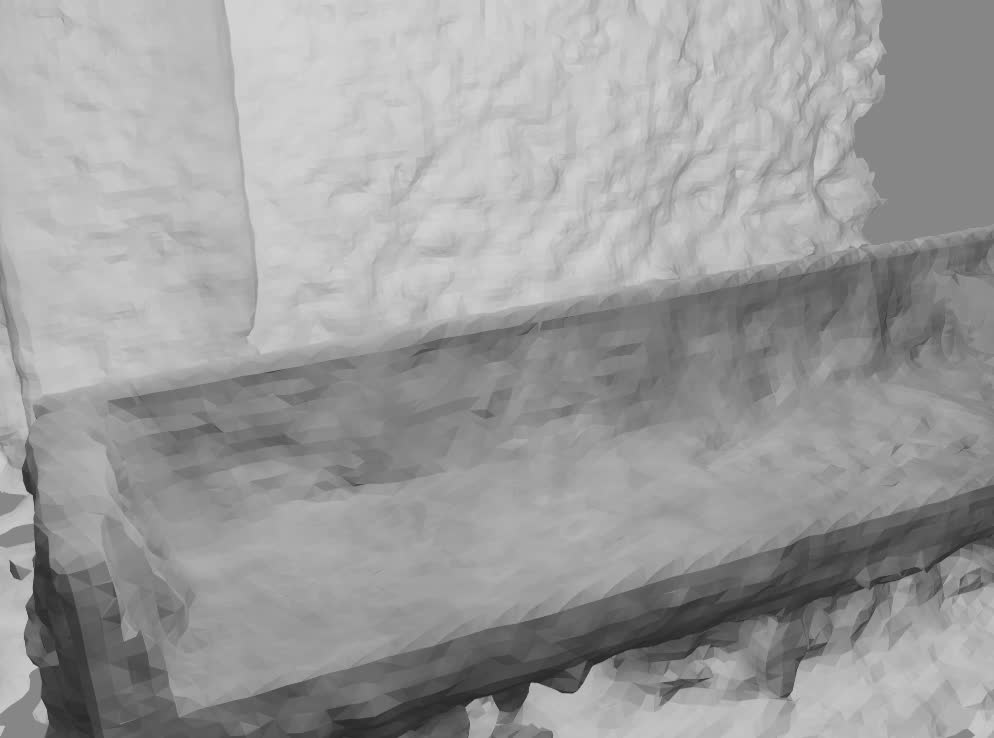}&
        \includegraphics[width=\mywidthmesh, height=\myheightmesh]{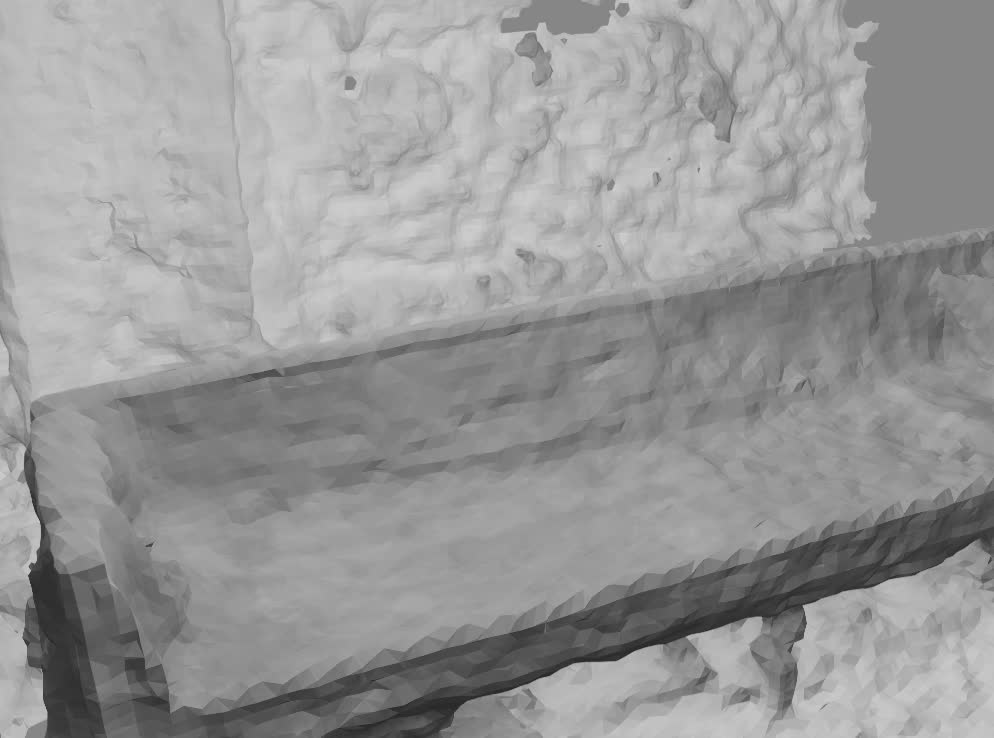}&
        \includegraphics[width=\mywidthmesh, height=\myheightmesh]{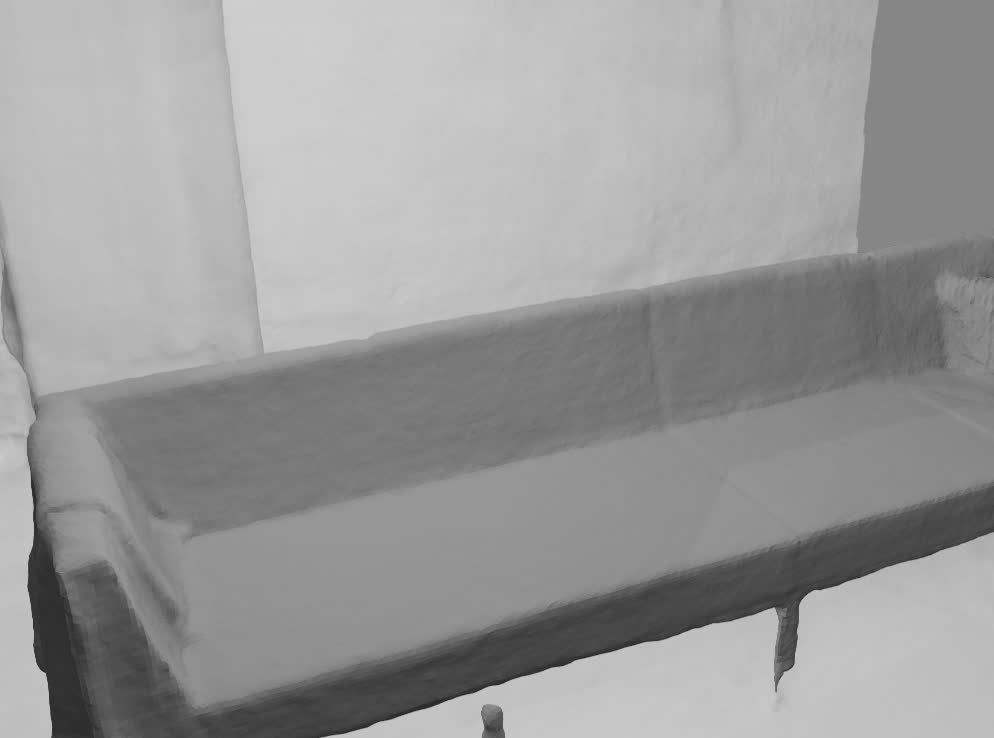}&
        \includegraphics[width=\mywidthmesh, height=\myheightmesh]{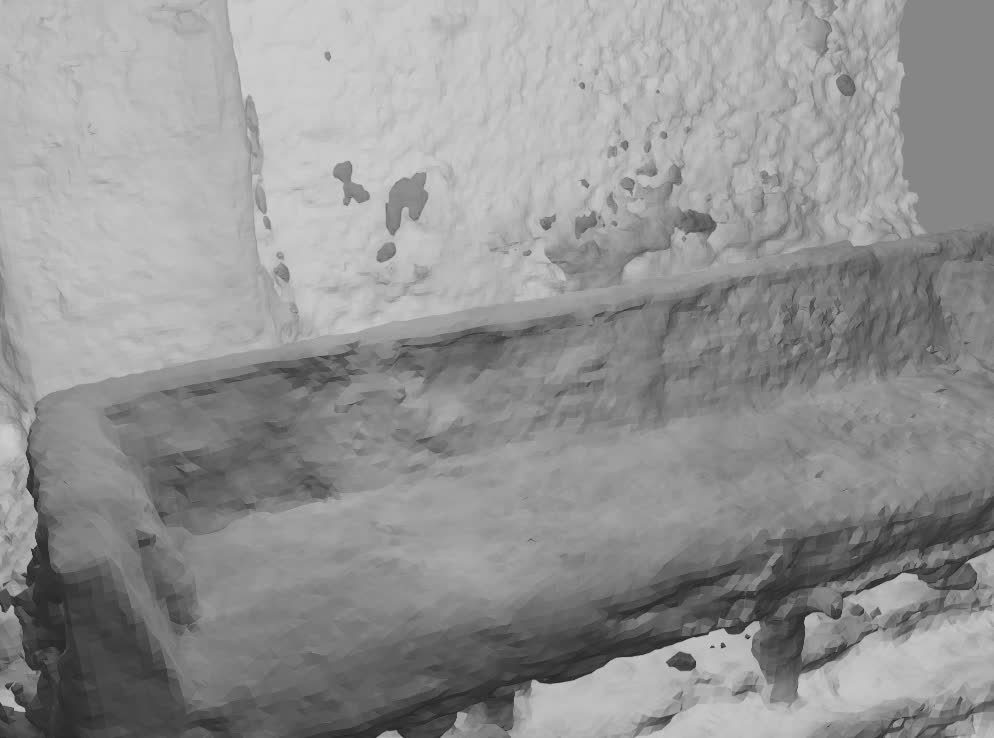}&
        \includegraphics[width=\mywidthmesh, height=\myheightmesh]{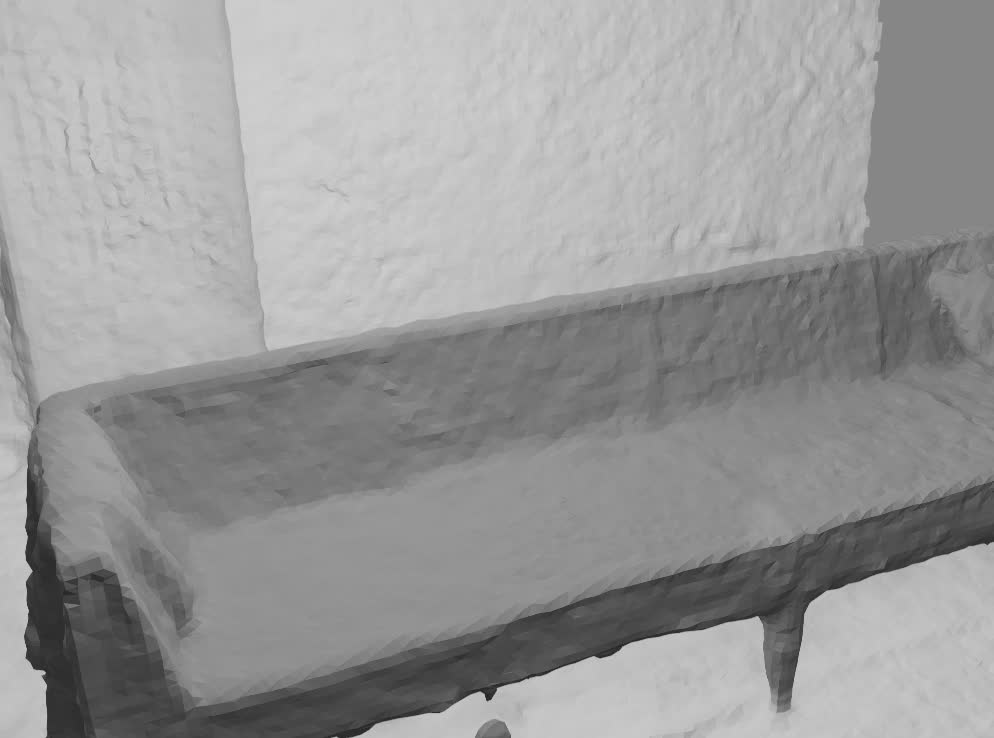}&
        \includegraphics[width=\mywidthmesh, height=\myheightmesh]{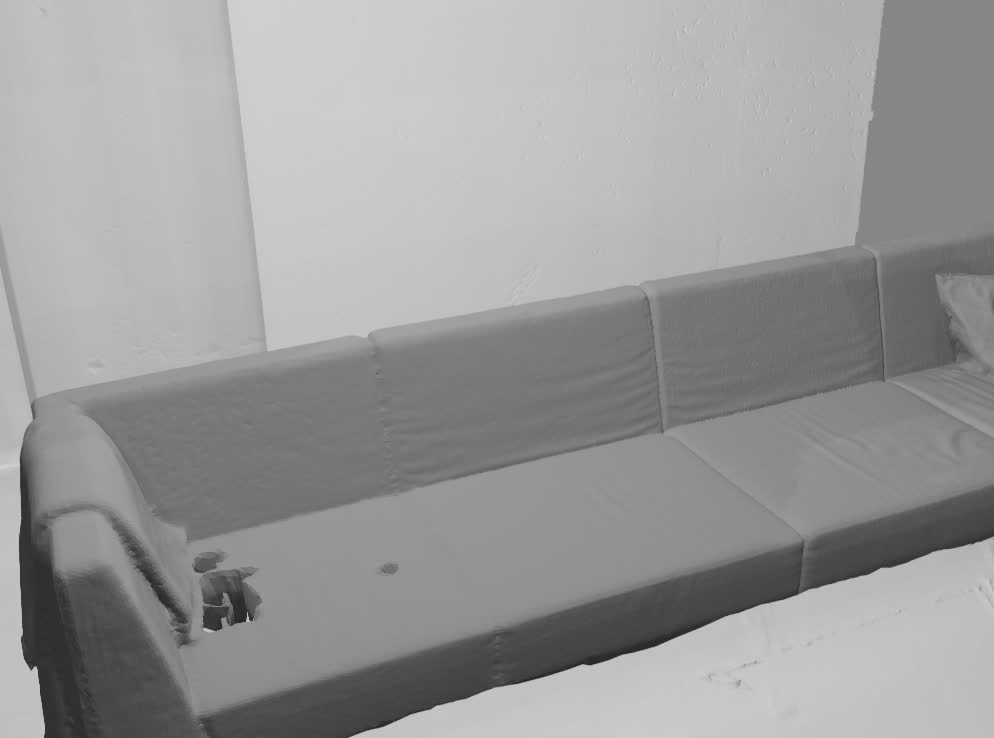}\\
        Nerfacto~\cite{nerfstudio} & Depth-Nerfacto~\cite{nerfstudio} & MonoSDF~\cite{Yu2022MonoSDF} & Splatfacto~\cite{kerbl20233d,nerfstudio} & Ours & iPhone GT\\
    \end{tabular}
  \caption{\textbf{Qualitative comparison on mesh reconstruction.} Comparison of baseline methods on sequences from the MuSHRoom dataset.}
 \label{fig:supp-different-method-mesh}
\end{figure*}

\input{figures/supplement/image_cmp_tex/image_files_mod}
\input{figures/supplement/image_cmp_tex/image_files2_mod}
\input{figures/supplement/image_cmp_tex/image_files3_mod}


\end{document}